\pdfoutput=1

\documentclass[11pt]{article}

\usepackage[final]{acl}
\usepackage{pifont}
\usepackage{times}
\usepackage{latexsym}
\usepackage{multirow}
\usepackage[T1]{fontenc}

\usepackage[utf8]{inputenc}

\usepackage{microtype}

\usepackage{inconsolata}

\usepackage{graphicx}
\usepackage{booktabs}
\usepackage{subfigure}
\usepackage[figuresright]{rotating}
\usepackage{booktabs}
\usepackage{diagbox}
\usepackage{color, colortbl}
\usepackage{amsmath}
\definecolor{LightCyan}{rgb}{0.88,1,1}

\usepackage{enumitem}
\usepackage{amssymb}

\usepackage{tablefootnote}

\newcommand{\Jiayi}[1]{\textcolor{red}{Jiayi}}
\newcommand{\Hongyi}[1]{\textcolor{red}{Hongyi}}
\newcommand{\Henry}[1]{\textcolor{red}{Henry}}

\title{KV Cache Compression, But What Must We Give in Return? \\
A Comprehensive Benchmark of Long Context Capable Approaches
}

 \author{
 \textbf{Jiayi Yuan\textsuperscript{$\ast$1}},
 \textbf{Hongyi Liu\textsuperscript{$\ast$1}},
 \textbf{Shaochen (Henry) Zhong\textsuperscript{$\ast$1}},
\\
 \textbf{Yu-Neng Chuang\textsuperscript{1}},
 \textbf{Songchen Li\textsuperscript{1}},
 \textbf{Guanchu Wang\textsuperscript{1}},
 \textbf{Duy Le\textsuperscript{1, 3}},
 \textbf{Hongye Jin\textsuperscript{2}},
\\
 \textbf{Vipin Chaudhary\textsuperscript{3}},
 \textbf{Zhaozhuo Xu\textsuperscript{4}},
 \textbf{Zirui Liu\textsuperscript{$\clubsuit$ 5}},
 \textbf{Xia Hu\textsuperscript{1}}
\\
\\
 \textsuperscript{1}Rice University,
 \textsuperscript{2}Texas A\&M University,
 \textsuperscript{3}Case Western Reserve University,\\
 \textsuperscript{4}Stevens Institute of Technology,
 \textsuperscript{5}University of Minnesota Twin Cities
\\
}

\begin{document}

\maketitle
\def\thefootnote{$\ast$}\footnotetext{Equal contribution, order determined by rolling dices.}\def\thefootnote{\arabic{footnote}}
\def\thefootnote{$\clubsuit$}\footnotetext{Project lead and correspond to Shaochen
 (Henry) Zhong \texttt{<henry.zhong@rice.edu>} and Zirui Liu \texttt{<zrliu@umn.edu>}.}\def\thefootnote{\arabic{footnote}}

\begin{abstract}
Long context capability is a crucial competency for large language models (LLMs) as it mitigates the human struggle to digest long-form texts. This capability enables complex task-solving scenarios such as book summarization, code assistance, and many more tasks that are traditionally manpower-intensive. However, transformer-based LLMs face significant challenges with long context input due to the growing size of the KV cache and the intrinsic complexity of attending to extended inputs; where multiple schools of efficiency-driven approaches — such as KV cache quantization, token dropping, prompt compression, linear-time sequence models, and hybrid architectures — have been proposed to produce efficient yet long context-capable models. Despite these advancements, no existing work has comprehensively benchmarked these methods in a reasonably aligned environment. \quad In this work, we fill this gap by providing a taxonomy of current methods and evaluating 10+ state-of-the-art approaches across seven categories of long context tasks. Our work reveals numerous previously unknown phenomena and offers insights — as well as a friendly workbench — for the future development of long context-capable LLMs. The source code is available at \url{https://github.com/henryzhongsc/longctx_bench}.

\end{abstract}

\section{Introduction}

Large Language Models (LLMs) have gained significant popularity and recognition due to their exceptional generalizability across a wide range of intellectual tasks. Like any other tool, their most precious utility is demonstrated when they enable us to accomplish tasks beyond our innate capabilities~\citep{gpt3,taylor2022galactica,yuan2023large}. For instance, while driving nails with bare hands is impractical, a hammer makes it feasible. Similarly, humans struggle with digesting and retaining long information, making it essential for LLMs to bridge this gap. The need for long-context capable LLMs is almost universally agreed upon, with different LLM service providers racing to launch models with even greater context lengths. For example, Google's Gemini 1.5 supports a context length of 128K tokens~\citep{reid2024gemini}, and Anthropic's Claude 3 offers a context length of 200K tokens.

However, this powerful long context capability comes with significantly higher costs. In long context scenarios, the key-value cache (KV cache) — which stores attention keys and values during generation to prevent re-computation — becomes the new memory and speed bottlenecks, as its size grows linearly with the number of tokens in the batch. 
For instance, a 500B model with a batch size of 128 and a context length of 8,192 typically requires a 3TB KV cache, imposing a substantial processing burden even on the most advanced hardware solutions~\citep{pope2023eff_scale_tf}. 
Similarly, in open-source models like QWen~\citep{bai2023qwen}, the KV cache size for a 4K context is 0.91 GB, whereas, for a 100K context, it is 22.8 GB~\citep{fu2024challenges} — which is a non-negligible growth regardless of the serving scenario. 
Given the limited memory space available for serving the model, supporting longer contexts usually requires reducing the number of requests that can be processed, leading to higher inference costs.

\begin{table*}[t!]
  \centering
  \caption{Featured methods in our benchmark. ``KV Cache Complexity'' is the complexity w.r.t. the number of input tokens. ``Sys. Supports'' refers to the availability of custom CUDA kernels to support fast serving. ``N/A'' means it can be directly accelerated by existing infrastructure.
  We note that the ``No'' system support for $\mathrm{H_2O}$~\citep{h2o} means it lacks the FlashAttention~\citep{dao2022flashattention} compatible CUDA kernels, making it unsuitable for direct use in an online setting. However, it still offers performance benefits when used in the off-load setting.}
  \label{tab:overview}
  \resizebox{.9\textwidth}{!}{%
  \begin{tabular}{lccc} 
    \toprule
    \textbf{Method} & \textbf{Taxonomy} & \textbf{KV Cache Complexity} & \textbf{Sys. Supports?}  \\ 
    \midrule
    Mamba~\citep{gu2023mamba} & \multirow{3}{*}{Linear-time Model} & \multirow{3}{*}{KV cache free} & Yes \\
    Mamba 2~\citep{dao2024mamba2} &  & & Yes \\
    RWKV~\citep{peng2023rwkv} & & & Yes \\ 
    \midrule
    \multirow{2}{*}{RecurrentGemma~\citep{botev2024recurrentgemma}} & Linear-time Model & \multirow{2}{*}{Constant} & \multirow{2}{*}{Yes} \\ 
    & + Local Attention & &\\
    \midrule
    StreamingLLM~\citep{streamllm} & \multirow{3}{*}{Token Dropping} & \multirow{3}{*}{Constant} & Yes \\
    $\mathrm{H_2O}$~\citep{h2o} & & & No \\
    InfLLM~\citep{xiao2024infllm} & & & Yes \\ 
    \midrule
    LLMLingua2~\citep{jiang2023llmlingua} & Prompt Compression & Constant & N/A \\
    \midrule
    FlexGen~\citep{sheng2023flexgen} & \multirow{2}{*}{Quantization} & \multirow{2}{*}{Linear} & Yes \\
    KIVI~\citep{kivi}  & & & Yes \\
    \bottomrule
    \vspace{-2em}
  \end{tabular}}
\end{table*}

Naturally, many efficiency-driven approaches have been proposed to enable LLMs to handle long contexts with reduced resource burdens, with a healthy selection of them featured in Table~\ref{tab:overview}. These approaches range from quantizing the KV cache into lower precision formats \citep{sheng2023flexgen, zhao2024atom, kivi}, evicting unimportant tokens to maintain a constant KV cache size \citep{streamllm, h2o}, compressing long prompt into a shorter input \citep{jiang2023llmlingua1, jiang2023llmlingua, chuang2024learning}, or exploring KV cache-free architectural designs \citep{gu2023mamba, peng2023rwkv, yang2023gated, qin2024hgrn2} and its hybrids with transformers \citep{de2024griffin, lieber2024jamba}. However, to the best of our knowledge, \textbf{no prior art has provided a comprehensive benchmark to analyze the performance retention of different long context-capable compression methods}\footnote{Due to the lack of directly related work, we provide a brief walkthrough of loosely related arts — which are often long context datasets evaluated on vanilla baseline models with limited focus on compression methods — in Appendix~\ref{app: related works}.} (which is also non-trivial to setup; more on this in Section~\ref{sec_bench_setup}). To fill this gap, we aim to answer the following question:

\emph{How do different long context-capable approaches perform under different long context tasks?}

This benchmark offers an accessible and reproducible pipeline to evaluate a diverse set of modern long-context compression methods from various schools of thought. It assesses these methods against multiple tasks requiring different long-context capabilities. Our main contributions are summarized as follows:

\begin{itemize}[leftmargin=*, noitemsep, topsep=0pt]
    \item \textbf{Comprehensive benchmarking, detailed analysis, and actionable insights:} We provide a comprehensive evaluation report that covers 10+ long context-capable efficient approaches under 65 different settings, against 7 categories of long context tasks \citep{mohtashami2023passkey, reid2024gemini, bai2023longbench}. We then walk through how to digest such mass results and provide analyses and discussion upon many previously unknown phenomena. Finally, we offer several actionable insights for future research advancement. 
    \item \textbf{Minimalistic, reproducible, yet extensible platform:} Given the non-trivial effort to set up the evaluation pipeline, we open source our benchmark implementations for future scholars. We intensionally make our code base in a minimalistic fashion for easier hacking and reproducing needs, yet we keep it extensible to include alternative or future-coming approaches that are not under in our already extensive, but certainly not exhaustive, benchmark coverage.
\end{itemize}

\section{Reviewing Different Schools of Efficient Long Context Handling Approcahes}
\label{sec_review}

Before going into the details of the experiment, we will briefly introduce different schools of long context-capable approaches and their corresponding exemplary methods. In Table~\ref{tab:overview}, we present a comprehensive overview of the school of long context optimization methods, including their KV cache complexities and the current support for system-level optimization. RNN-based models do not have a KV cache. Mixed models, token dropping methods, and prompt compression methods have fixed-size KV caches, which are independently configured by each method. Quantization methods compress the KV cache by a proportion; thus, the KV cache complexity still increases linearly with sequence length. Regarding system support scenarios, to the best of our knowledge, most methods have varying levels of system-level optimization, whereas some token-dropping methods are still under-optimized. More on this in Section~\ref{sec: challenge}.

\subsection{Linear-Time Sequence Models and Mixed Architecture}

There is a growing body of recent works that have developed linear-time sequence models, such as Mamba~\cite{gu2023mamba}, Mamba2~\citep{dao2024mamba2}, RWKV~\cite{peng2023rwkv}, HGRN~\cite{qin2024hgrn2}, MEGA~\cite{ma2022mega}, GLA~\cite{yang2023gated}, and RetNet~\cite{retnet}.
The fundamental difference between linear-time sequence models and transformers lies in how they handle context. Linear-time sequence models compress the context into a smaller state, whereas transformers store the entire context within attention mechanisms.
During the auto-regressive inference, every time the model generates a new token, transformers will ``review'' all previous tokens by explicitly storing the entire context (i.e., KV cache).
In contrast, there is no ``reviewing'' mechanism in linear-time sequence models, as they explicitly mix the input tokens into finite states.

From the above analysis, it is expected that pure linear-time sequence models are not well-suited for retrieval-related tasks, as they mix key information with other tokens. 
Thus, another line of work is to combine the linear-time sequence models and transformers.
For example, Griffin~\cite{de2024griffin} and RecurrentGemma~\cite{botev2024recurrentgemma} combine input-dependent RNNs with local attention; and Jamba~\cite{lieber2024jamba} combines full attention layers and Mamba layers.

\subsection{Quantization}

A simple yet effective approach to reducing the size of KV cache to enable a larger context is to quantize the floating-point numbers (FPN) in the KV cache using fewer bits. Specifically, the $B$-bit integer quantization-dequantization process can be expressed as:

\vspace{-1.5em}
\begin{align*}
    Q(\mathbf{X}) = \lfloor \frac{\mathbf{X} - z_X}{s_X} \rceil,~~
    \mathbf{X}' = Q(\mathbf{X}) \cdot s_X + z_X,
\end{align*}
\vspace{-1em}

where $z_X=\min \mathbf{X}$ is the zero-point, $s_X=(\max \mathbf{X} - \min \mathbf{X})/(2^B - 1)$ is the scaling factor, and $\lfloor\cdot\rceil$ is the rounding operation. 

FlexGen~\cite{sheng2023flexgen} utilized group-wise quantization, achieving 4bit quantization compared to standard 16bit with minimal accuracy loss. Following this, several other quantization methods have been proposed specifically for the KV cache~\cite{zhao2024atom, yang2024no, dong2024qaq}. Recently, KIVI~\cite{kivi} and KVQuant~\cite{hooper2024kvquant} advanced KV cache quantization to even lower bits by introducing per-channel quantization, which involves grouping tensor elements along the channel dimension, based on the discovery of channel outliers in the key cache. Following this finding, some other works continue to optimize this process~\cite{kang2024gear, duanmu2024skvq, zandieh2024qjl}. Furthermore, based on these findings, the latest research has pushed quantization to 1bit~\cite{zhang2024kv}.
The transformer-based LLM inference workflow involves two
stages: i) \emph{prefill stage}, where the input prompt is used to generate KV cache and the first output token; and
ii) \emph{decoding stage}, where the model uses and updates KV cache to generate the next token one by one.
\textbf{We emphasize that for all KV cache quantization methods evaluated in this paper, the quantized KV cache is not used in prefill time}. That means that KV cache quantization only affects the decoding phase.

\subsection{Token Dropping}


Based on the observation that attention scores are highly sparse, token dropping-based methods drop the unimportant token — or similar attention components — from the KV cache \cite{h2o, streamllm, xiao2024infllm, li2024qllm, li2024snapkv, liu2024scissorhands, ge2023model, jiang2024minference}. 
\textbf{Token dropping-based methods fall into two main categories: dropping tokens during prefill or dropping tokens after prefill.}
Dropping tokens during prefill means that tokens are dropped while generating the KV cache. 
In contrast, dropping tokens after prefill means generating the full KV cache first, then removing the unimportant tokens from it.
Given transformers inference process typically involves two phases, i.e., prefill and decoding,
\textbf{while dropping tokens during prefill can typically enable longer sequence length and faster prefill speed, we note that dropping tokens after prefill consistently yields better results across various settings}.
This is because many token-dropping methods rely on accurate attention scores to determine token importance, which benefits from generating the full KV cache first. In our benchmark, methods that drop tokens during prefill include StreamingLLM \citep{streamllm} and InfLLM \citep{xiao2024infllm}, where $\mathrm{H_2O}$ \citep{h2o} represents methods that drop tokens after prefill. We closely follow the official or endorsed implementation of each method, with more details shared in Appendix \ref{app_method_specific}.

\subsection{Prompt Compression}

\paragraph{Soft Prompt Compression}
Most existing work focuses on converting lengthy prompts into trainable soft prompts optimized with specific LLMs. One approach uses knowledge distillation to transform hard prompts into soft prompts~\cite{wingate2022prompt}. Another leverages LLM summarization to condense prompts by segmenting and compressing information~\cite{chevalier2023adapting}. Gist Token~\cite{mu2023learning} creates customized prefix soft prompts via a virtual soft prompt predictor. However, these methods are often model-or-even-task-specific, requiring training tailored to specific LLMs, and therefore come with limited adaptability. In this work, we focus on general compression methods for fair comparison with other KV cache compression approaches.

\vspace{-0.5em}
\paragraph{Natural Language Prompt Compression}

Methods like LLMLingua family~\citep{jiang2023llmlingua, jiang2023llmlingua1} enhance LLM performance on long-context tasks by converting long prompts into short prompts while maintaining their natural language format, and thus naturally adaptable (and often even transferable) to all LLMs. LLMLingua employs a budget controller to dynamically allocate compression ratios to different prompt parts, ensuring semantic integrity. Unlike LLMLingua's general approach, some hard prompt compression methods, like Nano-Capsulator~\citep{chuang2024learning}, provide task-specific compression to preserve long prompt performance and are therefore excluded in our benchmark.

\subsection{Other Schools of Thought: Linear Attention, Merging, and More.}

Other than the above-featured approaches, several notable avenues for efficient long context handling include \textit{linear attention} and \textit{merging}. Linear Attention is a well-explored area of transformer modification with many impactful prior arts like LinearAttention \citep{katharopoulos2020linearattention}, MetaFormer \citep{yu2022metaformer}, LinFormer \citep{wang2020linformer} and more, with most of them mainly focus on vision or natural language understanding tasks. To the best of our knowledge, Infini-Attention by \citet{munkhdalai2024infiniattention} is likely one of the most impactful linear attention approaches under the LLM context. 

KV cache merging is also a popular approach due to the mainstream adaptation of GQA \citep{ainslie2023gqa}, GQA and MQA \citep{shazeer2019mqa} conduct merging at the transformer head dimension to enable KV cache reuse. Similar cache-sharing strategies have been developed at the layer or token levels \citep{sun2024yoco, brandon2024cla, wu2024lckv,nawrot2024dmc}. Most of the techniques proposed under this category require intervention during the pre-training stage.

Unfortunately, we are unable to feature these schools of thought, since our work requires scaled-up open-source models in such designs to be available in the first place. With the lack of such availability, we cannot feature them \textit{per se} in our evaluation. However, we are able to feature Mamba 2 \citep{dao2024mamba2} — a model family with a generalized linear attention mechanism — at the 2.7B scale and thus provide some relevant results. We also direct our readers' attention to some recent attention variants like MLA \citep{deepseekv2}.

\section{Benchmarking}
\label{sec_bench}

Benchmarking such a variety of methods in a reasonable manner requires significant effort in terms of experiment design, execution, and computational resources. We first introduce the datasets and methods covered, along with the justifications for their selection. Then, we detail the experiment setup and explain how to interpret our experiment reports. Finally, we analyze the reported results by highlighting some interesting phenomena and providing insights for future scholars. All experiments are conducted on one or more 80G NVIDIA A100 GPUs under DGXA100 servers.

\newcounter{mpFootnoteValueSaver}

\begin{table*}[ht!]
\centering
\caption{Performance of KV cache quantization, token dropping, prompt compression, RNNs, and hybrid methods on our benchmark. For methods with residual full precision inputs like KIVI, we calculate the ``Comp. Ratio'' against 10k input length. 
``LB Avg.'' refers to the average results on LongBench. Results are abbreviated; please refer to Appendix~\ref{app: more exp} for our full report.}

\resizebox{\textwidth}{!}{
\begin{tabular}{c|l|c|cccccc|cc}
\toprule
\textbf{Model}      & \textbf{Method} & Comp. Ratio & Single. QA & Multi. QA & Summ. & Few-shot & Synthetic & Code & \textbf{LB Avg.} & \textbf{Needle} \\
\midrule
\multirow{20}{*}{\rotatebox[origin=c]{90}{Meta-Llama-3-8B-Instruct}} & Baseline & 1.00$\times$ & 36.8 & 34.8 & 26.8 & 69.1 & 67.0 & 54.2 & 45.2 & 100.0 \\
        & \cellcolor{LightCyan}KIVI-2bit & \cellcolor{LightCyan}5.05$\times$ & \cellcolor{LightCyan}36.2 & \cellcolor{LightCyan}34.8 & \cellcolor{LightCyan}26.4 & \cellcolor{LightCyan}69.2 & \cellcolor{LightCyan}67.5 & \cellcolor{LightCyan}48.8 & \cellcolor{LightCyan}44.3 & \cellcolor{LightCyan}100.0 \\
        & KIVI-4bit & 3.11$\times$ & 36.8 & 35.0 & 26.9 & 69.3 & 66.5 & 54.7 & 45.3 & 100.0 \\
        & \cellcolor{LightCyan}FlexGen-4bit & \cellcolor{LightCyan}3.20$\times$ & \cellcolor{LightCyan}36.5 & \cellcolor{LightCyan}32.4 & \cellcolor{LightCyan}26.4 & \cellcolor{LightCyan}68.6 & \cellcolor{LightCyan}65.5 & \cellcolor{LightCyan}55.2 & \cellcolor{LightCyan}44.5 & \cellcolor{LightCyan}100.0 \\
        & InfLLM-2x & 2.00$\times$ & 31.8 & 30.8 & 25.7 & 67.6 & 57.5 & 55.8 & 42.5 & 22.7 \\
        & \cellcolor{LightCyan}InfLLM-4x & \cellcolor{LightCyan}4.00$\times$ & \cellcolor{LightCyan}27.1 & \cellcolor{LightCyan}24.7 & \cellcolor{LightCyan}25.0 & \cellcolor{LightCyan}63.9 & \cellcolor{LightCyan}37.5 & \cellcolor{LightCyan}57.6 & \cellcolor{LightCyan}38.3 & \cellcolor{LightCyan}20.7 \\
        & InfLLM-6x & 6.00$\times$ & 24.4 & 23.4 & 24.3 & 61.1 & 29.5 & 59.2 & 36.5 & 24.7 \\
        & \cellcolor{LightCyan}InfLLM-8x & \cellcolor{LightCyan}8.00$\times$ & \cellcolor{LightCyan}21.0 & \cellcolor{LightCyan}21.0 & \cellcolor{LightCyan}23.7 & \cellcolor{LightCyan}60.3 & \cellcolor{LightCyan}18.0 & \cellcolor{LightCyan}59.9 & \cellcolor{LightCyan}34.4 & \cellcolor{LightCyan}22.0 \\
        & StreamLLM-2x & 2.00$\times$ & 26.1 & 28.8 & 24.6 & 66.5 & 34.0 & 55.6 & 38.9 & 29.0 \\
        & \cellcolor{LightCyan}StreamLLM-4x & \cellcolor{LightCyan}4.00$\times$ & \cellcolor{LightCyan}20.5 & \cellcolor{LightCyan}22.2 & \cellcolor{LightCyan}22.7 & \cellcolor{LightCyan}62.2 & \cellcolor{LightCyan}21.0 & \cellcolor{LightCyan}56.1 & \cellcolor{LightCyan}34.4 & \cellcolor{LightCyan}22.3 \\
        & StreamLLM-6x & 6.00$\times$ & 17.4 & 18.7 & 21.4 & 60.1 & 14.5 & 59.0 & 32.3 & 18.0 \\
        & \cellcolor{LightCyan}StreamLLM-8x & \cellcolor{LightCyan}8.00$\times$ & \cellcolor{LightCyan}15.7 & \cellcolor{LightCyan}18.0 & \cellcolor{LightCyan}20.5 & \cellcolor{LightCyan}55.9 & \cellcolor{LightCyan}8.0 & \cellcolor{LightCyan}58.1 & \cellcolor{LightCyan}30.3 & \cellcolor{LightCyan}18.0 \\

        & $\mathrm{H_2O}$-2x & 2.00$\times$ & 35.8 & 34.8 & 25.4 & 69.1 & 66.0 & 54.4 & 44.7 & 100.0 \\
        & \cellcolor{LightCyan}$\mathrm{H_2O}$-4x & \cellcolor{LightCyan}4.00$\times$ & \cellcolor{LightCyan}35.0 & \cellcolor{LightCyan}35.1 & \cellcolor{LightCyan}23.6 & \cellcolor{LightCyan}69.0 & \cellcolor{LightCyan}66.0 & \cellcolor{LightCyan}53.2 & \cellcolor{LightCyan}44.0 & \cellcolor{LightCyan}100.0 \\
        & $\mathrm{H_2O}$-6x & 6.00$\times$ & 33.9 & 35.1 & 22.7 & 69.1 & 66.0 & 53.1 & 43.6 & 100.0 \\
        & \cellcolor{LightCyan}$\mathrm{H_2O}$-8x & \cellcolor{LightCyan}8.00$\times$ & \cellcolor{LightCyan}33.7 & \cellcolor{LightCyan}35.0 & \cellcolor{LightCyan}22.2 & \cellcolor{LightCyan}69.1 & \cellcolor{LightCyan}65.5 & \cellcolor{LightCyan}52.7 & \cellcolor{LightCyan}43.4 & \cellcolor{LightCyan}100.0 \\
        
        & LLMLingua2-2x & 2.00$\times$ & 29.4 & 31.5 & 24.1 & 38.6 & 68.0 & 31.9 & 33.5 & 51.3 \\
        & \cellcolor{LightCyan}LLMLingua2-4x & \cellcolor{LightCyan}4.00$\times$ & \cellcolor{LightCyan}26.5 & \cellcolor{LightCyan}30.8 & \cellcolor{LightCyan}24.1 & \cellcolor{LightCyan}39.3 & \cellcolor{LightCyan}22.5 & \cellcolor{LightCyan}32.2 & \cellcolor{LightCyan}29.9 & \cellcolor{LightCyan}8.3 \\
        & LLMLingua2-6x & 6.00$\times$ & 25.8 & 26.4 & 23.4 & 37.9 & 18.0 & 31.3 & 28.1 & 0.7 \\
        &\cellcolor{LightCyan}LLMLingua2-8x & \cellcolor{LightCyan}8.00$\times$ & \cellcolor{LightCyan}24.0 & \cellcolor{LightCyan}25.4 & \cellcolor{LightCyan}22.9 & \cellcolor{LightCyan}36.9 & \cellcolor{LightCyan}13.0 & \cellcolor{LightCyan}31.9 & \cellcolor{LightCyan}26.9 & \cellcolor{LightCyan}0.0 \\

\midrule
\multirow{20}{*}{\rotatebox[origin=c]{90}{Mistral-7B-Instruct-v0.2}} & Baseline & 1.00$\times$ & 32.5 & 25.8 & 27.9 & 66.7 & 89.3 & 54.0 & 43.8 & 99.0 \\
        & \cellcolor{LightCyan}KIVI-2bit & \cellcolor{LightCyan}5.05$\times$ & \cellcolor{LightCyan}31.3 & \cellcolor{LightCyan}24.7 & \cellcolor{LightCyan}27.6 & \cellcolor{LightCyan}66.8 & \cellcolor{LightCyan}80.8 & \cellcolor{LightCyan}53.7 & \cellcolor{LightCyan}42.6 & \cellcolor{LightCyan}99.0 \\
        & KIVI-4bit & 3.11$\times$ & 32.3 & 25.8 & 27.9 & 66.9 & 89.4 & 54.0 & 43.8 & 99.0 \\
        & \cellcolor{LightCyan}FlexGen-4bit & \cellcolor{LightCyan}3.20$\times$ & \cellcolor{LightCyan}33.0 & \cellcolor{LightCyan}24.4 & \cellcolor{LightCyan}27.8 & \cellcolor{LightCyan}66.2 & \cellcolor{LightCyan}83.0 & \cellcolor{LightCyan}53.7 & \cellcolor{LightCyan}43.0 & \cellcolor{LightCyan}98.3 \\
        & InfLLM-2x & 2.00$\times$ & 30.7 & 24.8 & 26.8 & 65.1 & 65.8 & 54.2 & 41.1 & 64.3 \\
        & \cellcolor{LightCyan}InfLLM-4x & \cellcolor{LightCyan}4.00$\times$ & \cellcolor{LightCyan}25.4 & \cellcolor{LightCyan}23.7 & \cellcolor{LightCyan}25.5 & \cellcolor{LightCyan}63.4 & \cellcolor{LightCyan}41.4 & \cellcolor{LightCyan}54.0 & \cellcolor{LightCyan}37.5 & \cellcolor{LightCyan}29.7 \\
        & InfLLM-6x & 6.00$\times$ & 23.8 & 21.0 & 25.0 & 61.6 & 32.6 & 53.4 & 35.6 & 32.3 \\
        & \cellcolor{LightCyan}InfLLM-8x & \cellcolor{LightCyan}8.00$\times$ & \cellcolor{LightCyan}22.2 & \cellcolor{LightCyan}19.6 & \cellcolor{LightCyan}24.3 & \cellcolor{LightCyan}62.0 & \cellcolor{LightCyan}26.2 & \cellcolor{LightCyan}53.8 & \cellcolor{LightCyan}34.5 & \cellcolor{LightCyan}27.0 \\
        & StreamLLM-2x & 2.00$\times$ & 24.6 & 22.0 & 25.3 & 64.5 & 47.1 & 53.0 & 37.5 & 54.7 \\
        & \cellcolor{LightCyan}StreamLLM-4x & \cellcolor{LightCyan}4.00$\times$ & \cellcolor{LightCyan}20.1 & \cellcolor{LightCyan}19.9 & \cellcolor{LightCyan}23.3 & \cellcolor{LightCyan}61.3 & \cellcolor{LightCyan}31.6 & \cellcolor{LightCyan}53.9 & \cellcolor{LightCyan}34.2 & \cellcolor{LightCyan}32.0 \\
        & StreamLLM-6x & 6.00$\times$ & 18.2 & 16.0 & 22.1 & 59.6 & 25.3 & 54.9 & 32.2 & 25.0 \\
        & \cellcolor{LightCyan}StreamLLM-8x & \cellcolor{LightCyan}8.00$\times$ & \cellcolor{LightCyan}17.0 & \cellcolor{LightCyan}15.2 & \cellcolor{LightCyan}21.4 & \cellcolor{LightCyan}58.3 & \cellcolor{LightCyan}16.9 & \cellcolor{LightCyan}54.9 & \cellcolor{LightCyan}30.8 & \cellcolor{LightCyan}19.3 \\
        & $\mathrm{H_2O}$-2x & 2.00$\times$ & 31.9 & 25.4 & 26.8 & 66.8 & 87.7 & 53.8 & 43.2 & 97.3 \\
        & \cellcolor{LightCyan}$\mathrm{H_2O}$-4x & \cellcolor{LightCyan}4.00$\times$ & \cellcolor{LightCyan}30.4 & \cellcolor{LightCyan}23.9 & \cellcolor{LightCyan}25.1 & \cellcolor{LightCyan}67.2 & \cellcolor{LightCyan}82.9 & \cellcolor{LightCyan}53.1 & \cellcolor{LightCyan}41.9 & \cellcolor{LightCyan}93.3 \\
        & $\mathrm{H_2O}$-6x & 6.00$\times$ & 29.0 & 22.8 & 24.3 & 66.9 & 82.0 & 52.5 & 41.1 & 85.7 \\
        & \cellcolor{LightCyan}$\mathrm{H_2O}$-8x & \cellcolor{LightCyan}8.00$\times$ & \cellcolor{LightCyan}27.8 & \cellcolor{LightCyan}21.8 & \cellcolor{LightCyan}23.9 & \cellcolor{LightCyan}67.0 & \cellcolor{LightCyan}79.5 & \cellcolor{LightCyan}52.3 & \cellcolor{LightCyan}40.4 & \cellcolor{LightCyan}80.0 \\
        & LLMLingua2-2x & 2.00$\times$ & 28.6 & 23.0 & 26.4 & 45.6 & 54.9 & 31.7 & 32.6 & 41.7 \\
        & \cellcolor{LightCyan}LLMLingua2-4x & \cellcolor{LightCyan}4.00$\times$ & \cellcolor{LightCyan}25.0 & \cellcolor{LightCyan}21.3 & \cellcolor{LightCyan}24.6 & \cellcolor{LightCyan}39.2 & \cellcolor{LightCyan}14.0 & \cellcolor{LightCyan}33.1 & \cellcolor{LightCyan}27.4 & \cellcolor{LightCyan}9.7 \\
        & LLMLingua2-6x & 6.00$\times$ & 21.2 & 17.4 & 23.3 & 38.9 & 8.9 & 34.7 & 25.4 & 0.0 \\
        & \cellcolor{LightCyan}LLMLingua2-8x & \cellcolor{LightCyan}8.00$\times$ & \cellcolor{LightCyan}19.6 & \cellcolor{LightCyan}16.0 & \cellcolor{LightCyan}22.9 & \cellcolor{LightCyan}38.5 & \cellcolor{LightCyan}8.0 & \cellcolor{LightCyan}35.5 & \cellcolor{LightCyan}24.7 & \cellcolor{LightCyan}0.0 \\

\midrule
\multirow{3}{*}{Mamba} 
        & Mamba-2.8B &-& 7.3 & 6.3 & 19.1 & 39.0 & 1.2 & 47.6 & 20.8 & 10.7 \\
        & \cellcolor{LightCyan}Mamba-Chat-2.8B & \cellcolor{LightCyan}- & \cellcolor{LightCyan}9.2 & \cellcolor{LightCyan}6.9 & \cellcolor{LightCyan}21.2 & \cellcolor{LightCyan}37.5 & \cellcolor{LightCyan}3.7 & \cellcolor{LightCyan}47.7 & \cellcolor{LightCyan}21.6 &\cellcolor{LightCyan}10.7 \\
        & Mamba2-2.7B &-& 7.5 & 6.7 & 21.0 & 40.5 & 4.1 & 49.9 & 22.1 & 9.0 \\
\midrule
\multirow{1}{*}{RWKV} 
        & \cellcolor{LightCyan}RWKV-5-World-7B &\cellcolor{LightCyan}-& \cellcolor{LightCyan}9.8 & \cellcolor{LightCyan}5.4 & \cellcolor{LightCyan}18.5 & \cellcolor{LightCyan}52.4 & \cellcolor{LightCyan}4.5 & \cellcolor{LightCyan}34.0 & \cellcolor{LightCyan}22.1 & \cellcolor{LightCyan}3.7 \\
\midrule
\multirow{2}{*}{R-Gemma} 
        & R-Gemma-2B-it & - & 18.1 & 8.3 & 20.9 & 46.3 & 4.0 & 53.7 & 26.1 & 23.3 \\
        & \cellcolor{LightCyan}R-Gemma-9B-it &\cellcolor{LightCyan}-& \cellcolor{LightCyan}24.5 & \cellcolor{LightCyan}21.9 & \cellcolor{LightCyan}21.9 & \cellcolor{LightCyan}54.5 & \cellcolor{LightCyan}9.0 & \cellcolor{LightCyan}60.8 & \cellcolor{LightCyan}33.2 & \cellcolor{LightCyan}26.7 \\
\bottomrule
\end{tabular}
}
\label{tab:main_table}
\end{table*}

\begin{figure*}[h]
\vspace{-1em}
\setlength{\abovecaptionskip}{0mm}
\setlength{\belowcaptionskip}{0mm}
\centering
\subfigcapskip=-4mm
\!\!\!\!\!\!
\subfigure[]{
\centering
	\begin{minipage}[t]{0.24\linewidth}
		\includegraphics[width=0.99\linewidth]{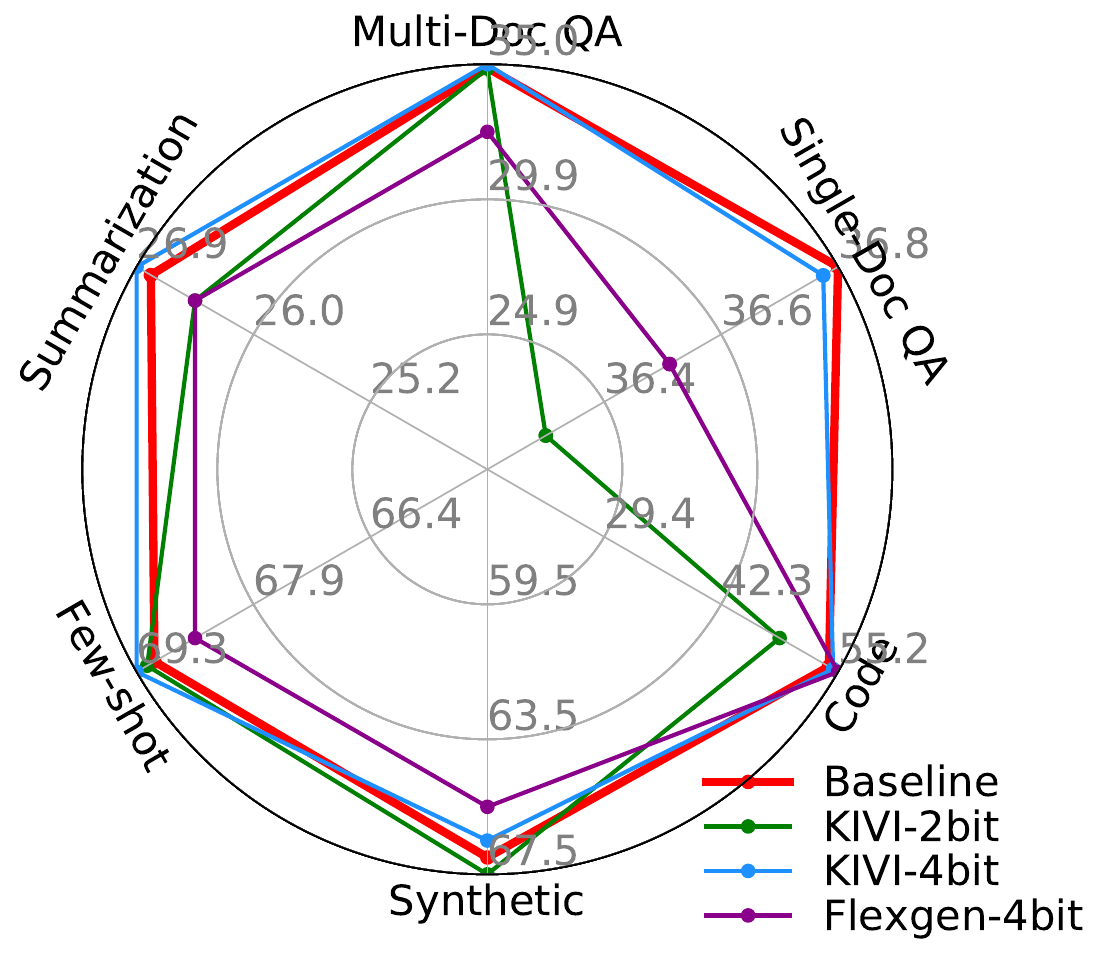}
	\end{minipage}%
}
\subfigure[]{
\centering
	\begin{minipage}[t]{0.248\linewidth}
		\includegraphics[width=0.99\linewidth]{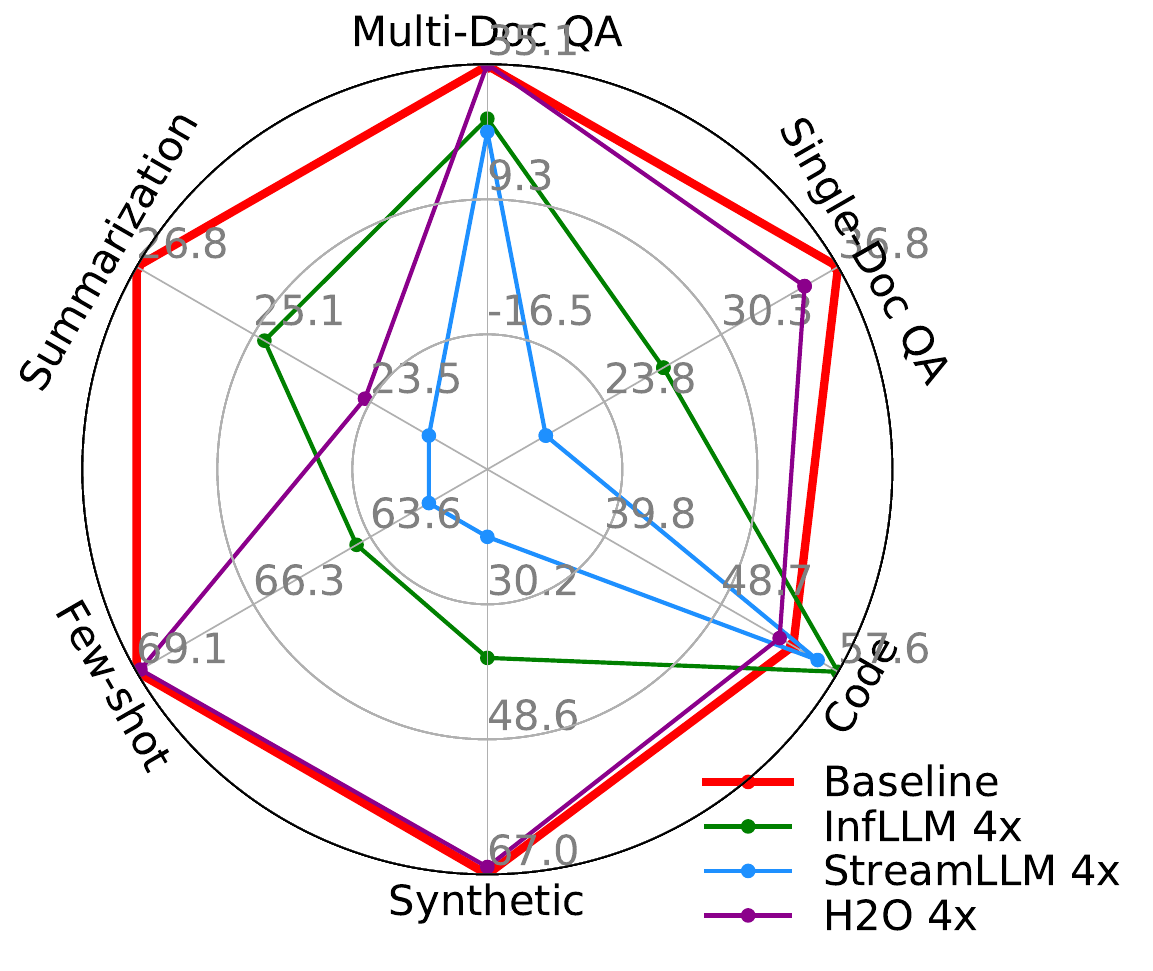}
	\end{minipage}%
}
\!\!\!\!
\subfigure[]{
\centering
	\begin{minipage}[t]{0.254\linewidth}
		\includegraphics[width=0.99\linewidth]{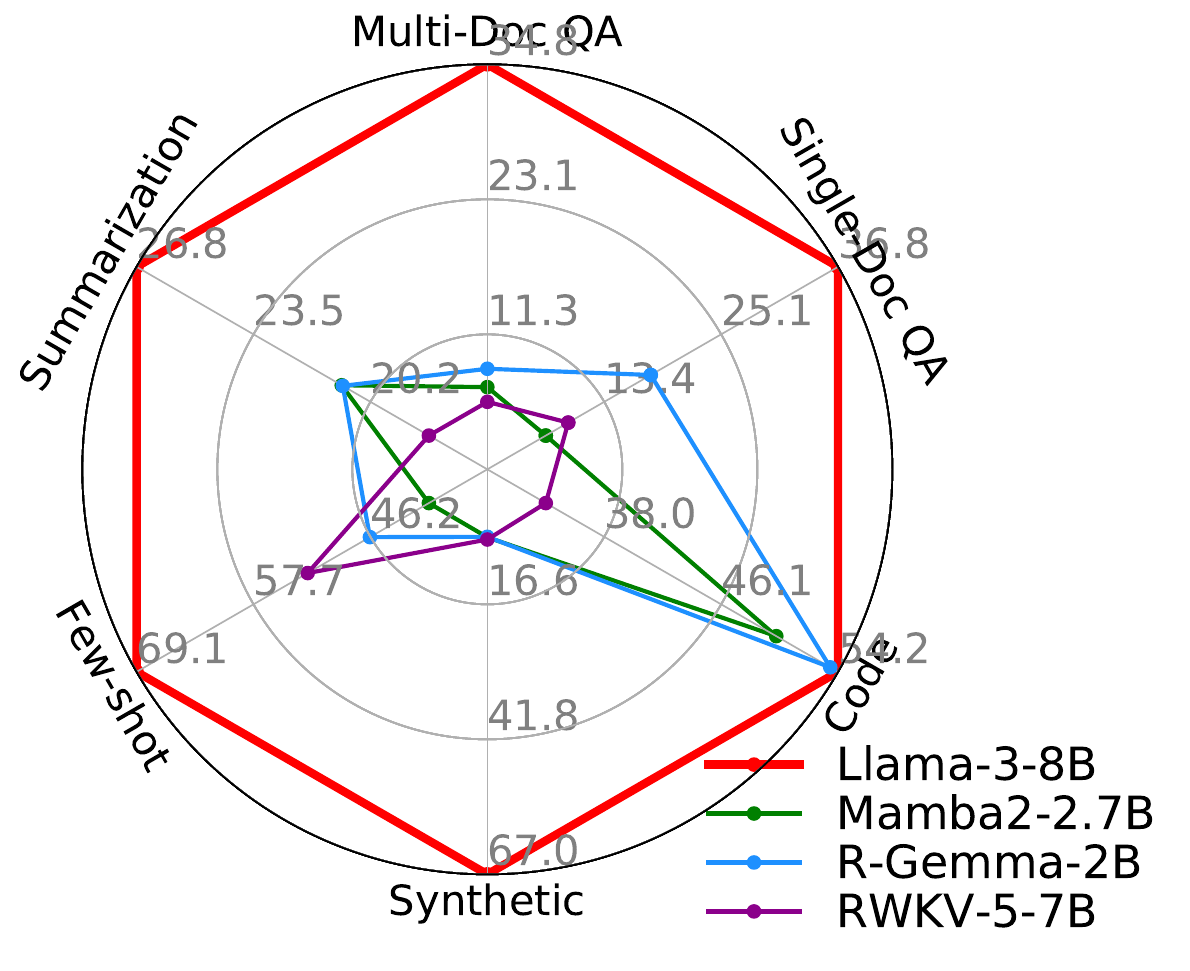}
	\end{minipage}%
}
\!\!\!\!\!\!
\subfigure[]{
\centering
	\begin{minipage}[t]{0.244\linewidth}
		\includegraphics[width=0.99\linewidth]{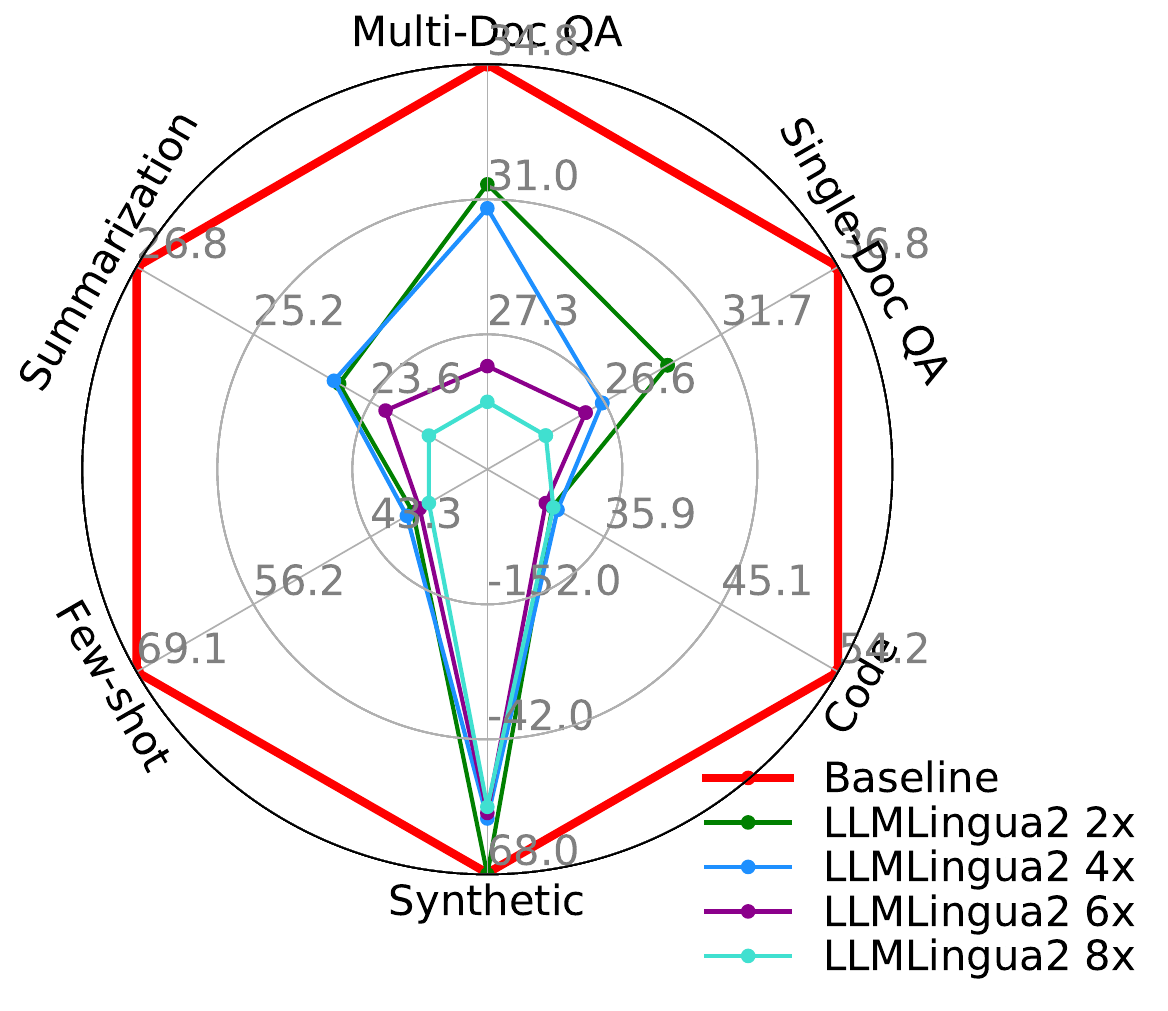}
	\end{minipage}%
}
\caption{The rador plot of different methods (a) Llama-3-8B Llama-3-8B w./ Quant. (b) Llama-3-8B w./ Token Dropping (c) Linear-time sequence models and mixed Architecture (d) Llama-3-8B w./ Prompt compression.}
\label{fig:radar}
\end{figure*}

\begin{figure*}[h]
\vspace{-1em}
\setlength{\abovecaptionskip}{0mm}
\setlength{\belowcaptionskip}{0mm}
\centering
\subfigcapskip=-2mm
\!\!\!\!\!\!
\subfigure[Llama-3-8B-Instruct]{
\centering
	\begin{minipage}[t]{0.27\linewidth}
		\includegraphics[width=0.99\linewidth]{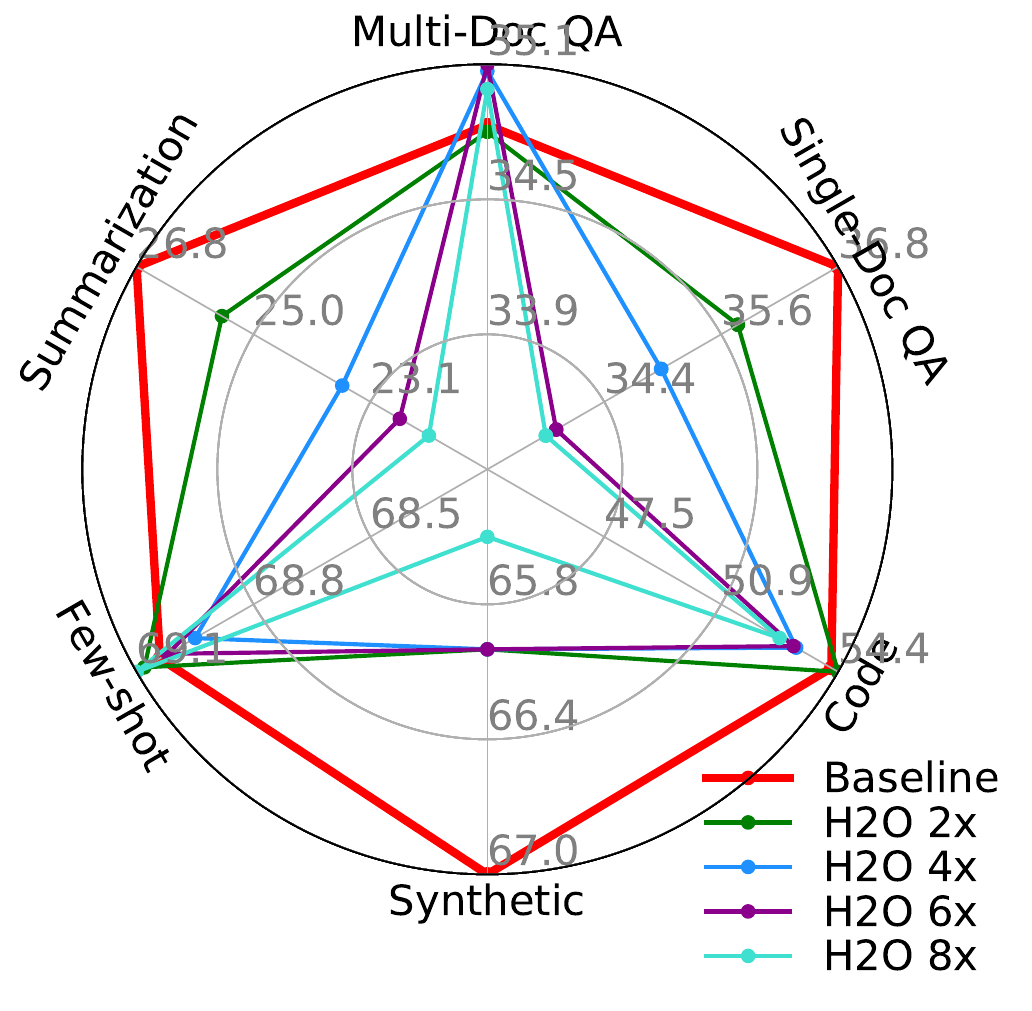}
	\end{minipage}%
}
\subfigure[Mistral-7B-v0.2-Instruct]{
\centering
	\begin{minipage}[t]{0.27\linewidth}
		\includegraphics[width=0.99\linewidth]{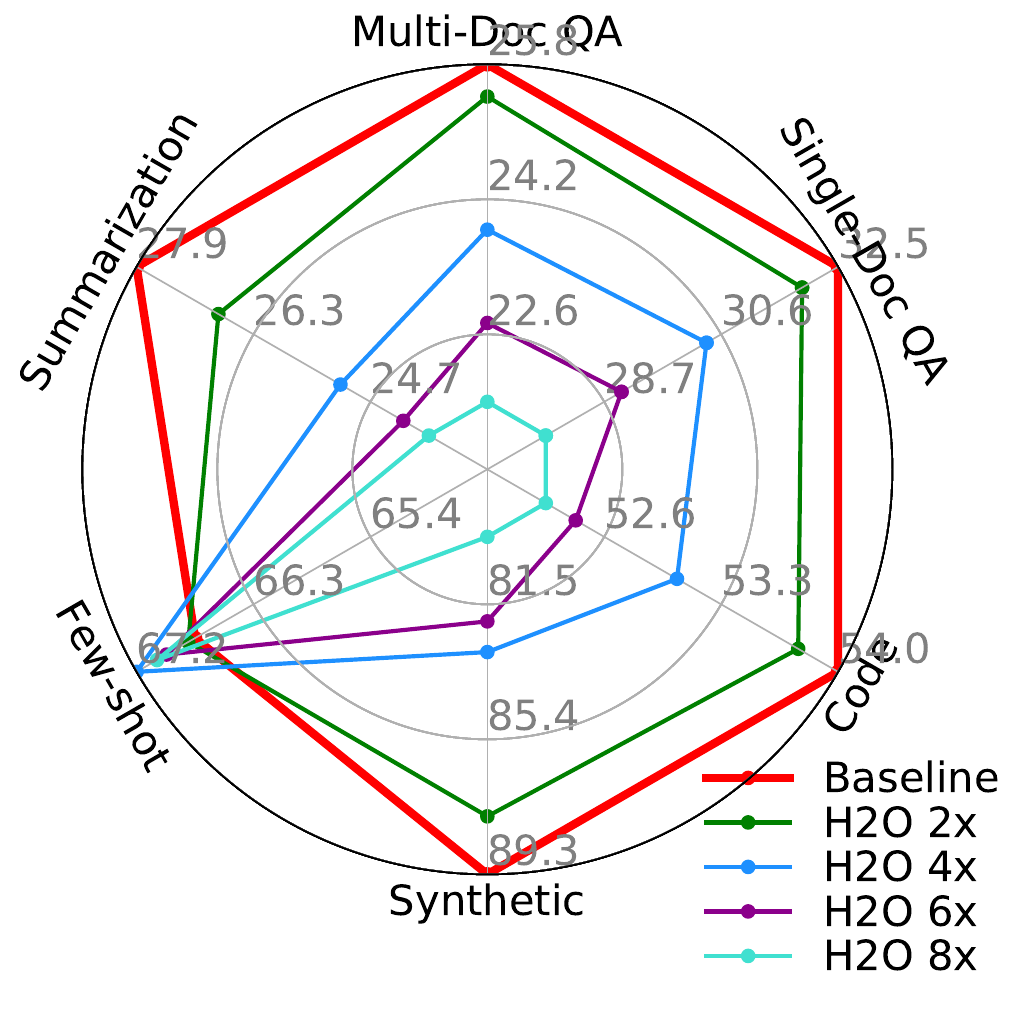}
	\end{minipage}%
}
\!\!\!\!
\subfigure[LongChat-7B-v1.5-32K]{
\centering
	\begin{minipage}[t]{0.27\linewidth}
		\includegraphics[width=0.99\linewidth]{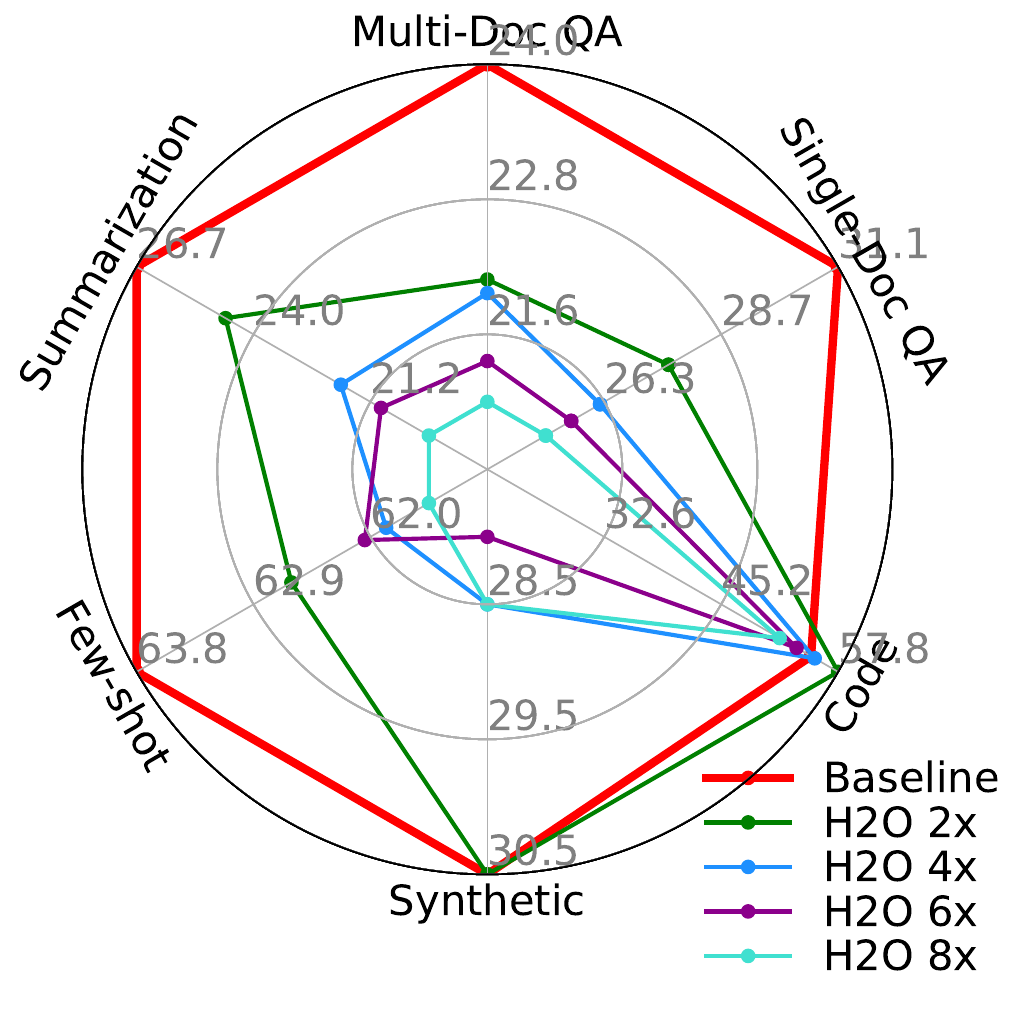}
	\end{minipage}%
}
\caption{ $\mathrm{H_2O}$ with different compression ratios on three commonly used LLMs.}
\vspace{-1em}
\label{fig:radar-h2o}
\end{figure*}

\begin{figure*}[h]
\setlength{\abovecaptionskip}{0mm}
\setlength{\belowcaptionskip}{0mm}
\centering
\subfigcapskip=-2mm

\begin{minipage}{0.95\linewidth}
    
\subfigure[\!Llama-3-8B-Instruct Baseline]{
\centering
        \includegraphics[width=0.3\linewidth]{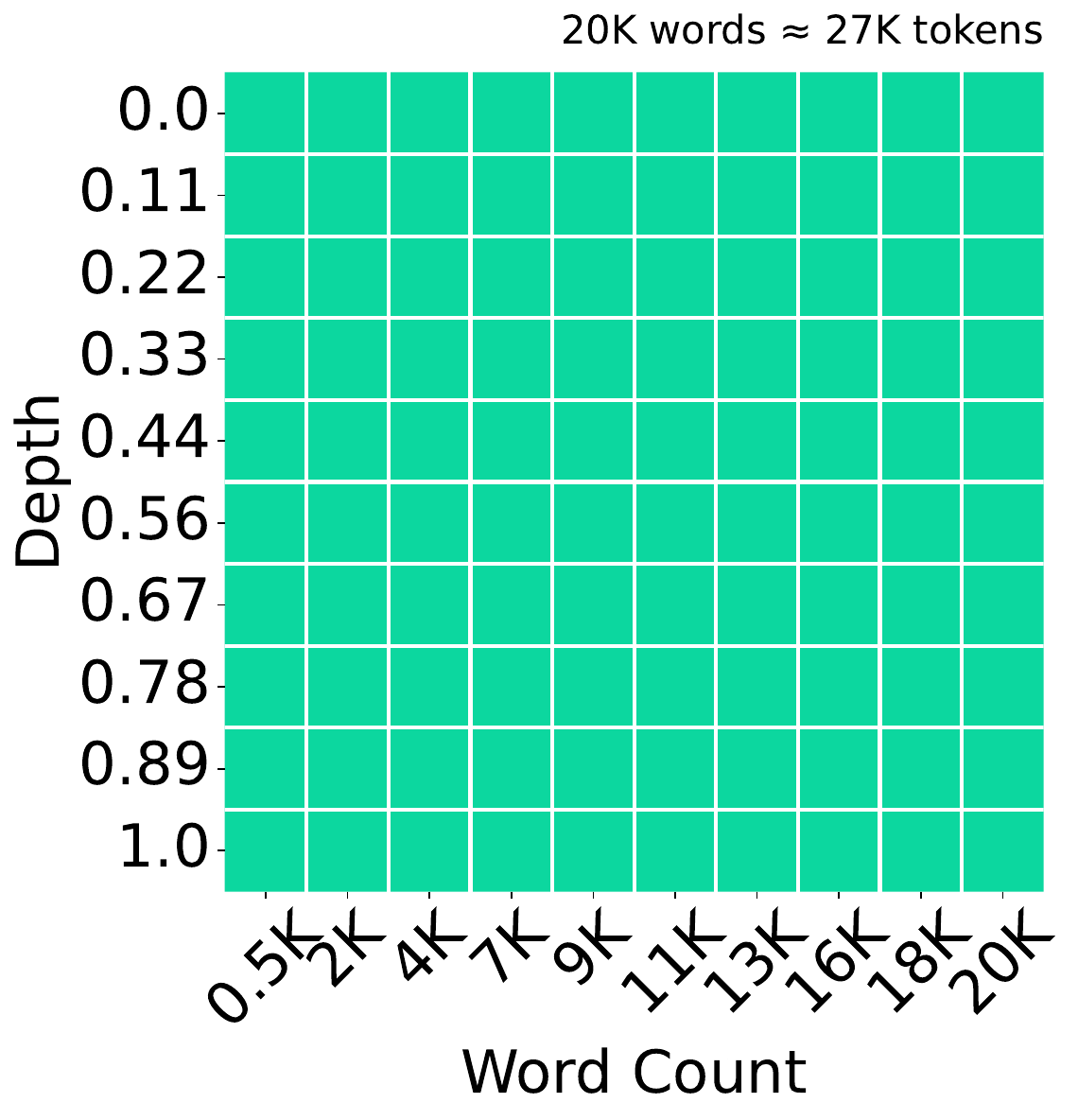}
}
\subfigure[\!Llama-3-8B-Instruct + KIVI-2bit]{
\centering
        \includegraphics[width=0.3\linewidth]{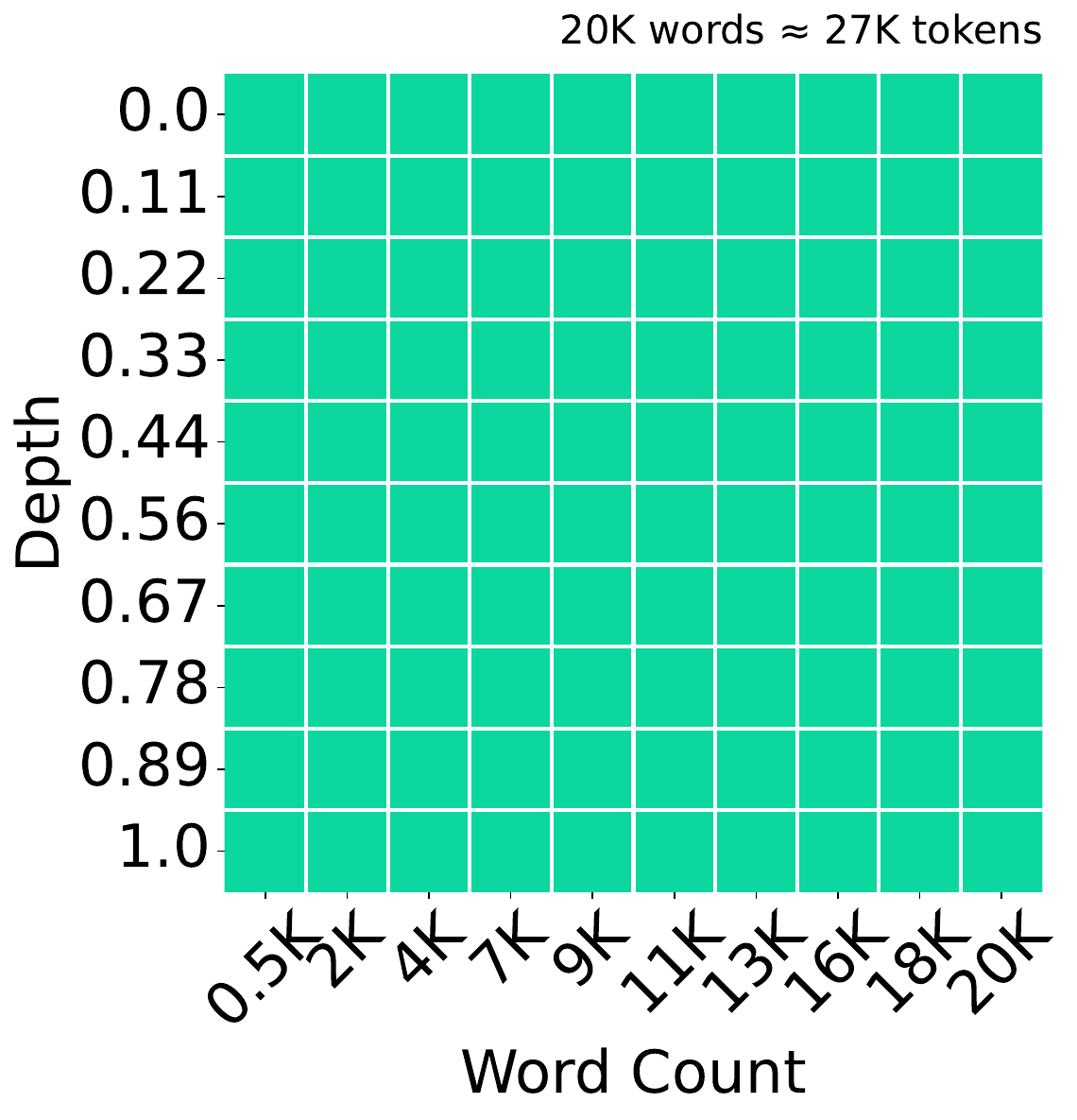}
}
\subfigure[\!Llama-3-8B-Instruct + InfLLM 4x]{
\centering
        \includegraphics[width=0.3\linewidth]{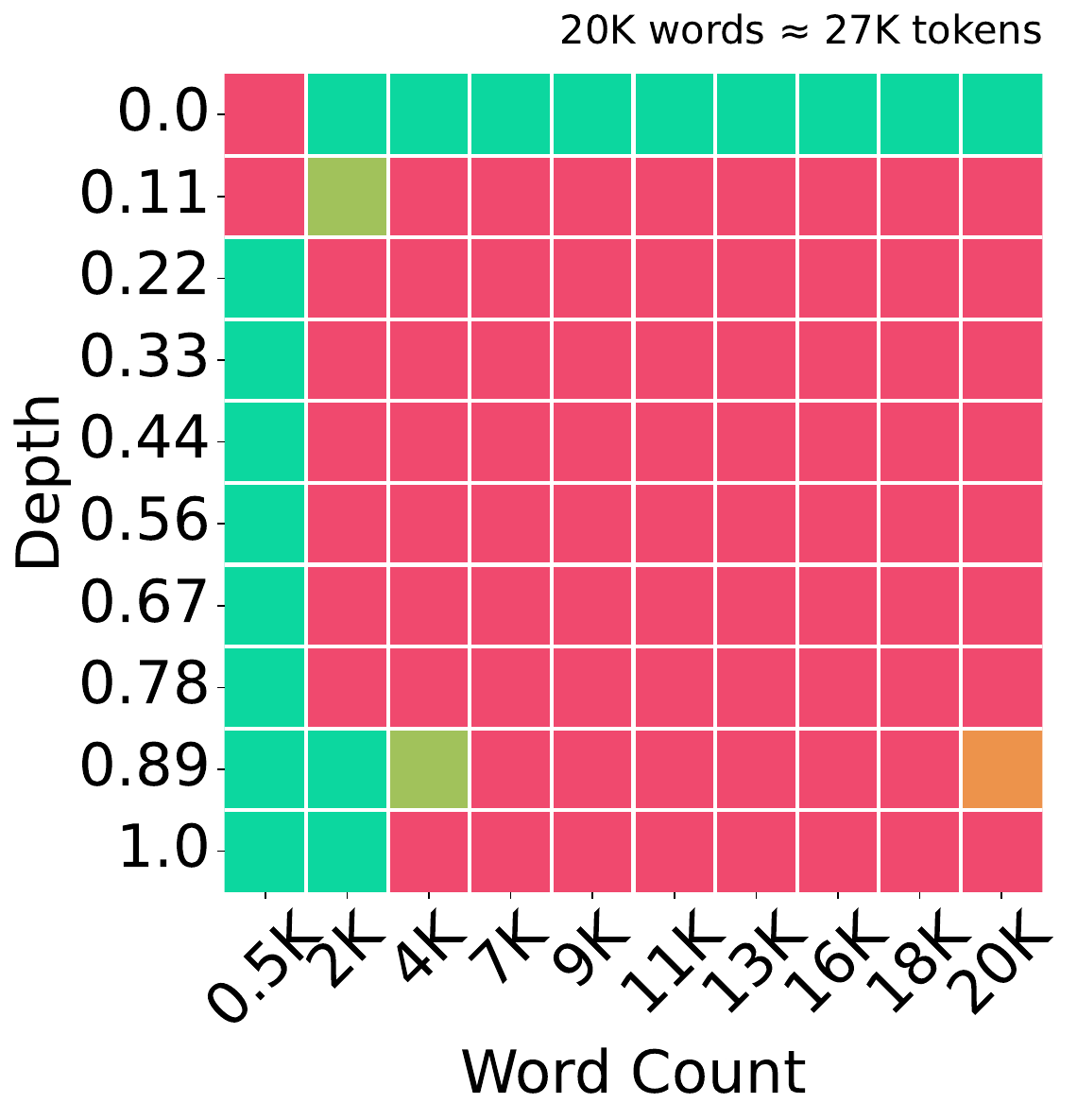}
}

\subfigure[\!LLMLingua2-4x]{
\centering
        \includegraphics[width=0.3\linewidth]{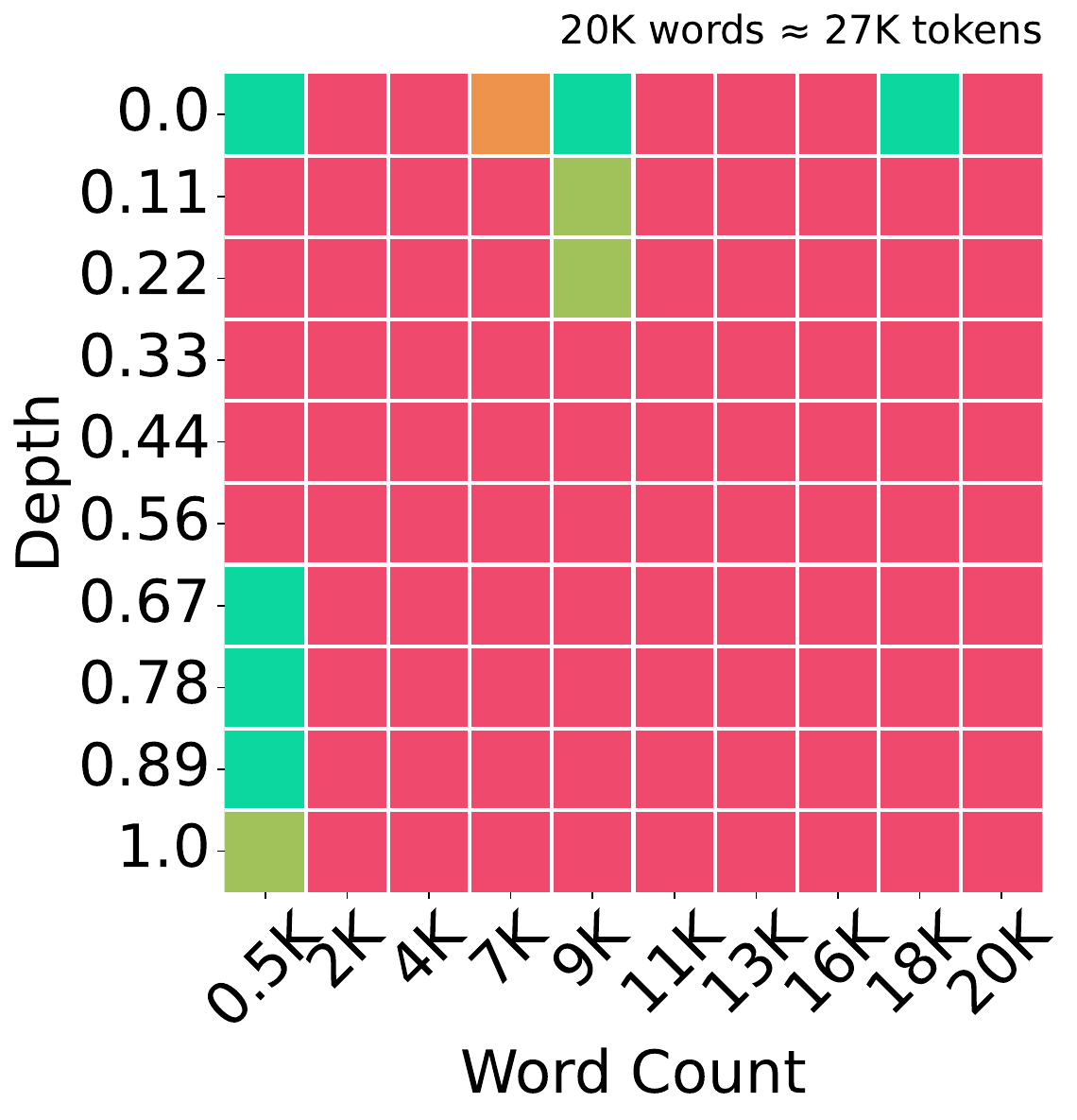}
}
\subfigure[\!Mamba-2.8B]{
\centering
        \includegraphics[width=0.3\linewidth]{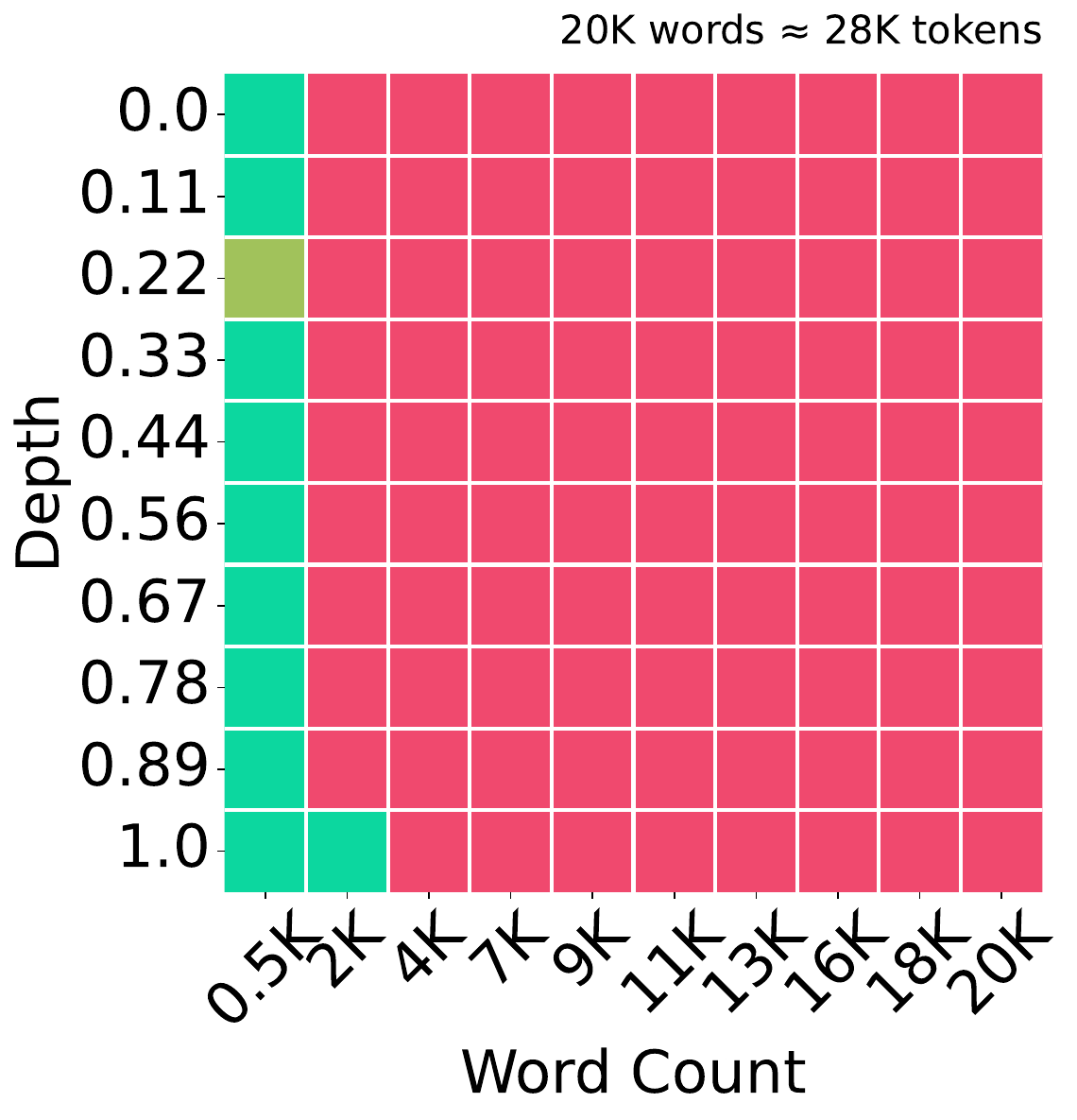}
}
\subfigure[\!RecurrentGemma-9B-it]{
\centering
        \includegraphics[width=0.3\linewidth]{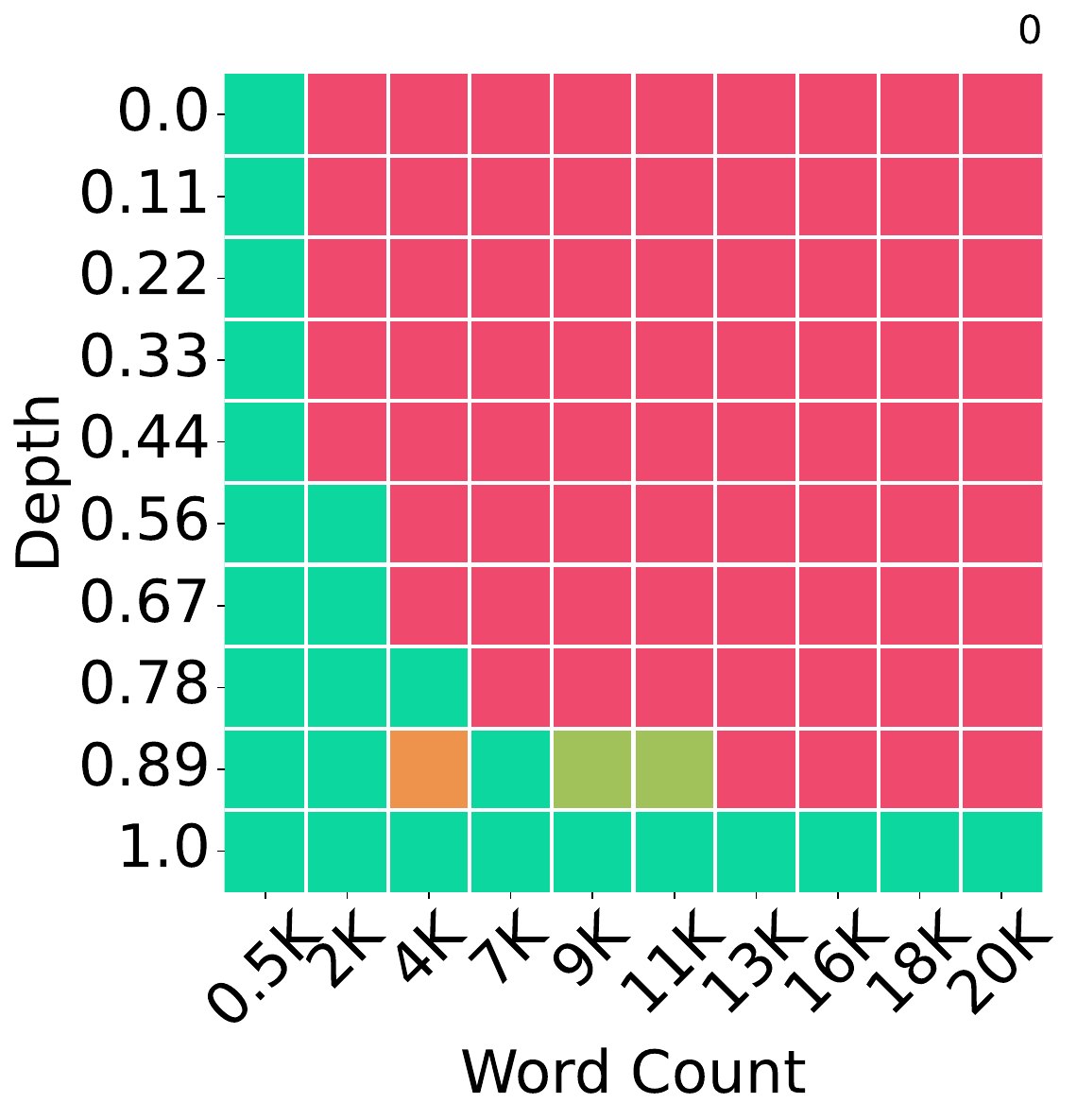}
}

\end{minipage}
\hspace{-1em}
\begin{minipage}{0.04\linewidth}
    \centering
    \includegraphics[width=1.25\linewidth]{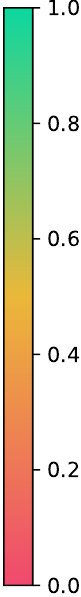}
\end{minipage}

\caption{Needle-in-a-Haystack results on Llama-3-8B-Instruct, linear-time sequence models, and mixed architectures. The best method in each school of approaches is featured with comparable compression ratios. The same length of input might convert to different numbers of tokens per different models, as noted in the upper right corners.}
\label{fig:needle_main}
\end{figure*}

\subsection{Coverage}
\label{sec_bench_coverage}

\paragraph{Tasks and Models.}

We focus on \textbf{16 different long context tasks under 7 major categories}, each requiring different long context handling abilities and covering key application scenarios. We provide a brief walkthrough of each task category as follows:
(1) \emph{Single-doc QA}, which tests the long context understanding ability with longer documents.
(2) \emph{Multi-Doc QA}, which needs to extract and combine information from several documents to obtain the answer;
(3) \emph{Summarization}, which requires a global understanding of the whole context;
(4) \emph{Few-shot Learning}, which is a practical setting requiring long-context understanding over provided examples;
(5) \emph{Synthetic Task}, which is designed to test the model’s ability on specific scenarios and patterns;
(6) \emph{Code Completion}, which is designed to test the model’s long-context ability in code auto-completion tasks; 
(7) \emph{Needle-in-a-Haystack Test}, which involves finding specific information within a large volume of text.

For categories (1)-(6), we directly adopt them from the LongBench dataset~\citep{bai2023longbench}. For the (7) Needle-in-a-Haystack Test, we largely follow the format of the original passkey retrieval task \citep{mohtashami2023passkey} while including some modern modifications set forward by \href{https://github.com/gkamradt/LLMTest_NeedleInAHaystack}{Arize-ai} and the technical report of Gemini 1.5 \citep{reid2024gemini}. We refer our readers to Appendix \ref{app: detailed datasets} for further details.

For models, we elect to cover \textbf{3 representative transformer-based LLMs and 3 pure or hybrid linear-time sequence model families}. For transformer-based LLMs, we opt for Mistral-7b-Instruct-v0.2~\citep{jiang2023mistral}, Longchat-7B-v1.5-32k~\citep{li2023longchat} and Llama-3-8B-Instruct~\citep{llama3modelcard} to provide a coverage of SOTA long-context capable model as well as the most recent progress of open-source LLMs. For linear-time sequence models and their hybrids, we evaluated Mamba-2.8B \citep{gu2023mamba}, Mamba2-2.7b \citep{dao2024mamba2}, Mamba-Chat-2.8B \citep{haven2023mambachat}, RWKV-5-World \citep{peng2023rwkv}, and RecurrentGemma-2b/9b-Instruct~\citep{botev2024recurrentgemma}. We refer readers to Appendix \ref{app: detailed setup} for more model-related details.
\textbf{}
\vspace{-0.5em}
\paragraph{Methods and Hyperparameter Settings.}
As shown in Table \ref{tab:overview}, we select representative methods  ranging from KV cache-free to linear complexity KV cache. Apart from the linear-time sequence models and their hybrids introduced above, we opt for the following compression methods:
For \emph{quantization}, we adopt KIVI \citep{kivi}, INT4 per-token quantization in FlexGen \citep{sheng2023flexgen};
For \emph{Token dropping}, we adopt StreamingLLM \cite{streamllm}, $\mathrm{H_2O}$ \citep{h2o}, and InfLLM \cite{xiao2024infllm}.
For \emph{Prompt Compression}, we adopt LLMLingua2 \cite{jiang2023llmlingua}. We note that \textbf{although token dropping-based methods are usually designed with a constant KV cache size in mind, we modify them to adapt linear compression schemes for fair comparison with other methods.} We share more method-specific details in Appendix \ref{app_method_specific}.

\subsection{Experiment Setup and Report Digestion}
\label{sec_bench_setup}

Given the vastly different design principles employed in different schools of long context handling methods, it is, in fact, impossible to achieve a global alignment where all covered methods are considered fairly aligned against each other. For example, while KV cache quantization methods like FlexGen~\citep{sheng2023flexgen} can adapt to different data precision, they can never be aligned with any KV cache-free approaches like Mamba~\citep{gu2023mamba}. Similarly, token dropping approaches typically employ a constant size of kept tokens and evict everything else, making their compression gain dynamic against inputs of different lengths; and, again, not alignable with KV cache quantization methods nor KV cache-free approaches. Note that the abovementioned issues are merely some alignment hardships due to conflicts in different long contexts when handling schools. In reality, two long context-specific methods — even under the same school — can also bring further complications: e.g., KIVI~\citep{kivi} includes a full precision sliding window for the most recent tokens, while FlexGen~\citep{sheng2023flexgen} doesn't. Further, known that models like Mamba~\citep{gu2023mamba} and RWKV~\citep{peng2023rwkv} are typically pre-trained on open-source datasets, their architecture potentials cannot be fairly evaluated compared to models like Llama-3 — which are pretrained upon proprietary data corpus and done so with an overtrained recipe that has proven to be beneficiary. More on this in Appendix~\ref{app_ex_dis}.

As the best alternative, we opt to compress different methods towards a range of available target compression ratios shown in Table~\ref{tab:main_table}. For KV cache quantization methods, we derive such compression ratios by referring to the reduction in KV cache memory size against full precision KV cache. For token dropping approaches, we forgo their typical constant kept token setup and dynamically adjust the amount of evicted tokens upon the length of each input request. For hard prompt compression, we simply compress the final hard prompt to or below the target compression ratio. We keep KV cache-free methods in their vanilla forms as they often have a constant memory complexity. More in Appendix \ref{app_method_specific}.

With such efforts, our experiment report should be reasonably comparable among similar compression ratios. Though we emphasize that our additional alignment effort will not resolve the pretraining difference among different backbone models — where an aligned comparison here can only be done by training different models from scratch, which will induce drastic computation costs and can only provide coverage on fully transparent transformer-based LLMs like Pythia~\citep{biderman2023pythia}, OpenLLaMA~\citep{geng2023openllama}, or LLM360~\citep{liu2023llm360}, where weight-only open-source models like Llama~\citep{touvron2023llama2, llama3modelcard} and Mistral~\citep{jiang2023mistral} can not be included due to the lack of reproducible training procedure and resource.

\vspace{-0.7em}
\subsection{Results and Discussion}
\label{sec_bench_results}

We showcase our main results in a category-based fashion in Table~\ref{tab:main_table} and \textbf{refer our readers to Appendix \ref{app: more exp} for many more additional results}. Table~\ref{tab:main_table} highlights the per-task-category performance of different long context-capable methods on Meta-Llama-3-8B-Instruct \citep{llama3modelcard} and Mistral-7B-Instruct-v0.2 \citep{jiang2023mistral}, as well as several other covered linear and mixed models. Based on all of our obtained results, we made the following observations.

\paragraph{OB \ding{182} Keeping the prefill process uncompressed is crucial for performance maintenance.}

This is because the KV cache for all prompt tokens is generated during the prefill stage. If we apply any compression at this stage, it will make the representation of said prompt in later layers inaccurate due to lossy \texttt{forward()} activation, leading to worse results when generating the output tokens. For instance, KIVI \citep{kivi}, FlexGen \citep{sheng2023flexgen}, and $\mathrm{H_2O}$ \citep{h2o} do not employ any compression operation during the prefill stage, which often leads to much better results than methods which do compress within (or even before) the prefill stage, namely StreamingLLM \citep{streamllm}, InfLLM \citep{xiao2024infllm}, and LLMLingua2 \citep{jiang2023llmlingua}.

That being said, we note this observation is likely limited to ``long input'' type of tasks, as all evaluated tasks in our work are considered ``long input, short output'' (like passkey retrieval from \citet{mohtashami2023passkey}), but not ``long generation'' (like multi-round conversation~\cite{li2023camel, wu2023autogen}, fiction writing~\cite{yang2022re3}, or long code generation~\cite{roziere2023code}), where compressing the input during the prefill stage will naturally carry more influence than compression during the decoding stage. More on this in Section~\ref{sec:limitations}.

\vspace{-0.5em}
\paragraph{OB \ding{183} Quantization methods can often achieve reliable performance across all task categories, yet token dropping approaches excel on some specific types of tasks (e.g., coding).}

We find that KV cache quantization techniques like FlexGen \citep{sheng2023flexgen} and KIVI \citep{kivi} tend to perform decently across all evaluated tasks. This is an intuitive finding, given quantization techniques do not evict any token completely, avoiding the possibility of dropping task-influential tokens by accident (e.g., one can imagine forging tokens around the needle insertion in the needle-in-the-haystack tasks \citep{mohtashami2023passkey} will surely be damaging, especially if such eviction happens during the prefill stage). The trade-off of such globally acceptable performance of KV cache quantization methods is their memory footprints \emph{must} grow with the sequence length, unlike token dropping approaches or linear-time sequence models, where a constant memory footprint is possible.

On the other hand, several featured token dropping methods showcased excellent performance on some specific subtasks. For example, StreamingLLM \citep{streamllm} and $\mathrm{H_2O}$ \citep{h2o} tend to perform exceptionally well on code-related tasks, with Figure~\ref{fig:radar-h2o} and Figure~\ref{fig:radar-streamllm} demonstrating perfect performance retention across various compression ratios upon the majority of featured LLMs; whereas InfLLM \citep{xiao2024infllm} — another token dropping methods that basically does KV cache retrieval of middle tokens on top of StreamingLLM — tend to deliver a more steady performance across all tasks without drastic shortcoming, with an extra advantage of being stronger under the needle test than StreamingLLM. 

Conversely, hard prompt compression methods like LLMLingua2 \citep{jiang2023llmlingua} perform the worst on the needle test across all KV cache-required methods — which is, once again, a well-expected finding as if one deletes the needle information within the input, the LLM will certainly not be able to answer the retrieval-required question correctly. LLMLingua2 performs modestly behind all featured KV cache-required methods in terms of LongBench \citep{bai2023longbench} tasks, though with the advantage of being model agnostic and can be theoretically applicable to black-box models with limited access.

\vspace{-0.5em}
\paragraph{OB \ding{184} Mixing with attention can greatly improve the long context capability of linear-time sequence models.}

We observe that hybrid models like RecurrentGemma \citep{botev2024recurrentgemma} can result in good performance improvement over pure linear-time sequence models like Mamba \citep{gu2023mamba} or Mamba-Chat \citep{haven2023mambachat} in terms of all evaluated tasks (Table~\ref{tab:main_table}). This indicates the potential of hybrid architectures due to the promising performance gain with an often acceptable increase in memory footprint.

\vspace{-0.5em}
\paragraph{OB \ding{185} Needle-in-a-haystack test remains challenging for KV cache-free or prefill time compression methods.} As demonstrated in Figure~\ref{fig:needle_main}, which features the best methods from each school of approaches: KIVI by \citet{kivi} (quantization), InfLLM by \citet{xiao2024infllm} (token dropping), LLMLingua2 by \citet{jiang2023llmlingua} (prompt compression), Mamba-2.8B by \citet{gu2023mamba} (linear-time sequence models), and RecurrentGemma-9B-it by \citet{botev2024recurrentgemma} (mixed architectures), we observe that compression during prefill or KV cache-free methods often struggle to maintain good retrieval performance as the baseline methods. While we believe different architectural or method designs do play a role here, we emphasize that unaligned pretraining recipes among different models, as well as the disparity of model sizes, are also certainly some strong influencing factors. For example, while not featured in our work, LongMamba \citep{zhang2024longmamba} — a finetuned version of Mamba-2.8B \citep{gu2023mamba} with long context focuses - tends to have much better needle performance.

Additionally, we note that we purposely decide to feature InfLLM \citep{xiao2024infllm} instead of $\mathrm{H_2O}$ \citep{h2o} in Figure~\ref{fig:needle_main} as a representation of the token dropping school, despite $\mathrm{H_2O}$ having an objectively much better needle result in Table \ref{tab:main_table} (100\% vs 20.7\% for 4$\times$ compression). This decision is made because our needle test requires the model to correctly answer a 7-digit passkey, where the ending of the instruction prompt is \texttt{``What is the pass key? The pass key is ''} (Appendix \ref{app_dataset_needle}), leading the model-in-question likely to answer the first several digits of the passkey as the first generated token. This, combined with the fact that $\mathrm{H_2O}$ does not evict tokens during prefill time, often means an $\mathrm{H_2O}$-powered model can get the first several digits (usually at least three, due to the design of tokenizers) of the passkey right for free, as no compression has happened for decoding the first token, and most transformer-based baseline models — like the Llama-3-8B-Instruct featured in Figure~\ref{fig:needle_main} — are able to get the full 7-digit passkey right under no compression. We confirmed $\mathrm{H_2O}$'s perfect needle performance on Llama-3-8B-Instruct showed in Table \ref{tab:main_table} and Figure \ref{fig:needle_h2o_llama} is indeed more of a product of this prompt template and the 7-digit passkey task configuration instead of its innate excellence in retrieval capability; as should we expand the passkey length to 64-digit while keeping everything else the same, $\mathrm{H_2O}$'s performance drops drastically (from 100\% to 35.0\% for 4$\times$ compression), where methods like KIVI \citep{kivi} and InfLLM \citep{xiao2024infllm} tend not to experience such significant of a performance drop (100\% to 91.0\% for KIVI-2bit; 20.7\% to 19.0\% for InfLLM 4$\times$ compression), as shown in Figure \ref{fig:needle_kivi_llama_64}, \ref{fig:needle_h2o_llama_64}, and \ref{fig:needle_infllm_llama_64}.\quad \textbf{Due to page limitations, we analyze more observations in Appendix~\ref{app_ex_dis}.}
\section{Challenges and Opportunities}
\label{sec: challenge}

In this section, we share our insights regarding different long context challenges and highlight several opportunities derived from our benchmarking observations.

\vspace{-0.5em}
\paragraph{How to effectively reduce prefill time and footprint?}
Based on our empirical observations, most KV cache compression methods struggle to make the prefill stage efficient without compromising performance (\textbf{OB} \ding{182}), which calls for investments in more performant prefill-time compression methods. However, other than the performance requirement on accuracy-like metrics, prefill-time compression methods are entangled with non-trivial technical comparability challenges. Recall that FlashAttention (FA) \citep{dao2022flashattention} is inevitable during the prefill stage to improve hardware utility, with the key spirit of FA being to avoid the generation of a full attention matrix. Thus, methods that rely on the availability of a full attention matrix cannot be easily integrated. Therefore, we advocate future research on prefill-time compression methods with FA compatibility in mind.

\vspace{-0.5em}
\paragraph{How to build efficient yet long context-capable architectures?}
We empirically observe that pure linear-time sequence models that mix input tokens together struggle with information retrieval (\textbf{OB} \ding{184}), where some sort of attention mechanism provides visible improvements (\textbf{OB} \ding{183}). Therefore, an important future direction is to explore how to efficiently combine attention layers with linear-time sequence model layers and determine the optimal number of attention layers needed to achieve an ideal performance-efficiency balance.

\vspace{-0.5em}
\paragraph{How to cash-in real-world efficiency?}
Different methods often have varying levels of optimization while being comparable in theoretical efficiency, meaning whether a method is practically efficient in real-world application is highly related to factors like the \emph{Ease of Optimization} (e.g., quantization is well-studied and easy to optimize, while some unstructured methods will involve extra challenges~\citep{liu2023sparseland}) and \emph{Compatibility with Established Software or Hardware Frameworks} (e.g., compatibility with FlashAttention, as mentioned above). Based on these factors, it is challenging to provide a fair apple-to-apple comparison regarding efficiency. Researchers should keep this challenge in mind and develop efficient yet long context-capable methods.

\section{Conclusion}

Our benchmark fills a critical gap by evaluating 10+ methods across 65 settings, uncovering new insights on long context-capable approaches. We also provide a minimalistic and extensible package for reproducible research.

\section*{Acknowledgments}

\vspace{-0.3em}

This research was partially supported by NSF Awards ITE-2429680, IIS-2310260, OAC-2112606, and OAC-2117439. Furthermore, this work was supported by the US Department of Transportation (USDOT) Tier-1 University Transportation Center (UTC) Transportation Cybersecurity Center for Advanced Research and Education (CYBER-CARE) grant \#69A3552348332.

This work also made use of the High Performance Computing Resource in the Core Facility for Advanced Research Computing at Case Western Reserve University (CWRU). We give special thanks to the CWRU HPC team for their prompt and professional help and maintenance. The views and conclusions in this paper are those of the authors and do not represent the views of any funding or supporting agencies.

\section*{Limitations and Potential Risks}
\label{sec:limitations}

Despite our best efforts to cover a wide range of long context-capable approaches across many backbone models, our benchmark work will inevitably lack the inclusion of some eligible and interesting methods, certain worthwhile tasks, or particular setups that are reflective of our benchmarking goal due to limited manpower and computing resources. Specifically, we recognize that we only benchmark on models with $<$10B parameters\footnote{Though part of it is to align with linear-time sequence models, which are often $\leq$8B.} and our tasks are more focused on long input but not long generation, with the latter also being an important, though less mature aspect of long context evaluation due to the open-ended nature of prolonged generation tasks. 

In terms of potential risks, while we aim to provide a comprehensive view of feature methods and tasks, we caution our readers to directly adopt our empirical conclusion without proper evaluation under high-stake scenarios.

\bibliography{ref}

\begin{thebibliography}{81}
\providecommand{\natexlab}[1]{#1}

\bibitem[{AI@Meta(2024)}]{llama3modelcard}
AI@Meta. 2024.
\newblock \href {https://github.com/meta-llama/llama3/blob/main/MODEL_CARD.md} {Llama 3 model card}.

\bibitem[{Ainslie et~al.(2023)Ainslie, Lee-Thorp, de~Jong, Zemlyanskiy, Lebr{\'o}n, and Sanghai}]{ainslie2023gqa}
Joshua Ainslie, James Lee-Thorp, Michiel de~Jong, Yury Zemlyanskiy, Federico Lebr{\'o}n, and Sumit Sanghai. 2023.
\newblock Gqa: Training generalized multi-query transformer models from multi-head checkpoints.
\newblock \emph{arXiv preprint arXiv:2305.13245}.

\bibitem[{Bai et~al.(2023{\natexlab{a}})Bai, Bai, Chu, Cui, Dang, Deng, Fan, Ge, Han, Huang et~al.}]{bai2023qwen}
Jinze Bai, Shuai Bai, Yunfei Chu, Zeyu Cui, Kai Dang, Xiaodong Deng, Yang Fan, Wenbin Ge, Yu~Han, Fei Huang, et~al. 2023{\natexlab{a}}.
\newblock Qwen technical report.
\newblock \emph{arXiv preprint arXiv:2309.16609}.

\bibitem[{Bai et~al.(2023{\natexlab{b}})Bai, Lv, Zhang, Lyu, Tang, Huang, Du, Liu, Zeng, Hou et~al.}]{bai2023longbench}
Yushi Bai, Xin Lv, Jiajie Zhang, Hongchang Lyu, Jiankai Tang, Zhidian Huang, Zhengxiao Du, Xiao Liu, Aohan Zeng, Lei Hou, et~al. 2023{\natexlab{b}}.
\newblock Longbench: A bilingual, multitask benchmark for long context understanding.
\newblock \emph{arXiv preprint arXiv:2308.14508}.

\bibitem[{Biderman et~al.(2023)Biderman, Schoelkopf, Anthony, Bradley, O’Brien, Hallahan, Khan, Purohit, Prashanth, Raff et~al.}]{biderman2023pythia}
Stella Biderman, Hailey Schoelkopf, Quentin~Gregory Anthony, Herbie Bradley, Kyle O’Brien, Eric Hallahan, Mohammad~Aflah Khan, Shivanshu Purohit, USVSN~Sai Prashanth, Edward Raff, et~al. 2023.
\newblock Pythia: A suite for analyzing large language models across training and scaling.
\newblock In \emph{International Conference on Machine Learning}, pages 2397--2430. PMLR.

\bibitem[{Botev et~al.(2024)Botev, De, Smith, Fernando, Muraru, Haroun, Berrada, Pascanu, Sessa, Dadashi et~al.}]{botev2024recurrentgemma}
Aleksandar Botev, Soham De, Samuel~L Smith, Anushan Fernando, George-Cristian Muraru, Ruba Haroun, Leonard Berrada, Razvan Pascanu, Pier~Giuseppe Sessa, Robert Dadashi, et~al. 2024.
\newblock Recurrentgemma: Moving past transformers for efficient open language models.
\newblock \emph{arXiv preprint arXiv:2404.07839}.

\bibitem[{Brandon et~al.(2024)Brandon, Mishra, Nrusimha, Panda, and Kelly}]{brandon2024cla}
William Brandon, Mayank Mishra, Aniruddha Nrusimha, Rameswar Panda, and Jonathan~Ragan Kelly. 2024.
\newblock Reducing transformer key-value cache size with cross-layer attention.
\newblock \emph{arXiv preprint arXiv:2405.12981}.

\bibitem[{Brown et~al.(2020)Brown, Mann, Ryder, Subbiah, Kaplan, Dhariwal, Neelakantan, Shyam, Sastry, Askell et~al.}]{gpt3}
Tom Brown, Benjamin Mann, Nick Ryder, Melanie Subbiah, Jared~D Kaplan, Prafulla Dhariwal, Arvind Neelakantan, Pranav Shyam, Girish Sastry, Amanda Askell, et~al. 2020.
\newblock Language models are few-shot learners.
\newblock \emph{Advances in neural information processing systems}, 33:1877--1901.

\bibitem[{Chevalier et~al.(2023)Chevalier, Wettig, Ajith, and Chen}]{chevalier2023adapting}
Alexis Chevalier, Alexander Wettig, Anirudh Ajith, and Danqi Chen. 2023.
\newblock Adapting language models to compress contexts.
\newblock \emph{arXiv preprint arXiv:2305.14788}.

\bibitem[{Chiang et~al.(2023)Chiang, Li, Lin, Sheng, Wu, Zhang, Zheng, Zhuang, Zhuang, Gonzalez, Stoica, and Xing}]{vicuna2023}
Wei-Lin Chiang, Zhuohan Li, Zi~Lin, Ying Sheng, Zhanghao Wu, Hao Zhang, Lianmin Zheng, Siyuan Zhuang, Yonghao Zhuang, Joseph~E. Gonzalez, Ion Stoica, and Eric~P. Xing. 2023.
\newblock \href {https://lmsys.org/blog/2023-03-30-vicuna/} {Vicuna: An open-source chatbot impressing gpt-4 with 90\%* chatgpt quality}.

\bibitem[{Chiang et~al.(2024)Chiang, Zheng, Sheng, Angelopoulos, Li, Li, Zhang, Zhu, Jordan, Gonzalez et~al.}]{chiang2024chatbot}
Wei-Lin Chiang, Lianmin Zheng, Ying Sheng, Anastasios~Nikolas Angelopoulos, Tianle Li, Dacheng Li, Hao Zhang, Banghua Zhu, Michael Jordan, Joseph~E Gonzalez, et~al. 2024.
\newblock Chatbot arena: An open platform for evaluating llms by human preference.
\newblock \emph{arXiv preprint arXiv:2403.04132}.

\bibitem[{Chuang et~al.(2024)Chuang, Xing, Chang, Liu, Chen, and Hu}]{chuang2024learning}
Yu-Neng Chuang, Tianwei Xing, Chia-Yuan Chang, Zirui Liu, Xun Chen, and Xia Hu. 2024.
\newblock Learning to compress prompt in natural language formats.
\newblock \emph{arXiv preprint arXiv:2402.18700}.

\bibitem[{Dao et~al.(2022)Dao, Fu, Ermon, Rudra, and R{\'e}}]{dao2022flashattention}
Tri Dao, Dan Fu, Stefano Ermon, Atri Rudra, and Christopher R{\'e}. 2022.
\newblock Flashattention: Fast and memory-efficient exact attention with io-awareness.
\newblock \emph{Advances in Neural Information Processing Systems}, 35:16344--16359.

\bibitem[{Dao and Gu(2024)}]{dao2024mamba2}
Tri Dao and Albert Gu. 2024.
\newblock \href {https://openreview.net/forum?id=ztn8FCR1td} {Transformers are {SSM}s: Generalized models and efficient algorithms through structured state space duality}.
\newblock In \emph{Forty-first International Conference on Machine Learning}.

\bibitem[{De et~al.(2024)De, Smith, Fernando, Botev, Cristian-Muraru, Gu, Haroun, Berrada, Chen, Srinivasan et~al.}]{de2024griffin}
Soham De, Samuel~L Smith, Anushan Fernando, Aleksandar Botev, George Cristian-Muraru, Albert Gu, Ruba Haroun, Leonard Berrada, Yutian Chen, Srivatsan Srinivasan, et~al. 2024.
\newblock Griffin: Mixing gated linear recurrences with local attention for efficient language models.
\newblock \emph{arXiv preprint arXiv:2402.19427}.

\bibitem[{DeepSeek-AI(2024)}]{deepseekv2}
DeepSeek-AI. 2024.
\newblock \href {https://arxiv.org/abs/2405.04434} {Deepseek-v2: A strong, economical, and efficient mixture-of-experts language model}.
\newblock \emph{Preprint}, arXiv:2405.04434.

\bibitem[{Dong et~al.(2024)Dong, Cheng, Qin, and Wang}]{dong2024qaq}
Shichen Dong, Wen Cheng, Jiayu Qin, and Wei Wang. 2024.
\newblock Qaq: Quality adaptive quantization for llm kv cache.
\newblock \emph{arXiv preprint arXiv:2403.04643}.

\bibitem[{Duanmu et~al.(2024)Duanmu, Yuan, Li, Duan, Zhang, and Lin}]{duanmu2024skvq}
Haojie Duanmu, Zhihang Yuan, Xiuhong Li, Jiangfei Duan, Xingcheng Zhang, and Dahua Lin. 2024.
\newblock Skvq: Sliding-window key and value cache quantization for large language models.
\newblock \emph{arXiv preprint arXiv:2405.06219}.

\bibitem[{Fu(2024)}]{fu2024challenges}
Yao Fu. 2024.
\newblock Challenges in deploying long-context transformers: A theoretical peak performance analysis.
\newblock \emph{arXiv preprint arXiv:2405.08944}.

\bibitem[{Ge et~al.(2023)Ge, Zhang, Liu, Zhang, Han, and Gao}]{ge2023model}
Suyu Ge, Yunan Zhang, Liyuan Liu, Minjia Zhang, Jiawei Han, and Jianfeng Gao. 2023.
\newblock Model tells you what to discard: Adaptive kv cache compression for llms.
\newblock \emph{arXiv preprint arXiv:2310.01801}.

\bibitem[{Geng and Liu(2023)}]{geng2023openllama}
Xinyang Geng and Hao Liu. 2023.
\newblock \href {https://github.com/openlm-research/open_llama} {Openllama: An open reproduction of llama}.

\bibitem[{Golovneva et~al.(2024)Golovneva, Wang, Weston, and Sukhbaatar}]{golovneva2024contextual}
Olga Golovneva, Tianlu Wang, Jason Weston, and Sainbayar Sukhbaatar. 2024.
\newblock Contextual position encoding: Learning to count what's important.
\newblock \emph{arXiv preprint arXiv:2405.18719}.

\bibitem[{Gu and Dao(2023)}]{gu2023mamba}
Albert Gu and Tri Dao. 2023.
\newblock Mamba: Linear-time sequence modeling with selective state spaces.
\newblock \emph{arXiv preprint arXiv:2312.00752}.

\bibitem[{Hoffmann et~al.(2022)Hoffmann, Borgeaud, Mensch, Buchatskaya, Cai, Rutherford, Casas, Hendricks, Welbl, Clark et~al.}]{hoffmann2022scallinglaw}
Jordan Hoffmann, Sebastian Borgeaud, Arthur Mensch, Elena Buchatskaya, Trevor Cai, Eliza Rutherford, Diego de~Las Casas, Lisa~Anne Hendricks, Johannes Welbl, Aidan Clark, et~al. 2022.
\newblock Training compute-optimal large language models.
\newblock \emph{arXiv preprint arXiv:2203.15556}.

\bibitem[{Hooper et~al.(2024)Hooper, Kim, Mohammadzadeh, Mahoney, Shao, Keutzer, and Gholami}]{hooper2024kvquant}
Coleman Hooper, Sehoon Kim, Hiva Mohammadzadeh, Michael~W Mahoney, Yakun~Sophia Shao, Kurt Keutzer, and Amir Gholami. 2024.
\newblock Kvquant: Towards 10 million context length llm inference with kv cache quantization.
\newblock \emph{arXiv preprint arXiv:2401.18079}.

\bibitem[{Hsieh et~al.(2024)Hsieh, Sun, Kriman, Acharya, Rekesh, Jia, and Ginsburg}]{hsieh2024ruler}
Cheng-Ping Hsieh, Simeng Sun, Samuel Kriman, Shantanu Acharya, Dima Rekesh, Fei Jia, and Boris Ginsburg. 2024.
\newblock Ruler: What's the real context size of your long-context language models?
\newblock \emph{arXiv preprint arXiv:2404.06654}.

\bibitem[{Jiang et~al.(2023{\natexlab{a}})Jiang, Sablayrolles, Mensch, Bamford, Chaplot, de~las Casas, Bressand, Lengyel, Lample, Saulnier, Lavaud, Lachaux, Stock, Scao, Lavril, Wang, Lacroix, and Sayed}]{jiang2023mistral}
Albert~Q. Jiang, Alexandre Sablayrolles, Arthur Mensch, Chris Bamford, Devendra~Singh Chaplot, Diego de~las Casas, Florian Bressand, Gianna Lengyel, Guillaume Lample, Lucile Saulnier, Lélio~Renard Lavaud, Marie-Anne Lachaux, Pierre Stock, Teven~Le Scao, Thibaut Lavril, Thomas Wang, Timothée Lacroix, and William~El Sayed. 2023{\natexlab{a}}.
\newblock Mistral 7b.
\newblock \emph{arXiv}.

\bibitem[{Jiang et~al.(2024)Jiang, Li, Zhang, Wu, Luo, Ahn, Han, Abdi, Li, Lin, Yang, and Qiu}]{jiang2024minference}
Huiqiang Jiang, Yucheng Li, Chengruidong Zhang, Qianhui Wu, Xufang Luo, Surin Ahn, Zhenhua Han, Amir~H Abdi, Dongsheng Li, Chin-Yew Lin, Yuqing Yang, and Lili Qiu. 2024.
\newblock Minference 1.0: Accelerating pre-filling for long-context llms via dynamic sparse attention.
\newblock \emph{arXiv preprint arXiv:2407.02490}.

\bibitem[{Jiang et~al.(2023{\natexlab{b}})Jiang, Wu, Lin, Yang, and Qiu}]{jiang2023llmlingua1}
Huiqiang Jiang, Qianhui Wu, Chin-Yew Lin, Yuqing Yang, and Lili Qiu. 2023{\natexlab{b}}.
\newblock Llmlingua: Compressing prompts for accelerated inference of large language models.
\newblock \emph{arXiv preprint arXiv:2310.05736}.

\bibitem[{Kang et~al.(2024)Kang, Zhang, Kundu, Jeong, Liu, Krishna, and Zhao}]{kang2024gear}
Hao Kang, Qingru Zhang, Souvik Kundu, Geonhwa Jeong, Zaoxing Liu, Tushar Krishna, and Tuo Zhao. 2024.
\newblock Gear: An efficient kv cache compression recipefor near-lossless generative inference of llm.
\newblock \emph{arXiv preprint arXiv:2403.05527}.

\bibitem[{Katharopoulos et~al.(2020)Katharopoulos, Vyas, Pappas, and Fleuret}]{katharopoulos2020linearattention}
Angelos Katharopoulos, Apoorv Vyas, Nikolaos Pappas, and Fran{\c{c}}ois Fleuret. 2020.
\newblock Transformers are rnns: Fast autoregressive transformers with linear attention.
\newblock In \emph{International conference on machine learning}, pages 5156--5165. PMLR.

\bibitem[{Levy et~al.(2024)Levy, Jacoby, and Goldberg}]{levy2024twoneedles}
Mosh Levy, Alon Jacoby, and Yoav Goldberg. 2024.
\newblock Same task, more tokens: the impact of input length on the reasoning performance of large language models.
\newblock \emph{arXiv preprint arXiv:2402.14848}.

\bibitem[{Li et~al.(2023{\natexlab{a}})Li, Shao, Xie, Sheng, Zheng, Gonzalez, Stoica, Ma, and Zhang}]{li2023longchat}
Dacheng Li, Rulin Shao, Anze Xie, Ying Sheng, Lianmin Zheng, Joseph Gonzalez, Ion Stoica, Xuezhe Ma, and Hao Zhang. 2023{\natexlab{a}}.
\newblock How long can context length of open-source llms truly promise?
\newblock In \emph{NeurIPS 2023 Workshop on Instruction Tuning and Instruction Following}.

\bibitem[{Li et~al.(2023{\natexlab{b}})Li, Hammoud, Itani, Khizbullin, and Ghanem}]{li2023camel}
Guohao Li, Hasan Hammoud, Hani Itani, Dmitrii Khizbullin, and Bernard Ghanem. 2023{\natexlab{b}}.
\newblock Camel: Communicative agents for" mind" exploration of large language model society.
\newblock \emph{Advances in Neural Information Processing Systems}, 36:51991--52008.

\bibitem[{Li et~al.(2024{\natexlab{a}})Li, Shi, Jiang, Li, Xu, and Jia}]{li2024qllm}
Jingyao Li, Han Shi, Xin Jiang, Zhenguo Li, Hong Xu, and Jiaya Jia. 2024{\natexlab{a}}.
\newblock \href {https://arxiv.org/abs/2406.07528} {Quickllama: Query-aware inference acceleration for large language models}.
\newblock \emph{Preprint}, arXiv:2406.07528.

\bibitem[{Li et~al.(2024{\natexlab{b}})Li, Zhang, Do, Yue, and Chen}]{li2024long}
Tianle Li, Ge~Zhang, Quy~Duc Do, Xiang Yue, and Wenhu Chen. 2024{\natexlab{b}}.
\newblock Long-context llms struggle with long in-context learning.
\newblock \emph{arXiv preprint arXiv:2404.02060}.

\bibitem[{Li et~al.(2024{\natexlab{c}})Li, Huang, Yang, Venkitesh, Locatelli, Ye, Cai, Lewis, and Chen}]{li2024snapkv}
Yuhong Li, Yingbing Huang, Bowen Yang, Bharat Venkitesh, Acyr Locatelli, Hanchen Ye, Tianle Cai, Patrick Lewis, and Deming Chen. 2024{\natexlab{c}}.
\newblock Snapkv: Llm knows what you are looking for before generation.
\newblock \emph{arXiv preprint arXiv:2404.14469}.

\bibitem[{Lieber et~al.(2024)Lieber, Lenz, Bata, Cohen, Osin, Dalmedigos, Safahi, Meirom, Belinkov, Shalev-Shwartz et~al.}]{lieber2024jamba}
Opher Lieber, Barak Lenz, Hofit Bata, Gal Cohen, Jhonathan Osin, Itay Dalmedigos, Erez Safahi, Shaked Meirom, Yonatan Belinkov, Shai Shalev-Shwartz, et~al. 2024.
\newblock Jamba: A hybrid transformer-mamba language model.
\newblock \emph{arXiv preprint arXiv:2403.19887}.

\bibitem[{Liu and Wang(2023)}]{liu2023sparseland}
Shiwei Liu and Zhangyang Wang. 2023.
\newblock Ten lessons we have learned in the new" sparseland": A short handbook for sparse neural network researchers.
\newblock \emph{arXiv preprint arXiv:2302.02596}.

\bibitem[{Liu et~al.(2023)Liu, Qiao, Neiswanger, Wang, Tan, Tao, Li, Wang, Sun, Pangarkar et~al.}]{liu2023llm360}
Zhengzhong Liu, Aurick Qiao, Willie Neiswanger, Hongyi Wang, Bowen Tan, Tianhua Tao, Junbo Li, Yuqi Wang, Suqi Sun, Omkar Pangarkar, et~al. 2023.
\newblock Llm360: Towards fully transparent open-source llms.
\newblock \emph{arXiv preprint arXiv:2312.06550}.

\bibitem[{Liu et~al.(2024{\natexlab{a}})Liu, Desai, Liao, Wang, Xie, Xu, Kyrillidis, and Shrivastava}]{liu2024scissorhands}
Zichang Liu, Aditya Desai, Fangshuo Liao, Weitao Wang, Victor Xie, Zhaozhuo Xu, Anastasios Kyrillidis, and Anshumali Shrivastava. 2024{\natexlab{a}}.
\newblock Scissorhands: Exploiting the persistence of importance hypothesis for llm kv cache compression at test time.
\newblock \emph{Advances in Neural Information Processing Systems}, 36.

\bibitem[{Liu et~al.(2024{\natexlab{b}})Liu, Yuan, Jin, Zhong, Xu, Braverman, Chen, and Hu}]{kivi}
Zirui Liu, Jiayi Yuan, Hongye Jin, Shaochen Zhong, Zhaozhuo Xu, Vladimir Braverman, Beidi Chen, and Xia Hu. 2024{\natexlab{b}}.
\newblock Kivi: A tuning-free asymmetric 2bit quantization for kv cache.
\newblock \emph{arXiv preprint arXiv:2402.02750}.

\bibitem[{Ma et~al.(2022)Ma, Zhou, Kong, He, Gui, Neubig, May, and Zettlemoyer}]{ma2022mega}
Xuezhe Ma, Chunting Zhou, Xiang Kong, Junxian He, Liangke Gui, Graham Neubig, Jonathan May, and Luke Zettlemoyer. 2022.
\newblock Mega: moving average equipped gated attention.
\newblock \emph{arXiv preprint arXiv:2209.10655}.

\bibitem[{Mattern and Hohr(2023)}]{haven2023mambachat}
Justus Mattern and Konstantin Hohr. 2023.
\newblock \href {https://github.com/havenhq/mamba-chat} {Mamba-chat}.
\newblock GitHub.

\bibitem[{Mohtashami and Jaggi(2023)}]{mohtashami2023passkey}
Amirkeivan Mohtashami and Martin Jaggi. 2023.
\newblock Landmark attention: Random-access infinite context length for transformers.
\newblock \emph{arXiv preprint arXiv:2305.16300}.

\bibitem[{Mu et~al.(2023)Mu, Li, and Goodman}]{mu2023learning}
Jesse Mu, Xiang~Lisa Li, and Noah Goodman. 2023.
\newblock Learning to compress prompts with gist tokens.
\newblock \emph{arXiv preprint arXiv:2304.08467}.

\bibitem[{Munkhdalai et~al.(2024)Munkhdalai, Faruqui, and Gopal}]{munkhdalai2024infiniattention}
Tsendsuren Munkhdalai, Manaal Faruqui, and Siddharth Gopal. 2024.
\newblock Leave no context behind: Efficient infinite context transformers with infini-attention.
\newblock \emph{arXiv preprint arXiv:2404.07143}.

\bibitem[{Nawrot et~al.(2024)Nawrot, {\L}a{\'n}cucki, Chochowski, Tarjan, and Ponti}]{nawrot2024dmc}
Piotr Nawrot, Adrian {\L}a{\'n}cucki, Marcin Chochowski, David Tarjan, and Edoardo~M Ponti. 2024.
\newblock Dynamic memory compression: Retrofitting llms for accelerated inference.
\newblock \emph{arXiv preprint arXiv:2403.09636}.

\bibitem[{Pan et~al.(2024)Pan, Wu, Jiang, Xia, Luo, Zhang, Lin, Ruhle, Yang, Lin, Zhao, Qiu, and Zhang}]{jiang2023llmlingua}
Zhuoshi Pan, Qianhui Wu, Huiqiang Jiang, Menglin Xia, Xufang Luo, Jue Zhang, Qingwei Lin, Victor Ruhle, Yuqing Yang, Chin-Yew Lin, H.~Vicky Zhao, Lili Qiu, and Dongmei Zhang. 2024.
\newblock {LLML}ingua-2: Data distillation for efficient and faithful task-agnostic prompt compression.
\newblock In \emph{Findings of the Association for Computational Linguistics ACL 2024}.

\bibitem[{Peng et~al.(2023)Peng, Alcaide, Anthony, Albalak, Arcadinho, Cao, Cheng, Chung, Grella, GV et~al.}]{peng2023rwkv}
Bo~Peng, Eric Alcaide, Quentin Anthony, Alon Albalak, Samuel Arcadinho, Huanqi Cao, Xin Cheng, Michael Chung, Matteo Grella, Kranthi~Kiran GV, et~al. 2023.
\newblock Rwkv: Reinventing rnns for the transformer era.
\newblock \emph{arXiv preprint arXiv:2305.13048}.

\bibitem[{Pope et~al.(2023)Pope, Douglas, Chowdhery, Devlin, Bradbury, Heek, Xiao, Agrawal, and Dean}]{pope2023eff_scale_tf}
Reiner Pope, Sholto Douglas, Aakanksha Chowdhery, Jacob Devlin, James Bradbury, Jonathan Heek, Kefan Xiao, Shivani Agrawal, and Jeff Dean. 2023.
\newblock Efficiently scaling transformer inference.
\newblock \emph{Proceedings of Machine Learning and Systems}, 5.

\bibitem[{Qin et~al.(2024)Qin, Yang, Sun, Shen, Li, Sun, and Zhong}]{qin2024hgrn2}
Zhen Qin, Songlin Yang, Weixuan Sun, Xuyang Shen, Dong Li, Weigao Sun, and Yiran Zhong. 2024.
\newblock Hgrn2: Gated linear rnns with state expansion.
\newblock \emph{arXiv preprint arXiv:2404.07904}.

\bibitem[{Reid et~al.(2024)Reid, Savinov, Teplyashin, Lepikhin, Lillicrap, Alayrac, Soricut, Lazaridou, Firat, Schrittwieser et~al.}]{reid2024gemini}
Machel Reid, Nikolay Savinov, Denis Teplyashin, Dmitry Lepikhin, Timothy Lillicrap, Jean-baptiste Alayrac, Radu Soricut, Angeliki Lazaridou, Orhan Firat, Julian Schrittwieser, et~al. 2024.
\newblock Gemini 1.5: Unlocking multimodal understanding across millions of tokens of context.
\newblock \emph{arXiv preprint arXiv:2403.05530}.

\bibitem[{Roziere et~al.(2023)Roziere, Gehring, Gloeckle, Sootla, Gat, Tan, Adi, Liu, Remez, Rapin et~al.}]{roziere2023code}
Baptiste Roziere, Jonas Gehring, Fabian Gloeckle, Sten Sootla, Itai Gat, Xiaoqing~Ellen Tan, Yossi Adi, Jingyu Liu, Tal Remez, J{\'e}r{\'e}my Rapin, et~al. 2023.
\newblock Code llama: Open foundation models for code.
\newblock \emph{arXiv preprint arXiv:2308.12950}.

\bibitem[{Shazeer(2019)}]{shazeer2019mqa}
Noam Shazeer. 2019.
\newblock Fast transformer decoding: One write-head is all you need.
\newblock \emph{arXiv preprint arXiv:1911.02150}.

\bibitem[{Sheng et~al.(2023)Sheng, Zheng, Yuan, Li, Ryabinin, Chen, Liang, R{\'e}, Stoica, and Zhang}]{sheng2023flexgen}
Ying Sheng, Lianmin Zheng, Binhang Yuan, Zhuohan Li, Max Ryabinin, Beidi Chen, Percy Liang, Christopher R{\'e}, Ion Stoica, and Ce~Zhang. 2023.
\newblock Flexgen: High-throughput generative inference of large language models with a single gpu.
\newblock In \emph{International Conference on Machine Learning}, pages 31094--31116. PMLR.

\bibitem[{Su et~al.(2024)Su, Ahmed, Lu, Pan, Bo, and Liu}]{su2024roformer}
Jianlin Su, Murtadha Ahmed, Yu~Lu, Shengfeng Pan, Wen Bo, and Yunfeng Liu. 2024.
\newblock Roformer: Enhanced transformer with rotary position embedding.
\newblock \emph{Neurocomputing}, 568:127063.

\bibitem[{Sun et~al.()Sun, Dong, Huang, Ma, Xia, Xue, Wang, and Wei}]{retnet}
Yutao Sun, Li~Dong, Shaohan Huang, Shuming Ma, Yuqing Xia, Jilong Xue, Jianyong Wang, and Furu Wei.
\newblock Retentive network: A successor to transformer for large language models (2023).
\newblock \emph{URL http://arxiv. org/abs/2307.08621 v1}.

\bibitem[{Sun et~al.(2024)Sun, Dong, Zhu, Huang, Wang, Ma, Zhang, Wang, and Wei}]{sun2024yoco}
Yutao Sun, Li~Dong, Yi~Zhu, Shaohan Huang, Wenhui Wang, Shuming Ma, Quanlu Zhang, Jianyong Wang, and Furu Wei. 2024.
\newblock You only cache once: Decoder-decoder architectures for language models.
\newblock \emph{arXiv preprint arXiv:2405.05254}.

\bibitem[{Taylor et~al.(2022)Taylor, Kardas, Cucurull, Scialom, Hartshorn, Saravia, Poulton, Kerkez, and Stojnic}]{taylor2022galactica}
Ross Taylor, Marcin Kardas, Guillem Cucurull, Thomas Scialom, Anthony Hartshorn, Elvis Saravia, Andrew Poulton, Viktor Kerkez, and Robert Stojnic. 2022.
\newblock Galactica: A large language model for science.
\newblock \emph{arXiv preprint arXiv:2211.09085}.

\bibitem[{Touvron et~al.(2023)Touvron, Martin, Stone, Albert, Almahairi, Babaei, Bashlykov, Batra, Bhargava, Bhosale et~al.}]{touvron2023llama2}
Hugo Touvron, Louis Martin, Kevin Stone, Peter Albert, Amjad Almahairi, Yasmine Babaei, Nikolay Bashlykov, Soumya Batra, Prajjwal Bhargava, Shruti Bhosale, et~al. 2023.
\newblock \href {https://arxiv.org/abs/2307.09288} {Llama 2: Open foundation and fine-tuned chat models}.
\newblock \emph{Preprint}, arXiv:2307.09288.

\bibitem[{Wang et~al.(2024{\natexlab{a}})Wang, Ran, Tang, Chang, Chuang, Liu, Braverman, Liu, and Hu}]{wang2024assessing}
Guanchu Wang, Junhao Ran, Ruixiang Tang, Chia-Yuan Chang, Yu-Neng Chuang, Zirui Liu, Vladimir Braverman, Zhandong Liu, and Xia Hu. 2024{\natexlab{a}}.
\newblock Assessing and enhancing large language models in rare disease question-answering.
\newblock \emph{arXiv preprint arXiv:2408.08422}.

\bibitem[{Wang et~al.(2020)Wang, Li, Khabsa, Fang, and Ma}]{wang2020linformer}
Sinong Wang, Belinda~Z Li, Madian Khabsa, Han Fang, and Hao Ma. 2020.
\newblock Linformer: Self-attention with linear complexity.
\newblock \emph{arXiv preprint arXiv:2006.04768}.

\bibitem[{Wang et~al.(2024{\natexlab{b}})Wang, Yuan, Chuang, Wang, Liu, Cusick, Kulkarni, Ji, Ibrahim, and Hu}]{wang2024dhp}
Yicheng Wang, Jiayi Yuan, Yu-Neng Chuang, Zhuoer Wang, Yingchi Liu, Mark Cusick, Param Kulkarni, Zhengping Ji, Yasser Ibrahim, and Xia Hu. 2024{\natexlab{b}}.
\newblock Dhp benchmark: Are llms good nlg evaluators?
\newblock \emph{arXiv preprint arXiv:2408.13704}.

\bibitem[{Wingate et~al.(2022)Wingate, Shoeybi, and Sorensen}]{wingate2022prompt}
David Wingate, Mohammad Shoeybi, and Taylor Sorensen. 2022.
\newblock Prompt compression and contrastive conditioning for controllability and toxicity reduction in language models.
\newblock \emph{arXiv preprint arXiv:2210.03162}.

\bibitem[{Wu and Tu(2024)}]{wu2024lckv}
Haoyi Wu and Kewei Tu. 2024.
\newblock Layer-condensed kv cache for efficient inference of large language models.
\newblock \emph{arXiv preprint arXiv:2405.10637}.

\bibitem[{Wu et~al.(2023)Wu, Bansal, Zhang, Wu, Zhang, Zhu, Li, Jiang, Zhang, and Wang}]{wu2023autogen}
Qingyun Wu, Gagan Bansal, Jieyu Zhang, Yiran Wu, Shaokun Zhang, Erkang Zhu, Beibin Li, Li~Jiang, Xiaoyun Zhang, and Chi Wang. 2023.
\newblock Autogen: Enabling next-gen llm applications via multi-agent conversation framework.
\newblock \emph{arXiv preprint arXiv:2308.08155}.

\bibitem[{Xiao et~al.(2024)Xiao, Zhang, Han, Xiao, Lin, Zhang, Liu, Han, and Sun}]{xiao2024infllm}
Chaojun Xiao, Pengle Zhang, Xu~Han, Guangxuan Xiao, Yankai Lin, Zhengyan Zhang, Zhiyuan Liu, Song Han, and Maosong Sun. 2024.
\newblock Infllm: Unveiling the intrinsic capacity of llms for understanding extremely long sequences with training-free memory.
\newblock \emph{arXiv preprint arXiv:2402.04617}.

\bibitem[{Xiao et~al.(2023)Xiao, Tian, Chen, Han, and Lewis}]{streamllm}
Guangxuan Xiao, Yuandong Tian, Beidi Chen, Song Han, and Mike Lewis. 2023.
\newblock Efficient streaming language models with attention sinks.
\newblock \emph{arXiv preprint arXiv:2309.17453}.

\bibitem[{Yang et~al.(2024)Yang, Kim, Bae, Kwon, Park, Yang, Kwon, and Lee}]{yang2024no}
June~Yong Yang, Byeongwook Kim, Jeongin Bae, Beomseok Kwon, Gunho Park, Eunho Yang, Se~Jung Kwon, and Dongsoo Lee. 2024.
\newblock No token left behind: Reliable kv cache compression via importance-aware mixed precision quantization.
\newblock \emph{arXiv preprint arXiv:2402.18096}.

\bibitem[{Yang et~al.(2022)Yang, Tian, Peng, and Klein}]{yang2022re3}
Kevin Yang, Yuandong Tian, Nanyun Peng, and Dan Klein. 2022.
\newblock Re3: Generating longer stories with recursive reprompting and revision.
\newblock \emph{arXiv preprint arXiv:2210.06774}.

\bibitem[{Yang et~al.(2023)Yang, Wang, Shen, Panda, and Kim}]{yang2023gated}
Songlin Yang, Bailin Wang, Yikang Shen, Rameswar Panda, and Yoon Kim. 2023.
\newblock Gated linear attention transformers with hardware-efficient training.
\newblock \emph{arXiv preprint arXiv:2312.06635}.

\bibitem[{Yu et~al.(2022)Yu, Luo, Zhou, Si, Zhou, Wang, Feng, and Yan}]{yu2022metaformer}
Weihao Yu, Mi~Luo, Pan Zhou, Chenyang Si, Yichen Zhou, Xinchao Wang, Jiashi Feng, and Shuicheng Yan. 2022.
\newblock Metaformer is actually what you need for vision.
\newblock In \emph{Proceedings of the IEEE/CVF conference on computer vision and pattern recognition}, pages 10819--10829.

\bibitem[{Yuan et~al.(2023)Yuan, Tang, Jiang, and Hu}]{yuan2023large}
Jiayi Yuan, Ruixiang Tang, Xiaoqian Jiang, and Xia Hu. 2023.
\newblock Large language models for healthcare data augmentation: An example on patient-trial matching.
\newblock In \emph{AMIA Annual Symposium Proceedings}, volume 2023, page 1324. American Medical Informatics Association.

\bibitem[{Zandieh et~al.(2024)Zandieh, Daliri, and Han}]{zandieh2024qjl}
Amir Zandieh, Majid Daliri, and Insu Han. 2024.
\newblock Qjl: 1-bit quantized jl transform for kv cache quantization with zero overhead.
\newblock \emph{arXiv preprint arXiv:2406.03482}.

\bibitem[{Zhang et~al.(2024{\natexlab{a}})Zhang, Li, Liu, yang, Liu, Chen, Luo, and Yang}]{zhang2024marathon}
Lei Zhang, Yunshui Li, Ziqiang Liu, Jiaxi yang, Junhao Liu, Longze Chen, Run Luo, and Min Yang. 2024{\natexlab{a}}.
\newblock \href {https://arxiv.org/abs/2312.09542} {Marathon: A race through the realm of long context with large language models}.
\newblock \emph{Preprint}, arXiv:2312.09542.

\bibitem[{Zhang(2024)}]{zhang2024longmamba}
Peiyuan Zhang. 2024.
\newblock Longmamba.
\newblock \url{https://github.com/jzhang38/LongMamba}.

\bibitem[{Zhang et~al.(2024{\natexlab{b}})Zhang, Yi, Xu, and Shrivastava}]{zhang2024kv}
Tianyi Zhang, Jonah Yi, Zhaozhuo Xu, and Anshumali Shrivastava. 2024{\natexlab{b}}.
\newblock Kv cache is 1 bit per channel: Efficient large language model inference with coupled quantization.
\newblock \emph{arXiv preprint arXiv:2405.03917}.

\bibitem[{Zhang et~al.(2024{\natexlab{c}})Zhang, Chen, Hu, Xu, Chen, Hao, Han, Thai, Wang, Liu et~al.}]{zhang2024inftybench}
Xinrong Zhang, Yingfa Chen, Shengding Hu, Zihang Xu, Junhao Chen, Moo~Khai Hao, Xu~Han, Zhen~Leng Thai, Shuo Wang, Zhiyuan Liu, et~al. 2024{\natexlab{c}}.
\newblock $\infty$bench: Extending long context evaluation beyond 100k tokens.
\newblock \emph{arXiv preprint arXiv:2402.13718}.

\bibitem[{Zhang et~al.(2024{\natexlab{d}})Zhang, Sheng, Zhou, Chen, Zheng, Cai, Song, Tian, R{\'e}, Barrett et~al.}]{h2o}
Zhenyu Zhang, Ying Sheng, Tianyi Zhou, Tianlong Chen, Lianmin Zheng, Ruisi Cai, Zhao Song, Yuandong Tian, Christopher R{\'e}, Clark Barrett, et~al. 2024{\natexlab{d}}.
\newblock H2o: Heavy-hitter oracle for efficient generative inference of large language models.
\newblock \emph{Advances in Neural Information Processing Systems}, 36.

\bibitem[{Zhao et~al.(2024)Zhao, Lin, Zhu, Ye, Chen, Zheng, Ceze, Krishnamurthy, Chen, and Kasikci}]{zhao2024atom}
Yilong Zhao, Chien-Yu Lin, Kan Zhu, Zihao Ye, Lequn Chen, Size Zheng, Luis Ceze, Arvind Krishnamurthy, Tianqi Chen, and Baris Kasikci. 2024.
\newblock Atom: Low-bit quantization for efficient and accurate llm serving.
\newblock \emph{Proceedings of Machine Learning and Systems}, 6:196--209.

\end{thebibliography}

\clearpage
\appendix

\section{Details about Datasets}
\label{app: detailed datasets}

\subsection{Details Regarding LongBench}

For the aforementioned task (1)-(6), we adopt the implementation and benchmark setting of LongBench~\cite{bai2023longbench}; here's a more detailed introduction of tasks.

The long context benchmarking tasks are categorized into several types: Multi-document QA, Single-document QA, Summarization, Few-shot learning, Synthetic tasks, and Code tasks. Each task has specific metrics for evaluation, such as the F1 score, ROUGE-L, and Accuracy. The average length of most tasks ranges from 5k to 15k, and each task has 200 datapoints, except for MultiFieldQA (150), LCC (500), and RepoBench-P (500).

Single-document QA tasks include MultiFieldQA, NarrativeQA, and Qasper, each requiring the comprehension and extraction of information from lengthy texts. Multi-document QA tasks like HotpotQA, 2WikiMQA, and Musique require answering questions based on multiple documents. Summarization tasks, such as GovReport, MultiNews, and QMSUM, involve condensing long documents into concise summaries evaluated using Rouge-L. Few-shot tasks, including TriviaQA, SAMSum, and TREC, provide limited examples to guide the model in answering questions or categorizing data. Synthetic tasks like PassageRetrieval and PassageCount simulate real-world scenarios where models must identify relevant paragraphs or count distinct passages within a repetitive text. Code tasks such as LCC and RepoBench-P assess the model's ability to predict subsequent lines of code in various programming languages, emphasizing the use of cross-file dependencies.

Overall, LongBench's diverse tasks are meticulously designed to push the boundaries of long-context processing, providing a robust benchmark for assessing advanced language models. 

In our benchmark, we purposely omit the results of PassageCount, as LLMs often do not count correctly even in relatively short contexts~\cite{golovneva2024contextual}. All models and methods exhibit poor performance (i.e., less than 10\% accuracy), making the average performance unreliable with such an outlier included.


\subsection{Details Regarding Needle-in-a-Haystack Test}
\label{app_dataset_needle}

Needle-in-a-haystack (NIAH) is a style of synthetically generated stress test aiming to evaluate the information retrieval capability of language models. NIAH tasks often introduce a piece of key information that is inserted into unrelated background texts of various lengths and at various positions. To the best of our knowledge, the first two widely adopted versions of this task are proposed by \citet{mohtashami2023passkey} and \href{https://github.com/gkamradt/LLMTest_NeedleInAHaystack}{Greg Kamradt}. Specifically,  \citet{mohtashami2023passkey} inserts a piece of key information formatted like \texttt{``The pass key is <PASS KEY>.  Remember it. <PASS KEY> is the pass key''} into the different lengths of unrelated background texts filled by repetition of \texttt{``The grass is green. The sky is blue. The sun is yellow. Here we go. There and back again.''} — this task is often known as the passkey retrieval task. Yet, Greg Kamradt's version of NIAH inserts a sentence like \texttt{``The best thing to do in San Francisco is eat a switch and sit in Dolores Park on a sunny day.''} Under both tasks, the LLM-in-question is then asked to answer a question that would require it to retrieve such a piece of inserted information successfully.

Given the vast variants of such NIAH tasks (\href{https://github.com/gkamradt/LLMTest_NeedleInAHaystack}{gkamradt}, \href{https://github.com/gkamradt/LLMTest_NeedleInAHaystack}{Arize-ai}, \citet{levy2024twoneedles, mohtashami2023passkey, reid2024gemini, hsieh2024ruler}) existing in the community, we clarify the formation of our needle task as the following, which largely follows the passkey retrieval prompt template of \citet{mohtashami2023passkey, wang2024assessing} but using 7-digit passkey and Paul Graham Essays\footnote{\url{https://paulgraham.com/articles.html}} as the background filler, as set forth in \href{https://github.com/gkamradt/LLMTest_NeedleInAHaystack}{Arize-ai} and \citet{reid2024gemini}:

\noindent {\small\texttt{There is an important info hidden inside a lot of irrelevant text.  Find it and memorize them.  I will quiz you about the important information there.}}

\noindent {\small\texttt{<prefix filled by Paul Graham Essays>}}

\noindent {\small\texttt{The pass key is <7-DIGIT PASS KEY>.  Remember it. <7-DIGIT PASS KEY> is the pass key.}}

\noindent {\small\texttt{<suffix filler>}}

\noindent {\small\texttt{What is the pass key?  The pass key is}}

\section{Detailed Experiment Setup}
\label{app: detailed setup}

\subsection{LongBench Setting}

For baseline (no compression) performance, we follow the truncation settings in the LongBench official implementation as below in Table~\ref{max_length}.

\begin{table}[h]
\centering
\caption{Maximal prompt length in LongBench of different benchmarks.}
\label{max_length}
\resizebox{.8\linewidth}{!}{

\begin{tabular}{lc} 
\toprule
Model & \texttt{max\_length}  \\ 
\midrule
Meta-Llama-3-8B-Instruct & 7,500 \\ 
Mistral-7B-Instruct-v0.2 & 31,500 \\
LongChat-7b-v1.5-32k & 31,500 \\ 
\bottomrule
\end{tabular}
}
\end{table}

We note that following the official implementation of LongBench \citep{bai2023longbench}, for prompts that exceed the \texttt{max\_length} specified in Table \ref{max_length}, they will be middle-truncated by preserving the first and last $\texttt{max\_length}/2$ tokens.

For a fair comparison, the LongBench inputs of prefill-time compression methods like InfLLM \citep{xiao2024infllm} and StreamingLLM \citep{streamllm} are not truncated, but their maximum cache budget is capped at the respective base model \texttt{max\_length} $\times$ compression ratio. Namely, suppose InfLLM is evaluated on LongBench tasks with a base model of Mistral-7B-Instruct-v0.2 and under a compression ratio of $2\times$, its maximum KV cache budget would be equivalent to $31,500/2 = 15,750$ tokens (or the full prompt length/compression ratio, if such given prompt has a lower length than 31,500 tokens). The difference lies in that methods like InfLLM can decide where to allocate such budget across the full, non-truncated prompt, whereas methods like $\mathrm{H_2O}$ \citep{h2o} and KIVI \citep{kivi} are only given the middle-truncated prompt at the first place due to such method do not conduct compression during the prefill stage. We refer our readers to Appendix \ref{app_method_specific} for detailed settings regarding each compression method.

\subsection{Needle-in-a-Haystack Setting}

Following the designs of \citet{mohtashami2023passkey} and \citet{hsieh2024ruler}, we adopt the passkey retrieval task formulated in Appendix~\ref{app_dataset_needle} as our needle test. For granularity, we evaluate the LLM-in-question against 10 different sequence lengths uniformly spanning from \texttt{512} to \texttt{20480} words and in 10 different depths from the start to the end of the input. For each length-depth combination, we iterate the test 3 times with 3 randomly generated \texttt{<7-DIGIT PASS KEY>}. We highlight the length of our needle test — \texttt{20480} — is in terms of the number of words, but not the number of tokens, as different models might employ tokenizers with different efficiency, where an aligned input construction should be maintained for proper cross model comparison (which is inevitable given the involvement of linear-time sequence models and their hybrids). \texttt{20480} words in our needle test usually converts to roughly 30.6k tokens with the tokenizer utilized in models like LongChat-7b-v1.5-32k \citep{li2023longchat}, but only 27.2k tokens in models like Meta-Llama-3-8B-Instruct \citep{llama3modelcard} with a more efficient tokenizer.

We evaluated our needle test against three popular transformer-based language models (Mistral-7b-Instruct-v0.2 \citep{jiang2023mistral}, LongChat-7B-v1.5-32K \citep{li2023longchat},  Llama-8B-Instruct \citep{llama3modelcard}) as well as several other liner-time sequence models and hybrid architectures mentioned in Section~\ref{sec_bench_coverage}. Given that Mistral-7b-Instruct-v0.2 and LongChat-7B-v1.5-32K come with a context window of 32k tokens, we feed our needle inputs into such models in a vanilla fashion, whereas for Llama-8B-Instruct, we enlarge its RoPE base theta ($\theta$) \citep{su2024roformer} setting to 32$\times$ of its original size due to its limited 8k off-the-shelf context window.


\subsection{Method-specific Setting}

\label{app_method_specific}

\paragraph{Linear-time sequence models and mixed architecture} In our paper, we benchmark five pure or hybrid linear-time sequence models. While such models can theoretically achieve infinite context lengths, model performance is still expected to degrade when the context length exceeds the effective context length, which is typically the length used during the pretraining phase. The context lengths used in benchmarking LongBench \citep{bai2023longbench} are provided in Table \ref{ctx-len}. For our Needle-in-a-Haystack task \citep{mohtashami2023passkey, hsieh2024ruler} defined in Appendix \ref{app_dataset_needle}, we uniformly set the maximum context length to \texttt{20480} words to ensure consistency and fair comparison across tasks.

\begin{table}[h]
\centering
\caption{Effective context length and model size of the five linear time sequence models benchmarked in our paper.}
\label{ctx-len}
\resizebox{.8\linewidth}{!}{

\begin{tabular}{lc} 
\toprule
Model                & Eff. context length~  \\ 
\midrule
Mamba-2.8B           & 2k               \\ 
Mamba2-2.7B           & 2k               \\ 
Mamba-Chat-2.8B      & 2k               \\ 
RWKV-5-World-7B      & 4k               \\ 
RecurrentGemma-2B-it & 8k               \\ 
RecurrentGemma-9B-it & 8k               \\
\bottomrule
\end{tabular}
}
\end{table}

\paragraph{Quantization}

We benchmark two popular KV cache quantization methods: one 2bit quantization (KIVI-2) and two 4bit quantizations (KIVI-4 and FlexGen). 
For KIVI \citep{kivi}, we use the official implementation\footnote{\url{https://github.com/jy-yuan/KIVI}}, and for FlexGen \citep{sheng2023flexgen}, we follow the group-wise quantization in the official codebase\footnote{\url{https://github.com/FMInference/FlexGen}}.  
The group size for both KIVI and FlexGen is set to 32. We further set the residual length, which is unique to KIVI, as 128.

\paragraph{Token Dropping}

We evaluate three popular token dropping methods used for handling long contexts: StreamingLLM~\citep{streamllm}, InfLLM~\citep{xiao2024infllm}, and $\mathrm{H_2O}$~\citep{h2o}. In $\mathrm{H_2O}$, there are two parameters for controlling the token dropping ratio: the heavy ratio and the recent ratio. The recent ratio controls the number of tokens preserved within the local window, while the heavy ratio controls the number of heavy-hitter tokens outside the local window. We set both the heavy ratio and recent ratio to the same values of 25\%, 12.5\%, 8.3\%, and 6.25\% of the total token length to achieve compression gains of 2$\times$, 4$\times$, 6$\times$, and 8$\times$, respectively, following the setting\footnote{\url{https://github.com/FMInference/H2O/blob/main/h2o_hf/README.md}} set forth in $\mathrm{H_2O}$'s official implementation\footnote{\url{https://github.com/FMInference/H2O}}.

We emphasize that under this linear compression scheme utilized in $\mathrm{H_2O}$, the KV cache size scales linearly with the input prompt length. On the other hand, StreamingLLM maintains a constant window size of ``attention sinks'' (i.e., front-most tokens) and recent tokens, making the size of the KV cache constant at all times (irrelevant to input length) in its original design. Thus, to hit a consistent compression rate that is reasonably comparable to other methods, we modify the total number of tokens retained in the StreamingLLM pipeline as the product of the target compression rate and the input length — i.e., for a prompt of 1,000 tokens, a StreamingLLM-empowered LLM with 2$\times$ compression rate would have a 500 tokens KV cache budget to distribute among its attention sink and most recent tokens. We ensure the ratio of attention sinks to recent tokens within the KV cache matches the ratio of 2\% and 98\%, according to its official configurations\footnote{e.g., \url{https://github.com/thunlp/InfLLM/blob/main/config/mistral-stream-llm.yaml}}. 

In addition to the attention sink and recent tokens, InfLLM \citep{xiao2024infllm} incorporates the most relevant tokens from the middle of the context into the kept KV cache. We, therefore, preserve the ratio of attention sinks, middle tokens, and recent tokens as 2\%, 32\%, and 66\%, respectively, again being faithful to its official configurations\footnote{e.g., \url{https://github.com/thunlp/InfLLM/blob/main/config/llama-3-inf-llm.yaml}}. 

We borrowed our implementations of StreamingLLM and InfLLM from InfLLM's \citep{xiao2024infllm} official repository\footnote{\url{https://github.com/thunlp/InfLLM}} as this is the official implementation of InfLLM, yet it is endorsed by the lead author of StreamingLLM due to overlapped authorships.

\paragraph{Prompt Compression}
We evaluate LLMLingua\footnote{\url{https://github.com/microsoft/LLMLingua}} \citep{jiang2023llmlingua} on four different compression rates. K$\times$ for $K \in \{2, 4, 6, 8\}$ denotes that the compressor is restricted to compress the length into 1/K of the original length of long inputs.

\section{Related Works}
\label{app: related works}

The evaluation of LLM has been well studied~\citep{chiang2024chatbot, wang2024dhp}. Given the importance of long context-capable LLMs, many related works try to quantify such capabilities, usually via means of purposing new, long context-focused datasets. For example, LongBench~\cite{bai2023longbench} — which is also utilized in our work — provides a bilingual, multitask benchmark for long context understanding. Datasets like InfiniBench~\citep{zhang2024inftybench}, LongICLBench~\citep{li2024long}, Marathon~\citep{zhang2024marathon}, and Ruler~\citep{hsieh2024ruler} all contribute their perspective in terms of long context evaluation via different collections of real or synthetic tasks.

Our work differs from the abovementioned prior arts as such arts mainly focus on producing long context evaluation datasets, where the included benchmarks — if any — are mostly evaluated on vanilla baseline models without any compression methods applied; where our work presents comprehensive results primarily highlighting \textbf{the comparison among different efficient but long context-capable approaches}. We cover 10+ long context-capable approaches under 60+ different settings. To the best of our knowledge, no prior art has benchmarked similar coverage of compression methods under a long context scenario as we do.

\section{Extended Experimental Results}
\label{app: more exp}

In this section, we present additional experimental results for LongBench and the needle tasks. 

\begin{figure*}[h]
\setlength{\abovecaptionskip}{0mm}
\setlength{\belowcaptionskip}{0mm}
\centering
\subfigcapskip=-2mm
\subfigure[Llama-3-8B-Instruct]{
\centering
	\begin{minipage}[t]{0.32\linewidth}
		\includegraphics[width=\linewidth]{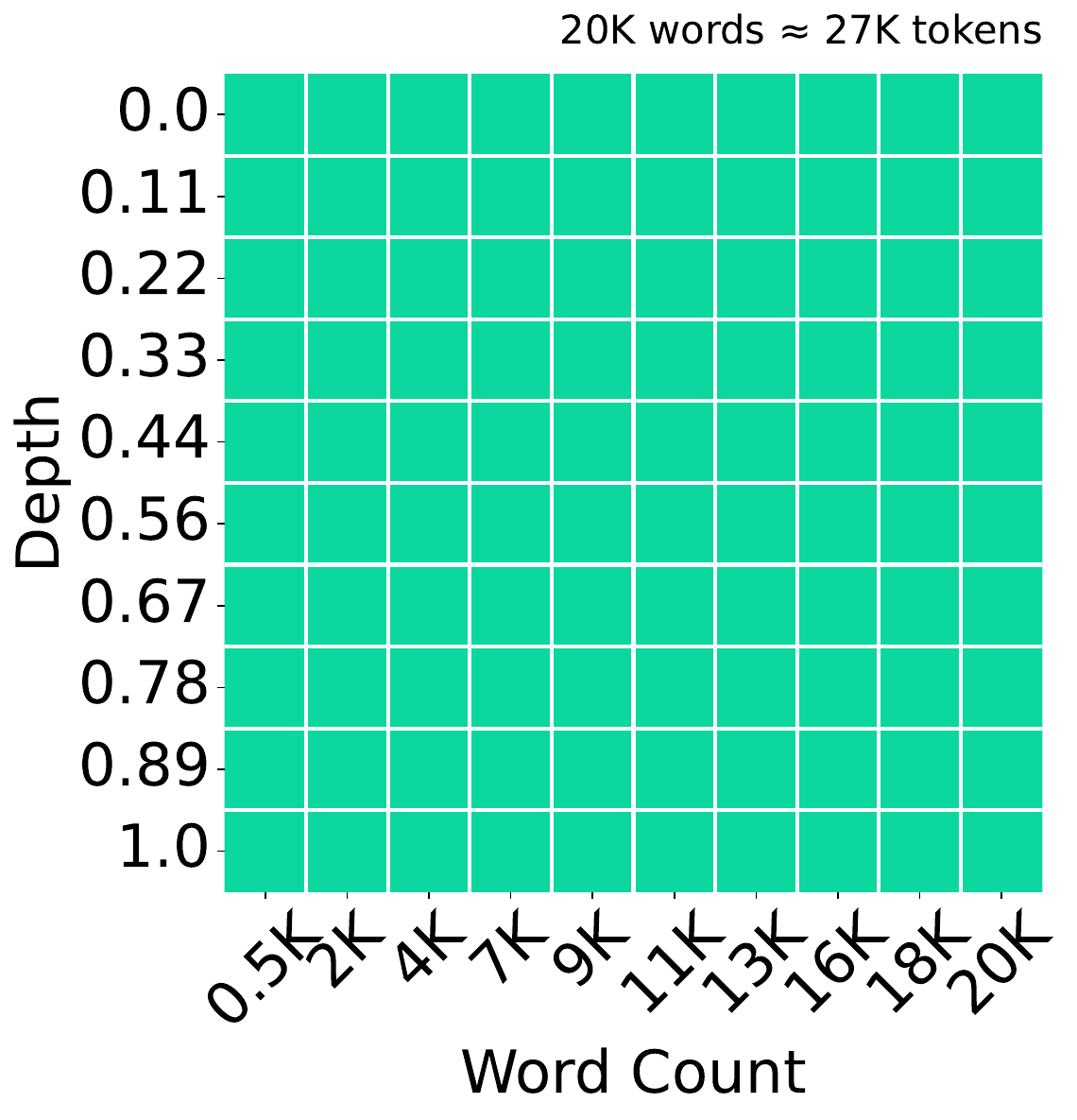}
	\end{minipage}%
}
\subfigure[Mistral-7B-v0.2-Instruct]{
\centering
	\begin{minipage}[t]{0.32\linewidth}
		\includegraphics[width=\linewidth]{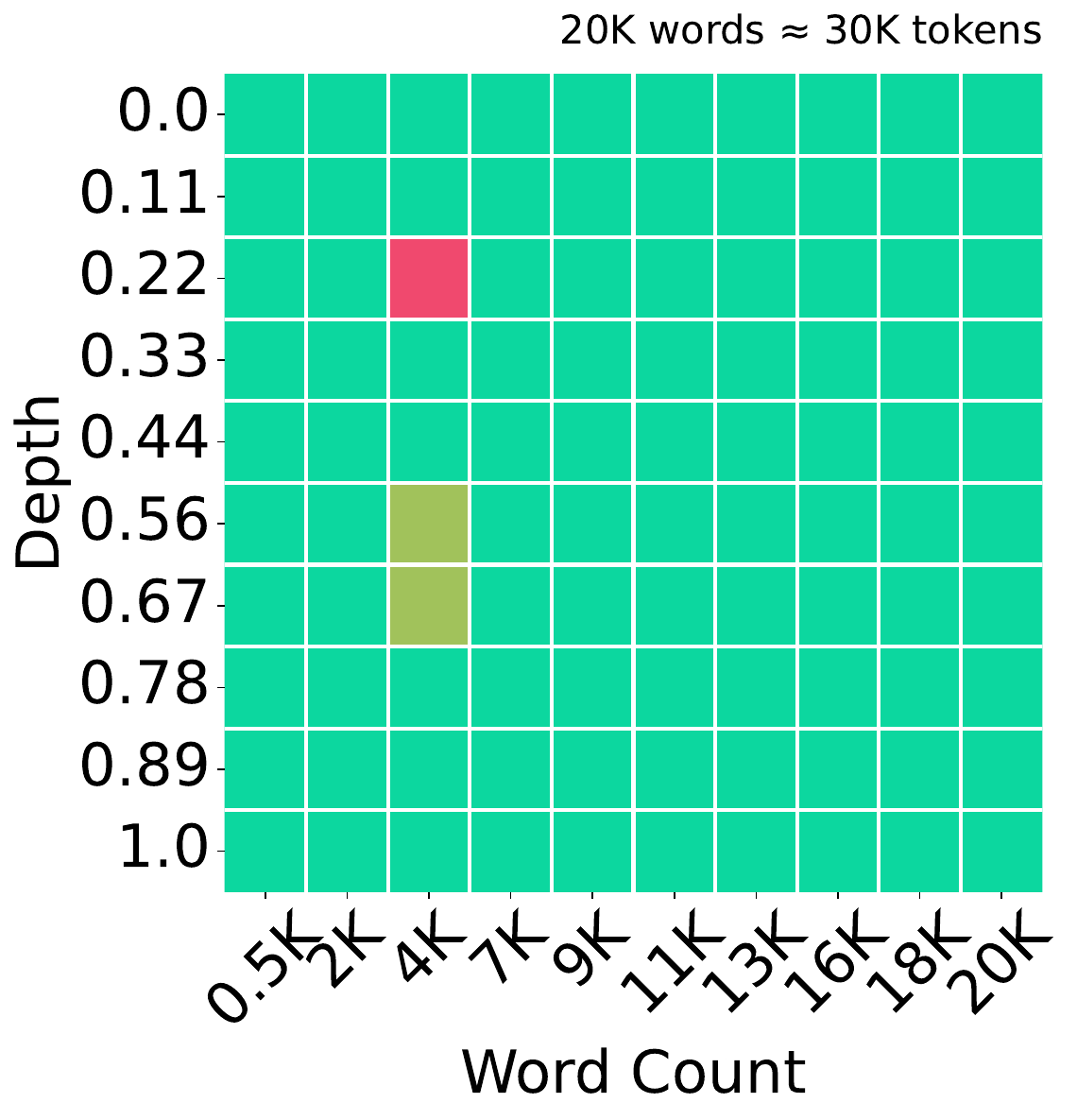}
	\end{minipage}%
}
\subfigure[LongChat-7B-v1.5-32k]{
\centering
	\begin{minipage}[t]{0.32\linewidth}
		\includegraphics[width=\linewidth]{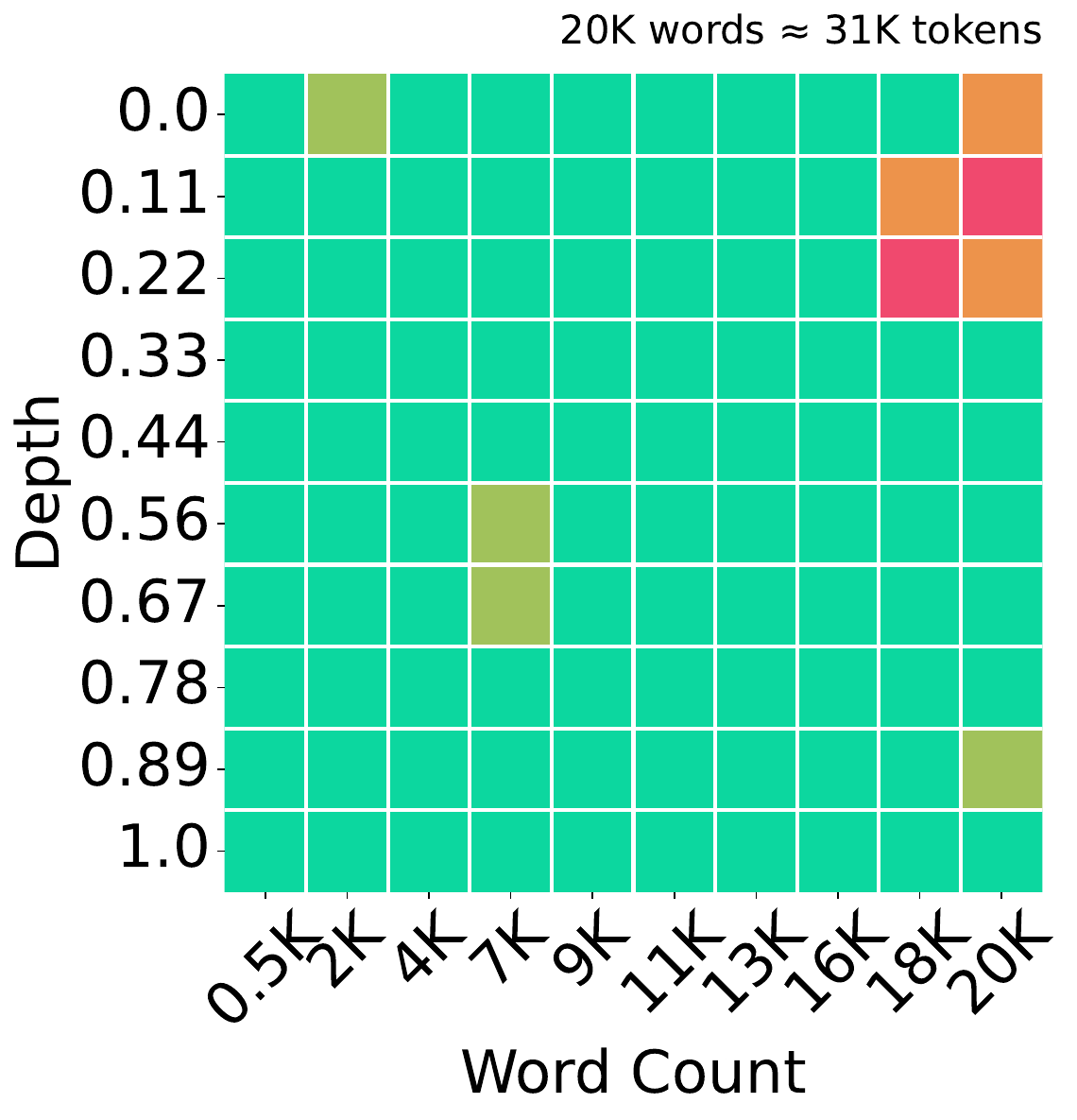}
	\end{minipage}%
}
\caption{FlexGen performance under needle test on three commonly used LLMs}
\label{fig:needle_flexgen}
\end{figure*}

\begin{figure*}[h]
\setlength{\abovecaptionskip}{0mm}
\setlength{\belowcaptionskip}{0mm}
\centering
\subfigcapskip=-2mm
\subfigure[2x Compression]{
\centering
	\begin{minipage}[t]{0.23\linewidth}
		\includegraphics[width=\linewidth]{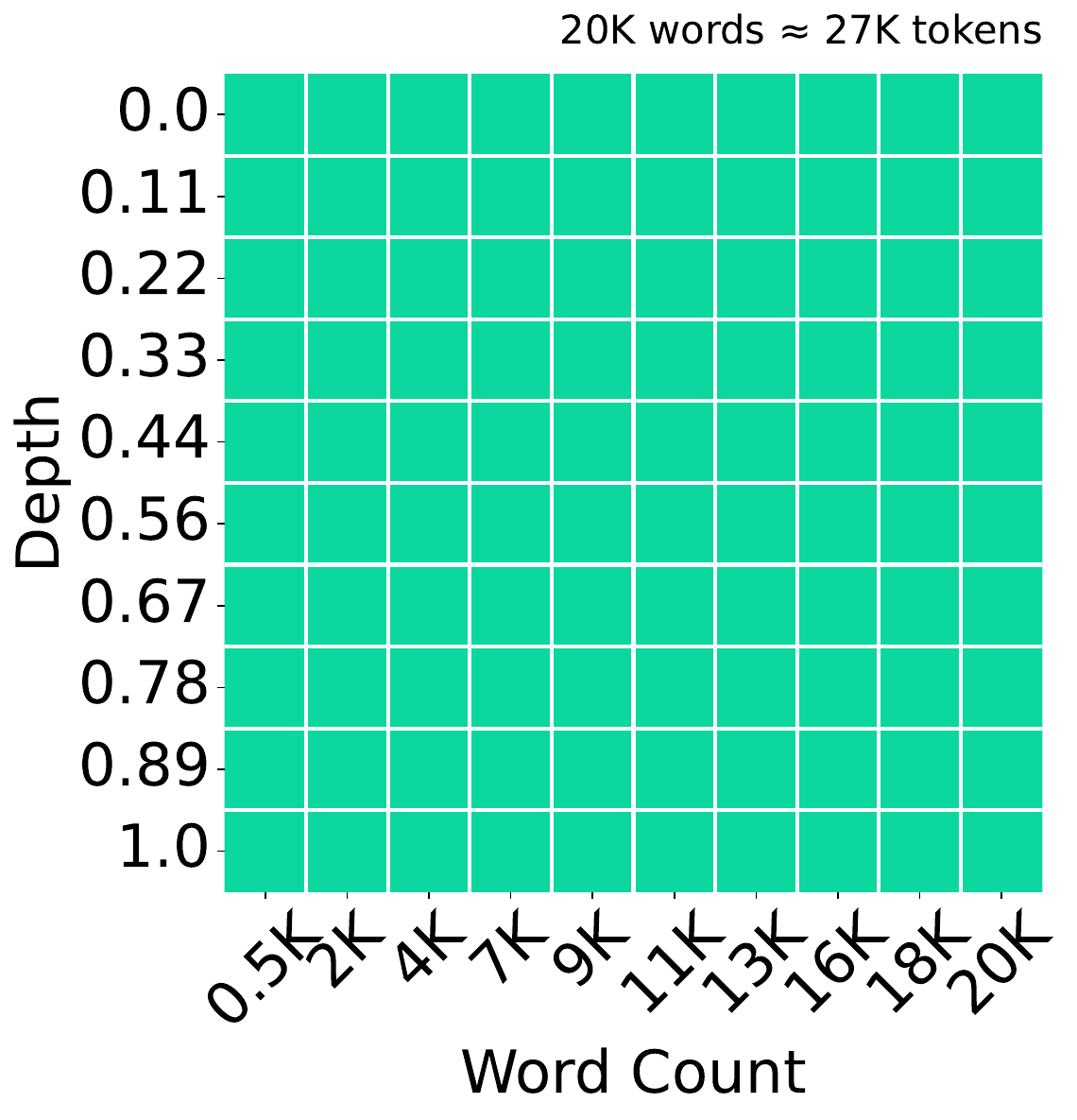}
	\end{minipage}%
}
\subfigure[4x Compression]{
\centering
	\begin{minipage}[t]{0.23\linewidth}
		\includegraphics[width=\linewidth]{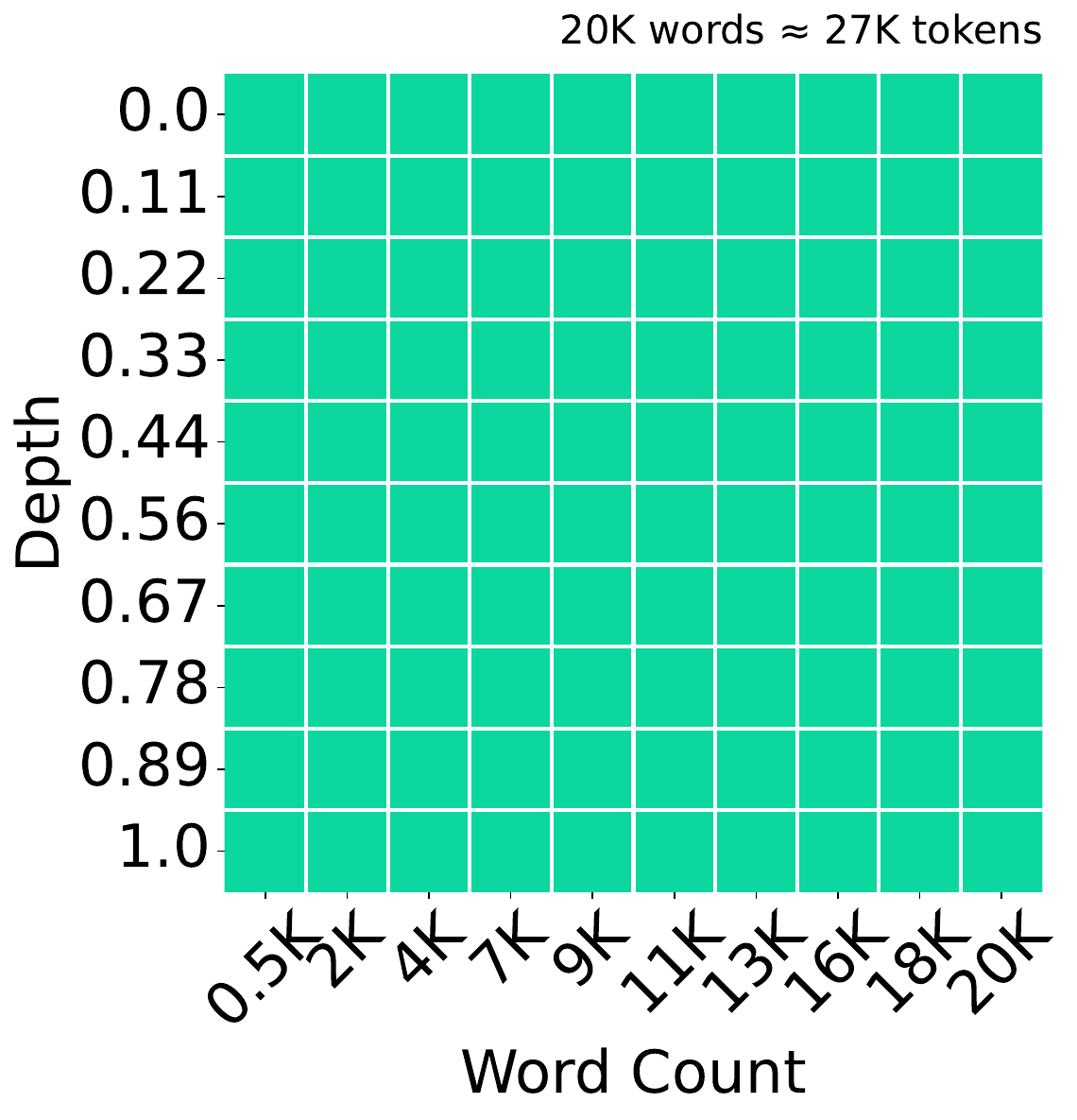}
	\end{minipage}%
}
\subfigure[6x Compression]{
\centering
	\begin{minipage}[t]{0.23\linewidth}
		\includegraphics[width=\linewidth]{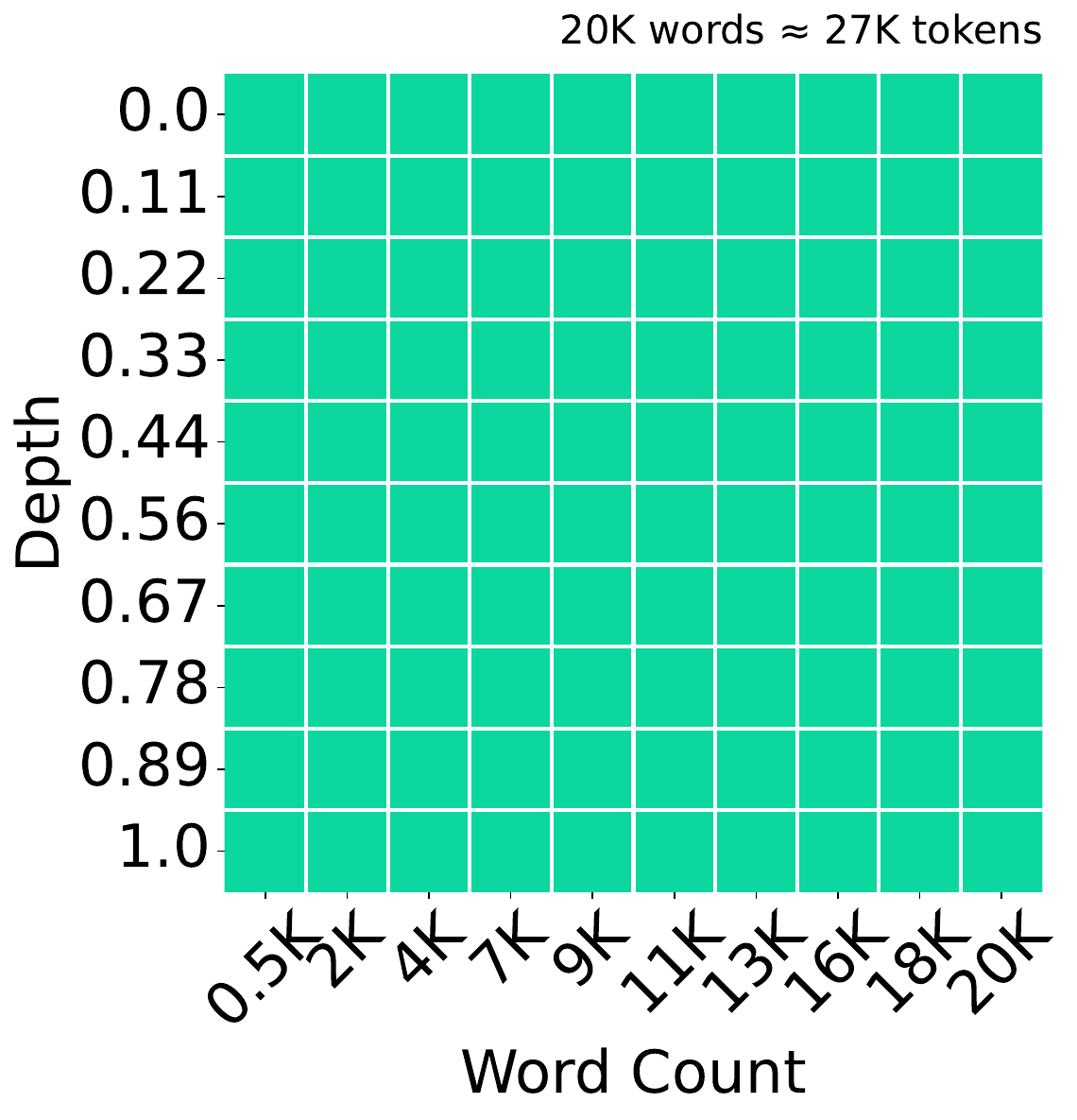}
	\end{minipage}%
}
\subfigure[8x Compression]{
\centering
	\begin{minipage}[t]{0.23\linewidth}
		\includegraphics[width=\linewidth]{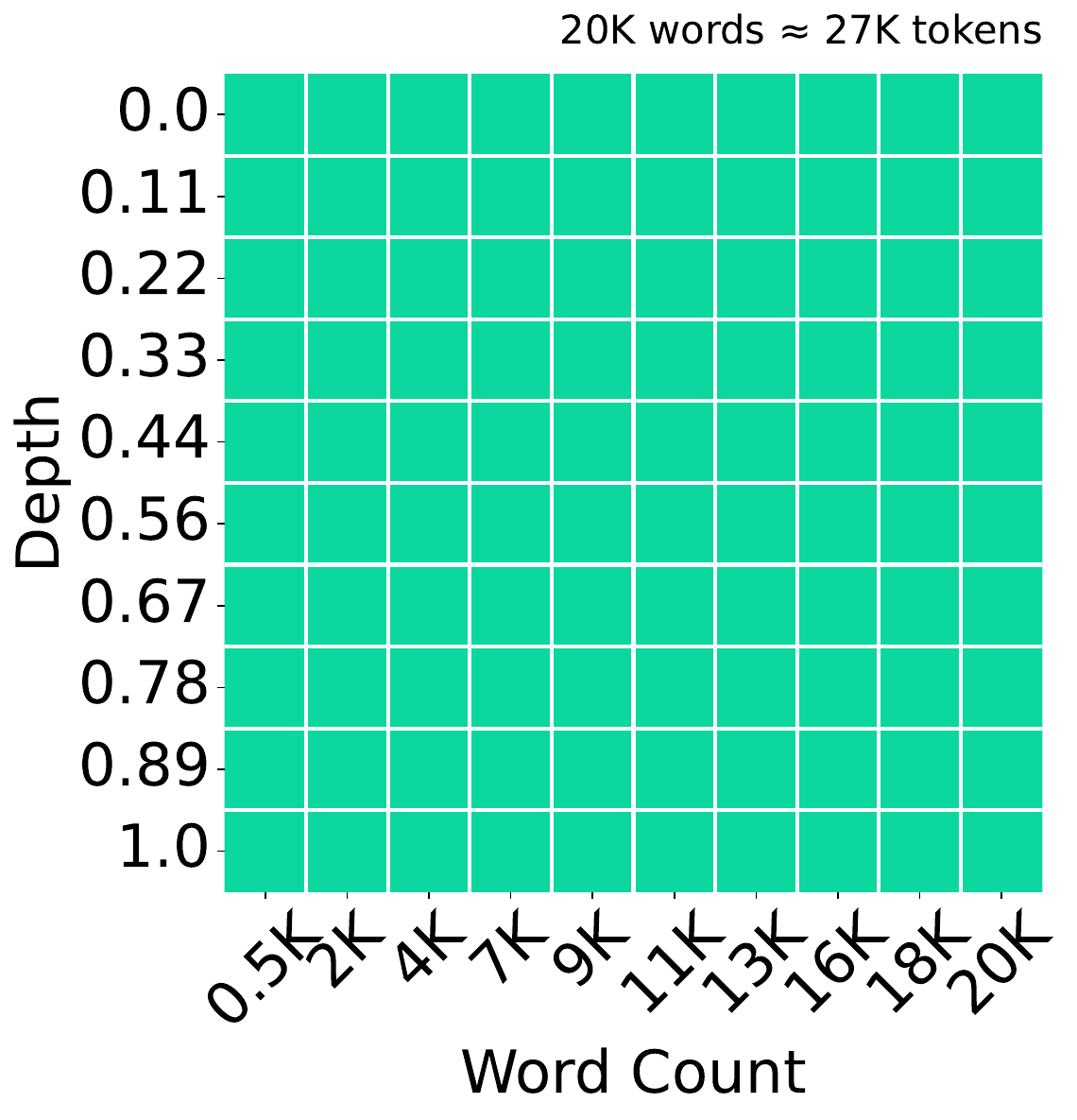}
	\end{minipage}%
}
\caption{$\mathrm{H_2O}$ on Llama-3-8B-Instruct with 4 different compression rates under needle test}
\label{fig:needle_h2o_llama}
\end{figure*}

\begin{figure*}[h]
\setlength{\abovecaptionskip}{0mm}
\setlength{\belowcaptionskip}{0mm}
\centering
\subfigcapskip=-2mm
\subfigure[2x Compression]{
\centering
	\begin{minipage}[t]{0.23\linewidth}
		\includegraphics[width=\linewidth]{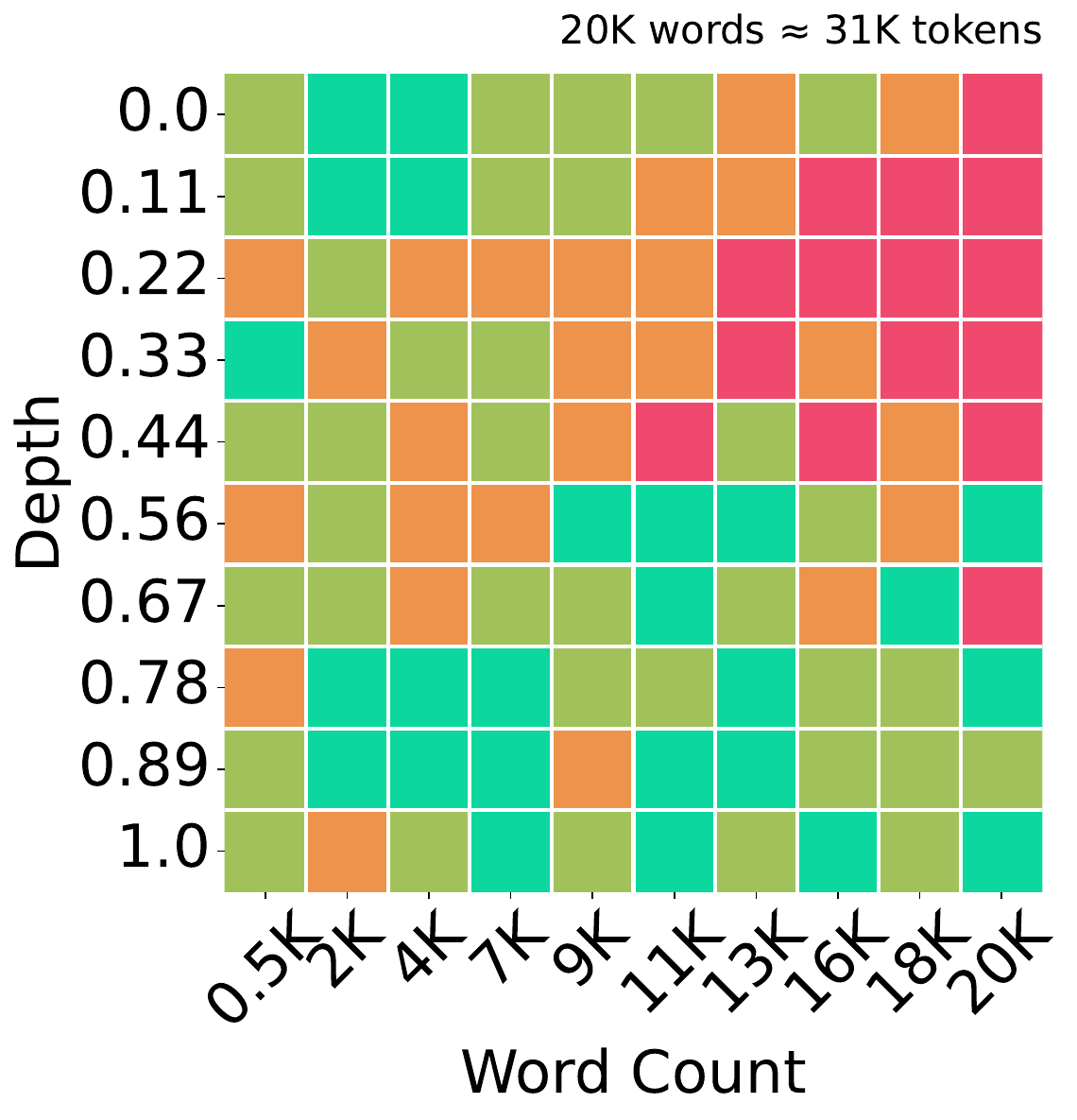}
	\end{minipage}%
}
\subfigure[4x Compression]{
\centering
	\begin{minipage}[t]{0.23\linewidth}
		\includegraphics[width=\linewidth]{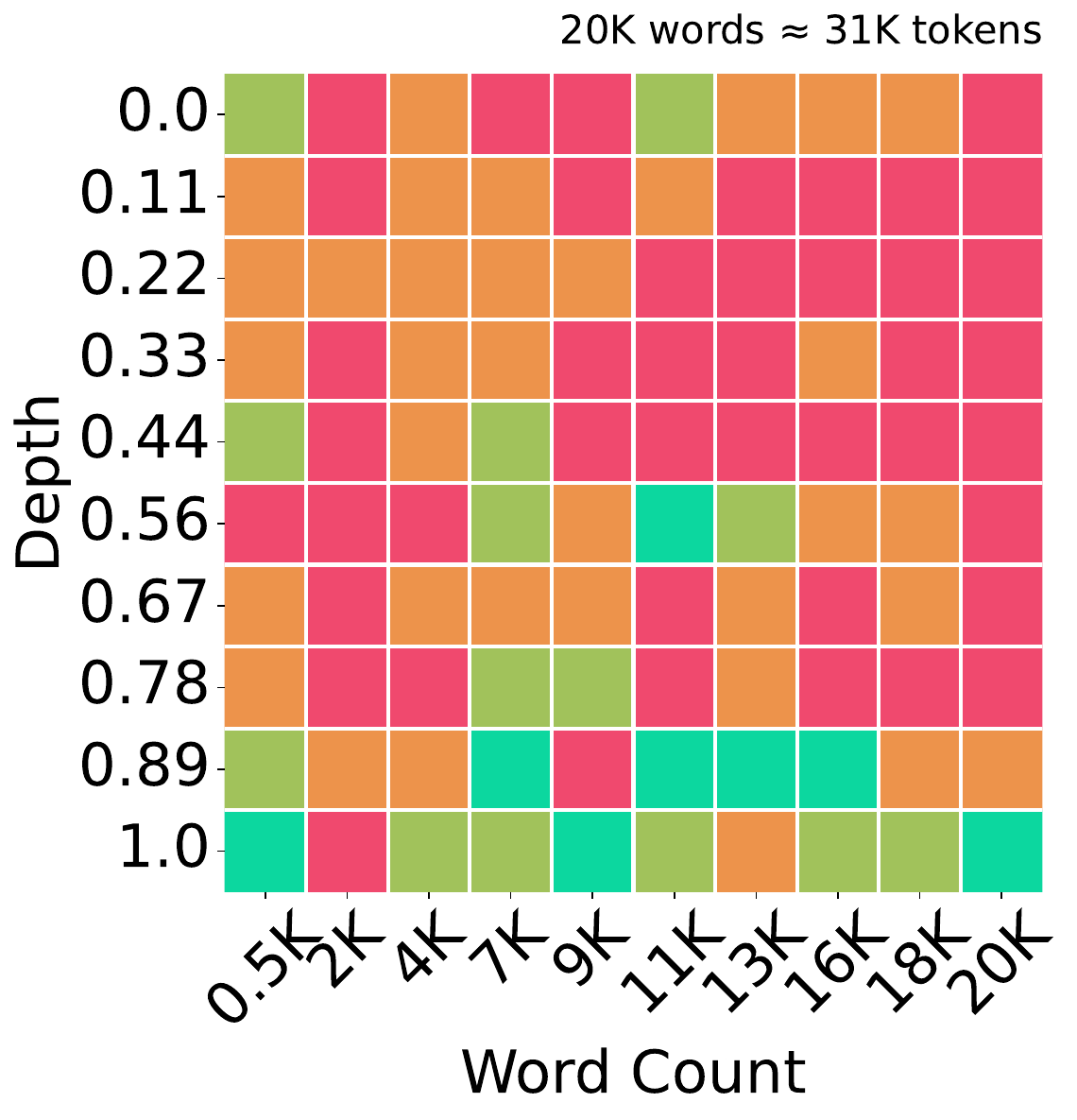}
	\end{minipage}%
}
\subfigure[6x Compression]{
\centering
	\begin{minipage}[t]{0.23\linewidth}
		\includegraphics[width=\linewidth]{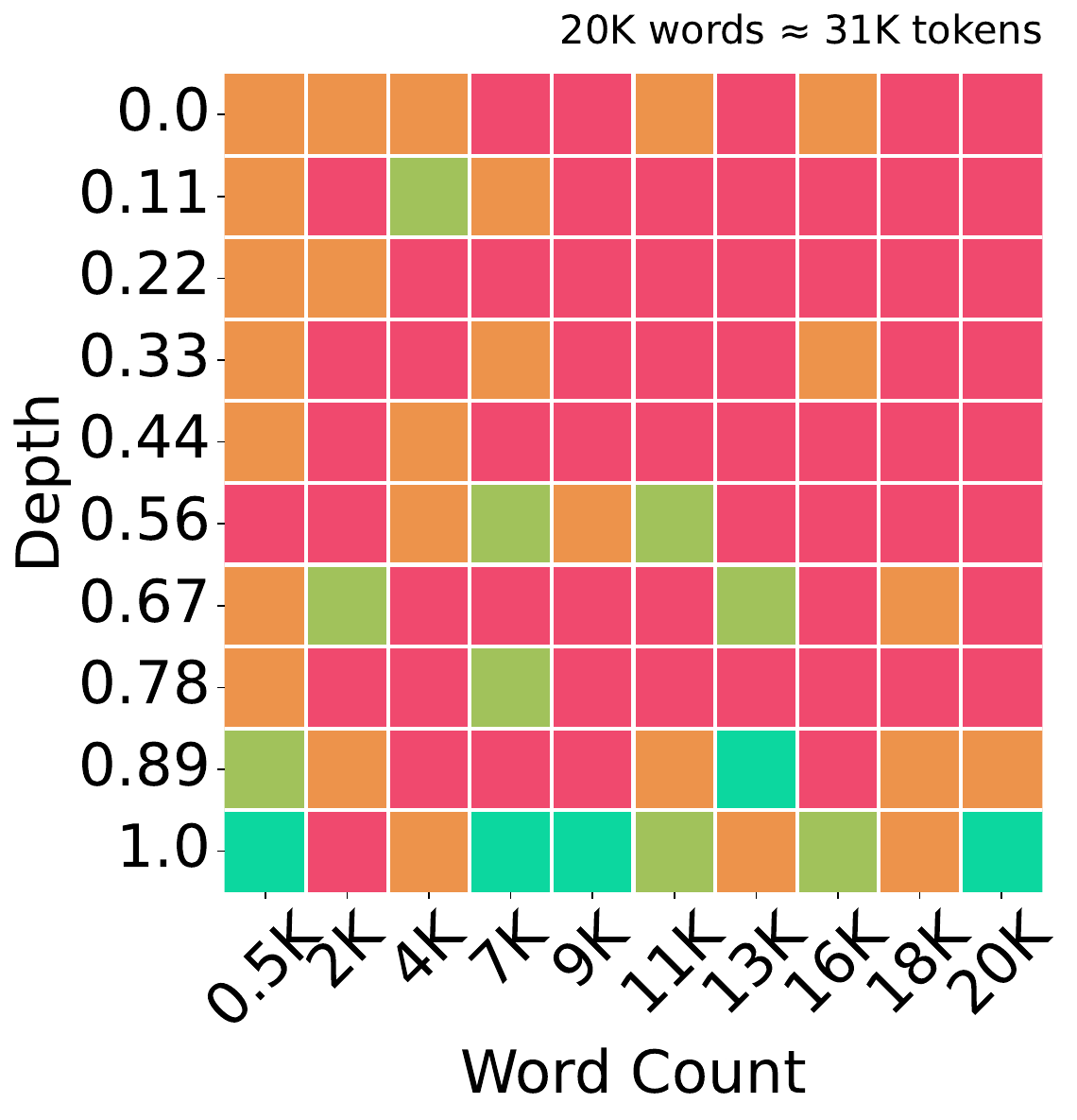}
	\end{minipage}%
}
\subfigure[8x Compression]{
\centering
	\begin{minipage}[t]{0.23\linewidth}
		\includegraphics[width=\linewidth]{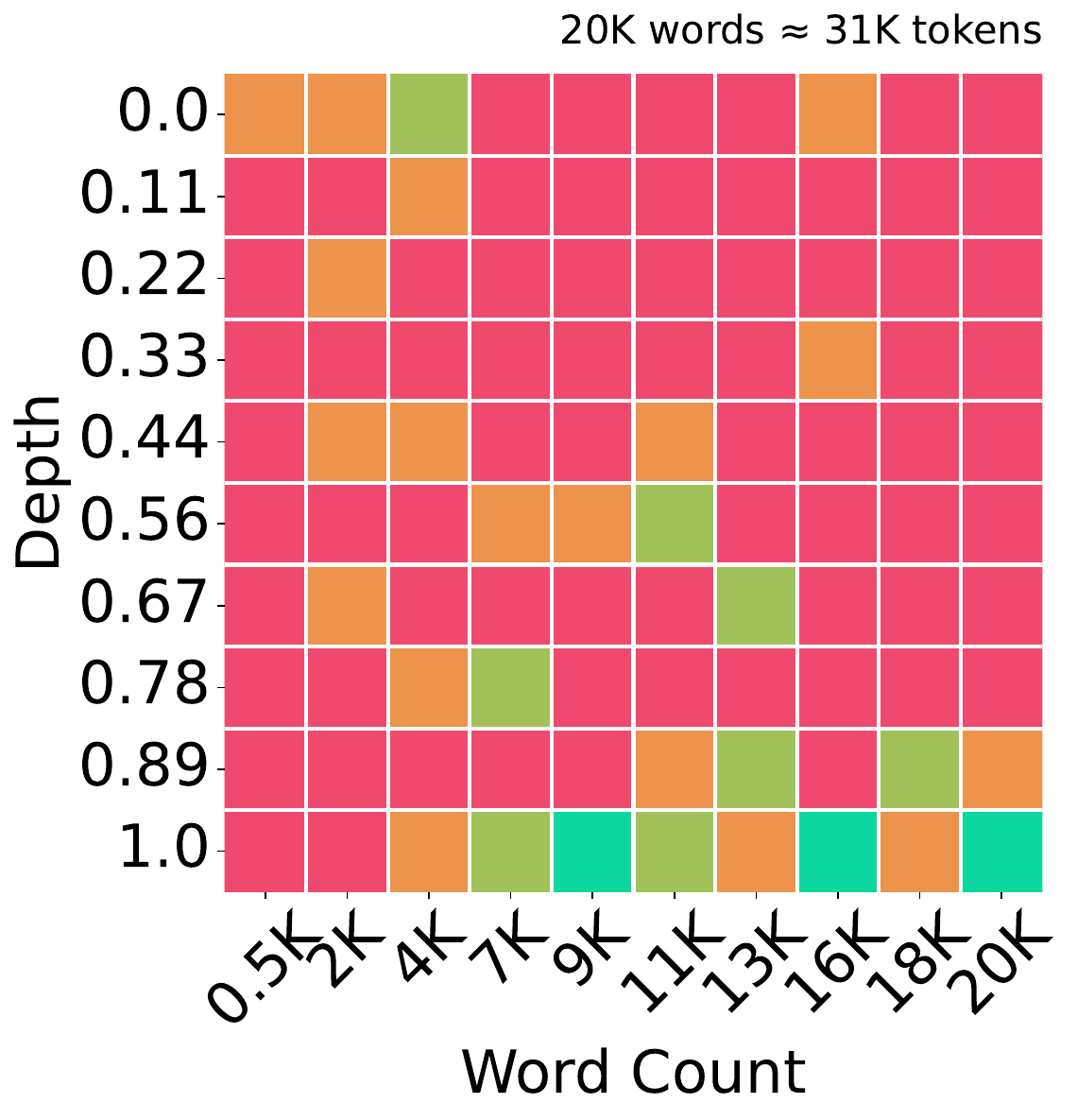}
	\end{minipage}%
}
\caption{$\mathrm{H_2O}$ on LongChat-7B-v1.5-32k with 4 different compression rates under needle test}
\label{fig:needle_h2o_longchat}
\end{figure*}

\begin{figure*}[h]
\setlength{\abovecaptionskip}{0mm}
\setlength{\belowcaptionskip}{0mm}
\centering
\subfigcapskip=-2mm
\subfigure[2x Compression]{
\centering
	\begin{minipage}[t]{0.23\linewidth}
		\includegraphics[width=\linewidth]{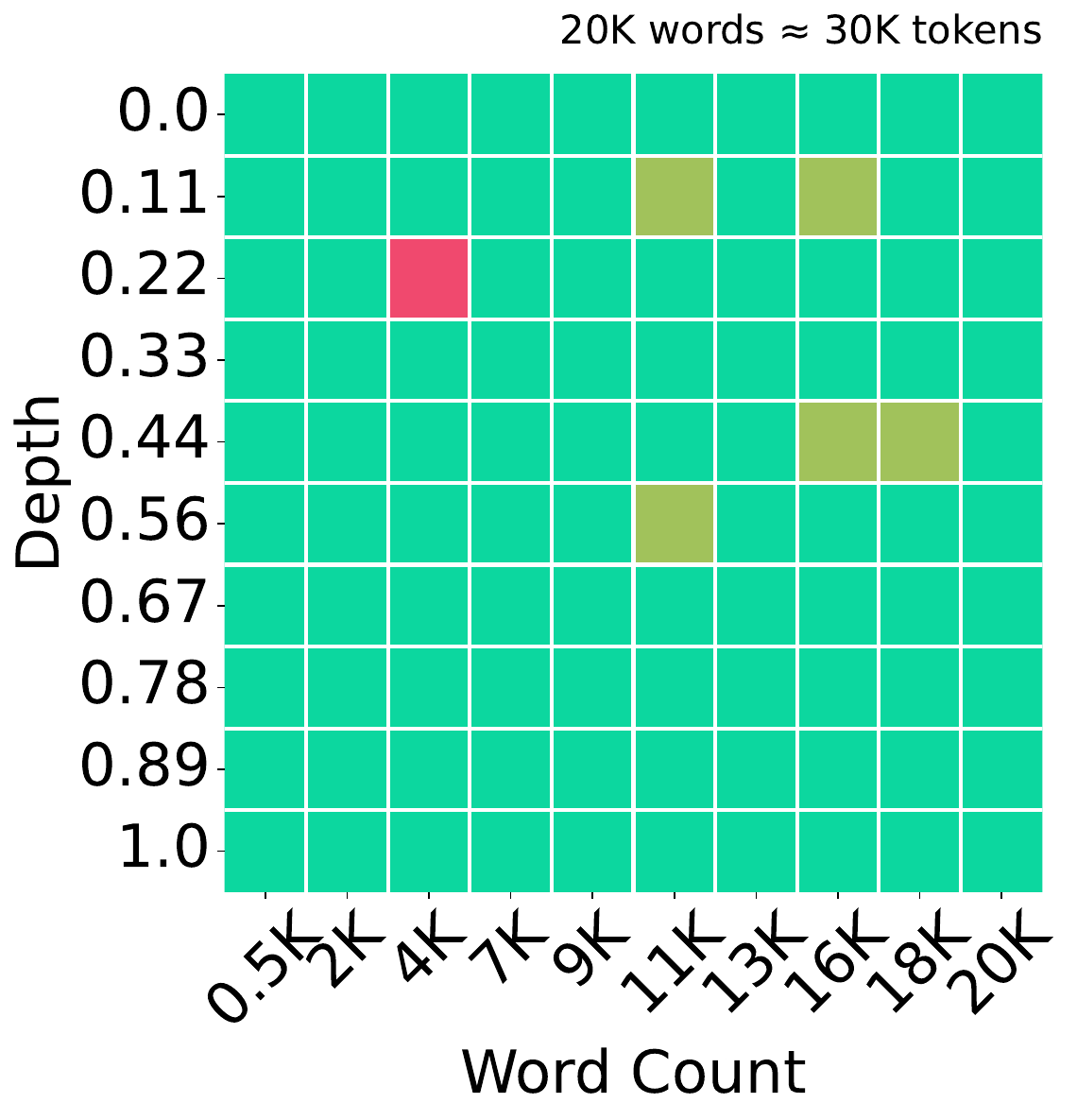}
	\end{minipage}%
}
\subfigure[4x Compression]{
\centering
	\begin{minipage}[t]{0.23\linewidth}
		\includegraphics[width=\linewidth]{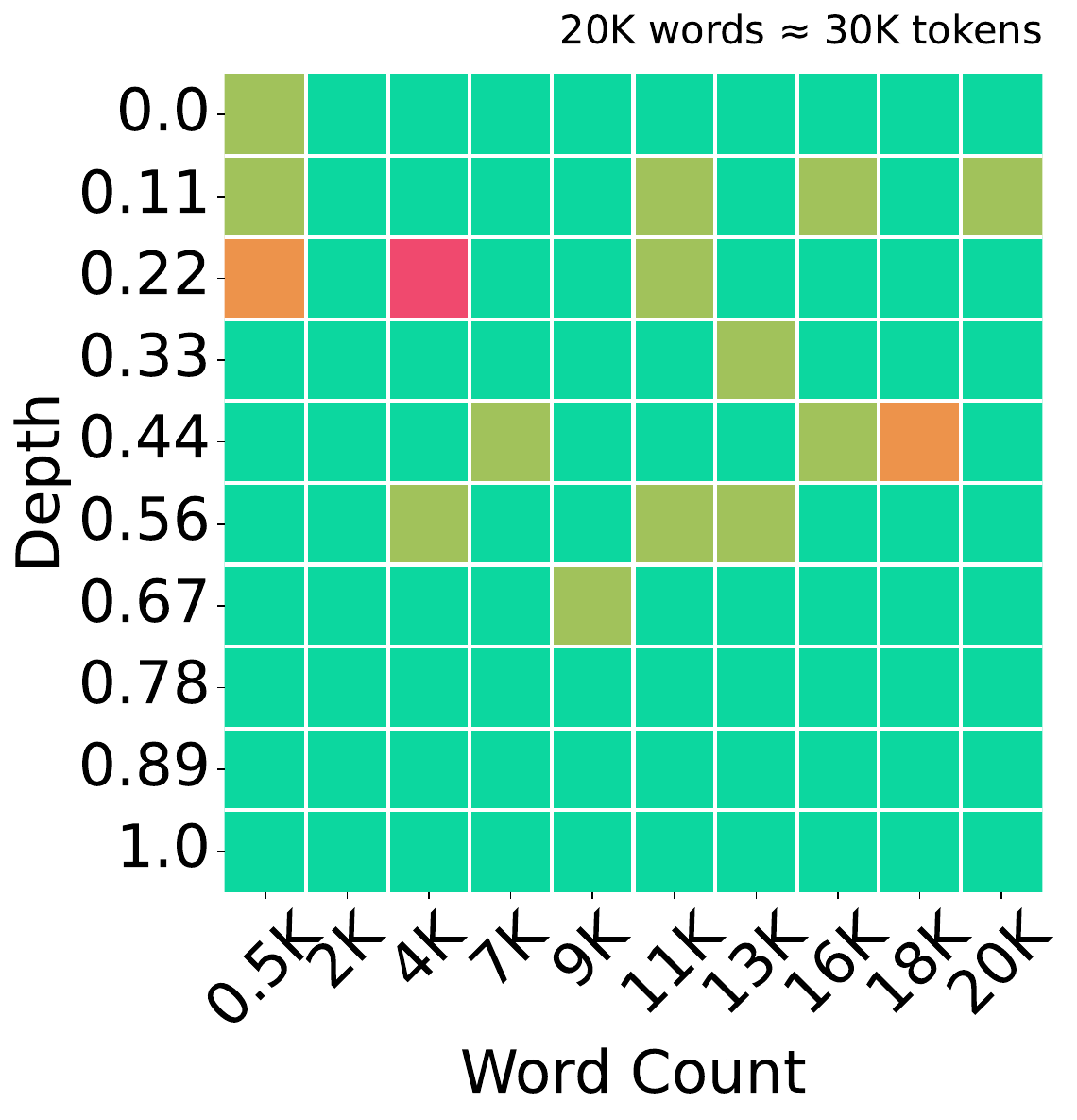}
	\end{minipage}%
}
\subfigure[6x Compression]{
\centering
	\begin{minipage}[t]{0.23\linewidth}
		\includegraphics[width=\linewidth]{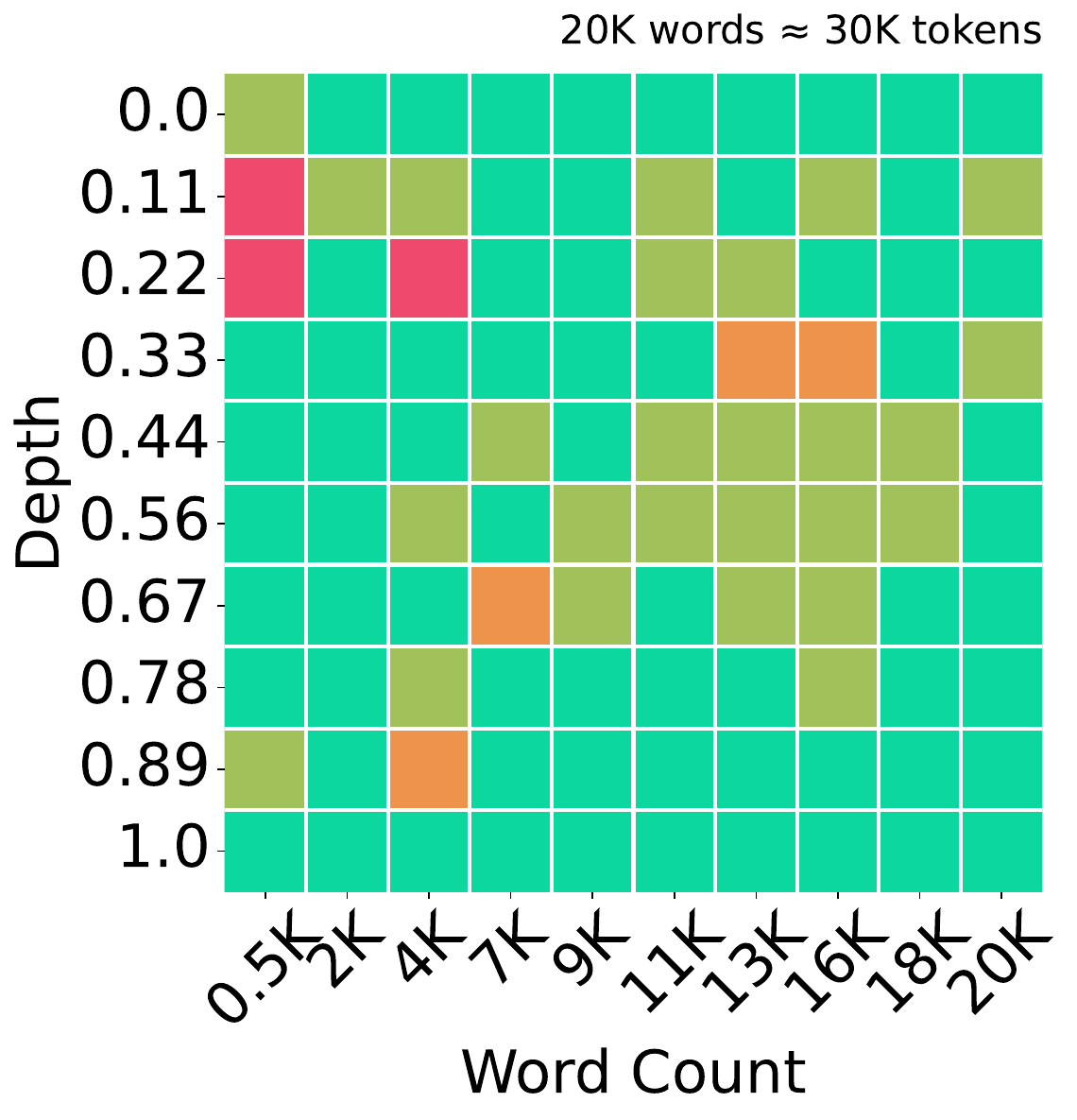}
	\end{minipage}%
}
\subfigure[8x Compression]{
\centering
	\begin{minipage}[t]{0.23\linewidth}
		\includegraphics[width=\linewidth]{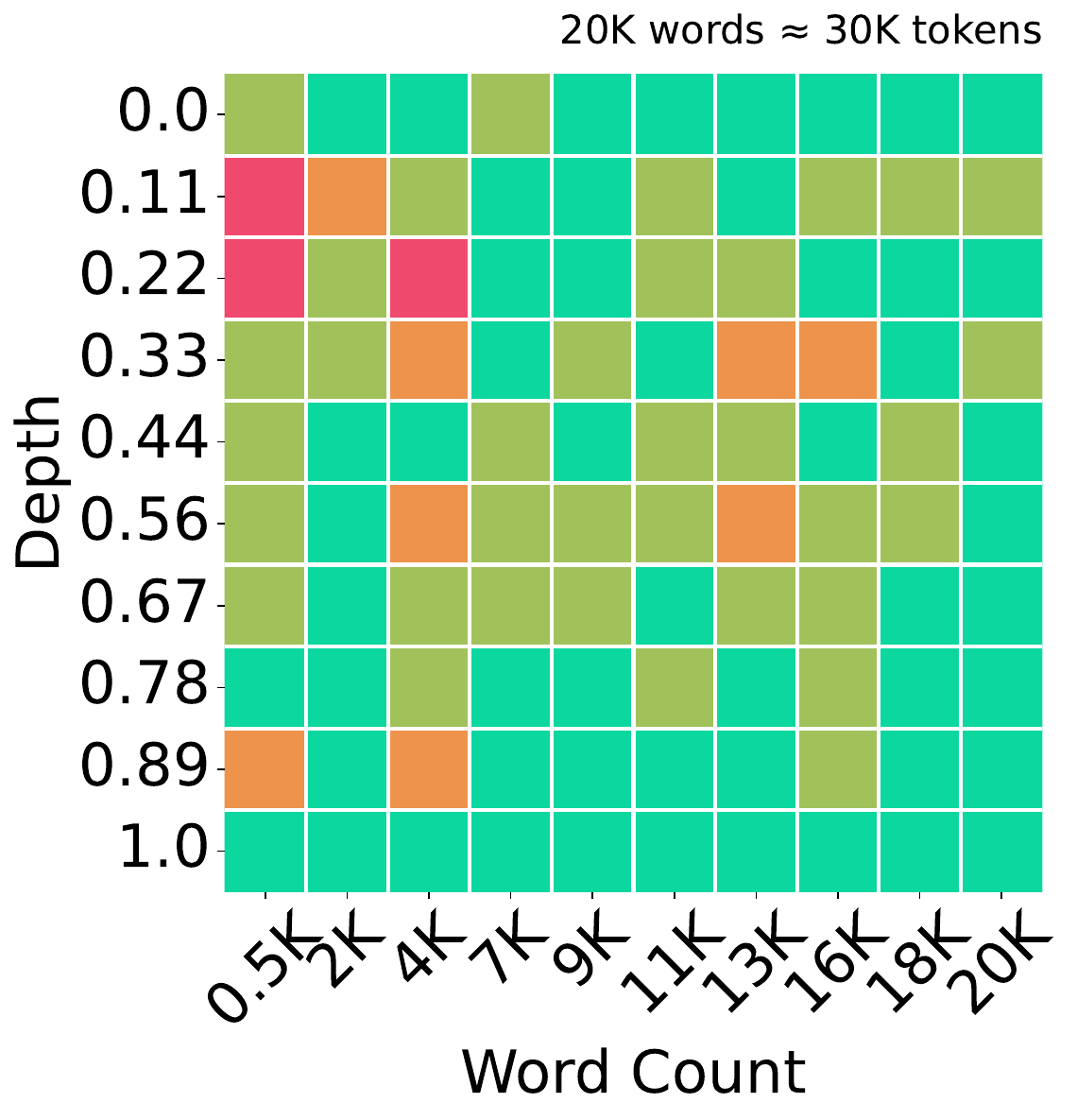}
	\end{minipage}%
}
\caption{$\mathrm{H_2O}$ on Mistral-7B-v0.2-Instruct with 4 different compression rates under needle test}
\label{fig:needle_h2o_mistral}
\end{figure*}

Table~\ref{tab:main_table_longchat} shows all the LongChat-7B results on LongBench and the needle experiment. We present FlexGen \citep{sheng2023flexgen} results on three different LLMs in Figure \ref{fig:needle_flexgen}. Additional $\mathrm{H_2O}$ \citep{h2o} results for different compression ratios on Llama-3-8B, LongChat-7B-v1.5, and Mistral-7B-v0.2 can be found in Figure~\ref{fig:needle_h2o_llama},~\ref{fig:needle_h2o_longchat} and~\ref{fig:needle_h2o_mistral} respectively. 

\begin{figure*}[h]
\setlength{\abovecaptionskip}{0mm}
\setlength{\belowcaptionskip}{0mm}
\centering
\subfigcapskip=-2mm
\subfigure[Llama-3-8B-Instruct]{
\centering
	\begin{minipage}[t]{0.32\linewidth}
		\includegraphics[width=\linewidth]{figures/needle/icl_raw_extend/llama.pdf}
	\end{minipage}%
}
\subfigure[Mistral-7B-v0.2-Instruct]{
\centering
	\begin{minipage}[t]{0.32\linewidth}
		\includegraphics[width=\linewidth]{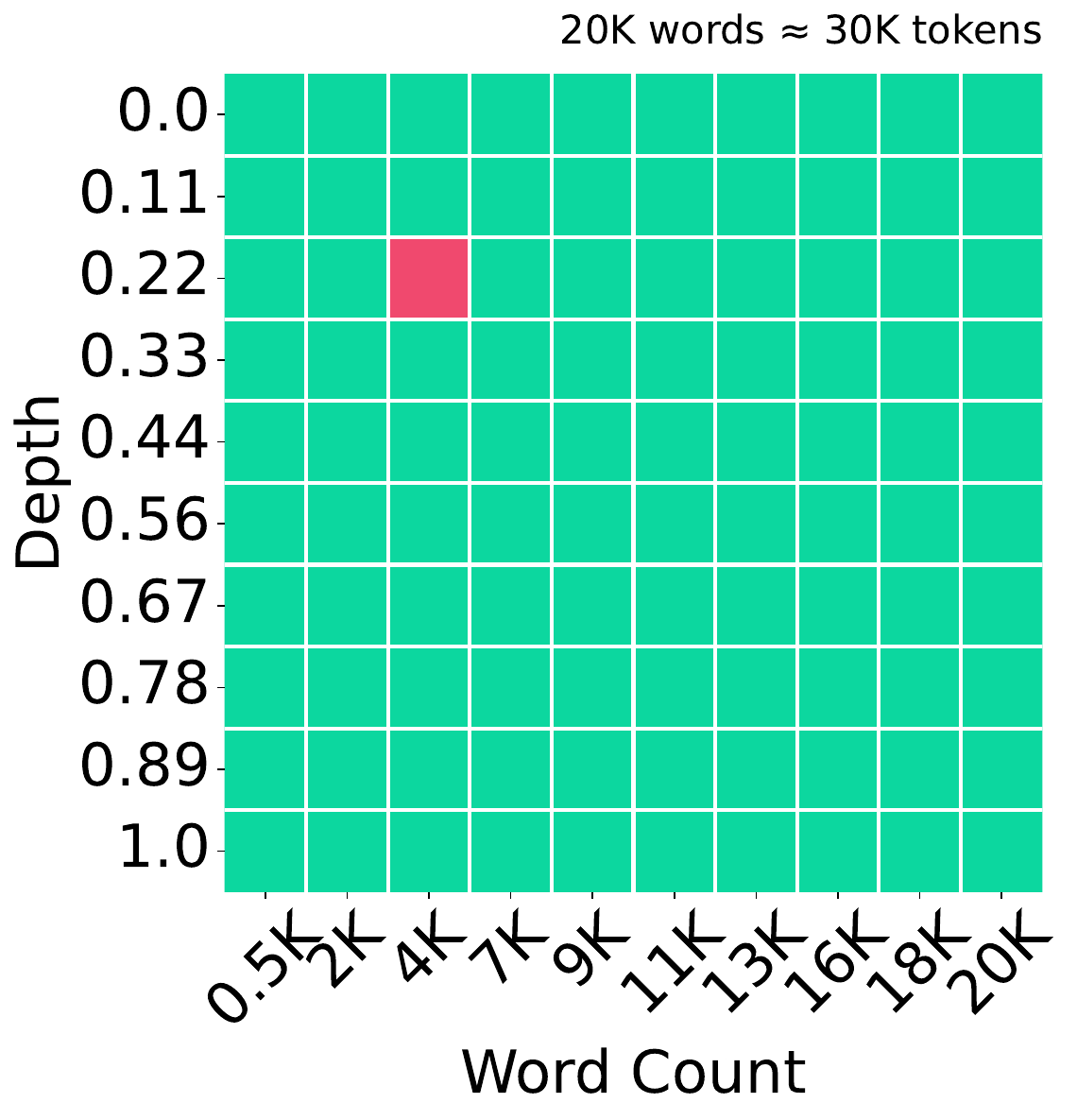}
	\end{minipage}%
}
\subfigure[LongChat-7B-v1.5-32k]{
\centering
	\begin{minipage}[t]{0.32\linewidth}
		\includegraphics[width=\linewidth]{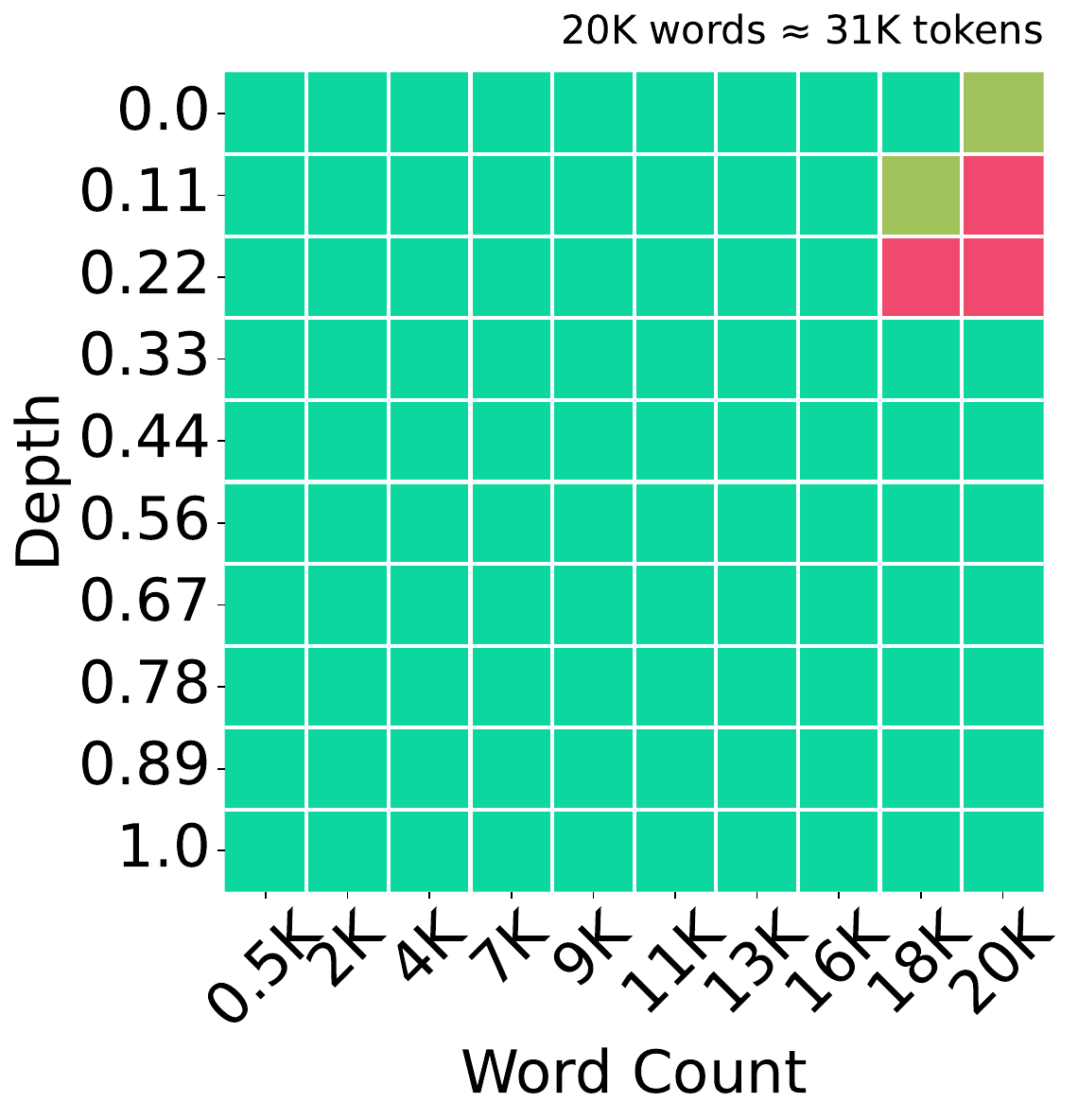}
	\end{minipage}%
}
\caption{Baseline performance under needle test on three commonly used LLMs}
\label{fig:needle_icl_raw_extend}
\end{figure*}

\begin{figure*}[h]
\setlength{\abovecaptionskip}{0mm}
\setlength{\belowcaptionskip}{0mm}
\centering
\subfigcapskip=-2mm
\subfigure[2x Compression]{
\centering
	\begin{minipage}[t]{0.23\linewidth}
		\includegraphics[width=\linewidth]{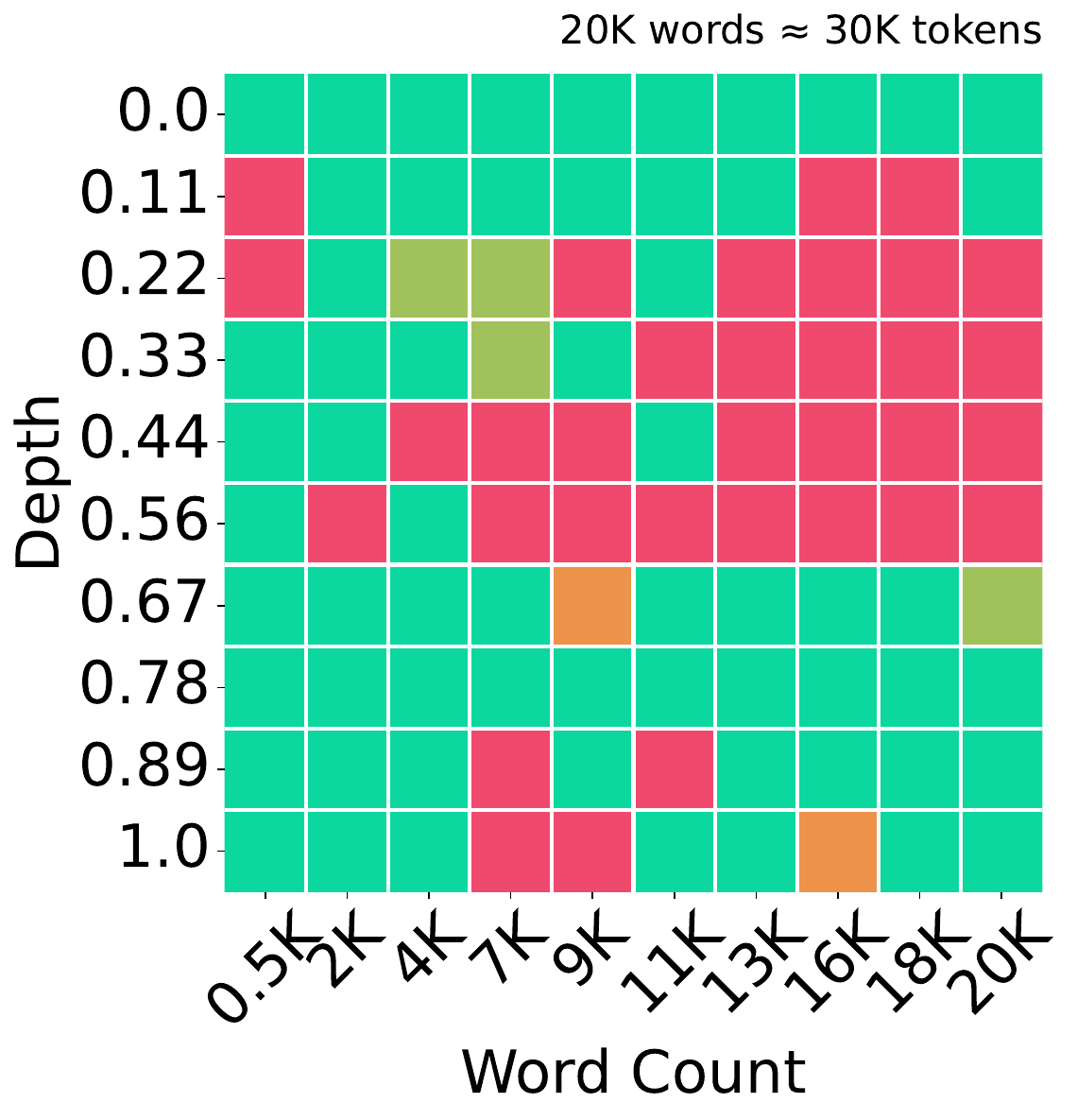}
	\end{minipage}%
}
\subfigure[4x Compression]{
\centering
	\begin{minipage}[t]{0.23\linewidth}
		\includegraphics[width=\linewidth]{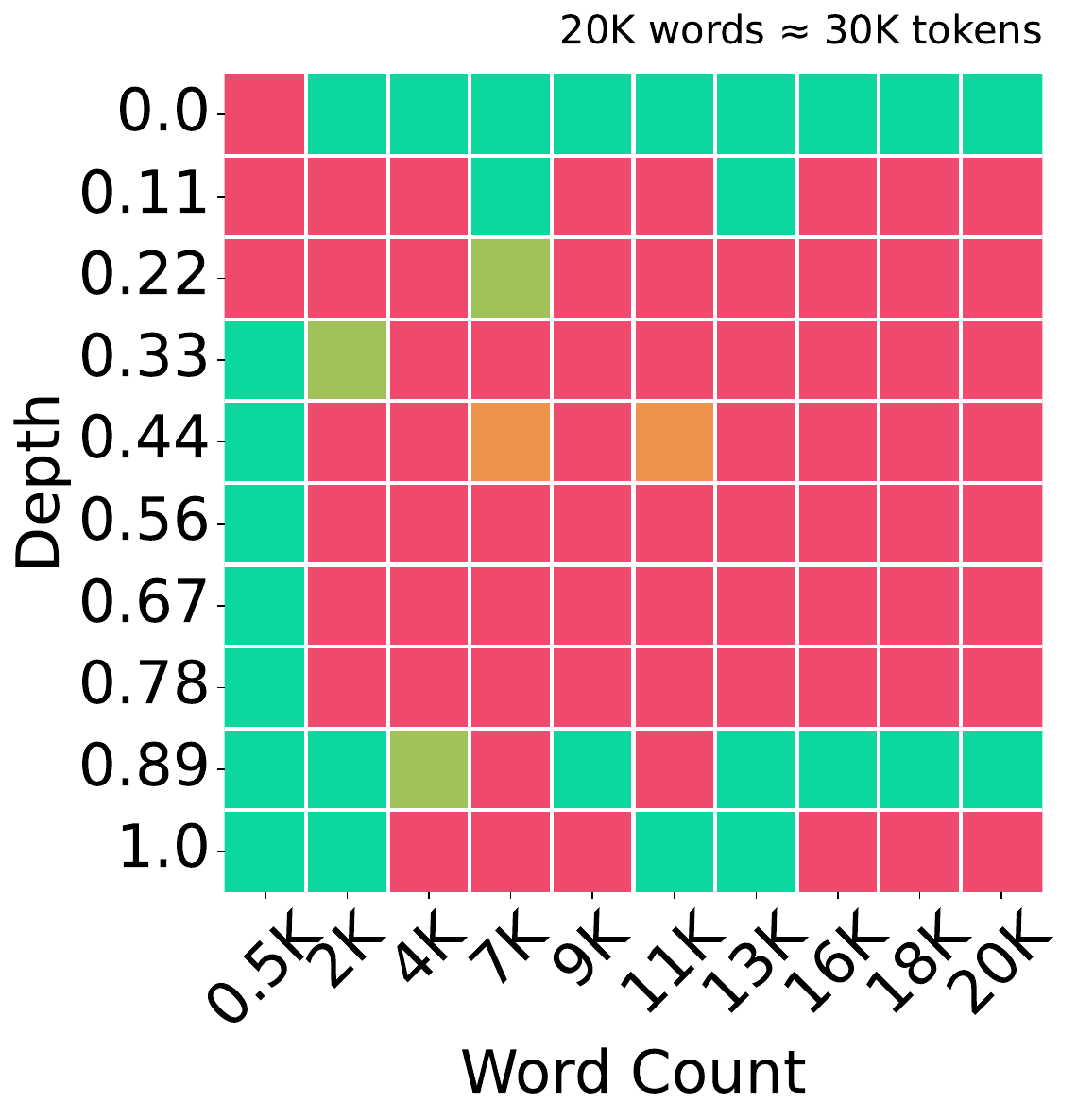}
	\end{minipage}%
}
\subfigure[6x Compression]{
\centering
	\begin{minipage}[t]{0.23\linewidth}
		\includegraphics[width=\linewidth]{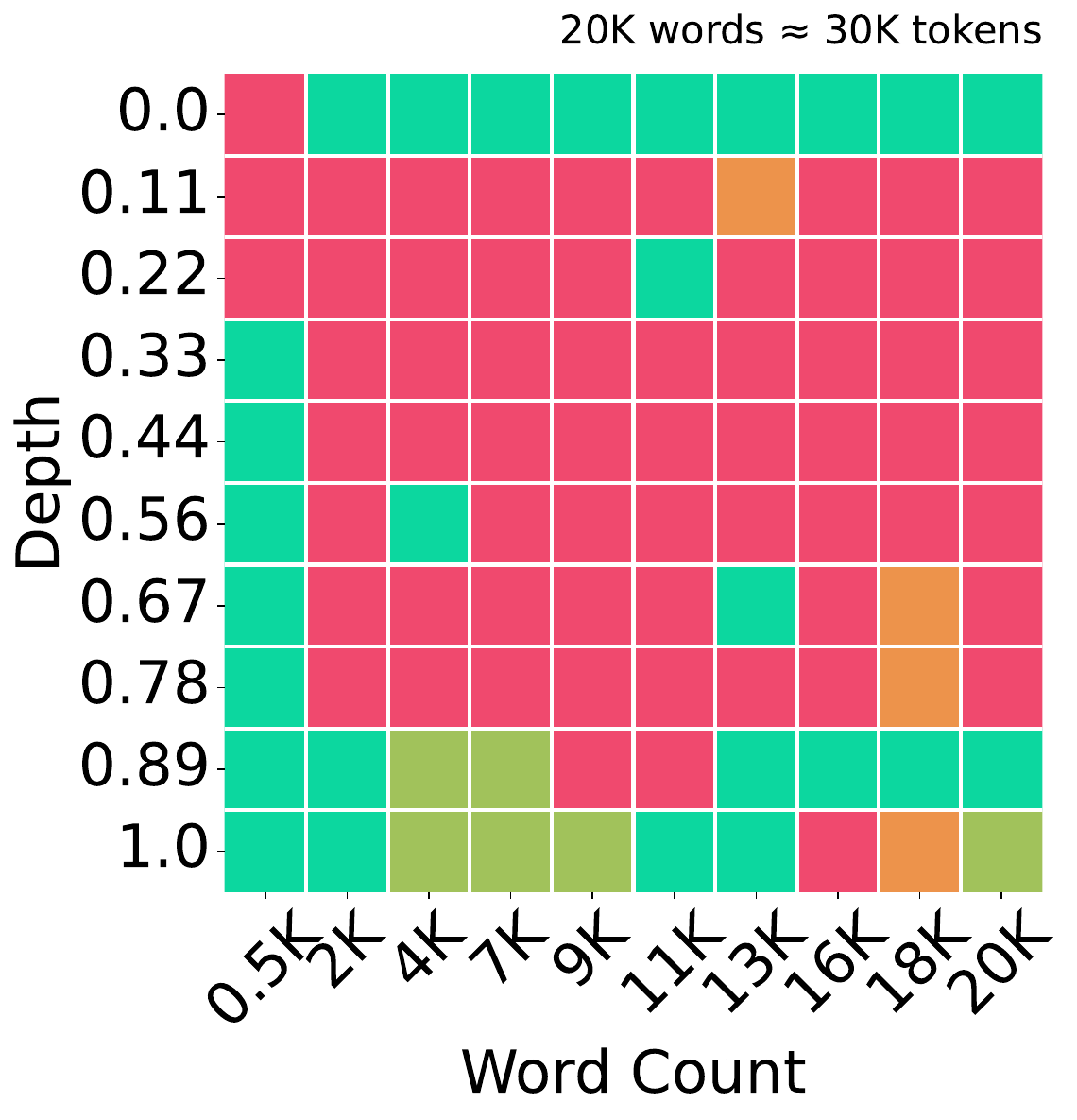}
	\end{minipage}%
}
\subfigure[8x Compression]{
\centering
	\begin{minipage}[t]{0.23\linewidth}
		\includegraphics[width=\linewidth]{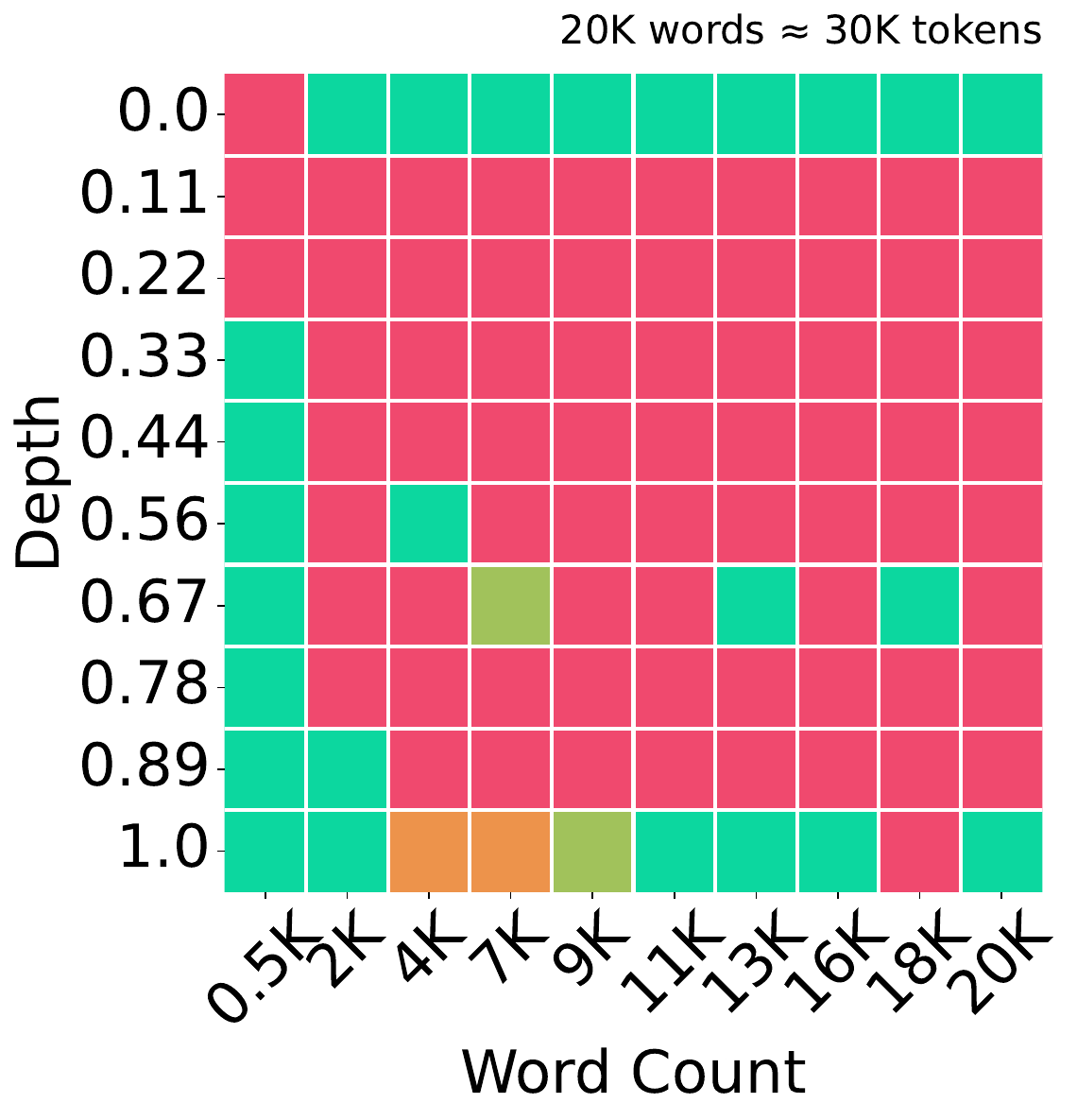}
	\end{minipage}%
}
\caption{InfLLM on Mistral-7B-v0.2-Instruct with 4 different compression rates under needle test}
\label{fig:needle_infllm_mistral}
\end{figure*}

\begin{figure*}[h]
\setlength{\abovecaptionskip}{0mm}
\setlength{\belowcaptionskip}{0mm}
\centering
\subfigcapskip=-4mm
\!\!\!\!\!\!
\subfigure[]{
\centering
	\begin{minipage}[t]{0.326\linewidth}
		\includegraphics[width=0.99\linewidth]{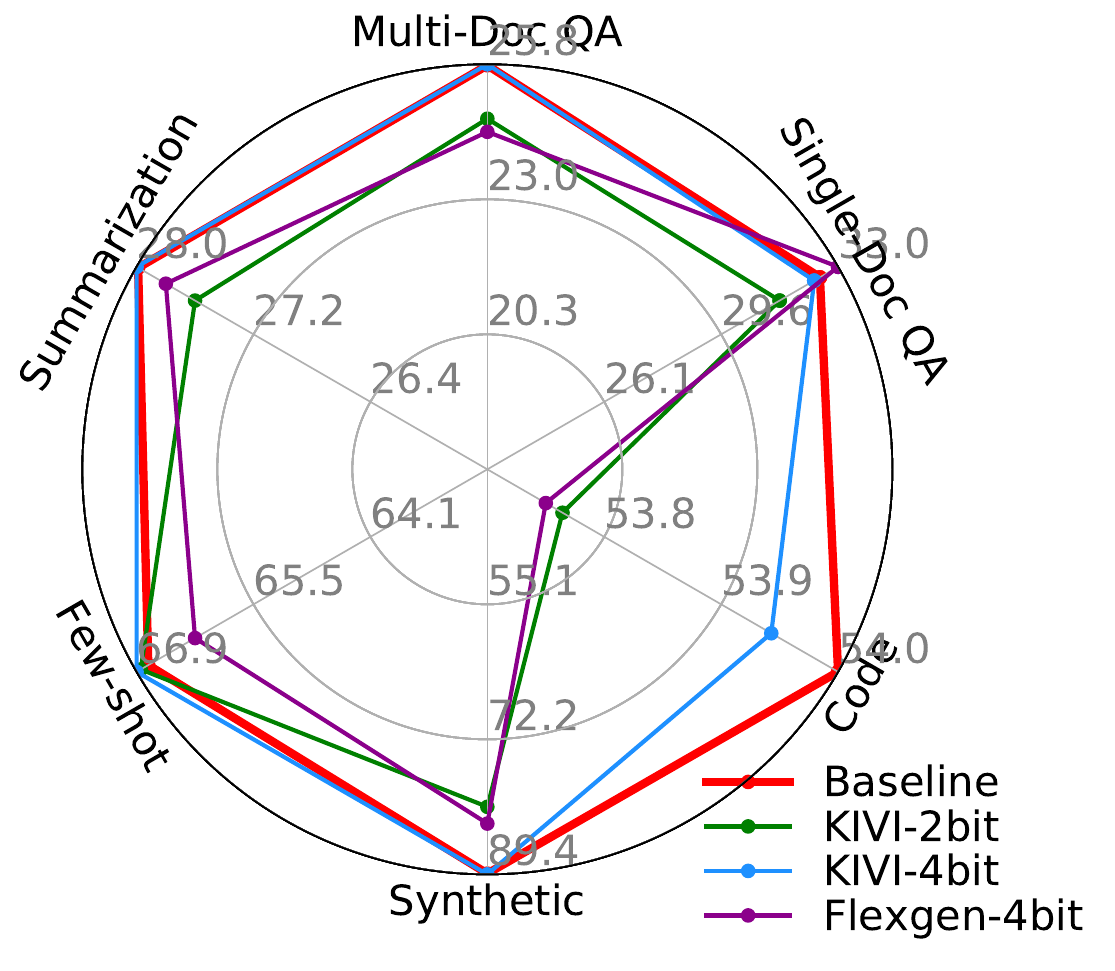}
	\end{minipage}%
}
\subfigure[]{
\centering
	\begin{minipage}[t]{0.336\linewidth}
		\includegraphics[width=0.99\linewidth]{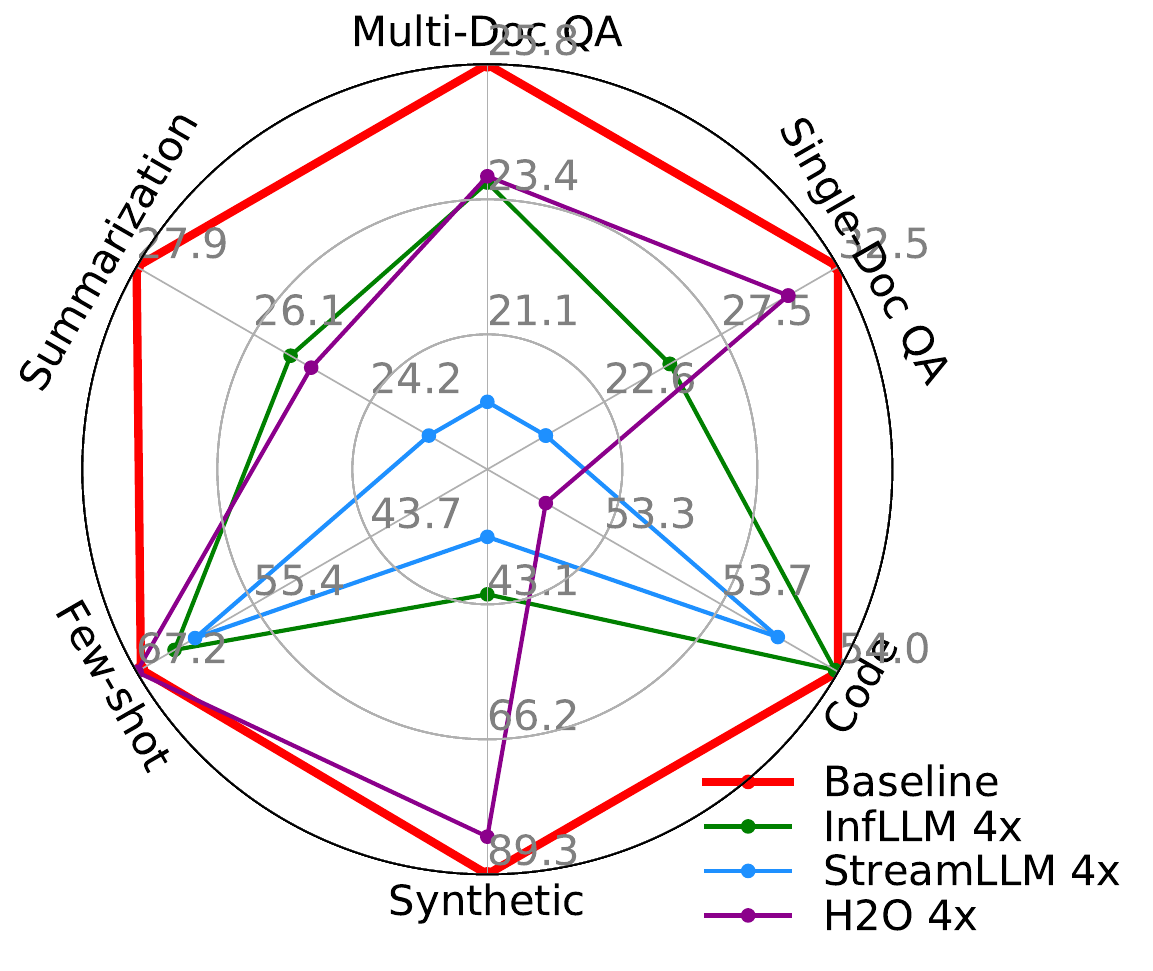}
	\end{minipage}%
}
\!\!\!\!
\subfigure[]{
\centering
	\begin{minipage}[t]{0.336\linewidth}
		\includegraphics[width=0.99\linewidth]{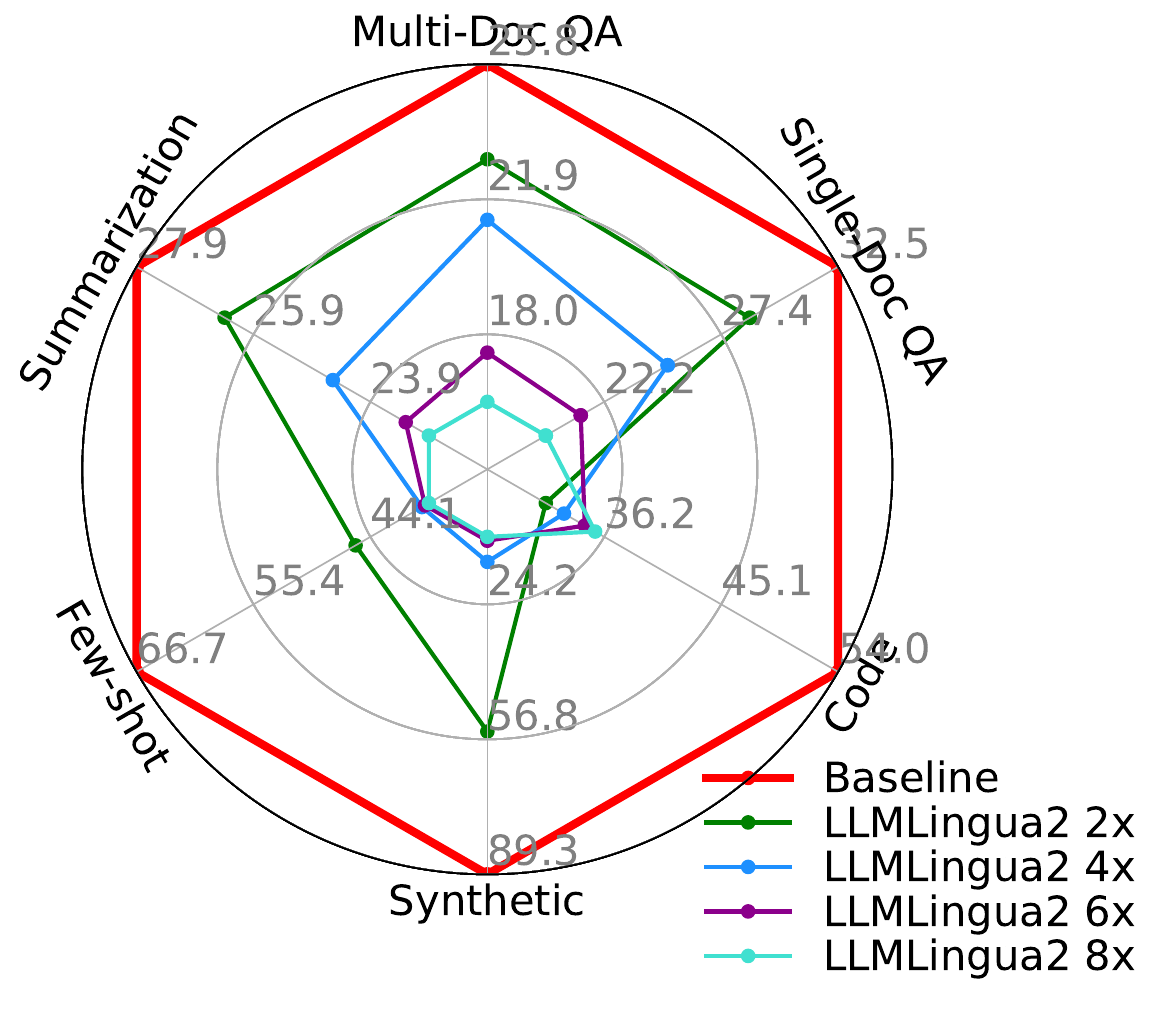}
	\end{minipage}%
}
\caption{Mistral-7B-v0.2-Instruct with different compression methods (a) with Quantization; (b) with Token Dropping (c) with prompt compression.}
\label{fig:radar-mistral}
\end{figure*}

\begin{figure*}[h]
\setlength{\abovecaptionskip}{0mm}
\setlength{\belowcaptionskip}{0mm}
\centering
\subfigcapskip=-4mm
\!\!\!\!\!\!
\subfigure[]{
\centering
	\begin{minipage}[t]{0.326\linewidth}
		\includegraphics[width=0.99\linewidth]{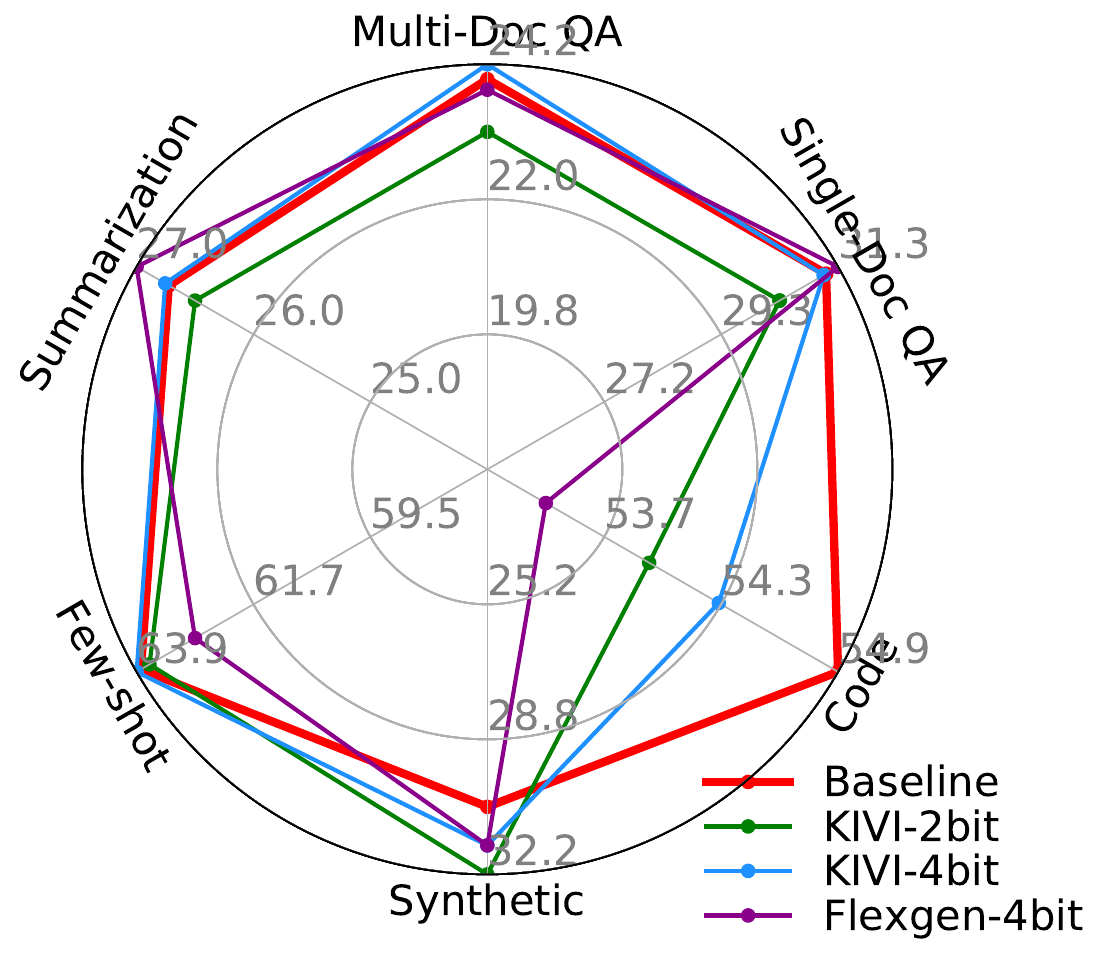}
	\end{minipage}%
}
\subfigure[]{
\centering
	\begin{minipage}[t]{0.336\linewidth}
		\includegraphics[width=0.99\linewidth]{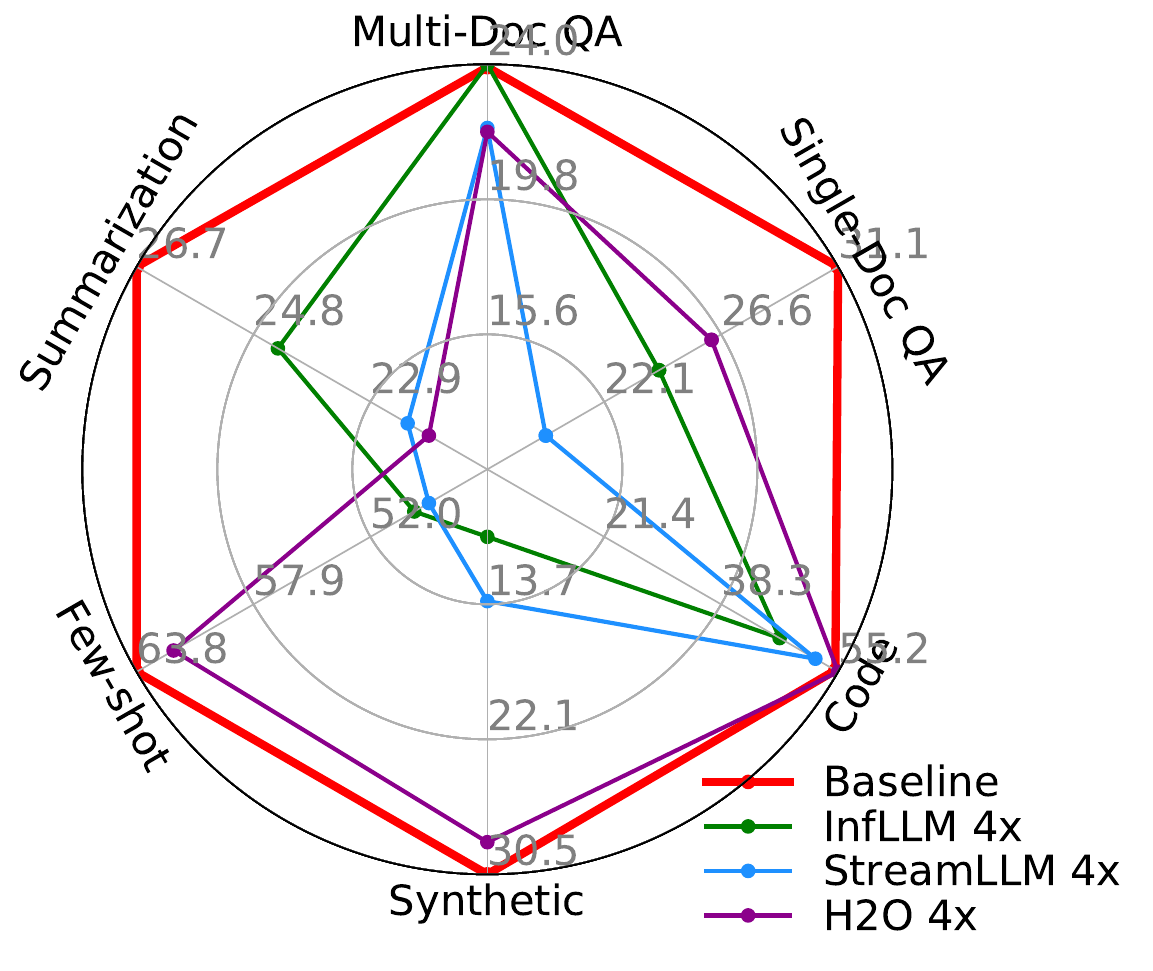}
	\end{minipage}%
}
\!\!\!\!
\subfigure[]{
\centering
	\begin{minipage}[t]{0.336\linewidth}
		\includegraphics[width=0.99\linewidth]{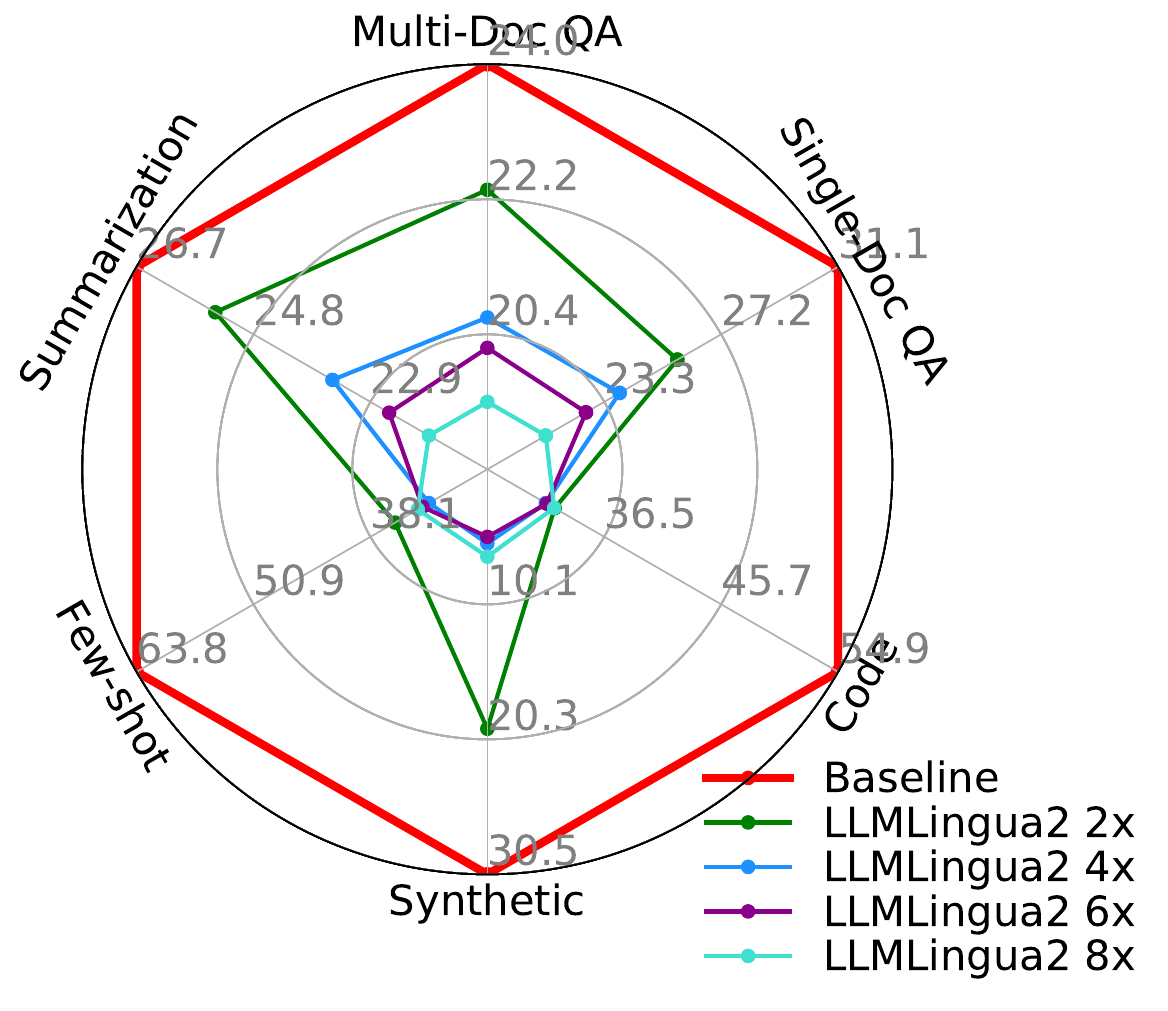}
	\end{minipage}%
}
\caption{LongChat-7B-v1.5-32K with different compression methods (a) with Quantization; (b) with Token Dropping (c) with prompt compression.}
\label{fig:radar-LongChat}
\end{figure*}

\begin{figure*}[h]
\setlength{\abovecaptionskip}{0mm}
\setlength{\belowcaptionskip}{0mm}
\centering
\subfigcapskip=-4mm
\!\!\!\!\!\!
\subfigure[Llama-3-8B-Instruct]{
\centering
	\begin{minipage}[t]{0.333\linewidth}
		\includegraphics[width=0.99\linewidth]{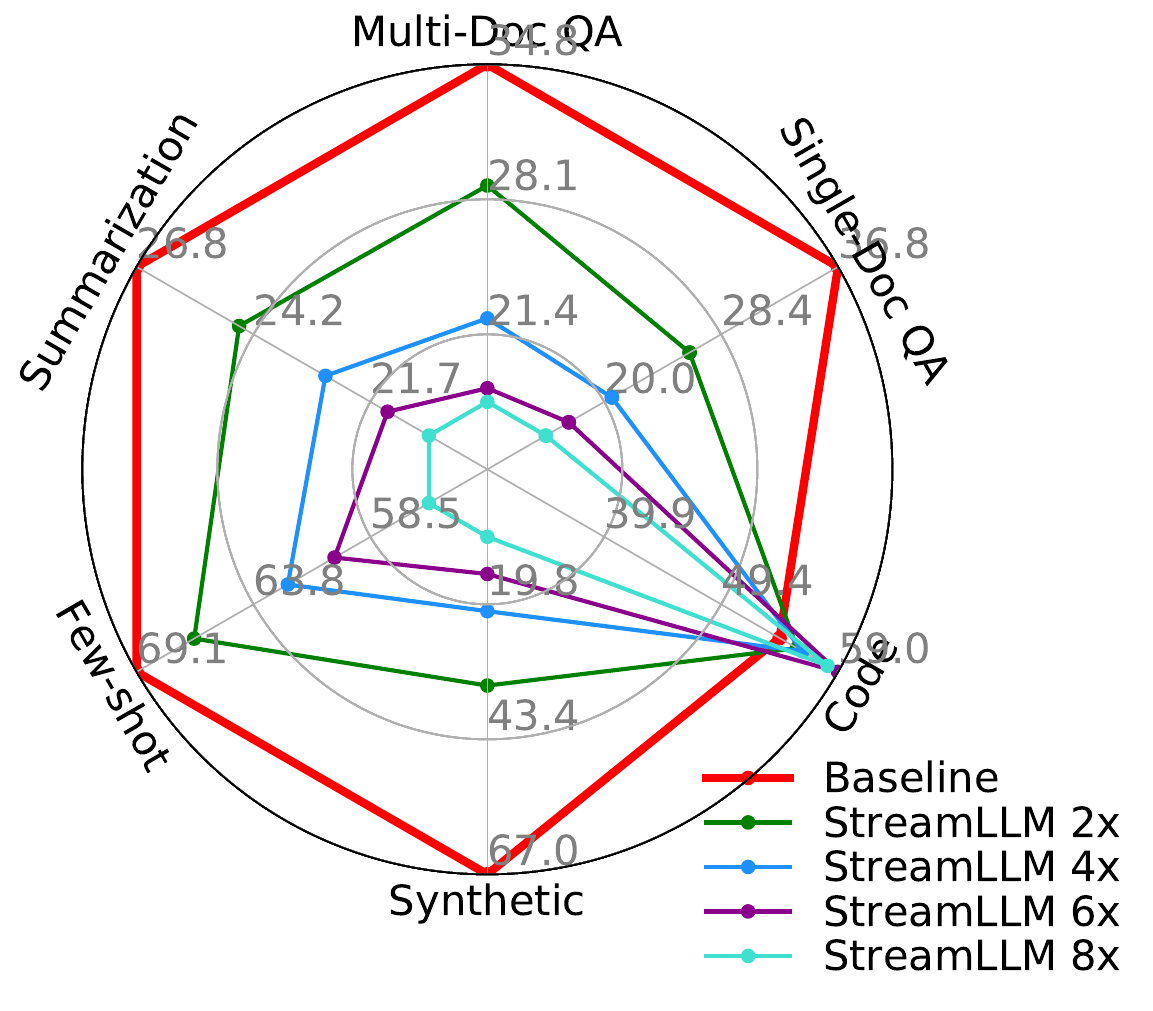}
	\end{minipage}%
}
\subfigure[Mistral-7B-v0.2-Instruct]{
\centering
	\begin{minipage}[t]{0.333\linewidth}
		\includegraphics[width=0.99\linewidth]{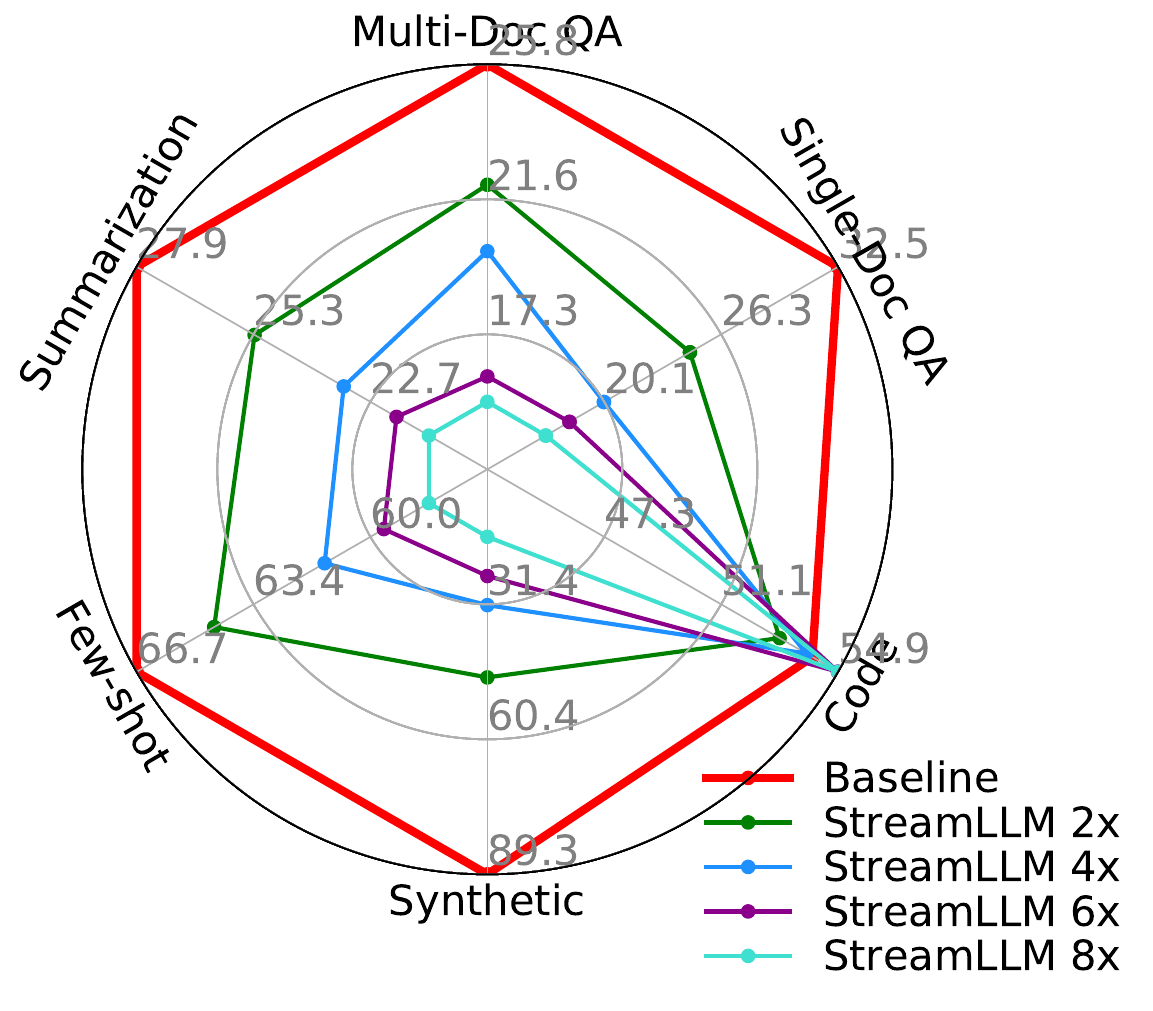}
	\end{minipage}%
}
\!\!\!\!
\subfigure[LongChat-7B-v1.5-32K]{
\centering
	\begin{minipage}[t]{0.333\linewidth}
		\includegraphics[width=0.99\linewidth]{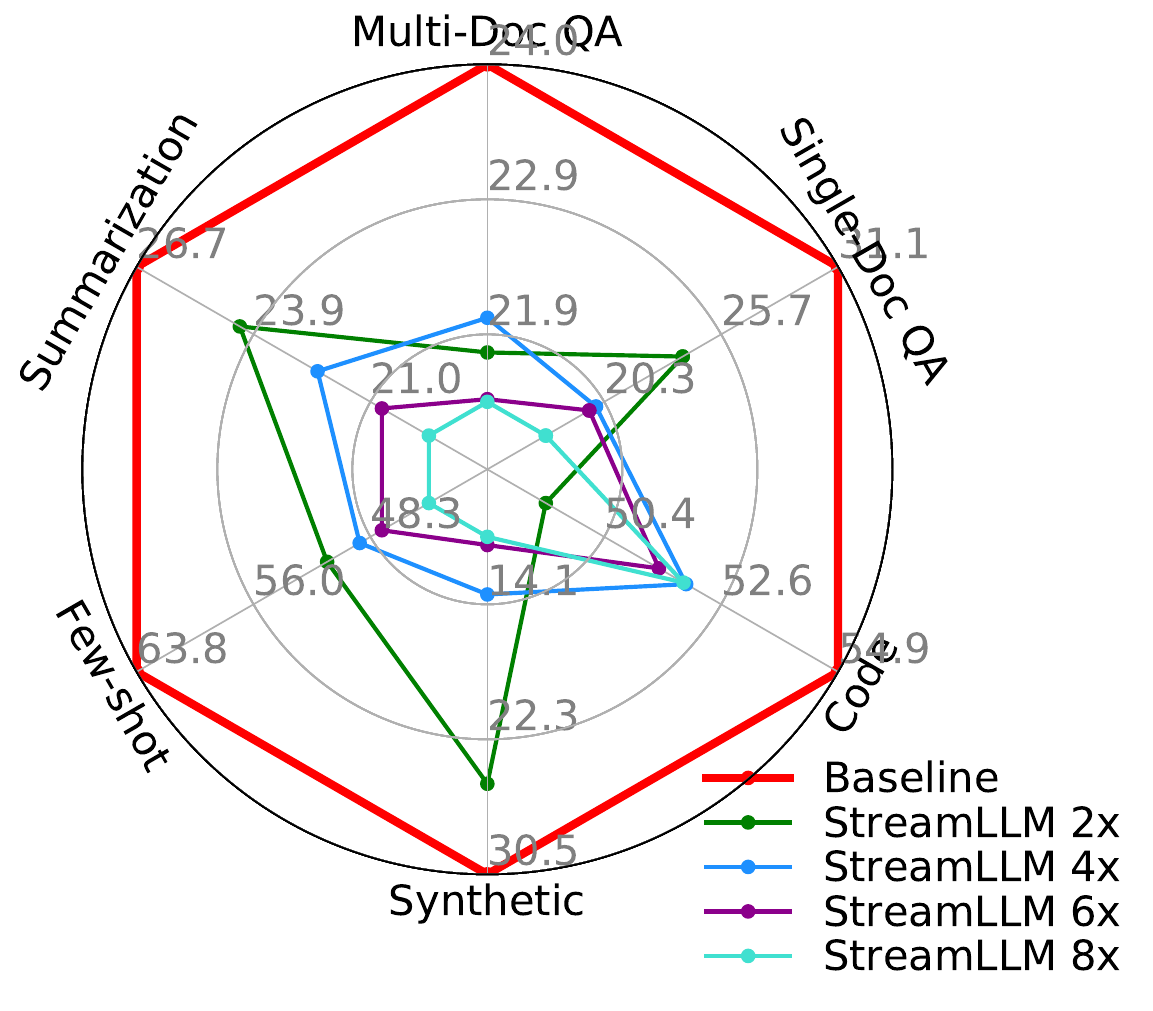}
	\end{minipage}%
}
\caption{StreamingLLM with different compression ratios on three commonly used LLMs.}
\label{fig:radar-streamllm}
\end{figure*}

\begin{figure*}[h]
\setlength{\abovecaptionskip}{0mm}
\setlength{\belowcaptionskip}{0mm}
\centering
\subfigcapskip=-4mm
\!\!\!\!\!\!
\subfigure[Llama-3-8B-Instruct]{
\centering
	\begin{minipage}[t]{0.333\linewidth}
		\includegraphics[width=0.99\linewidth]{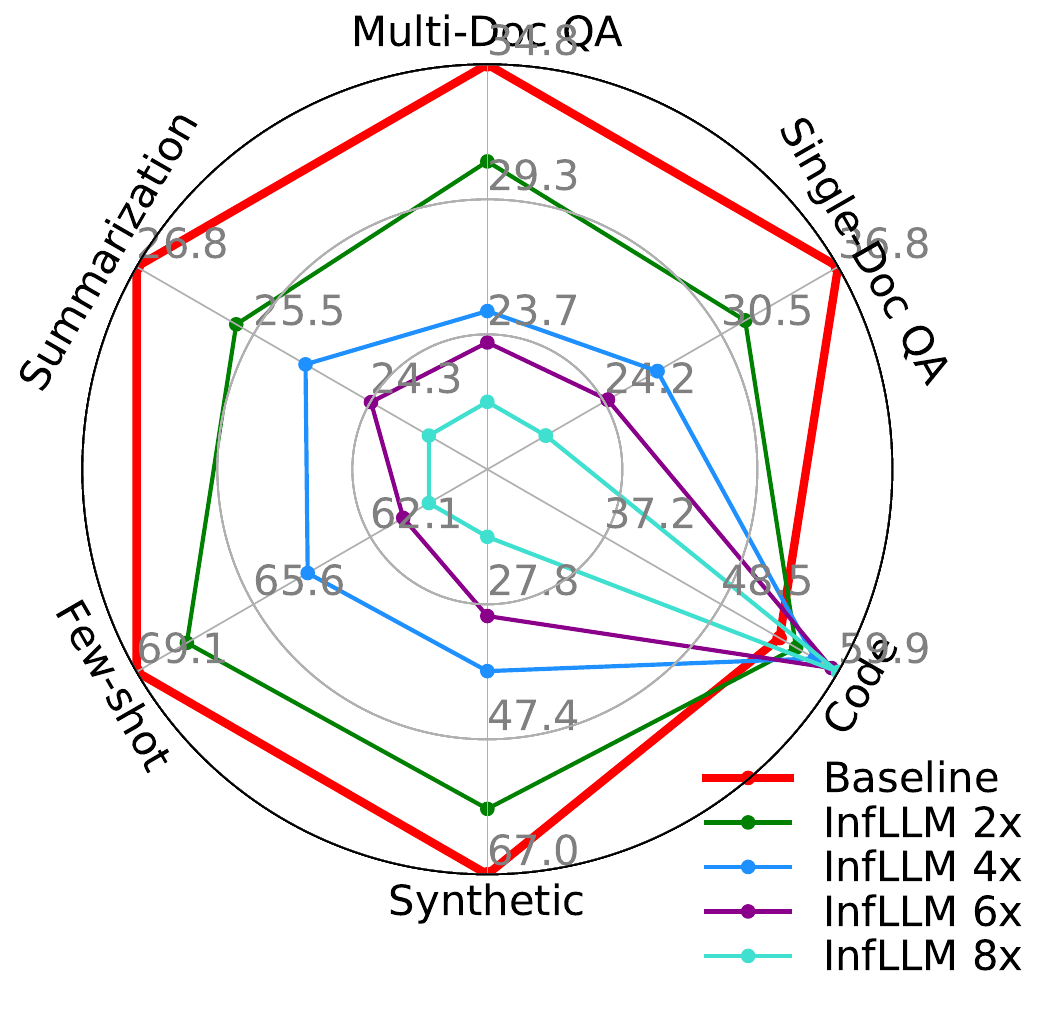}
	\end{minipage}%
}
\subfigure[Mistral-7B-v0.2-Instruct]{
\centering
	\begin{minipage}[t]{0.333\linewidth}
		\includegraphics[width=0.99\linewidth]{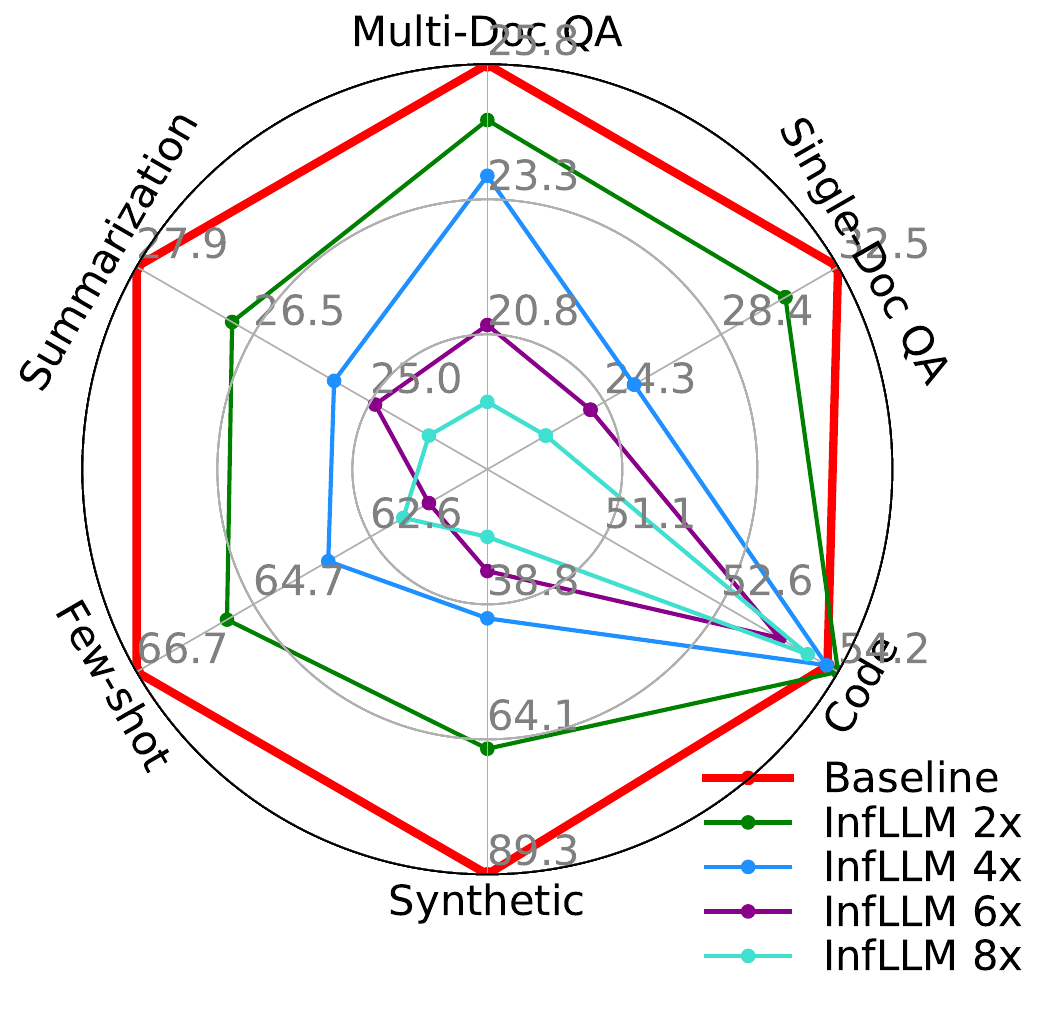}
	\end{minipage}%
}
\!\!\!\!
\subfigure[LongChat-7B-v1.5-32K]{
\centering
	\begin{minipage}[t]{0.333\linewidth}
		\includegraphics[width=0.99\linewidth]{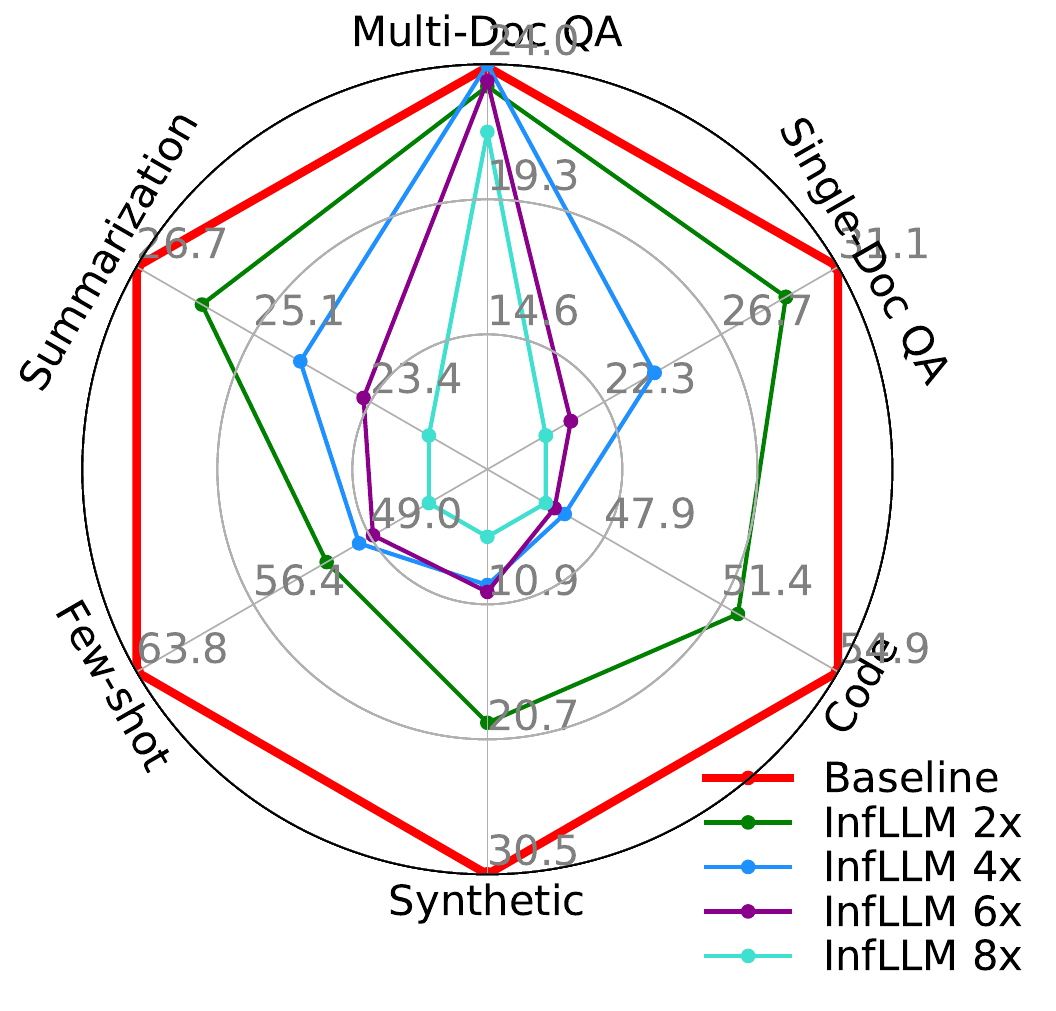}
	\end{minipage}%
}
\caption{InfLLM with different compression ratios on three commonly used LLMs.}
\label{fig:radar-infllm}
\end{figure*}

\begin{figure*}[h]
\setlength{\abovecaptionskip}{0mm}
\setlength{\belowcaptionskip}{0mm}
\centering
\!\!\!\!\!\!
\subfigure[Llama-3-8B-Instruct]{
\centering
	\begin{minipage}[t]{0.333\linewidth}
		\includegraphics[width=0.99\linewidth]{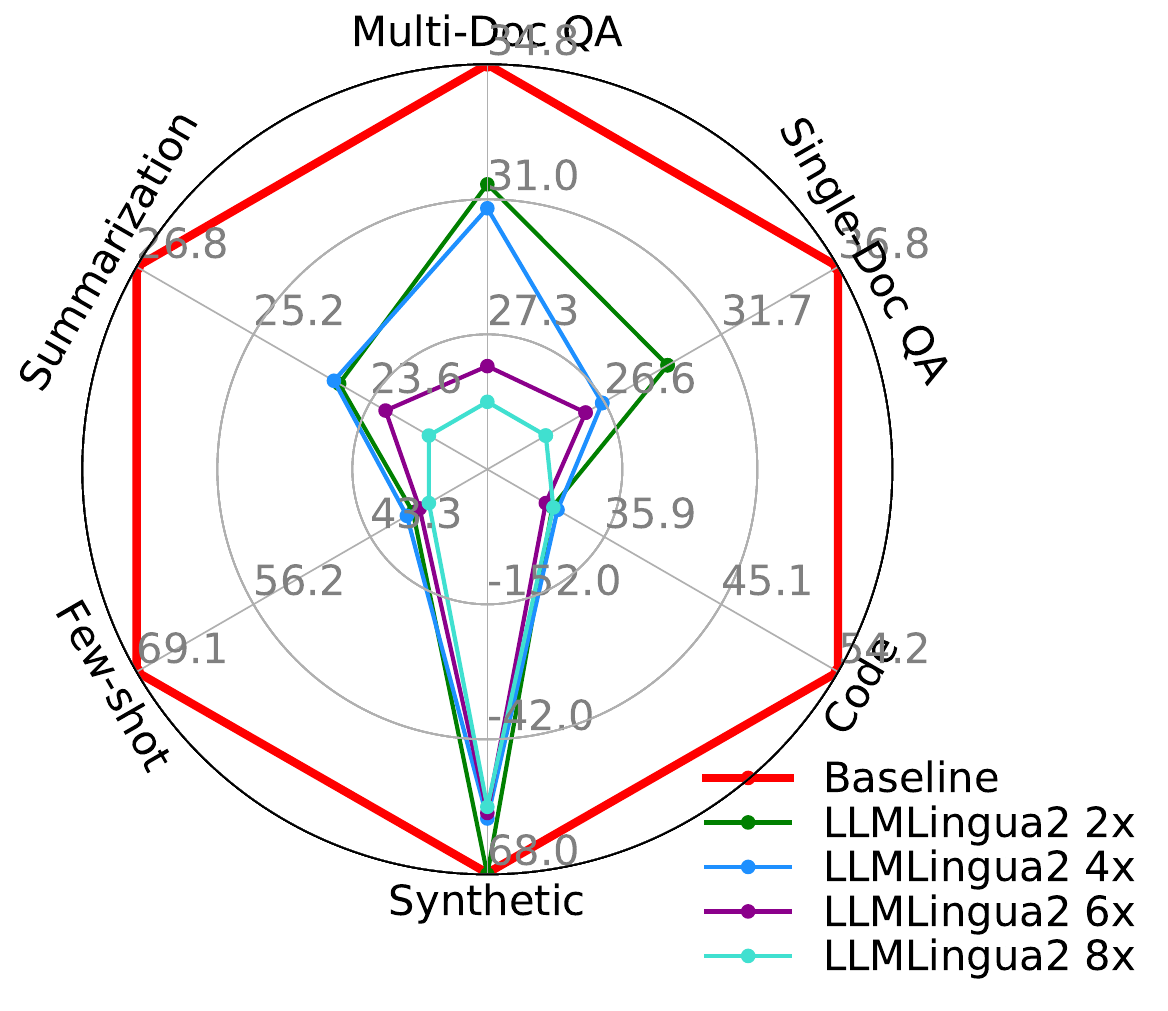}
	\end{minipage}%
}
\subfigure[Mistral-7B-v0.2-Instruct]{
\centering
	\begin{minipage}[t]{0.333\linewidth}
		\includegraphics[width=0.99\linewidth]{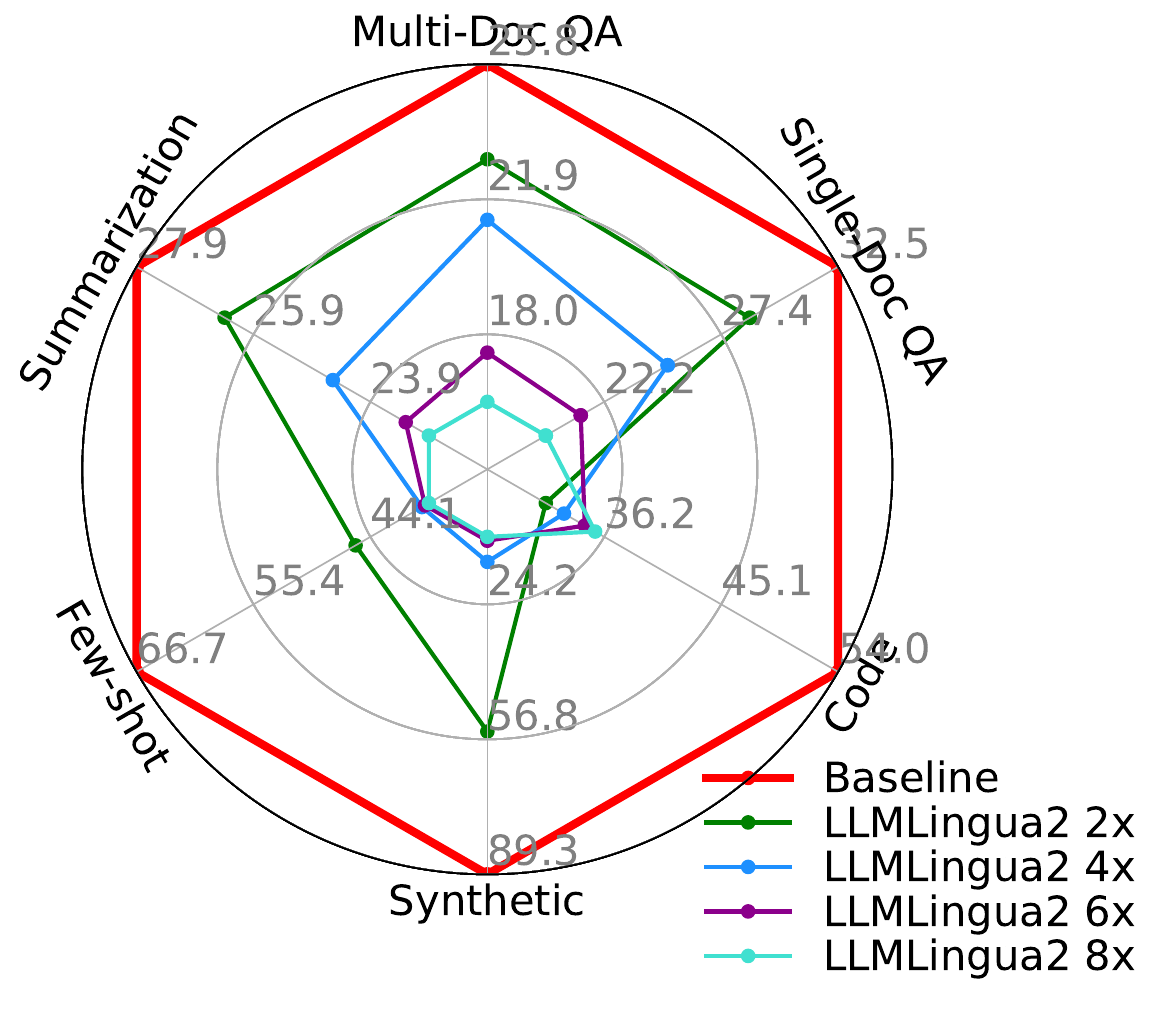}
	\end{minipage}%
}
\!\!\!\!
\subfigure[LongChat-7B-v1.5-32K]{
\centering
	\begin{minipage}[t]{0.333\linewidth}
		\includegraphics[width=0.99\linewidth]{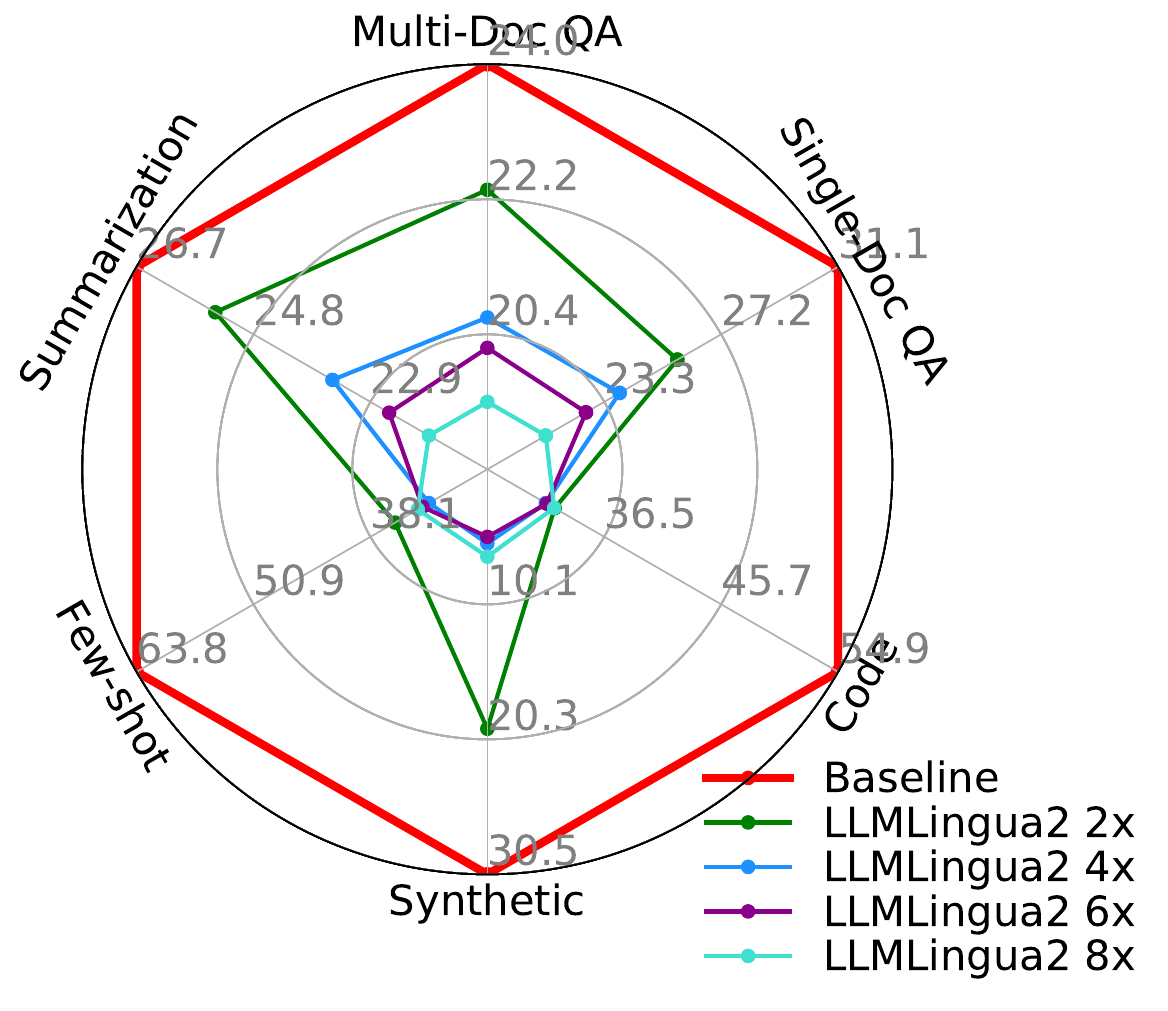}
	\end{minipage}%
}
\caption{LLMLingua with different compression ratios on three commonly used LLMs.}
\label{fig:radar-llmlingue}
\end{figure*}

We provide more visualization results on the needle task. For baseline performance for the three models in Figure~\ref{fig:needle_icl_raw_extend}. For InfLLM results on the LongChat and Mistral models, the results are listed in Figure~\ref{fig:needle_infllm_LongChat} and~\ref{fig:needle_infllm_mistral}. Figure~\ref{fig:radar-mistral} and~\ref{fig:radar-LongChat} show the performance of quantization, token dropping, and prompt compression on Mistral and LongChat, respectively. Figure~\ref{fig:radar-streamllm},~\ref{fig:radar-infllm} and~\ref{fig:radar-llmlingue} illustrates the effectiveness of different compression ratios across various subtasks in LongBench. 

Finally, Table~\ref{tab:appendix_llama},~\ref{tab:appendix_mistral},~\ref{tab:appendix_longchat} and~\ref{tab:appendix_rnn} show the detailed results for each task in LongBench. 

\begin{figure*}[h]
\setlength{\abovecaptionskip}{0mm}
\setlength{\belowcaptionskip}{0mm}
\centering
\subfigcapskip=-2mm
\subfigure[Llama-3-8B-Instruct (99.3\%)]{
\centering
	\begin{minipage}[t]{0.32\linewidth}
		\includegraphics[width=\linewidth]{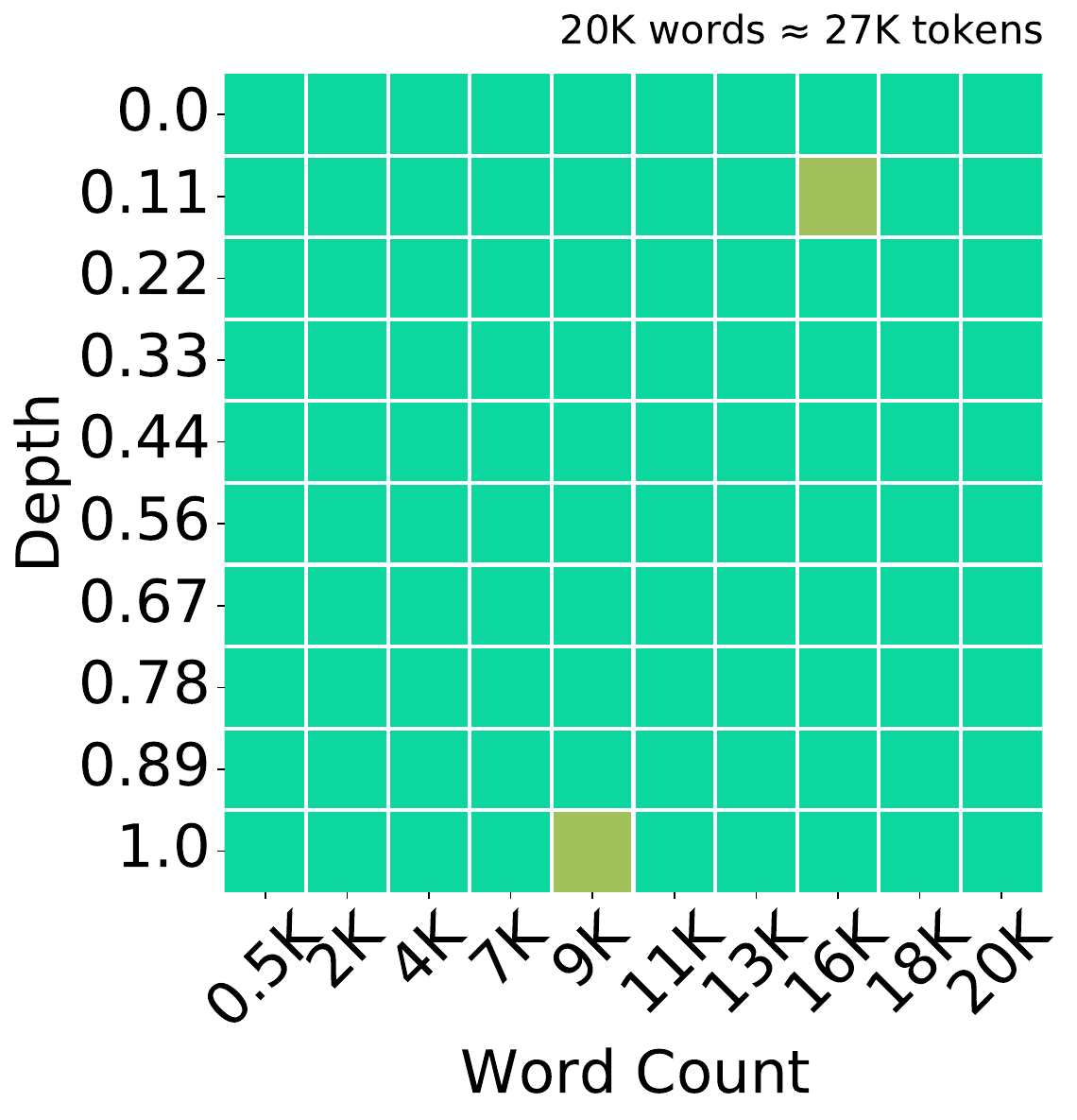}
	\end{minipage}%
}
\subfigure[Llama-3 + KIVI-4bit (99.3\%)]{
\centering
	\begin{minipage}[t]{0.32\linewidth}
		\includegraphics[width=\linewidth]{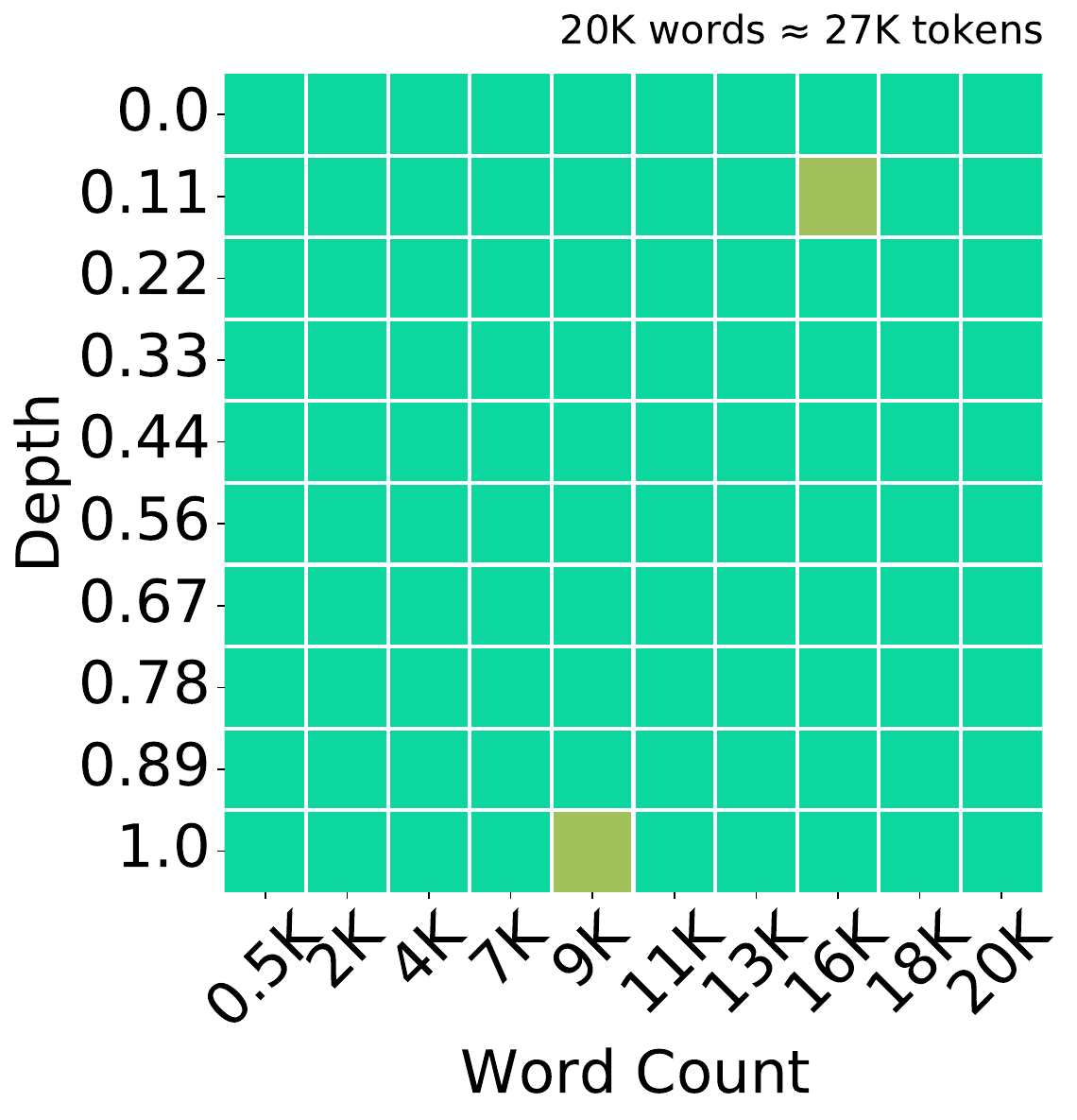}
	\end{minipage}%
}
\subfigure[Llama-3 + KIVI-2bit (91.0\%)]{
\centering
	\begin{minipage}[t]{0.32\linewidth}
		\includegraphics[width=\linewidth]{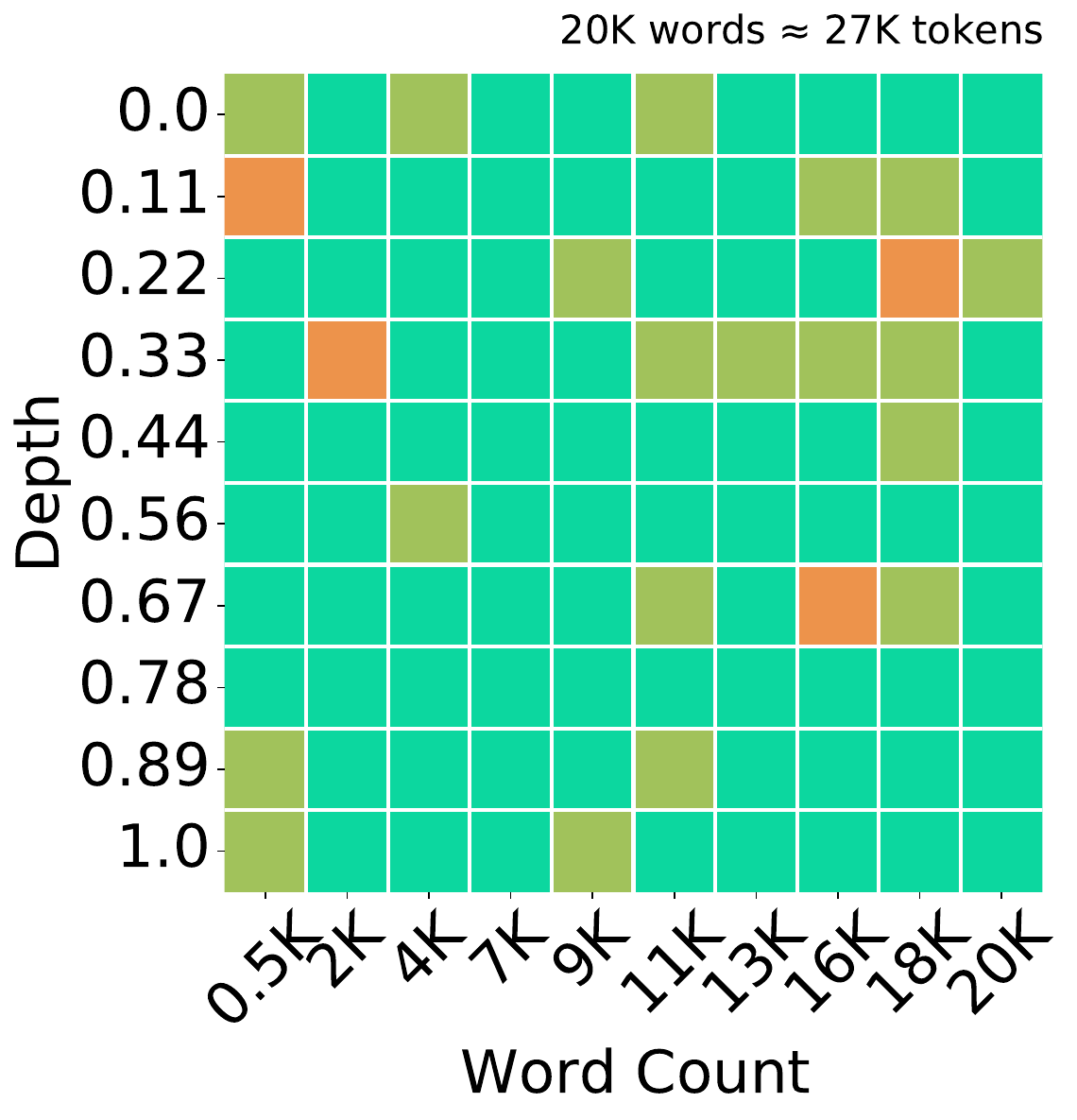}
	\end{minipage}%
}
\caption{Llama-3-8B-Instruct with no compression, as well as with 4bit and 2bit KIVI under needle test with 64-digit passkey. Overall accuracies are noted within parentheses.}
\label{fig:needle_kivi_llama_64}
\end{figure*}

\begin{figure*}[h]
\setlength{\abovecaptionskip}{0mm}
\setlength{\belowcaptionskip}{0mm}
\centering
\subfigcapskip=-2mm
\subfigure[2x Compression (67.3\%)]{
\centering
	\begin{minipage}[t]{0.23\linewidth}
		\includegraphics[width=\linewidth]{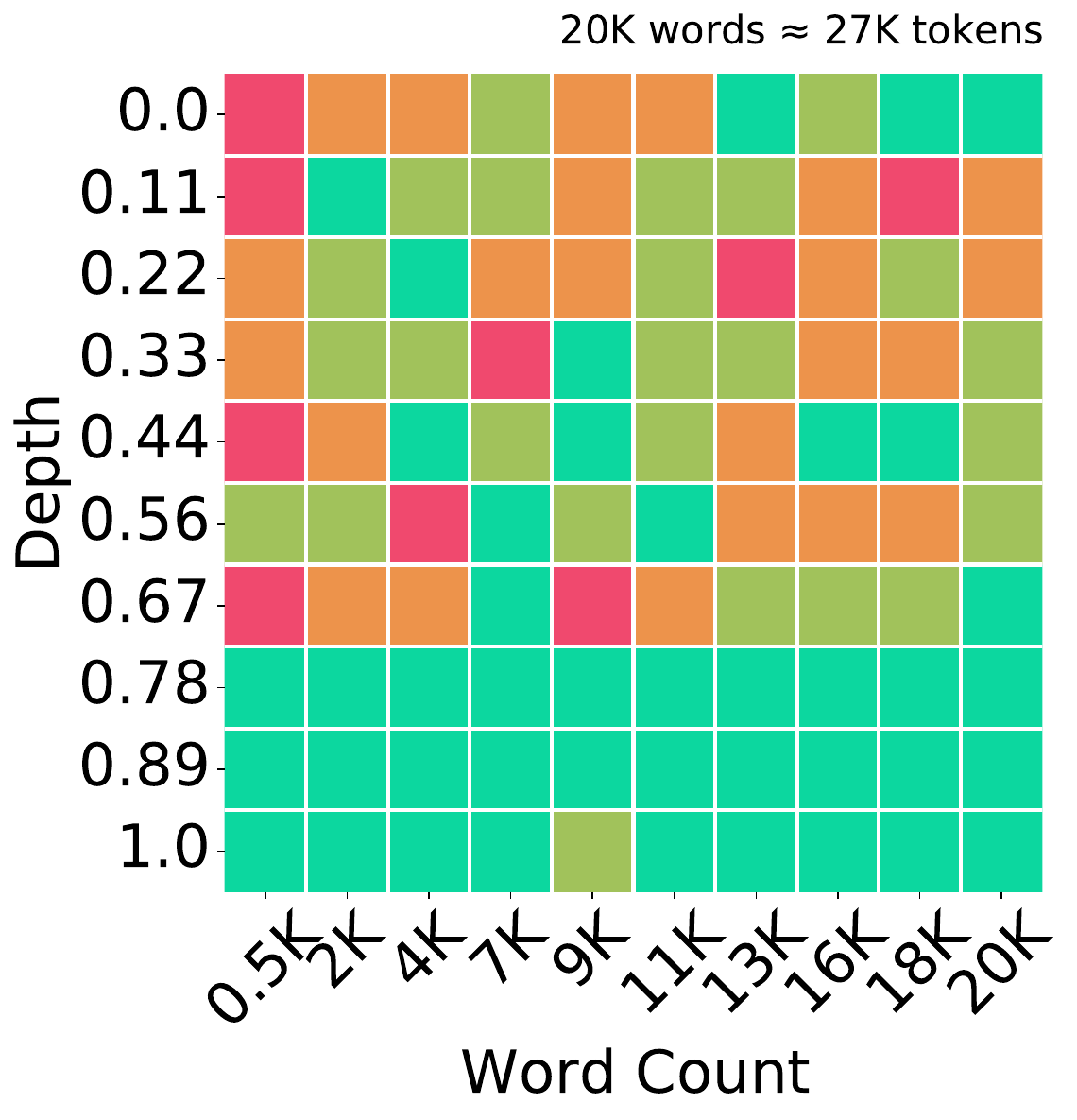}
	\end{minipage}%
}
\subfigure[4x Compression (35.0\%)]{
\centering
	\begin{minipage}[t]{0.23\linewidth}
		\includegraphics[width=\linewidth]{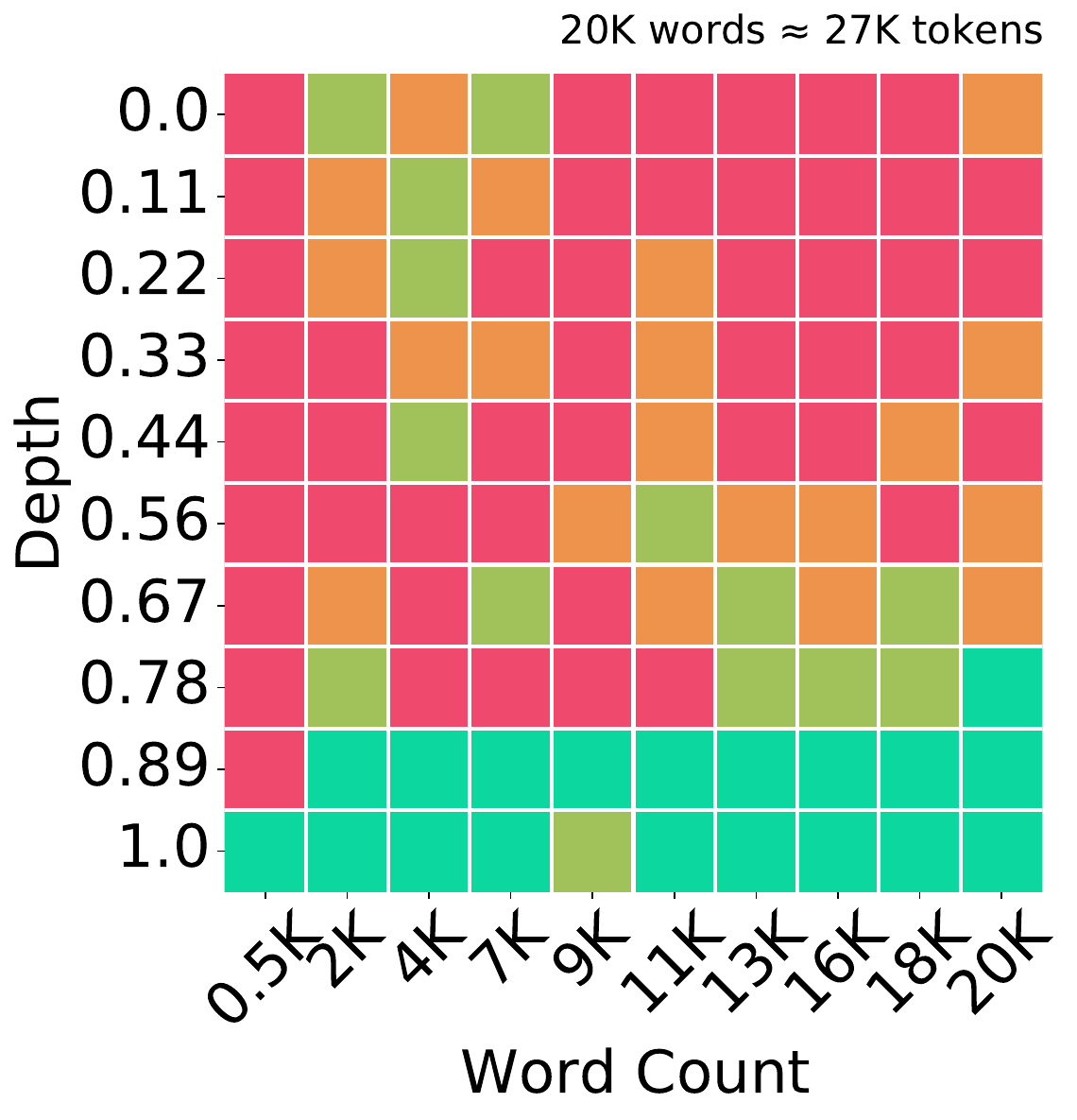}
	\end{minipage}%
}
\subfigure[6x Compression (22.0\%)]{
\centering
	\begin{minipage}[t]{0.23\linewidth}
		\includegraphics[width=\linewidth]{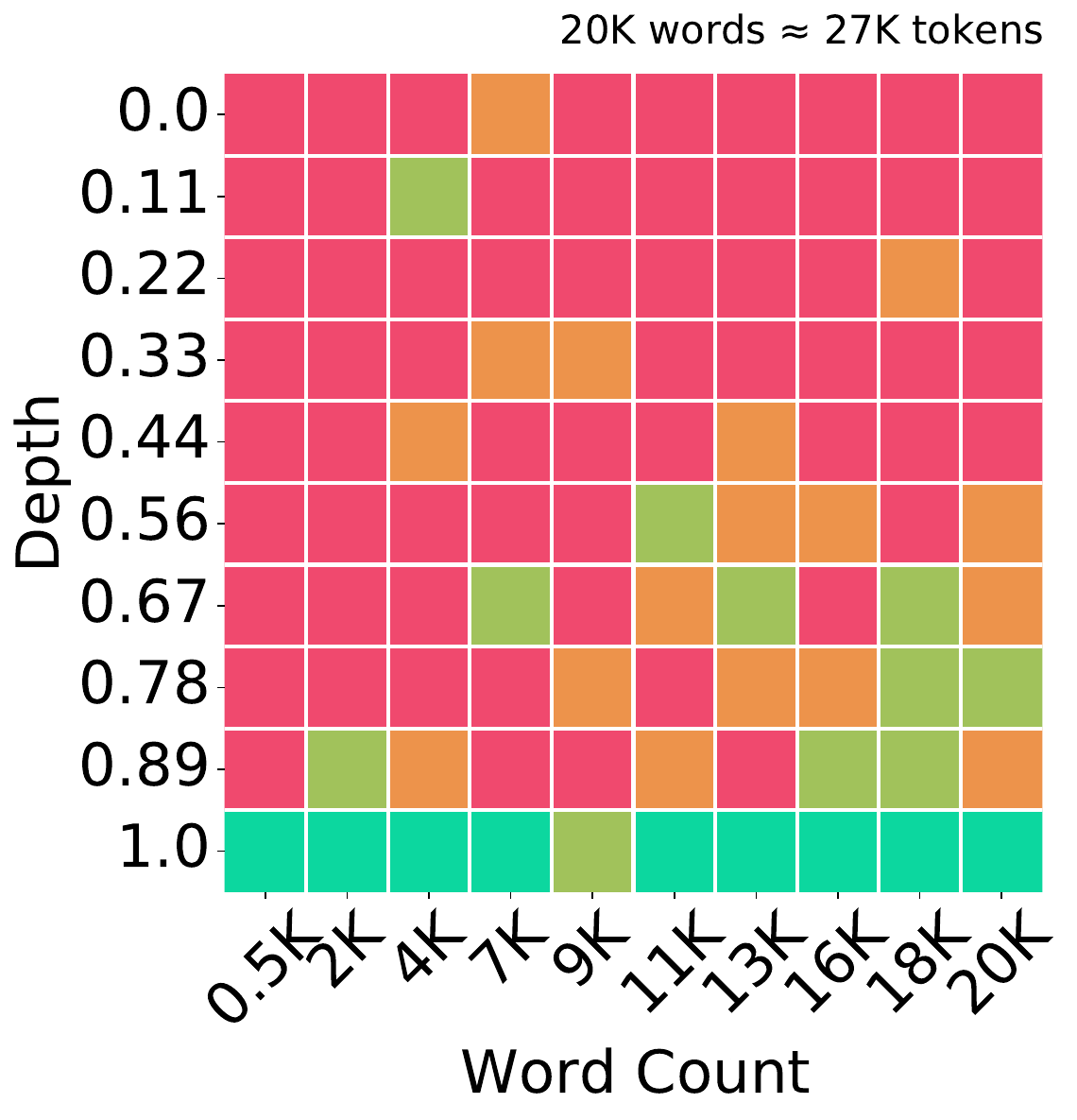}
	\end{minipage}%
}
\subfigure[8x Compression (18.7\%)]{
\centering
	\begin{minipage}[t]{0.23\linewidth}
		\includegraphics[width=\linewidth]{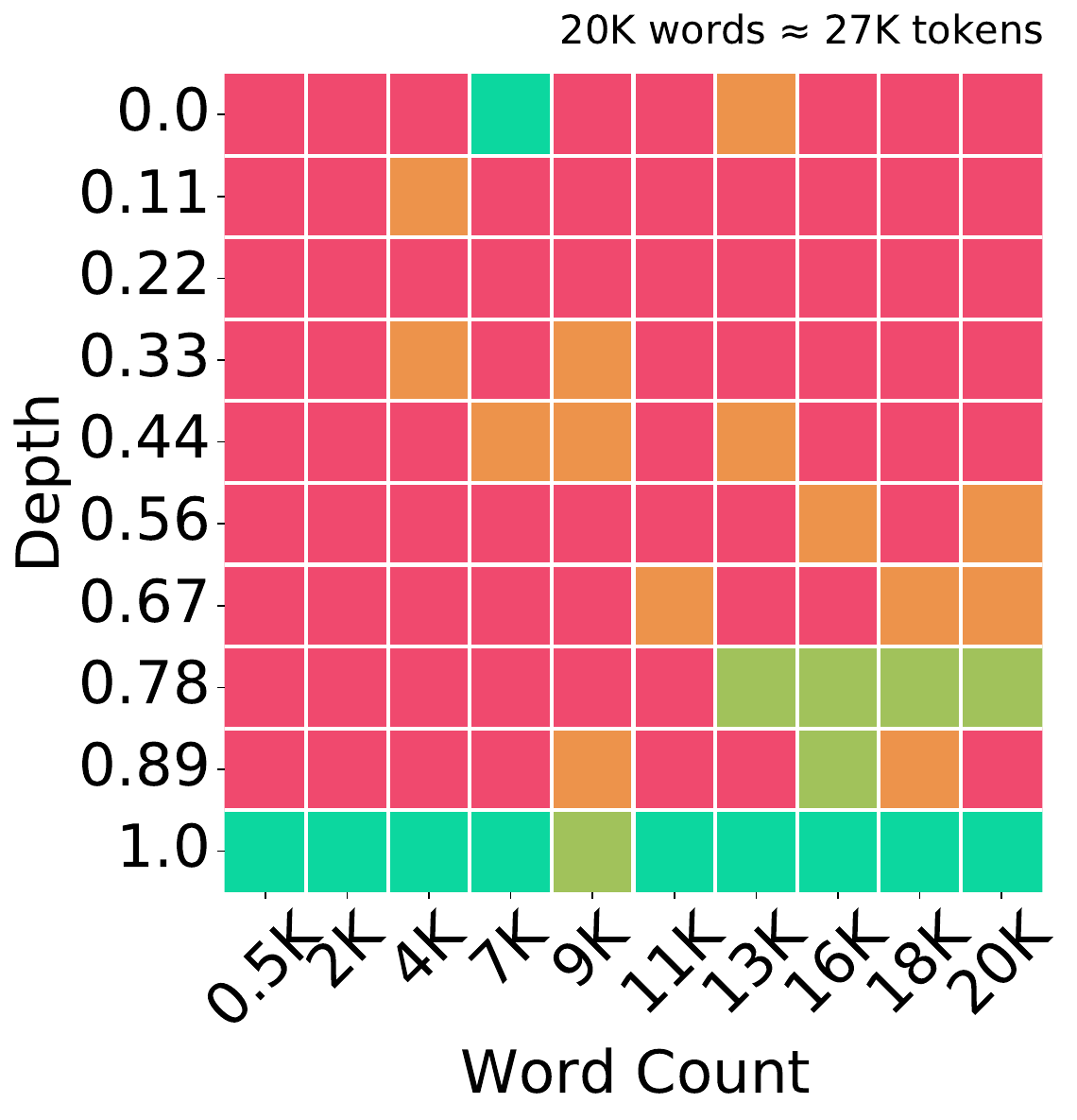}
	\end{minipage}%
}
\caption{$\mathrm{H_2O}$ on Llama-3-8B-Instruct with 4 different compression rates under needle test with 64-digit passkey. Overall accuracies are noted within parentheses.}
\label{fig:needle_h2o_llama_64}
\end{figure*}

\begin{figure*}[h]
\setlength{\abovecaptionskip}{0mm}
\setlength{\belowcaptionskip}{0mm}
\centering
\subfigcapskip=-2mm
\subfigure[2x Compression (17.7\%)]{
\centering
	\begin{minipage}[t]{0.23\linewidth}
		\includegraphics[width=\linewidth]{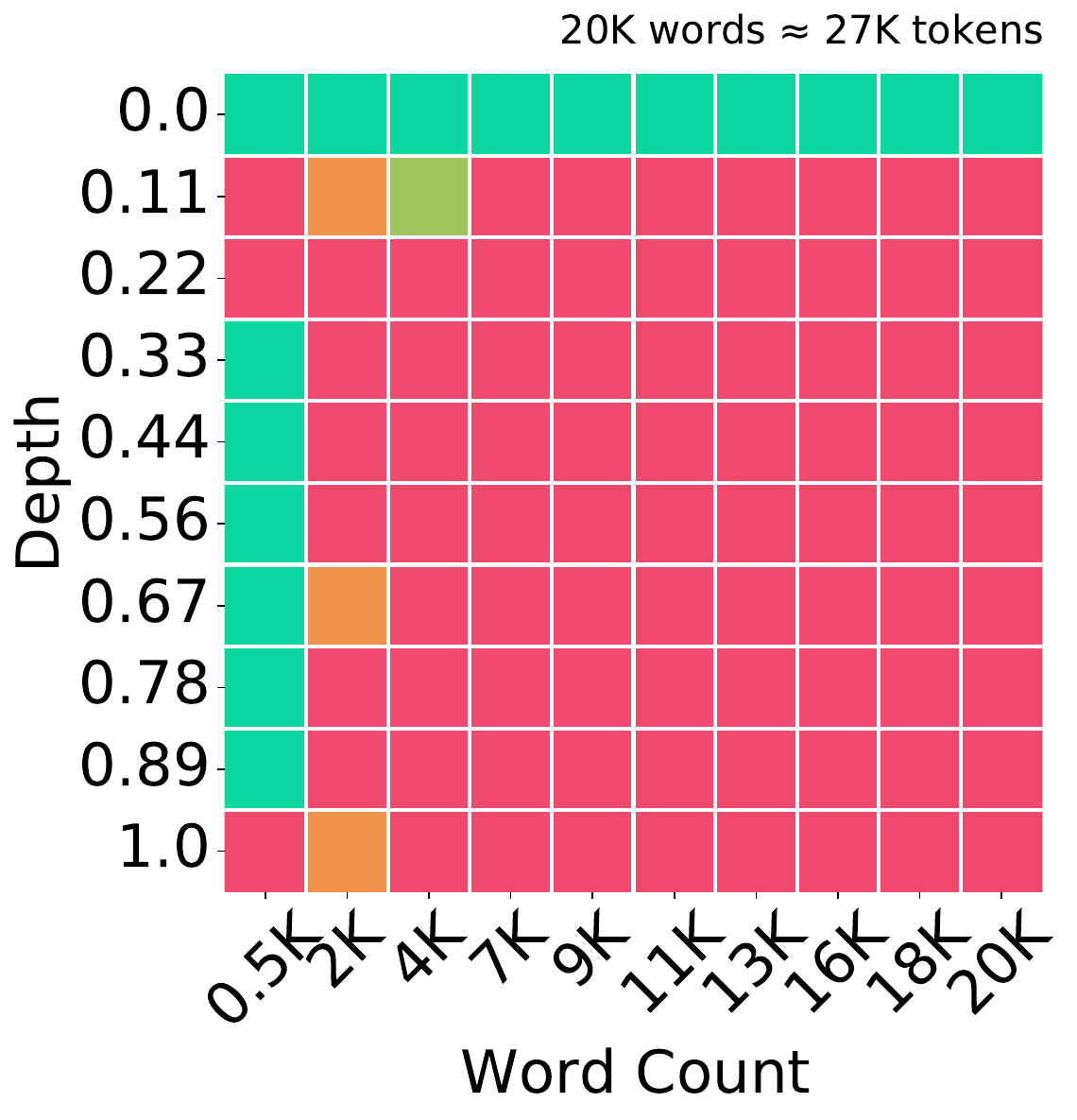}
	\end{minipage}%
}
\subfigure[4x Compression (19.0\%)]{
\centering
	\begin{minipage}[t]{0.23\linewidth}
		\includegraphics[width=\linewidth]{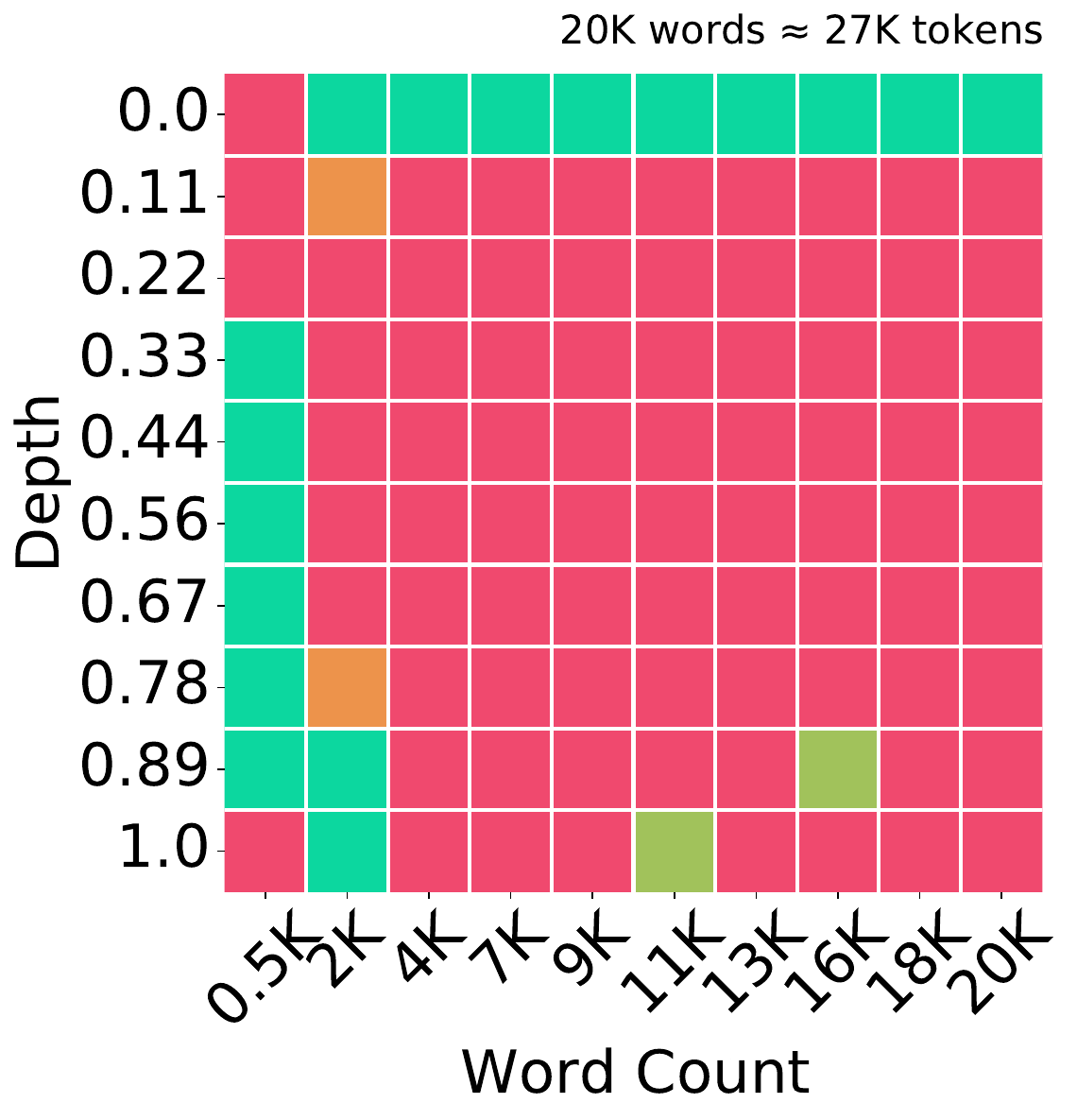}
	\end{minipage}%
}
\subfigure[6x Compression (19.0\%)]{
\centering
	\begin{minipage}[t]{0.23\linewidth}
		\includegraphics[width=\linewidth]{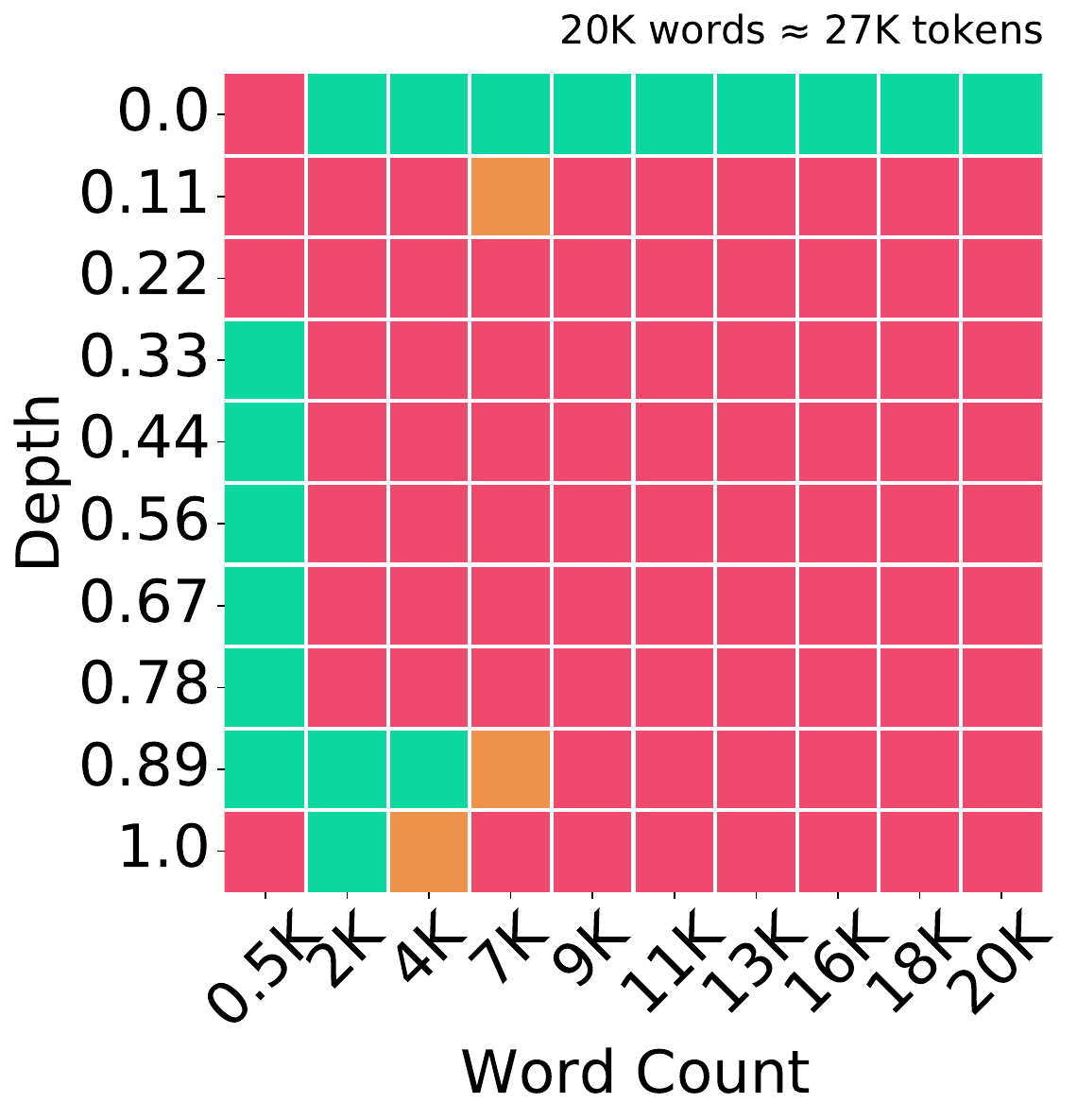}
	\end{minipage}%
}
\subfigure[8x Compression (19.3\%)]{
\centering
	\begin{minipage}[t]{0.23\linewidth}
		\includegraphics[width=\linewidth]{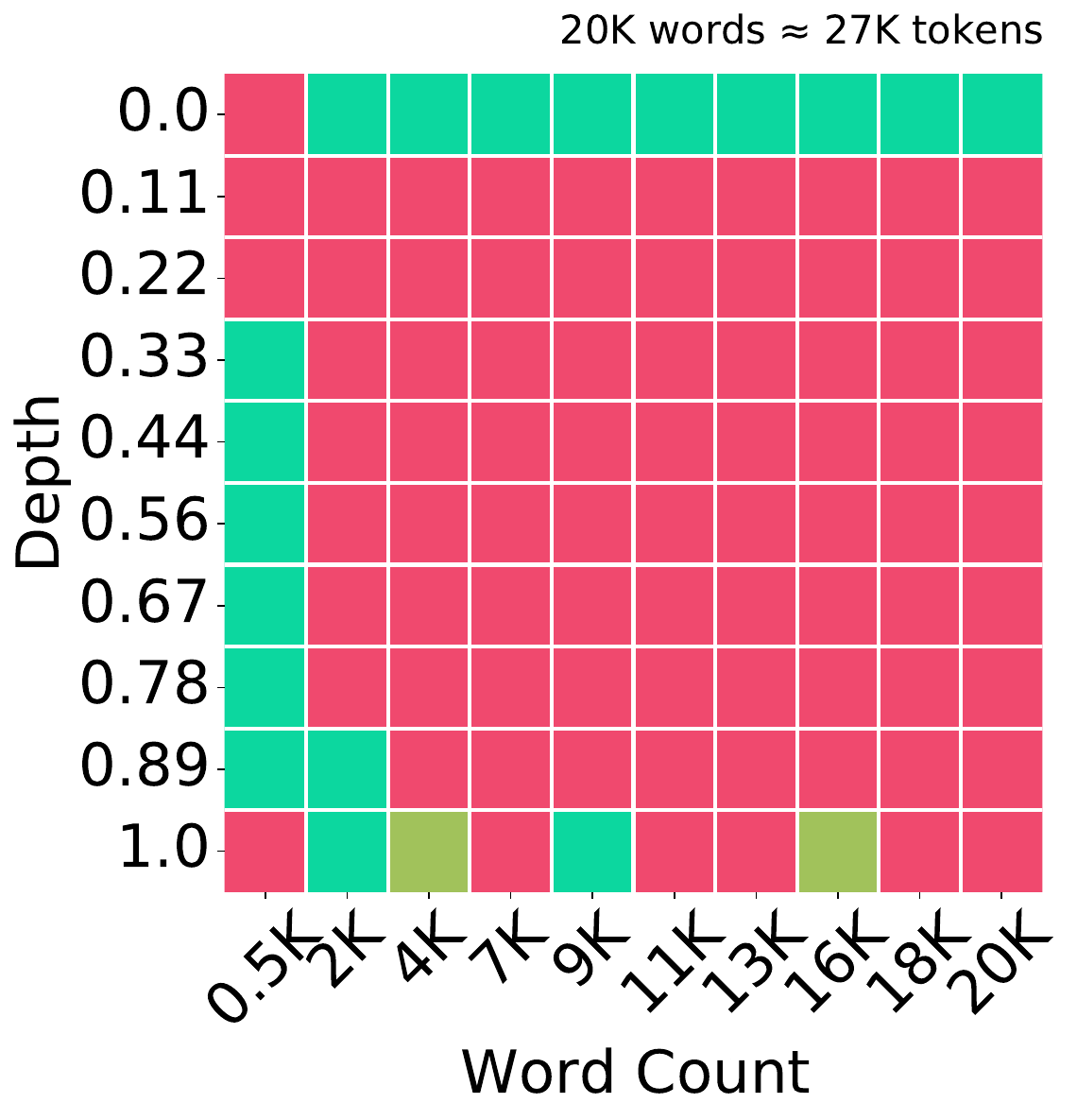}
	\end{minipage}%
}
\caption{InfLLM on Llama-3-8B-Instruct with 4 different compression rates under needle test with 64-digit passkey. Overall accuracies are noted within parentheses.}
\label{fig:needle_infllm_llama_64}
\end{figure*}

We additionally have Figure \ref{fig:needle_kivi_llama_64}, \ref{fig:needle_h2o_llama_64}, and \ref{fig:needle_infllm_llama_64} to showcase the performance drop of $\mathrm{H_2O}$ \citep{h2o} under the needle test with a 64-digit passkey as mentioned in OB \ding{185}, in comparison to other methods.

\section{Extended Results and Discussion}
\label{app_ex_dis}

\paragraph{A note on the ``overtraining'' recipe.}

In section~\ref{sec_bench_setup}, we briefly mentioned an ``overtrained recipe.'' This is mostly referring to Llama-3-8B, which is trained on 15T tokens and is way beyond the optimal point according to Chinchilla scaling law \citep{hoffmann2022scallinglaw}. This overtraining recipe is considered a main contributor to Llama-3's performance improvement.

We highlight it because most RNN/hybrid architectures are trying to outperform some fully transparent LLMs (ones we can reproduce the pretraining, in contrast to just having access to the trained weights) at a certain parameter scale with an identical training recipe — e.g., Mamba \citep{gu2023mamba} to Pythia \citep{biderman2023pythia} — where such LLMs-in-comparison do not employ this overtraining ingredient. This presents a gap in directly comparing the performance of weight-only open-sourced LLMs with fully transparent RNN/hybrids, and we alert our readers to be vigilant in drawing direct numerical comparisons.

\paragraph{A note on LLMLingua2 and coding tasks.}

LLMLingua2 \citep{jiang2023llmlingua} performs significantly worse in LCC compared to RopeBench-P, despite both being coding tasks (Table~\ref{tab:appendix_llama}, Table~\ref{tab:appendix_mistral}, and Table~\ref{tab:appendix_longchat}). We hypothesize this is because LCC is a single file code completion task, where RopeBench-P prefixes the code modules according to the important statements of a certain file at a repository level. This potentially gives RopeBench-P a natural ``outline,'' which can be favorable cues for hard prompt compression approaches like LLMLingua2 as these cues may drive the compression of different parts accordingly.

\paragraph{A note on Mamba-Chat.}

While Mamba-Chat \citep{haven2023mambachat} is presented as an instruction-tuned version of Mamba \citep{gu2023mamba}, it does not deliver much better performance than the original Mamba. Though much of this can be attributed to the particular instruction tuning recipe of Mamba-Chat, it suggests that supervised finetuning SSM models might require some extra considerations and careful monitoring.

\paragraph{A note on InfLLM with models utilizing condensing rotary embeddings.}

We noticed a significant performance improvement in InfLLM \citep{xiao2024infllm} on models utilizing condensing rotary embeddings (e.g., InfLLM on LongChat \citep{li2023longchat} in Table~\ref{tab:main_table_longchat} between the first and current version of our work). Upon investigation, we realize that the original InfLLM implementation does not take into account of the condensing rotary embedding technique\footnote{\url{https://lmsys.org/blog/2023-06-29-LongChat/}} (namely, \texttt{position\_ids}/ratio)  — a simple RoPE-variant~\citep{su2024roformer} with long context handling in mind, often utilized in LongChat and Vicuna \citep{vicuna2023} family of models — as InfLLM authors then shifted their focus out of the Vicuna family in their later versions. Upon updated implementation, we observe a decent performance boost on InfLLM with LongChat (Table~\ref{tab:main_table_longchat}).

\paragraph{A note on measured peak memory usage reports.}

During the rebuttal of this work, we promised our reviewers that we'd include real, code-measured, peak memory usage reports in the camera-ready. We then realized that HuggingFace Transformers involves some significant KV cache implementation changes around v4.42\footnote{\url{https://github.com/huggingface/transformers/pull/30536}}, resulting in up to 2x saving in memory consumption just by changing its versions. Transformers v4.42 is not available by the time of our submission and is in conflict with some of the environment requirements of our featured methods. For this reason, we will postpone sharing this report. We aim to provide an update in our repository once we are able to bring some of our featured methods to Transformers v4.45+ (where another major memory-related update\footnote{\url{https://github.com/huggingface/transformers/pull/31292}} has been done.

\paragraph{A note on having an alternative visualization than radar chart.}

Our work mainly employs radar charts to demonstrate the LongBench-related results. This decision is made based on the original choice of visualization utilized in the LongBench paper \citep{bai2023longbench}, and the fact that radar chart is one of the most space-efficient visualization options — a welcoming character when we are trying to feature multiple methods under different compression ratios against various datasets. That being said, we recognize that radar charts can sometimes be misleading due to amplifying the delta between different readings, as each apex of the radar chart is defined by the highest number in that regard. 

\begin{figure*}
\centerline{\includegraphics[width=1.1\linewidth]{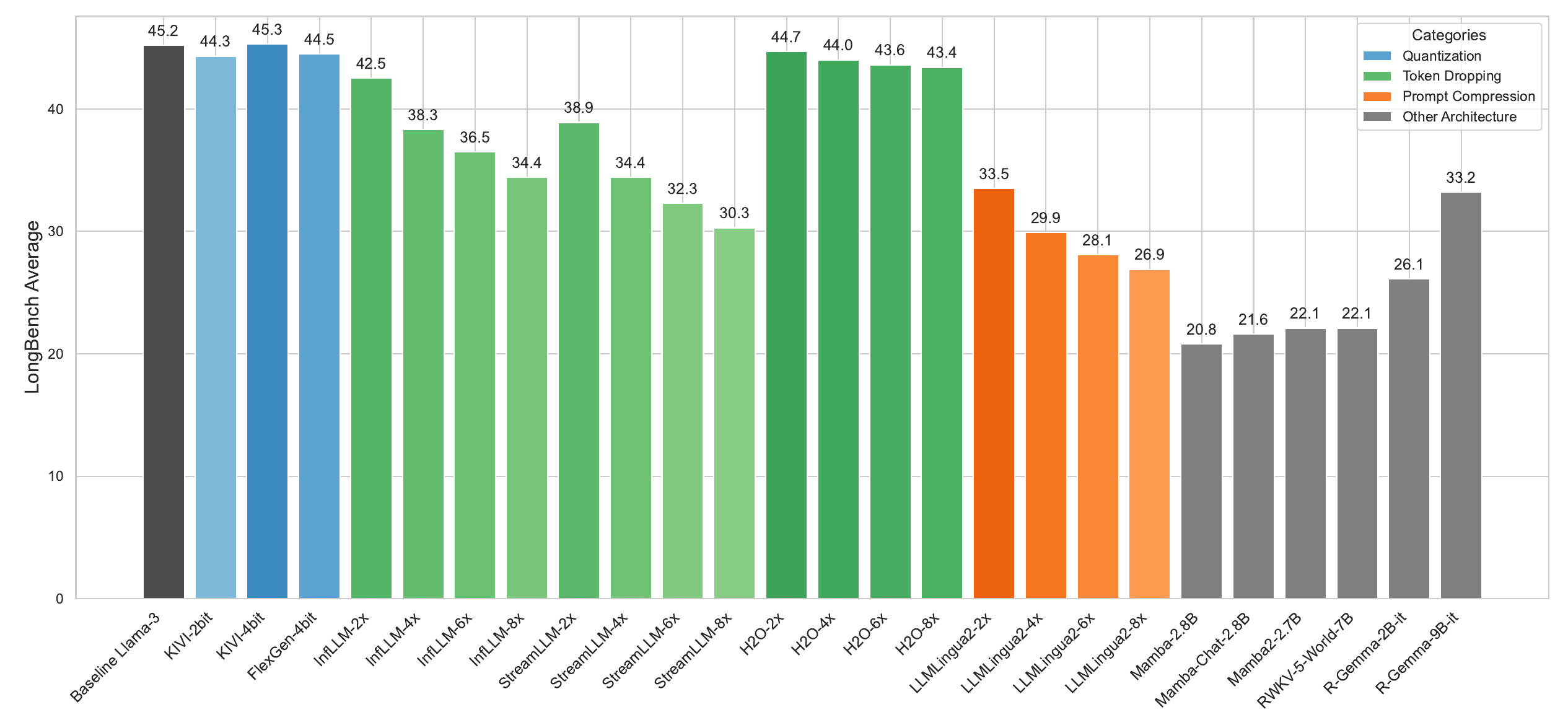}}
\caption{Performance of KV cache quantization, token dropping, prompt compression, and other architectures on LongBench.}
\label{fig:bar}
\vspace{-0.4cm}
\end{figure*}

Typically, a bar chart is the next best option and is free from the abovementioned concerns. However, a full bar chart plot for all the experiments we conducted would be too massive, as we are looking at roughly 20 method-compression ratio settings per LLM, where each method is tested against 7 categories of datasets. Here, we provide Figure~\ref{fig:bar}, a bar chart plot of Table~\ref{tab:main_table}. Specifically, the LongBench result was compared across different methods on Llama-3 and other model architectures.

\begin{table*}[h]
\centering
\caption{Performance of KV cache quantization, token dropping, and prompt compression methods on LongChat-7B in our benchmark.}
\resizebox{\textwidth}{!}{
\begin{tabular}{c|l|c|cccccc|cc}
\toprule
\textbf{Model} & \textbf{Method} & Comp. Ratio & Single. QA & Multi. QA & Summ. & Few-shot & Synthetic & Code & \textbf{LB Avg.} & \textbf{Needle} \\
\midrule
\multirow{20}{*}{\rotatebox[origin=c]{90}{longchat-7b-v1.5-32k}} 
        & Baseline & 1.00$\times$ & 31.1 & 23.9 & 26.7 & 63.8 & 30.5 & 54.9 & 38.5 & 96.3 \\
        & \cellcolor{LightCyan}KIVI-2bit & \cellcolor{LightCyan}5.05$\times$ & \cellcolor{LightCyan}30.3 & \cellcolor{LightCyan}23.1 & \cellcolor{LightCyan}26.5 & \cellcolor{LightCyan}63.6 & \cellcolor{LightCyan}32.2 & \cellcolor{LightCyan}53.9 & \cellcolor{LightCyan}38.0 & \cellcolor{LightCyan}85.3 \\
        & KIVI-4bit & 3.11$\times$ & 31.1 & 24.2 & 26.8 & 63.9 & 31.5 & 54.3 & 38.5 & 96.3 \\
        & \cellcolor{LightCyan}FlexGen-4bit & \cellcolor{LightCyan} 3.20$\times$ & \cellcolor{LightCyan}31.3 & \cellcolor{LightCyan}23.8 & \cellcolor{LightCyan}27.0 & \cellcolor{LightCyan}62.8 & \cellcolor{LightCyan}31.5 & \cellcolor{LightCyan}53.4 & \cellcolor{LightCyan}38.2 & \cellcolor{LightCyan}94.7 \\
        & InfLLM-2x & 2.00$\times$ & 29.2 & 23.3 & 25.8 & 51.8 & 19.5 & 51.9 & 34.2 & 58.7 \\
        & \cellcolor{LightCyan}InfLLM-4x & \cellcolor{LightCyan}4.00$\times$ & \cellcolor{LightCyan}24.2 & \cellcolor{LightCyan}24.0 & \cellcolor{LightCyan}24.4 & \cellcolor{LightCyan}49.7 & \cellcolor{LightCyan}9.5 & \cellcolor{LightCyan}46.8 & \cellcolor{LightCyan}31.4 & \cellcolor{LightCyan}34.0 \\
        & InfLLM-6x & 6.00$\times$ & 21.1 & 23.5 & 23.5 & 48.9 & 10.0 & 46.5 & 30.2 & 35.0 \\
        & \cellcolor{LightCyan}InfLLM-8x & \cellcolor{LightCyan}8.00$\times$ & \cellcolor{LightCyan}20.1 & \cellcolor{LightCyan}21.7 & \cellcolor{LightCyan}22.6 & \cellcolor{LightCyan}45.3 & \cellcolor{LightCyan}6.0 & \cellcolor{LightCyan}46.2 & \cellcolor{LightCyan}28.5 & \cellcolor{LightCyan}25.7 \\
        & StreamLLM-2x & 2.00$\times$ & 23.9 & 21.8 & 24.2 & 51.2 & 25.0 & 49.3 & 32.5 & 47.7 \\
        & \cellcolor{LightCyan}StreamLLM-4x & \cellcolor{LightCyan}4.00$\times$ & \cellcolor{LightCyan}19.9 & \cellcolor{LightCyan}22.1 & \cellcolor{LightCyan}22.3 & \cellcolor{LightCyan}49.0 & \cellcolor{LightCyan}13.5 & \cellcolor{LightCyan}52.0 & \cellcolor{LightCyan}30.5 & \cellcolor{LightCyan}32.3 \\
        & StreamLLM-6x & 6.00$\times$ & 19.6 & 21.4 & 20.7 & 47.5 & 10.5 & 51.4 & 29.4 & 21.3 \\
        & \cellcolor{LightCyan}StreamLLM-8x & \cellcolor{LightCyan}8.00$\times$ & \cellcolor{LightCyan}17.5 & \cellcolor{LightCyan}21.4 & \cellcolor{LightCyan}19.5 & \cellcolor{LightCyan}44.4 & \cellcolor{LightCyan}10.0 & \cellcolor{LightCyan}51.9 & \cellcolor{LightCyan}28.2 & \cellcolor{LightCyan}19.7 \\
        & $\mathrm{H_2O}$-2x & 2.00$\times$ &  27.6 & 22.1 & 24.6 & 62.6 & 30.5 & 57.8 & 37.1 & 56.7 \\
        & \cellcolor{LightCyan}$\mathrm{H_2O}$-4x & \cellcolor{LightCyan}4.00$\times$ & \cellcolor{LightCyan}26.2 & \cellcolor{LightCyan}21.9 & \cellcolor{LightCyan}21.9 & \cellcolor{LightCyan}61.9 & \cellcolor{LightCyan}28.5 & \cellcolor{LightCyan}55.2 & \cellcolor{LightCyan}35.7 & \cellcolor{LightCyan}28.7 \\
        & $\mathrm{H_2O}$-6x & 6.00$\times$ & 25.7 & 21.3 & 21.0 & 62.1 & 28.0 & 53.3 & 35.0 & 19.7 \\
        & \cellcolor{LightCyan}$\mathrm{H_2O}$-8x & \cellcolor{LightCyan}8.00$\times$ & \cellcolor{LightCyan}25.1 & \cellcolor{LightCyan}21.0 & \cellcolor{LightCyan}19.8 & \cellcolor{LightCyan}61.6 & \cellcolor{LightCyan}28.5 & \cellcolor{LightCyan}51.5 & \cellcolor{LightCyan}34.3 & \cellcolor{LightCyan}14.3 \\
        & LLMLingua2-2x & 2.00$\times$ & 25.7 & 22.3 & 25.4 & 35.4 & 19.5 & 32.6 & 27.4 & 28.7 \\
        &\cellcolor{LightCyan}LLMLingua2-4x & \cellcolor{LightCyan}4.00$\times$ & \cellcolor{LightCyan}23.8 & \cellcolor{LightCyan}20.6 & \cellcolor{LightCyan}23.5 & \cellcolor{LightCyan}31.6 & \cellcolor{LightCyan}5.5 & \cellcolor{LightCyan}31.9 & \cellcolor{LightCyan}24.5 & \cellcolor{LightCyan}3.3 \\
        & LLMLingua2-6x & 6.00$\times$ & 22.6 & 20.2 & 22.6 & 32.3 & 5.0 & 31.9 & 24.1 & 0.6 \\
        &\cellcolor{LightCyan}LLMLingua2-8x & \cellcolor{LightCyan}8.00$\times$ & \cellcolor{LightCyan}21.3 & \cellcolor{LightCyan}19.5 & \cellcolor{LightCyan}21.9 & \cellcolor{LightCyan}32.9 & \cellcolor{LightCyan}6.5 & \cellcolor{LightCyan}32.5 & \cellcolor{LightCyan}23.9 & \cellcolor{LightCyan}0.0 \\
\bottomrule
\end{tabular}
}
\label{tab:main_table_longchat}
\end{table*}

\begin{figure*}[h]
\setlength{\abovecaptionskip}{0mm}
\setlength{\belowcaptionskip}{0mm}
\centering
\subfigcapskip=-2mm

\subfigure[Llama-3 + KIVI-2bit]{
\centering
    \begin{minipage}[t]{0.3\linewidth}
        \includegraphics[width=\linewidth]{figures/needle/kivi/llama/2_bit.pdf}
    \end{minipage}%
}
\subfigure[Llama-3 + KIVI-4bit]{
\centering
    \begin{minipage}[t]{0.3\linewidth}
        \includegraphics[width=\linewidth]{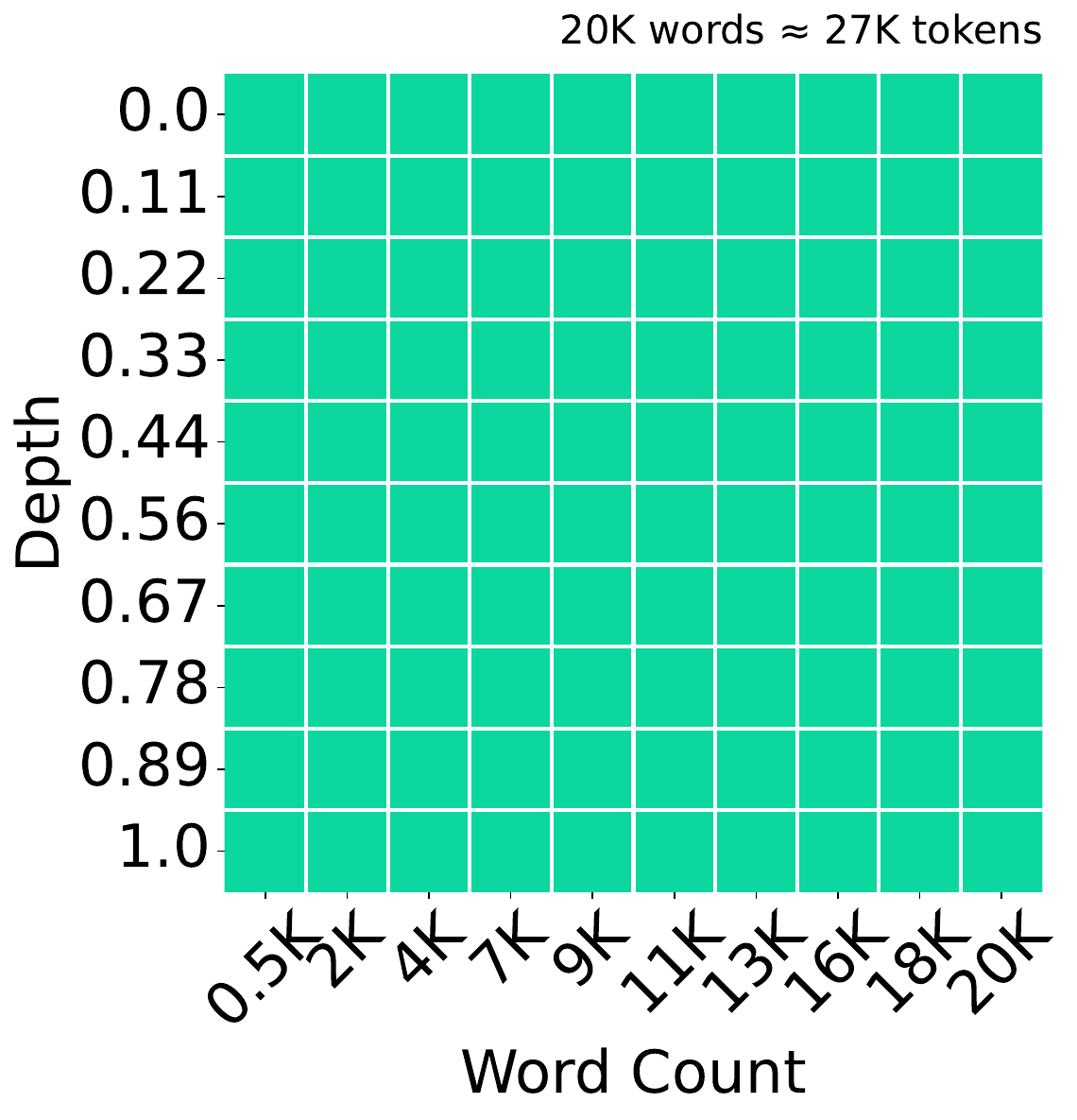}
    \end{minipage}%
}
\subfigure[LongChat-7B + KIVI-2bit]{
\centering
    \begin{minipage}[t]{0.3\linewidth}
        \includegraphics[width=\linewidth]{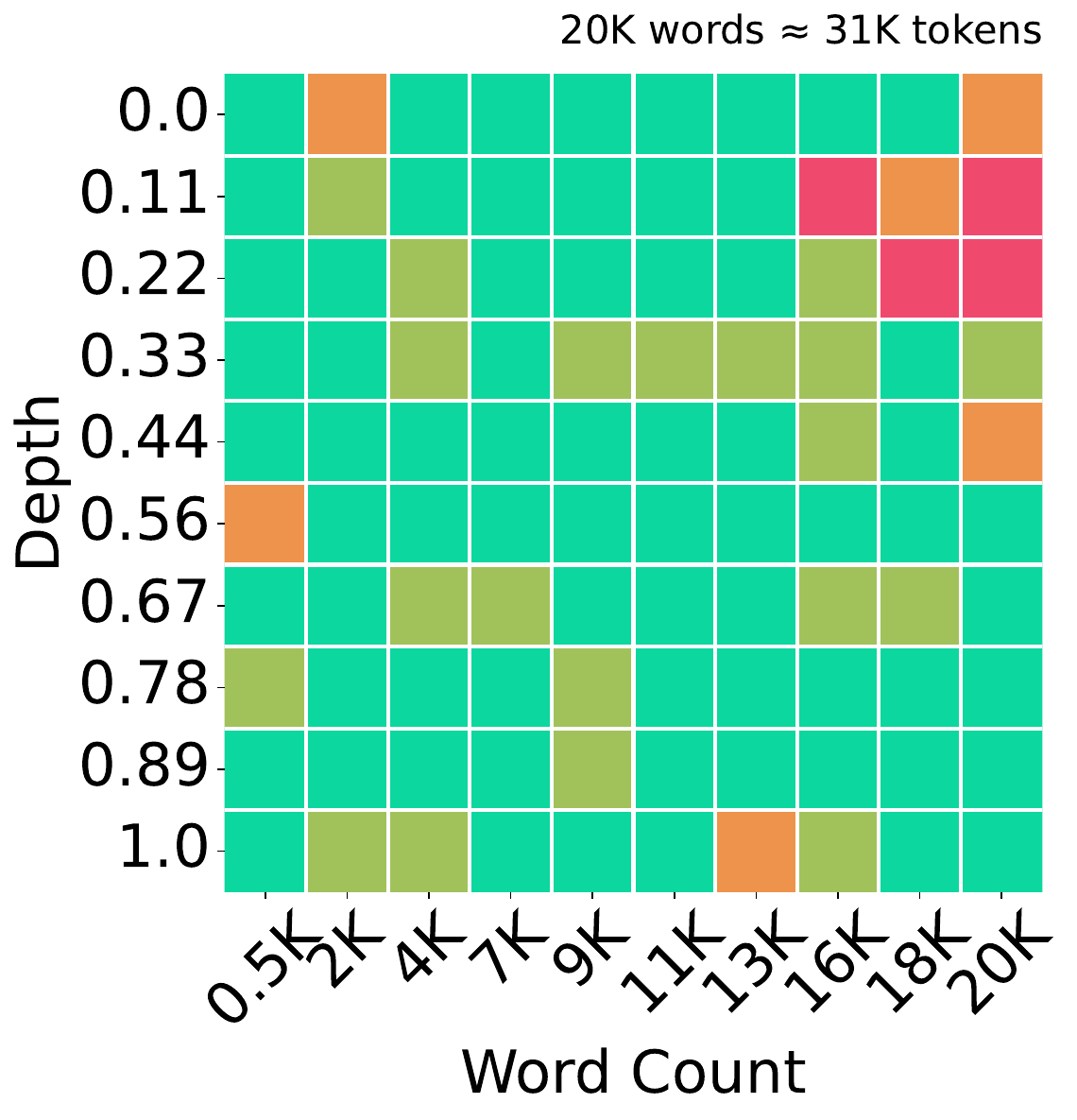}
    \end{minipage}%
}

\subfigure[LongChat-7B + KIVI-4bit]{
\centering
    \begin{minipage}[t]{0.3\linewidth}
        \includegraphics[width=\linewidth]{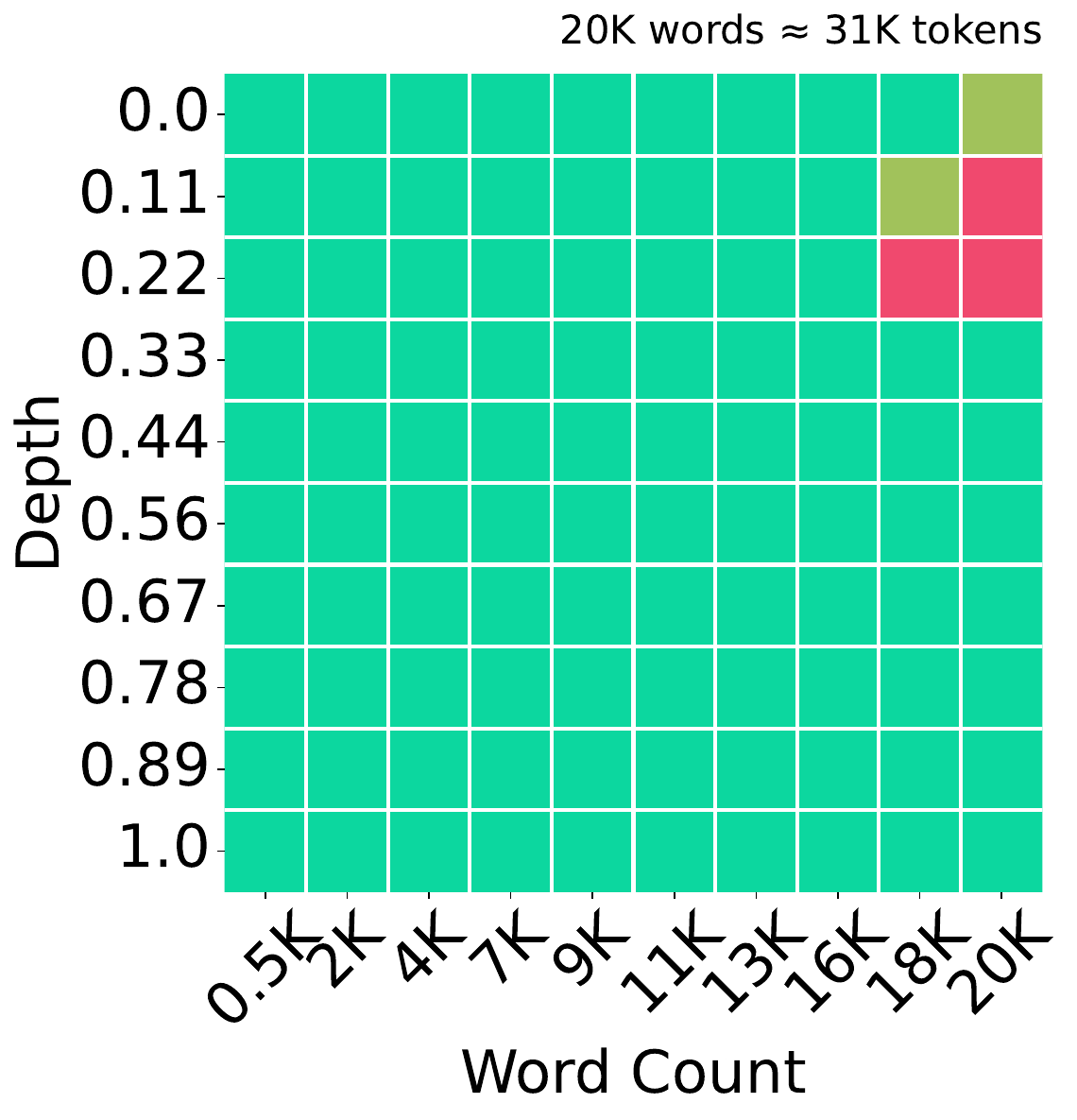}
    \end{minipage}%
}
\subfigure[Mistral-7B + KIVI-2bit]{
\centering
    \begin{minipage}[t]{0.3\linewidth}
        \includegraphics[width=\linewidth]{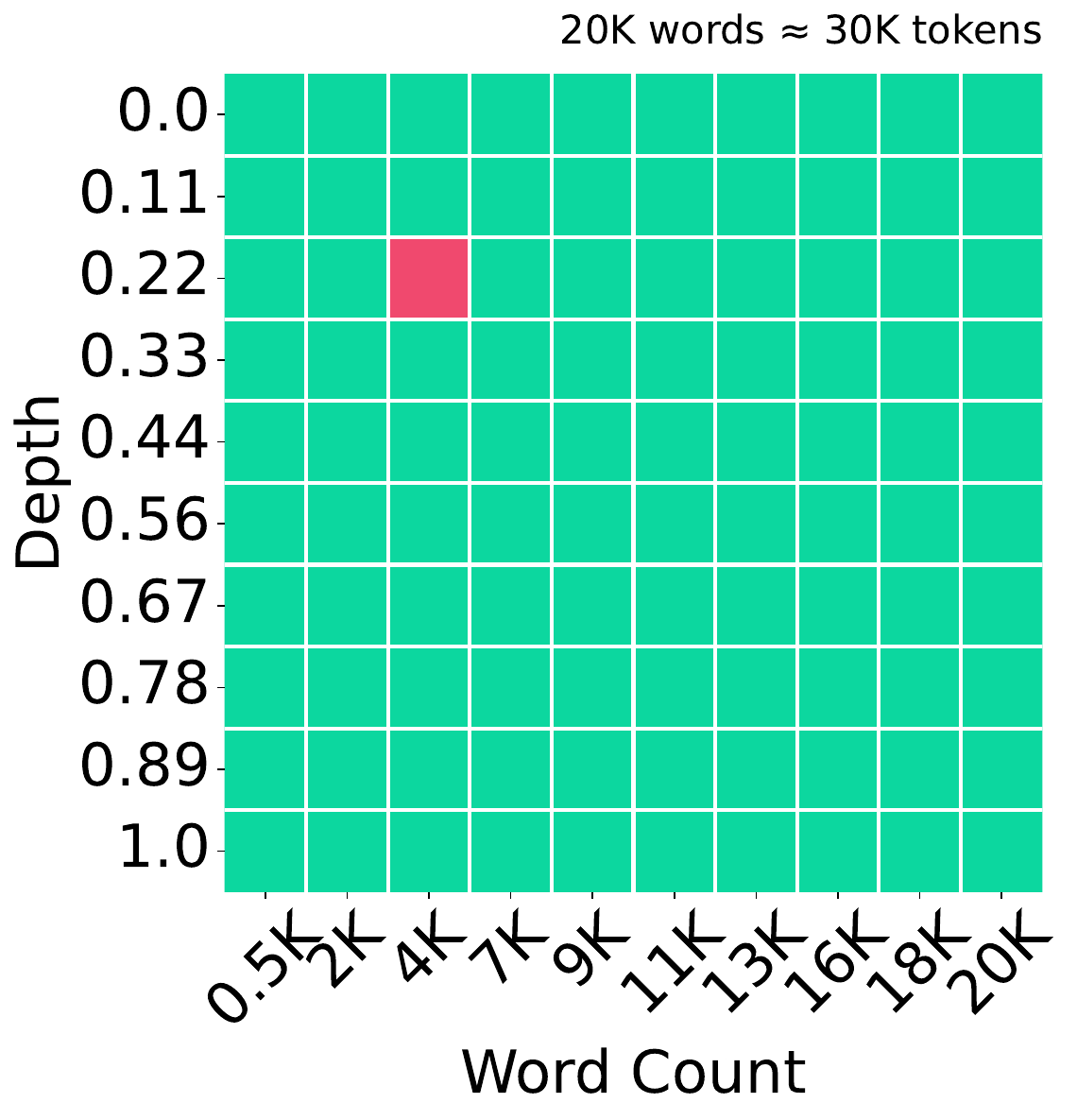}
    \end{minipage}%
}
\subfigure[Mistral-7B + KIVI-4bit]{
\centering
    \begin{minipage}[t]{0.3\linewidth}
        \includegraphics[width=\linewidth]{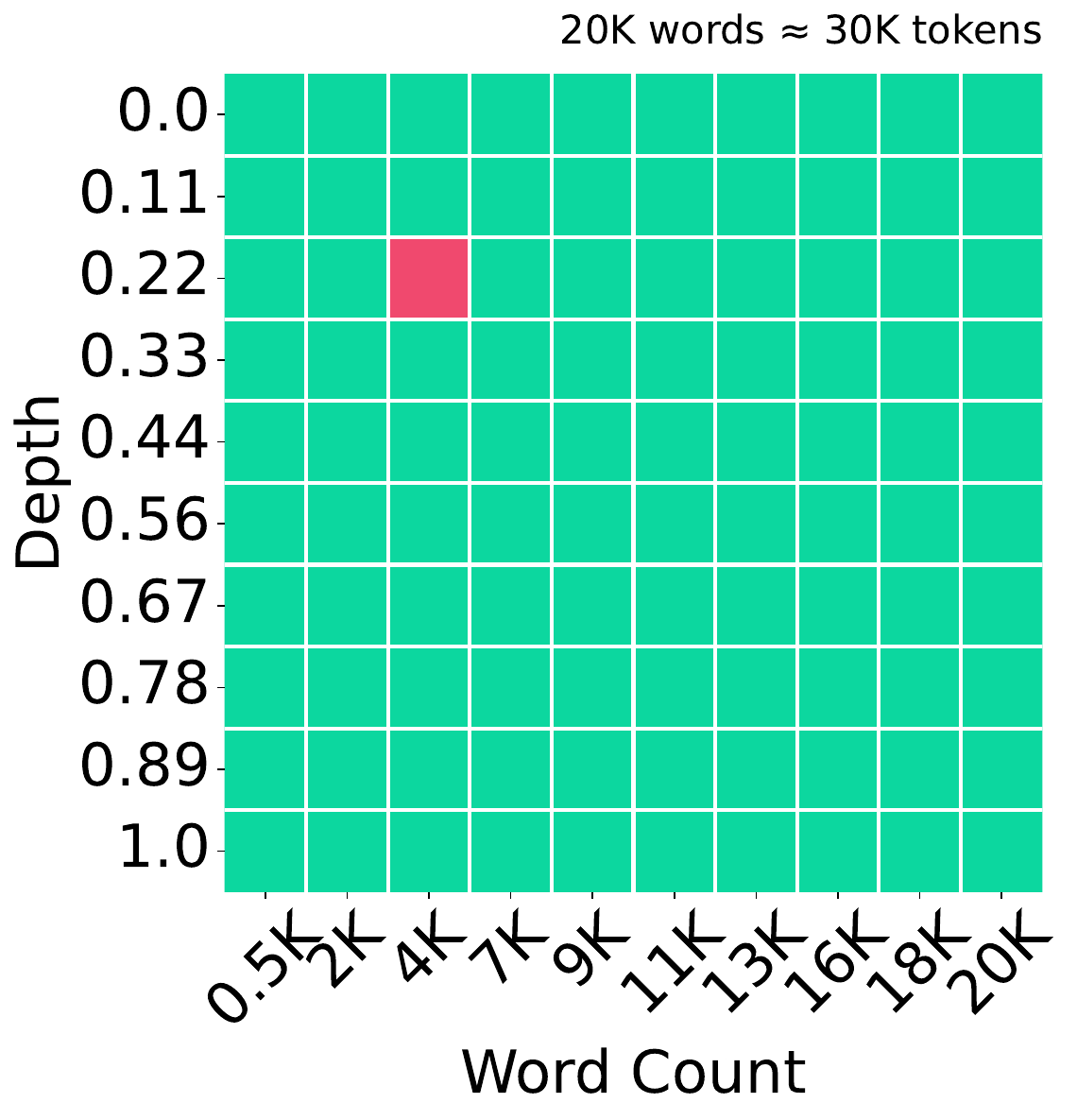}
    \end{minipage}%
}

\caption{KIVI performance under needle test on three commonly used LLMs with 2-bit and 4-bit quantization}
\label{fig:needle_kivi}
\end{figure*}

\begin{figure*}[h]
\setlength{\abovecaptionskip}{0mm}
\setlength{\belowcaptionskip}{0mm}
\centering
\subfigcapskip=-2mm
\subfigure[2x Compression]{
\centering
	\begin{minipage}[t]{0.23\linewidth}
		\includegraphics[width=\linewidth]{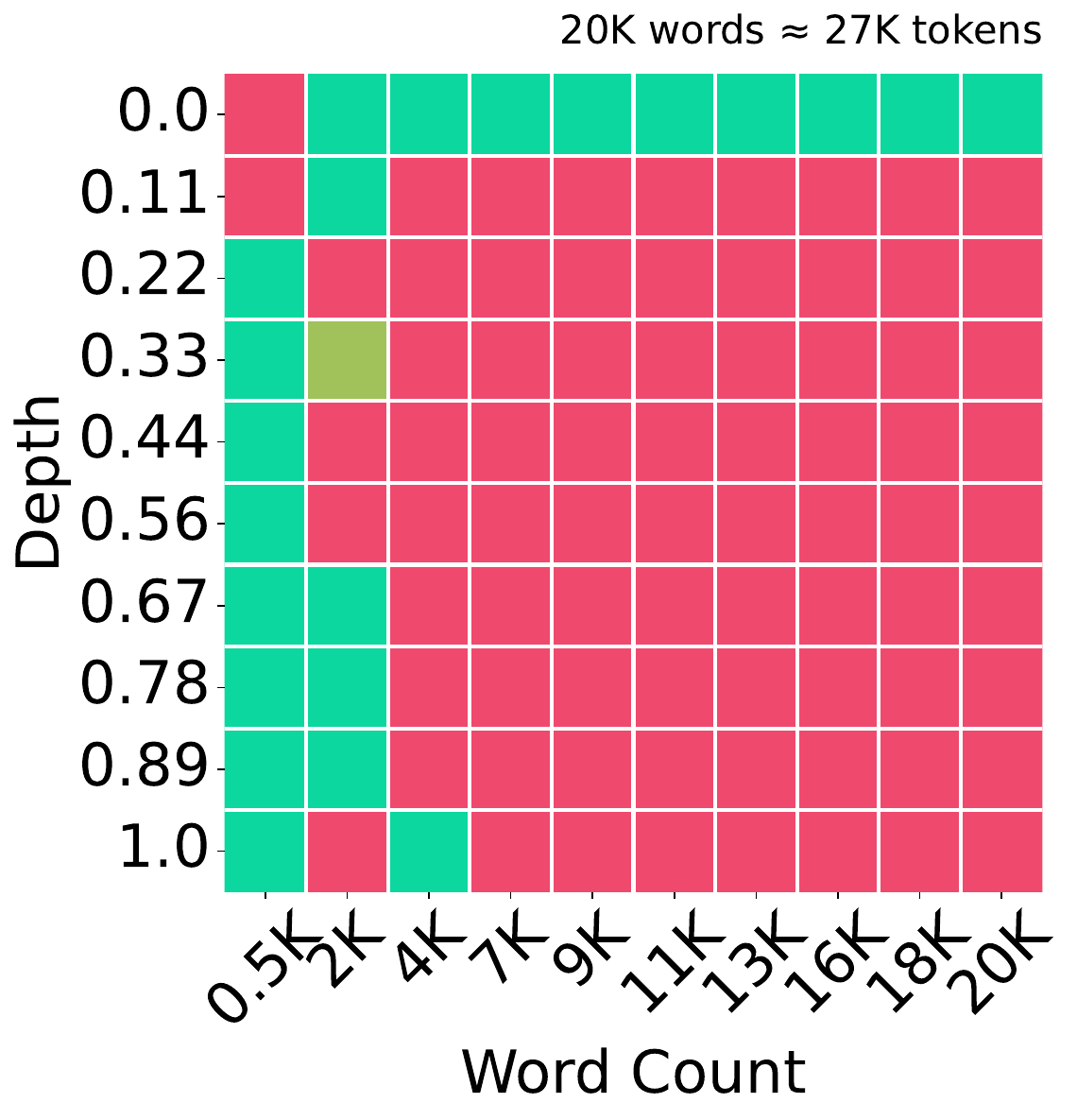}
	\end{minipage}%
}
\subfigure[4x Compression]{
\centering
	\begin{minipage}[t]{0.23\linewidth}
		\includegraphics[width=\linewidth]{figures/needle/infllm/llama/4x.pdf}
	\end{minipage}%
}
\subfigure[6x Compression]{
\centering
	\begin{minipage}[t]{0.23\linewidth}
		\includegraphics[width=\linewidth]{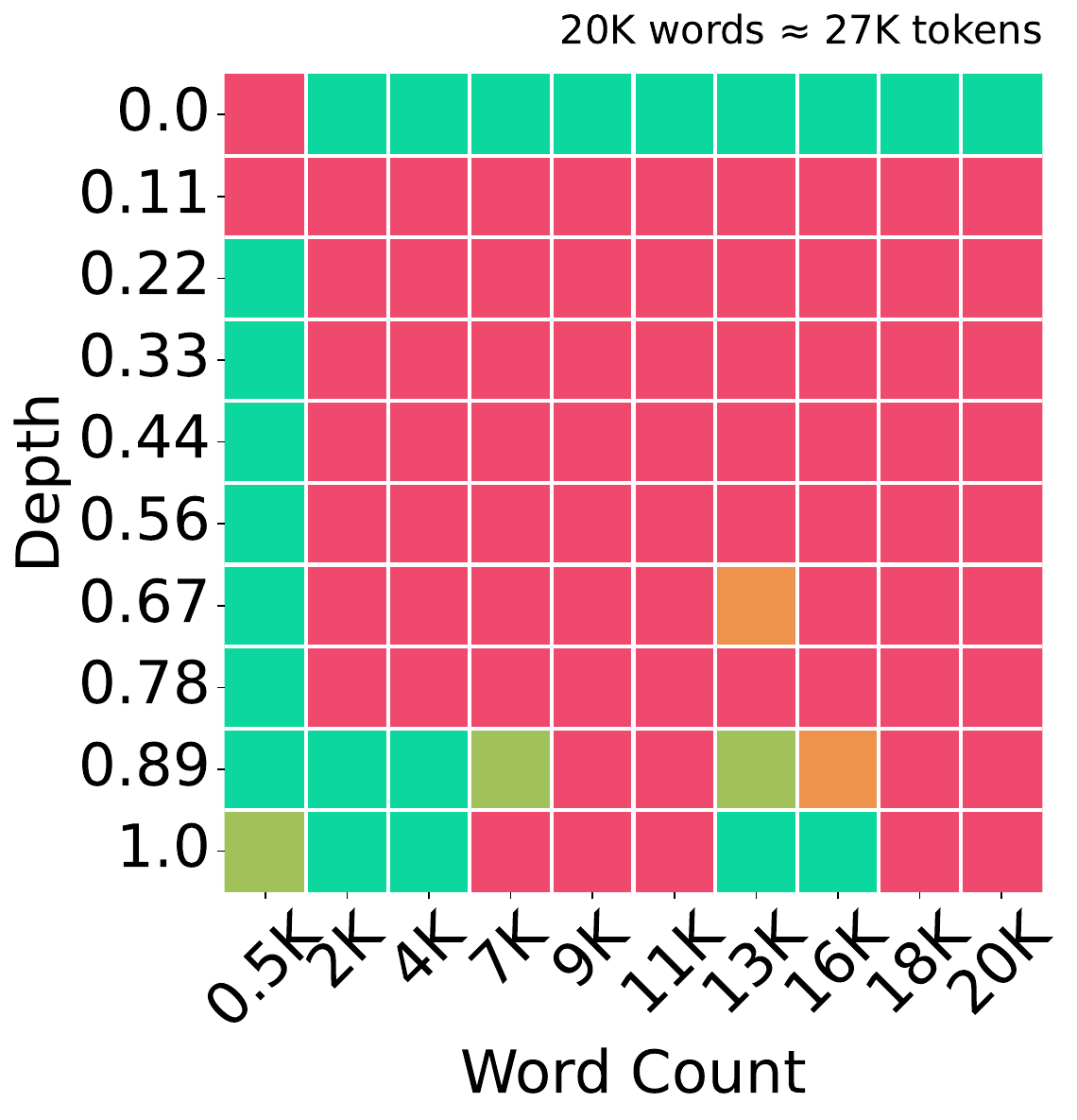}
	\end{minipage}%
}
\subfigure[8x Compression]{
\centering
	\begin{minipage}[t]{0.23\linewidth}
		\includegraphics[width=\linewidth]{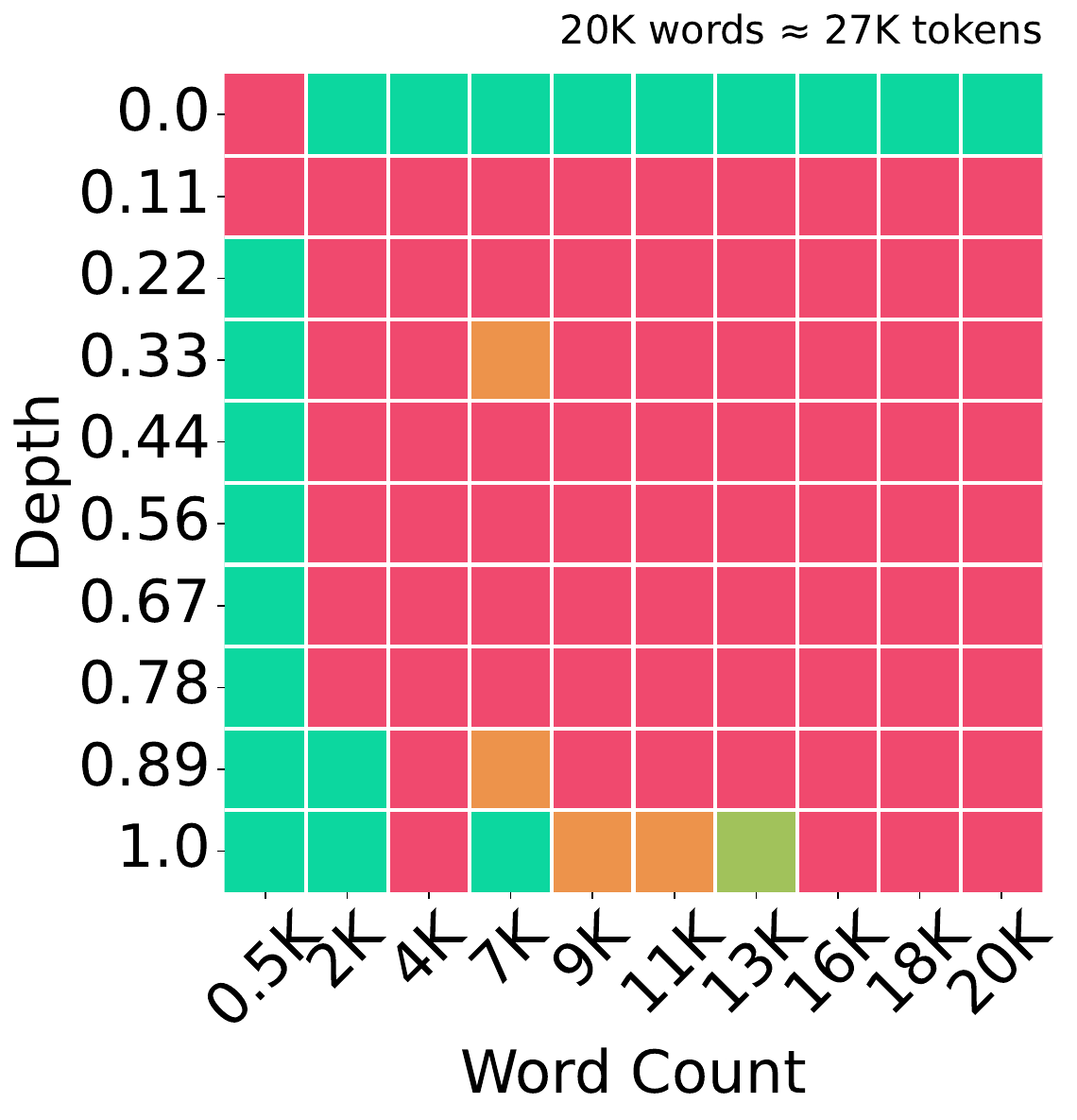}
	\end{minipage}%
}
\caption{InfLLM on Llama-3-8B-Instruct with 4 different compression rates under needle test}
\label{fig:needle_infllm_llama}
\end{figure*}

\begin{figure*}[h]
\setlength{\abovecaptionskip}{0mm}
\setlength{\belowcaptionskip}{0mm}
\centering
\subfigcapskip=-2mm
\subfigure[2x Compression]{
\centering
	\begin{minipage}[t]{0.23\linewidth}
		\includegraphics[width=\linewidth]{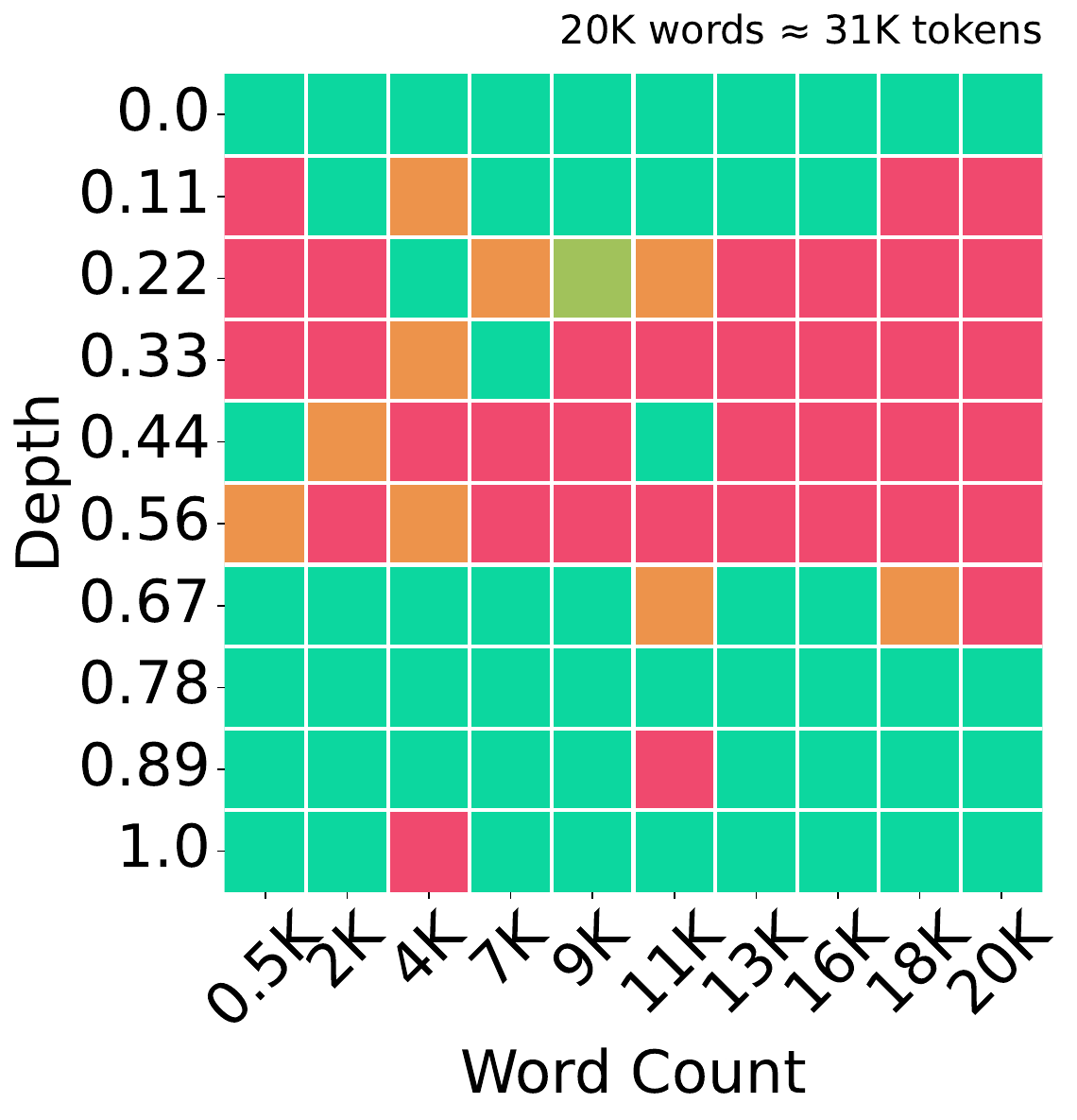}
	\end{minipage}%
}
\subfigure[4x Compression]{
\centering
	\begin{minipage}[t]{0.23\linewidth}
		\includegraphics[width=\linewidth]{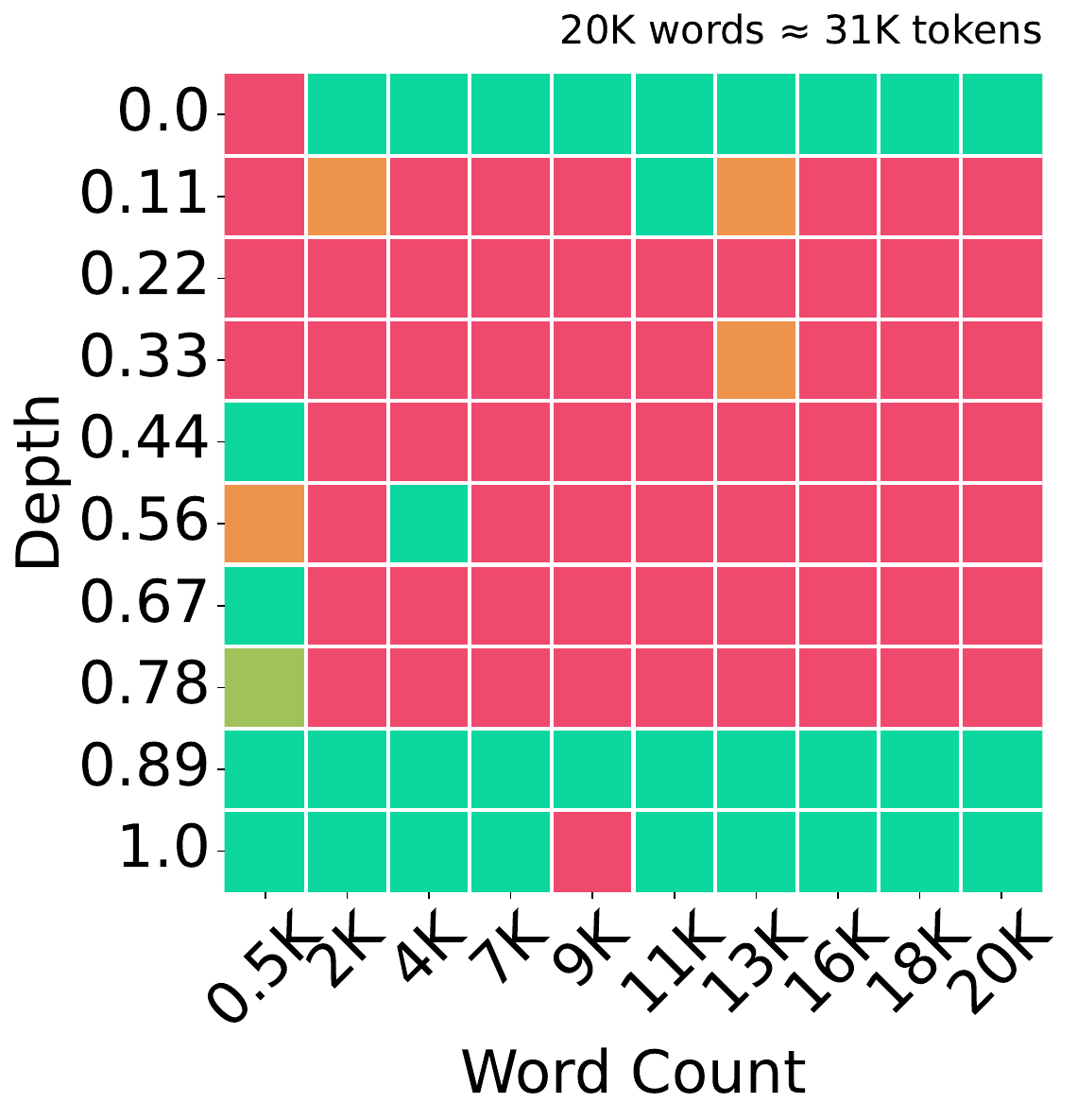}
	\end{minipage}%
}
\subfigure[6x Compression]{
\centering
	\begin{minipage}[t]{0.23\linewidth}
		\includegraphics[width=\linewidth]{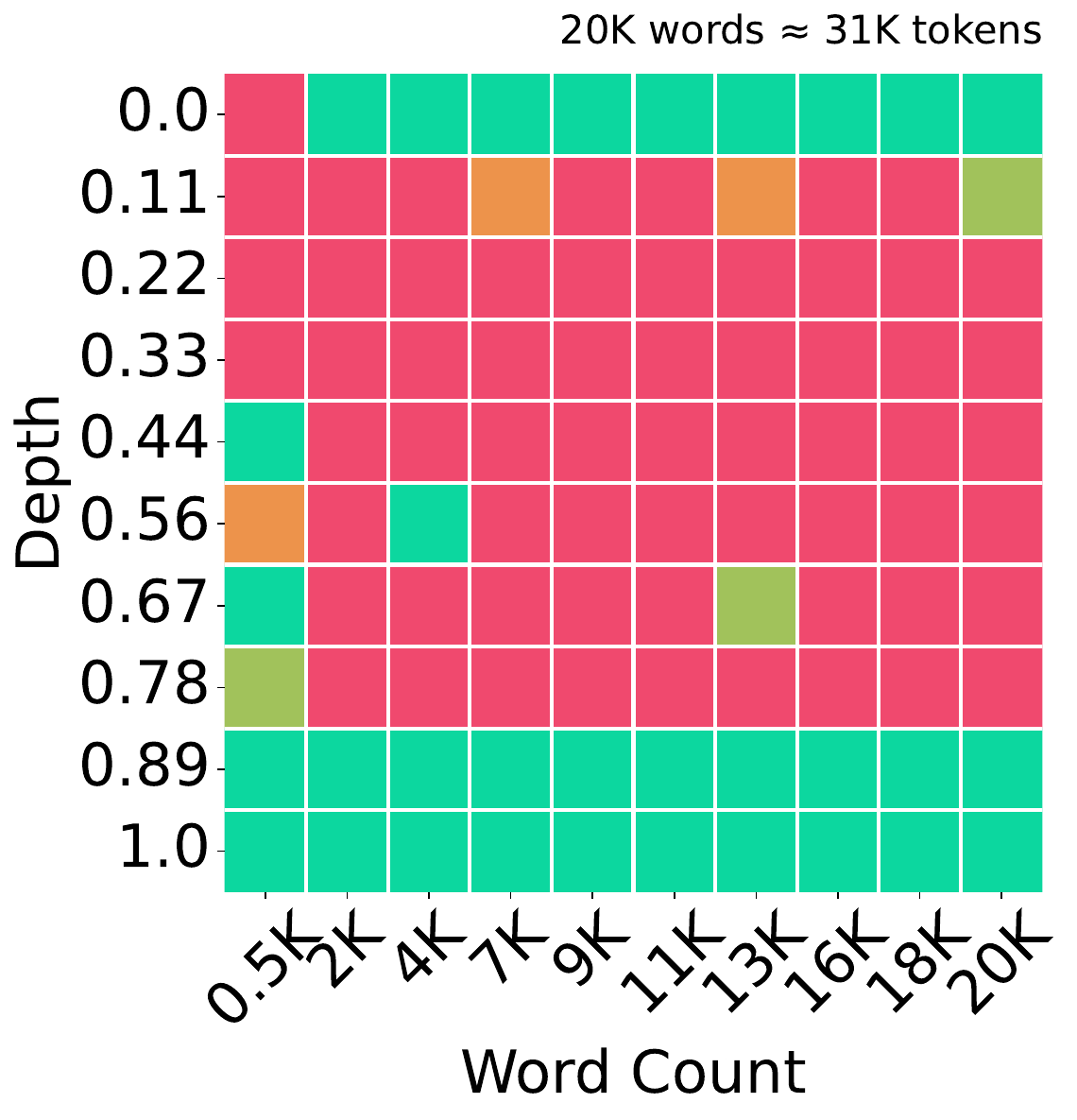}
	\end{minipage}%
}
\subfigure[8x Compression]{
\centering
	\begin{minipage}[t]{0.23\linewidth}
		\includegraphics[width=\linewidth]{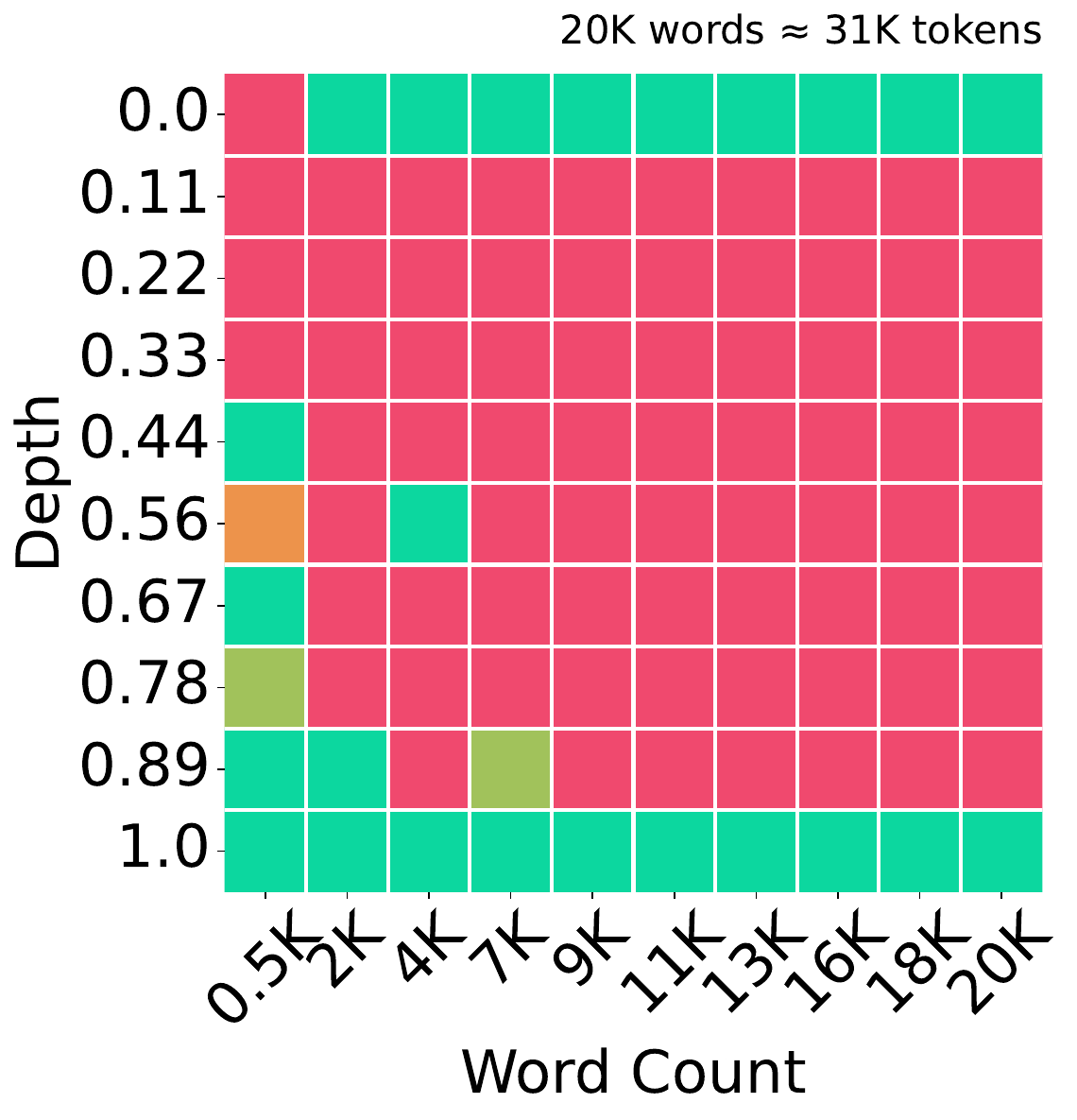}
	\end{minipage}%
}
\caption{InfLLM on LongChat-7B-v1.5-32k with 4 different compression rates under needle test}
\label{fig:needle_infllm_longchat}
\end{figure*}

\begin{figure*}[h]
\setlength{\abovecaptionskip}{0mm}
\setlength{\belowcaptionskip}{0mm}
\centering
\subfigcapskip=-2mm
\subfigure[2x Compression]{
\centering
	\begin{minipage}[t]{0.23\linewidth}
		\includegraphics[width=\linewidth]{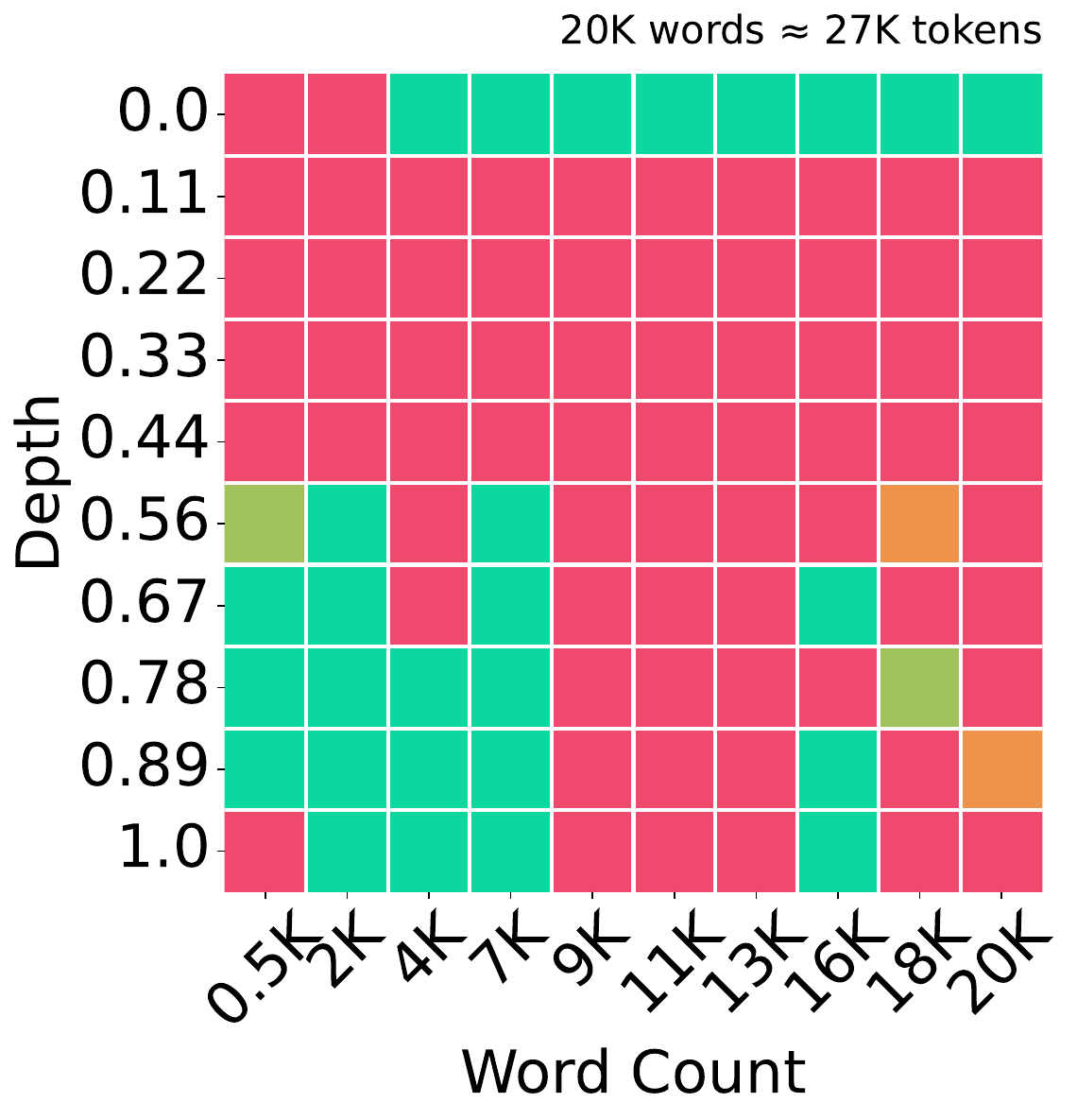}
	\end{minipage}%
}
\subfigure[4x Compression]{
\centering
	\begin{minipage}[t]{0.23\linewidth}
		\includegraphics[width=\linewidth]{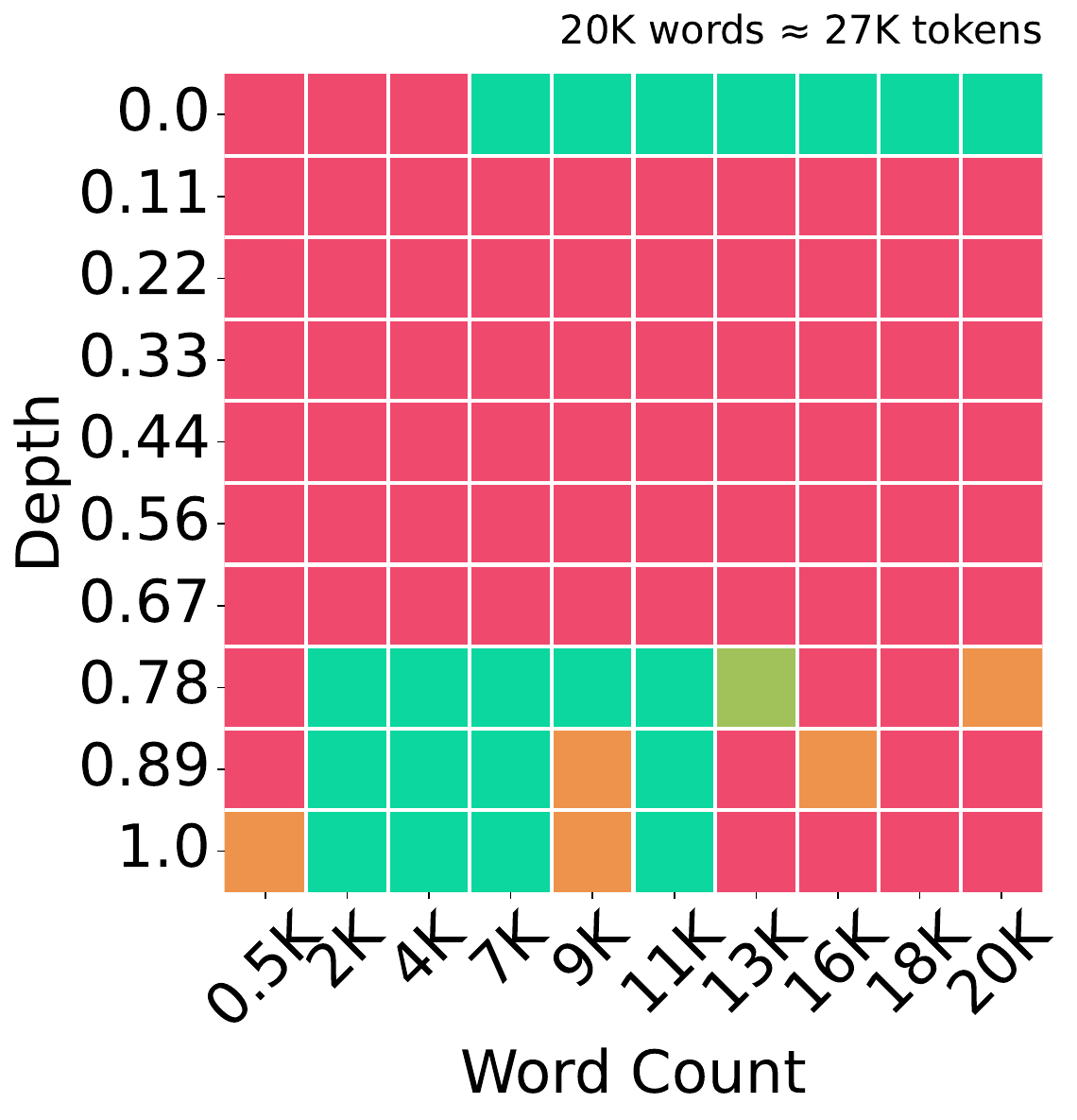}
	\end{minipage}%
}
\subfigure[6x Compression]{
\centering
	\begin{minipage}[t]{0.23\linewidth}
		\includegraphics[width=\linewidth]{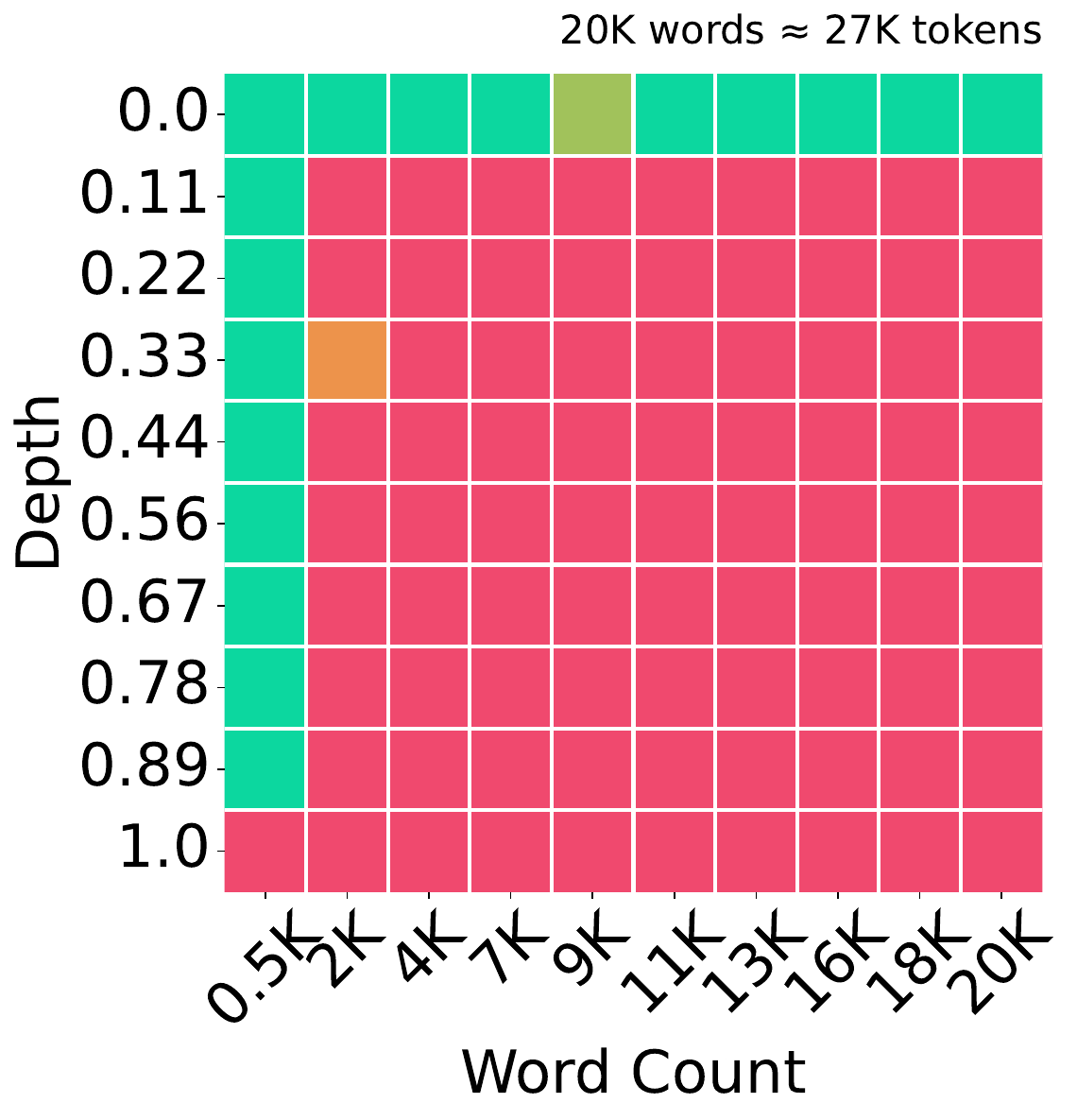}
	\end{minipage}%
}
\subfigure[8x Compression]{
\centering
	\begin{minipage}[t]{0.23\linewidth}
		\includegraphics[width=\linewidth]{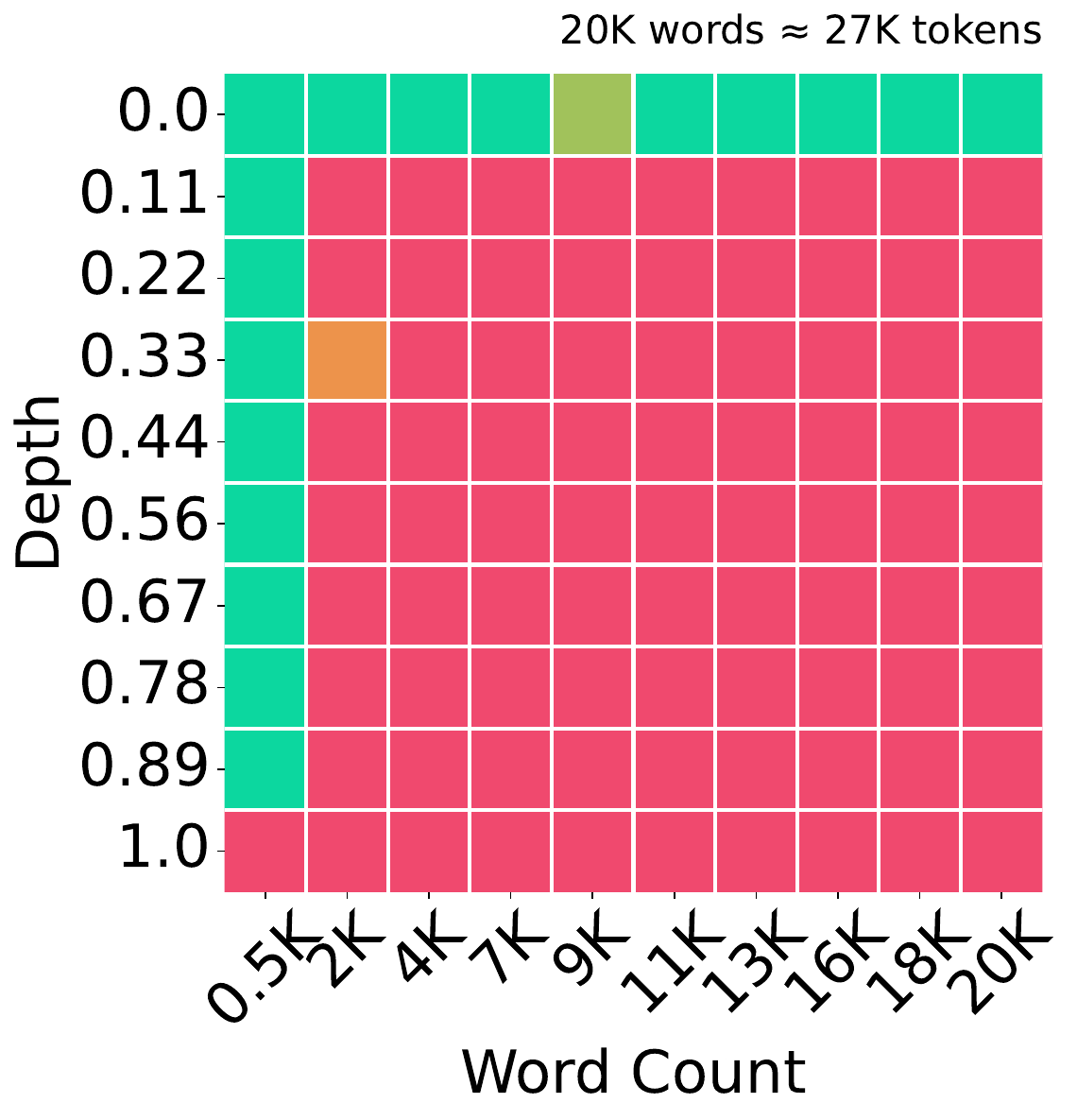}
	\end{minipage}%
}
\caption{StreamingLLM on Llama-3-8B-Instruct with 4 different compression rates under needle test}
\label{fig:needle_StreamingLLM_llama}
\end{figure*}

\begin{figure*}[h]
\setlength{\abovecaptionskip}{0mm}
\setlength{\belowcaptionskip}{0mm}
\centering
\subfigcapskip=-2mm
\subfigure[2x Compression]{
\centering
	\begin{minipage}[t]{0.23\linewidth}
		\includegraphics[width=\linewidth]{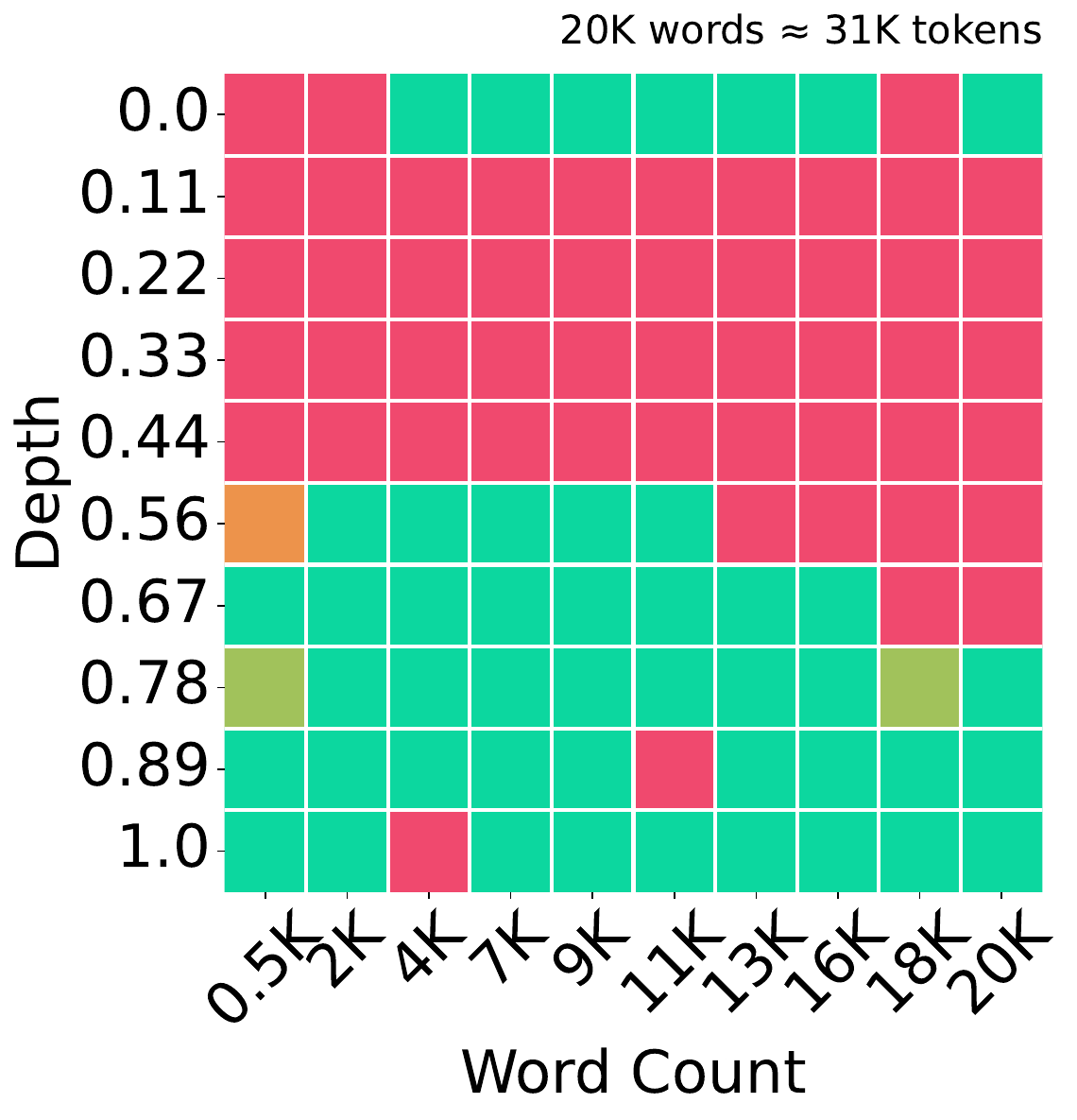}
	\end{minipage}%
}
\subfigure[4x Compression]{
\centering
	\begin{minipage}[t]{0.23\linewidth}
		\includegraphics[width=\linewidth]{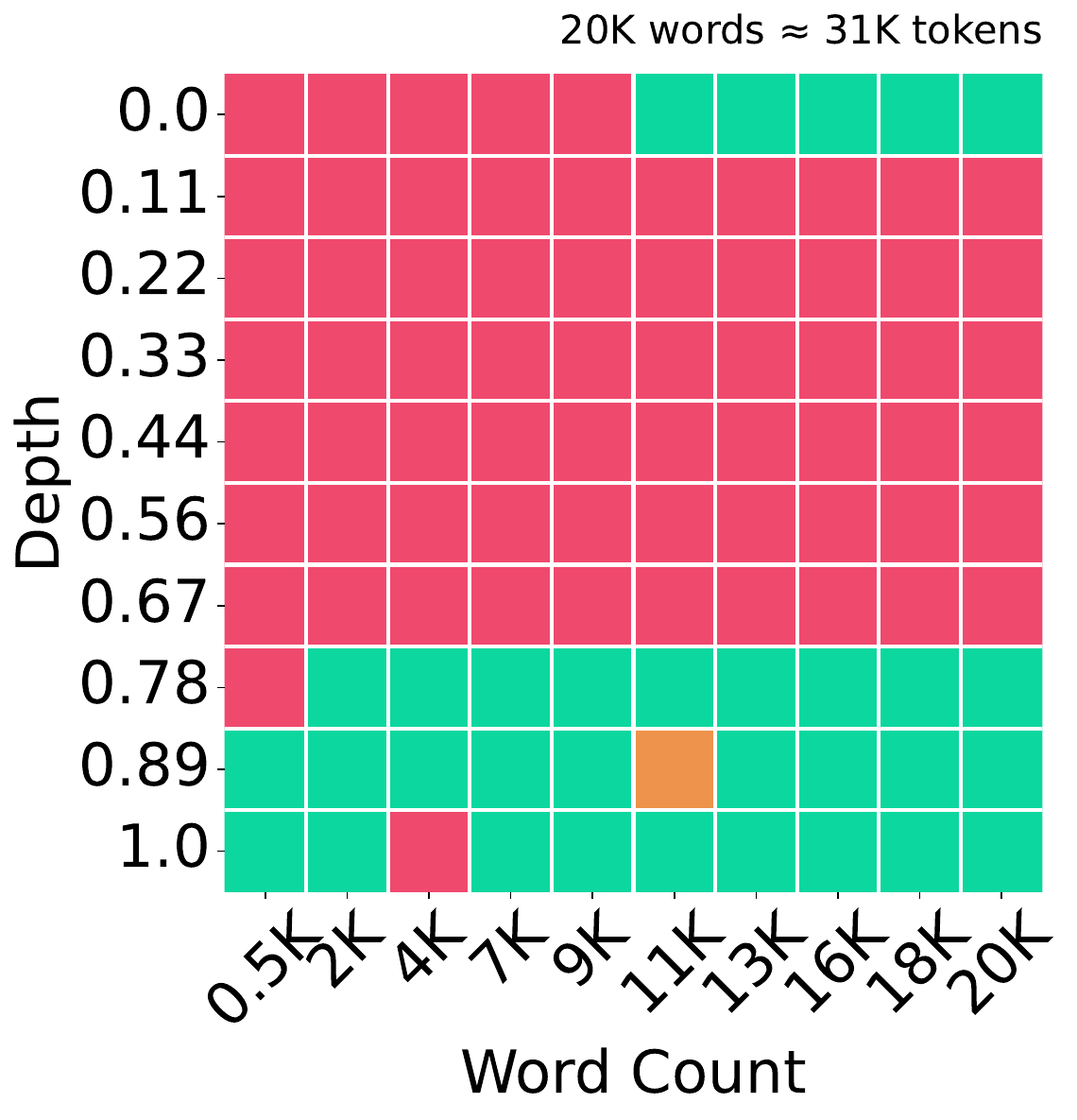}
	\end{minipage}%
}
\subfigure[6x Compression]{
\centering
	\begin{minipage}[t]{0.23\linewidth}
		\includegraphics[width=\linewidth]{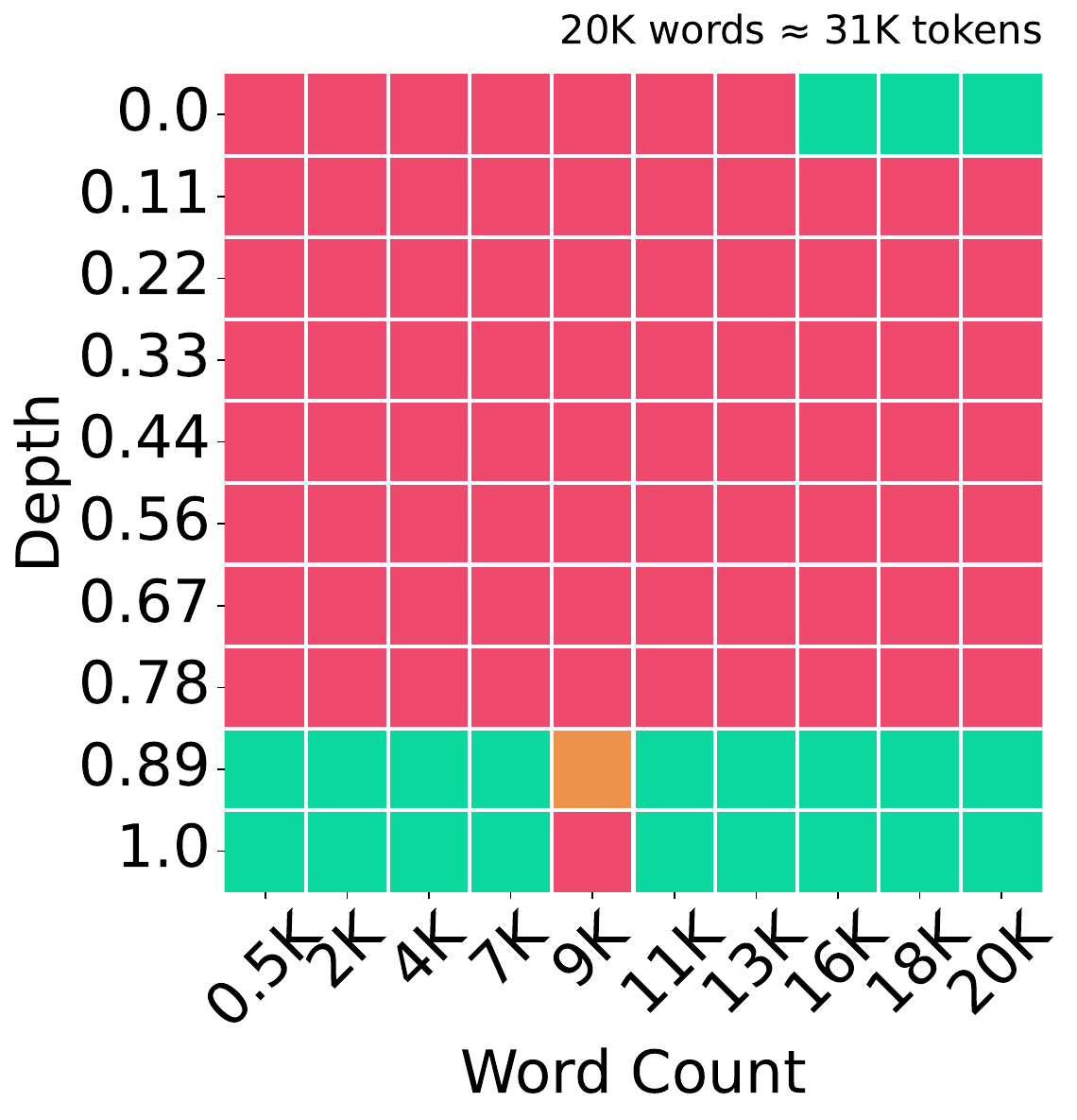}
	\end{minipage}%
}
\subfigure[8x Compression]{
\centering
	\begin{minipage}[t]{0.23\linewidth}
		\includegraphics[width=\linewidth]{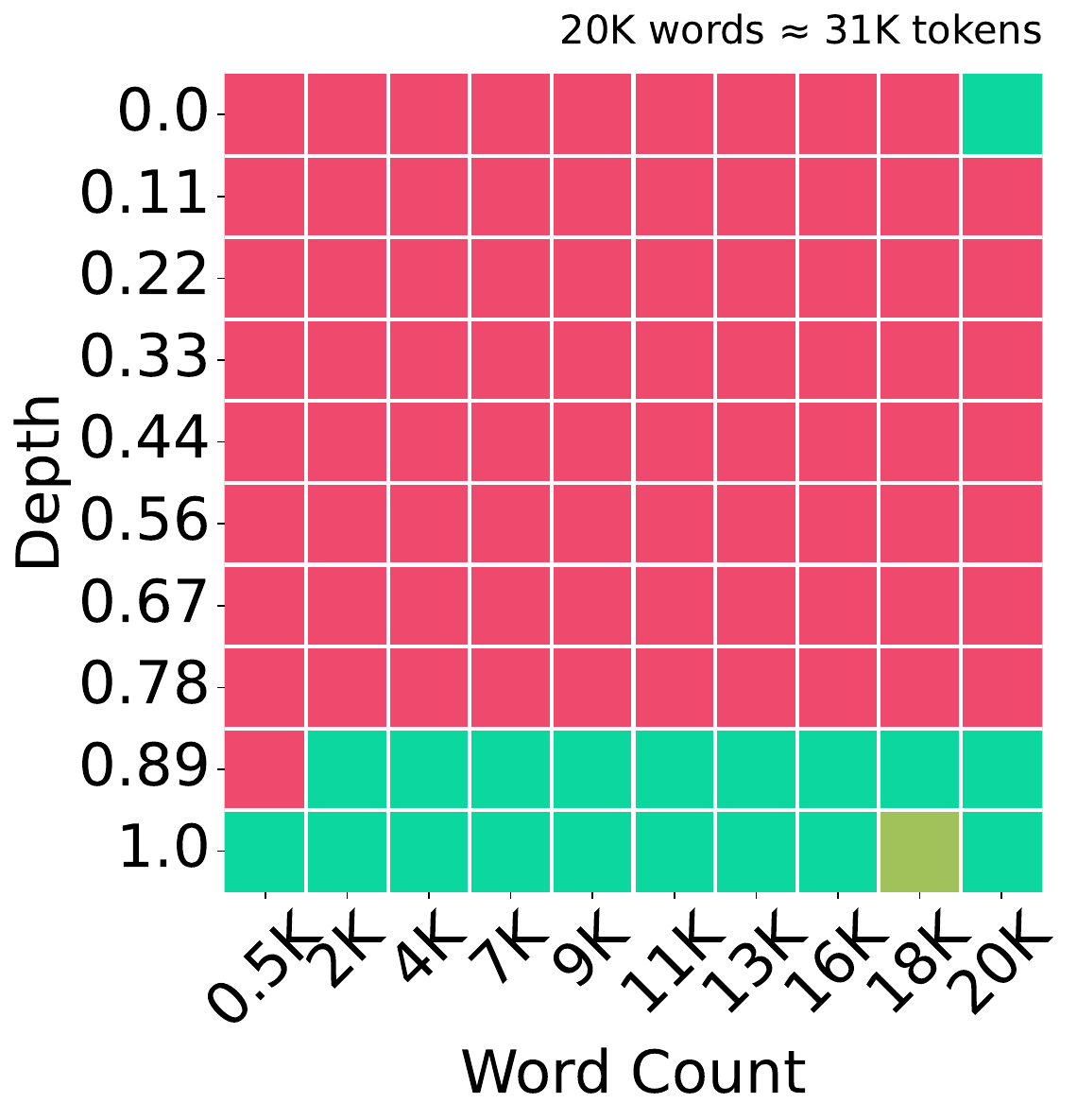}
	\end{minipage}%
}
\caption{StreamingLLM on LongChat-7B-v1.5-32k with 4 different compression rates under needle test}
\label{fig:needle_StreamingLLM_longchat}
\end{figure*}

\begin{figure*}[h]
\setlength{\abovecaptionskip}{0mm}
\setlength{\belowcaptionskip}{0mm}
\centering
\subfigcapskip=-2mm
\subfigure[2x Compression]{
\centering
	\begin{minipage}[t]{0.23\linewidth}
		\includegraphics[width=\linewidth]{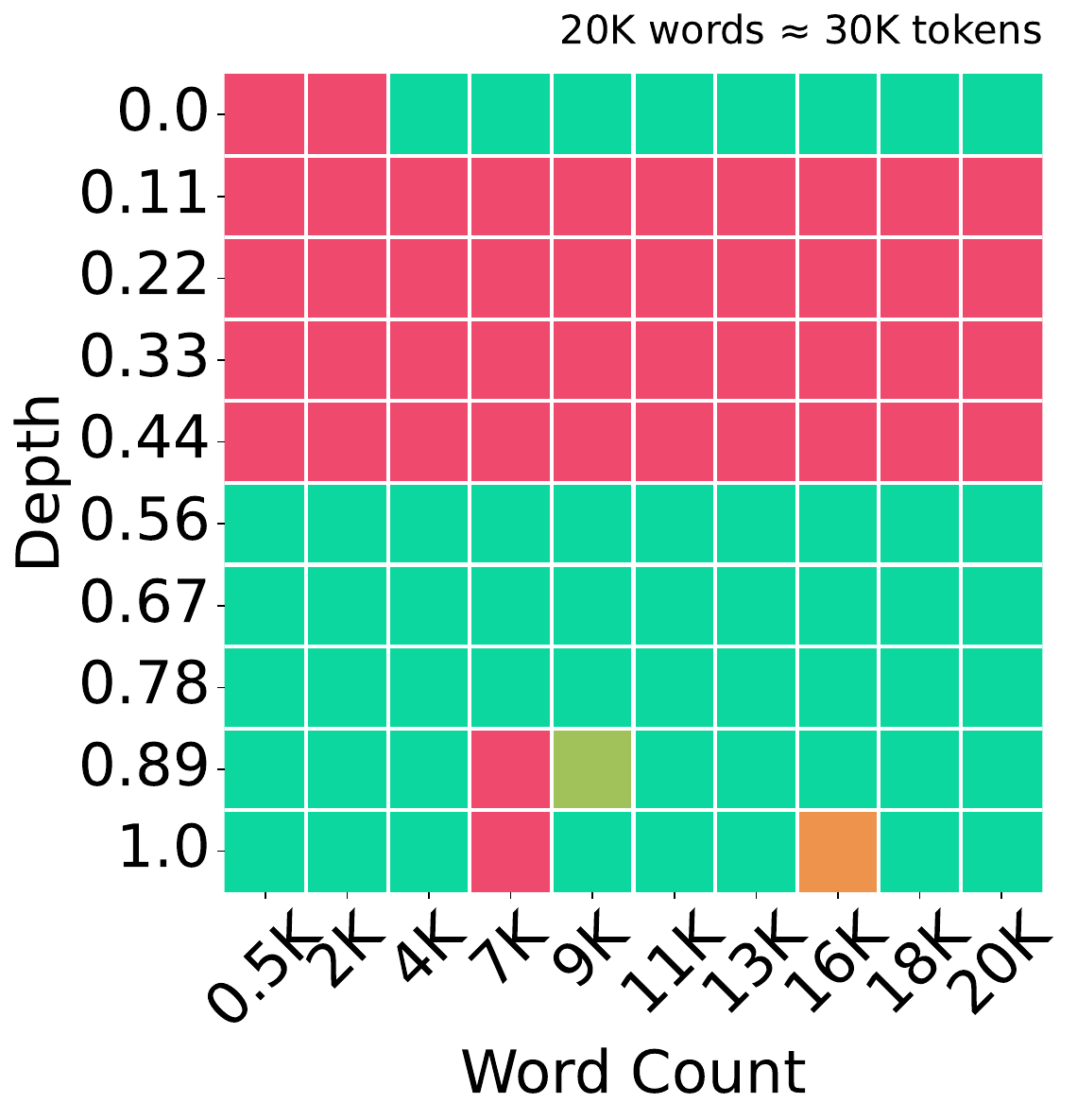}
	\end{minipage}%
}
\subfigure[4x Compression]{
\centering
	\begin{minipage}[t]{0.23\linewidth}
		\includegraphics[width=\linewidth]{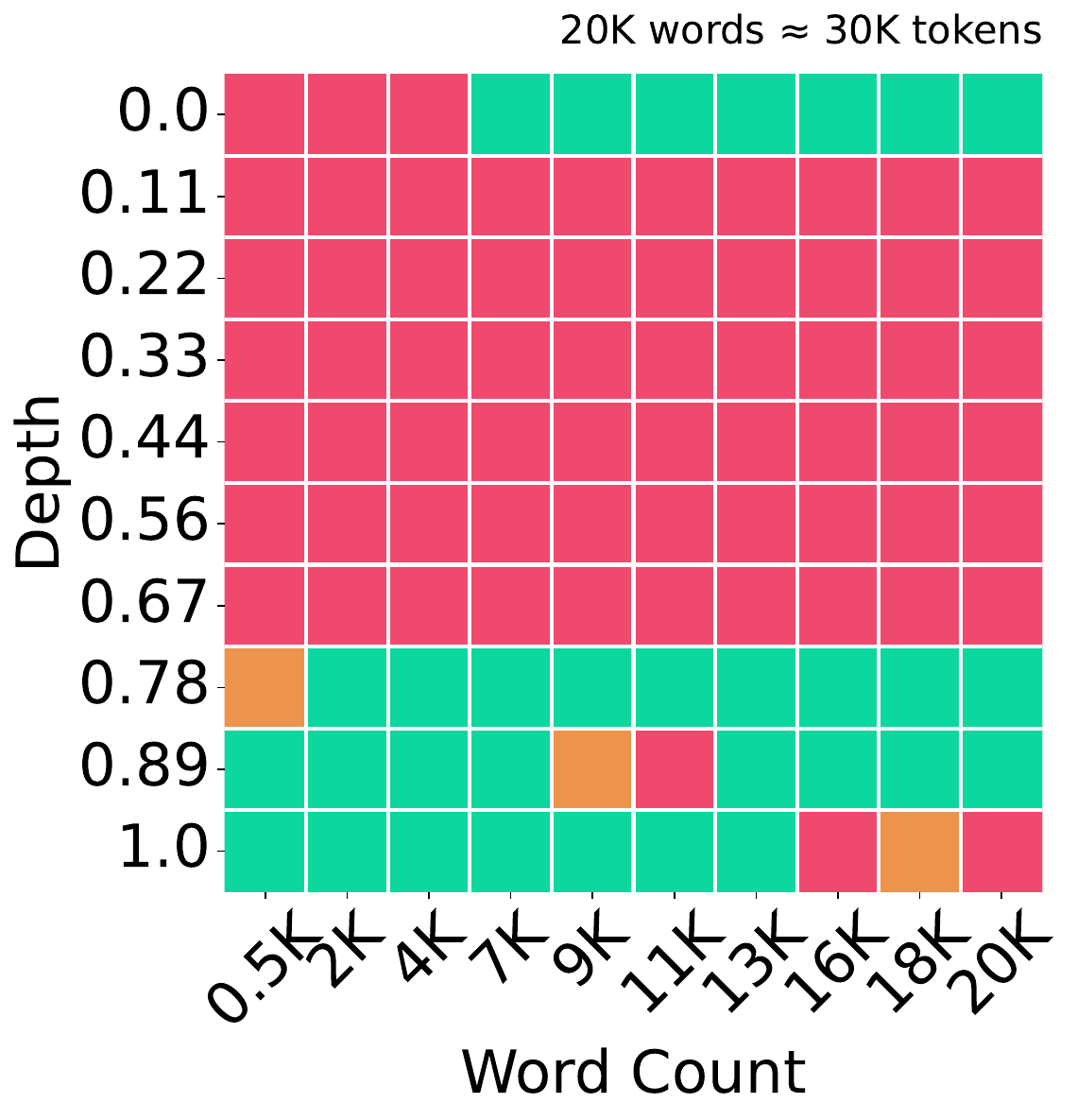}
	\end{minipage}%
}
\subfigure[6x Compression]{
\centering
	\begin{minipage}[t]{0.23\linewidth}
		\includegraphics[width=\linewidth]{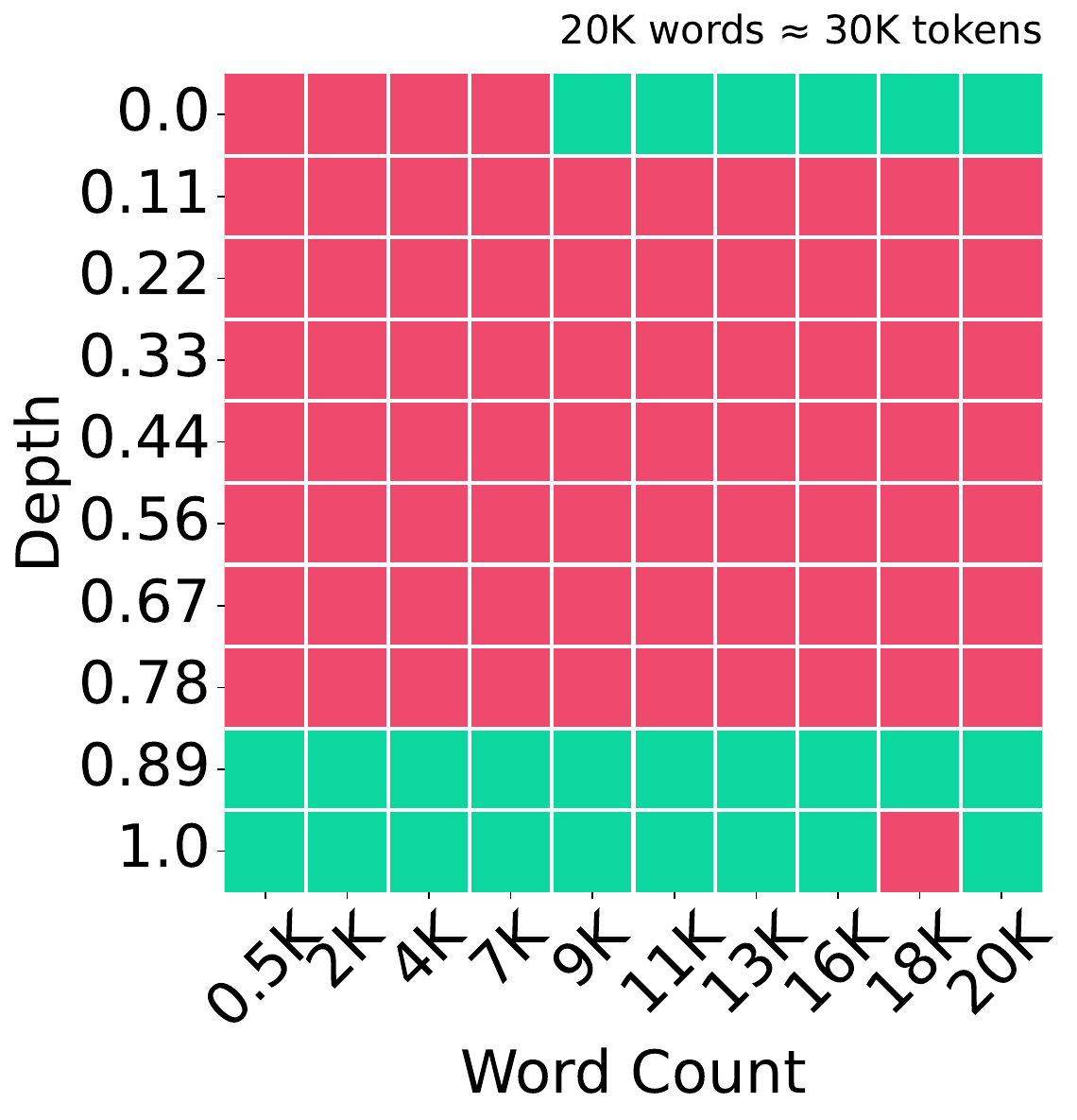}
	\end{minipage}%
}
\subfigure[8x Compression]{
\centering
	\begin{minipage}[t]{0.23\linewidth}
		\includegraphics[width=\linewidth]{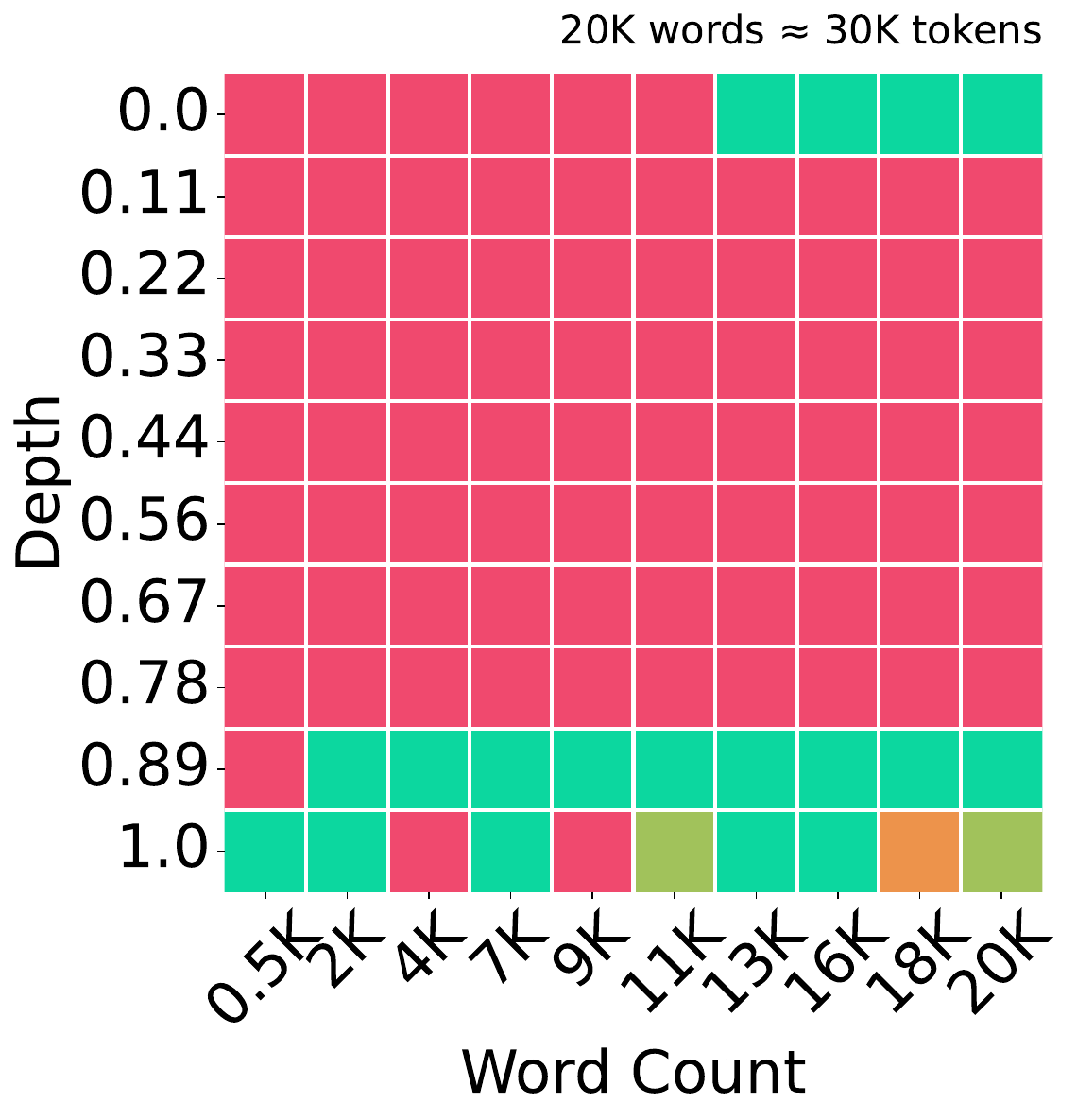}
	\end{minipage}%
}
\caption{StreamingLLM on Mistral-7B-v0.2-Instruct with 4 different compression rates under needle test}
\label{fig:needle_streamllm_mistral}
\end{figure*}

\begin{figure*}[h]
\setlength{\abovecaptionskip}{0mm}
\setlength{\belowcaptionskip}{0mm}
\centering
\subfigcapskip=-2mm
\subfigure[2x Compression]{
\centering
	\begin{minipage}[t]{0.23\linewidth}
		\includegraphics[width=\linewidth]{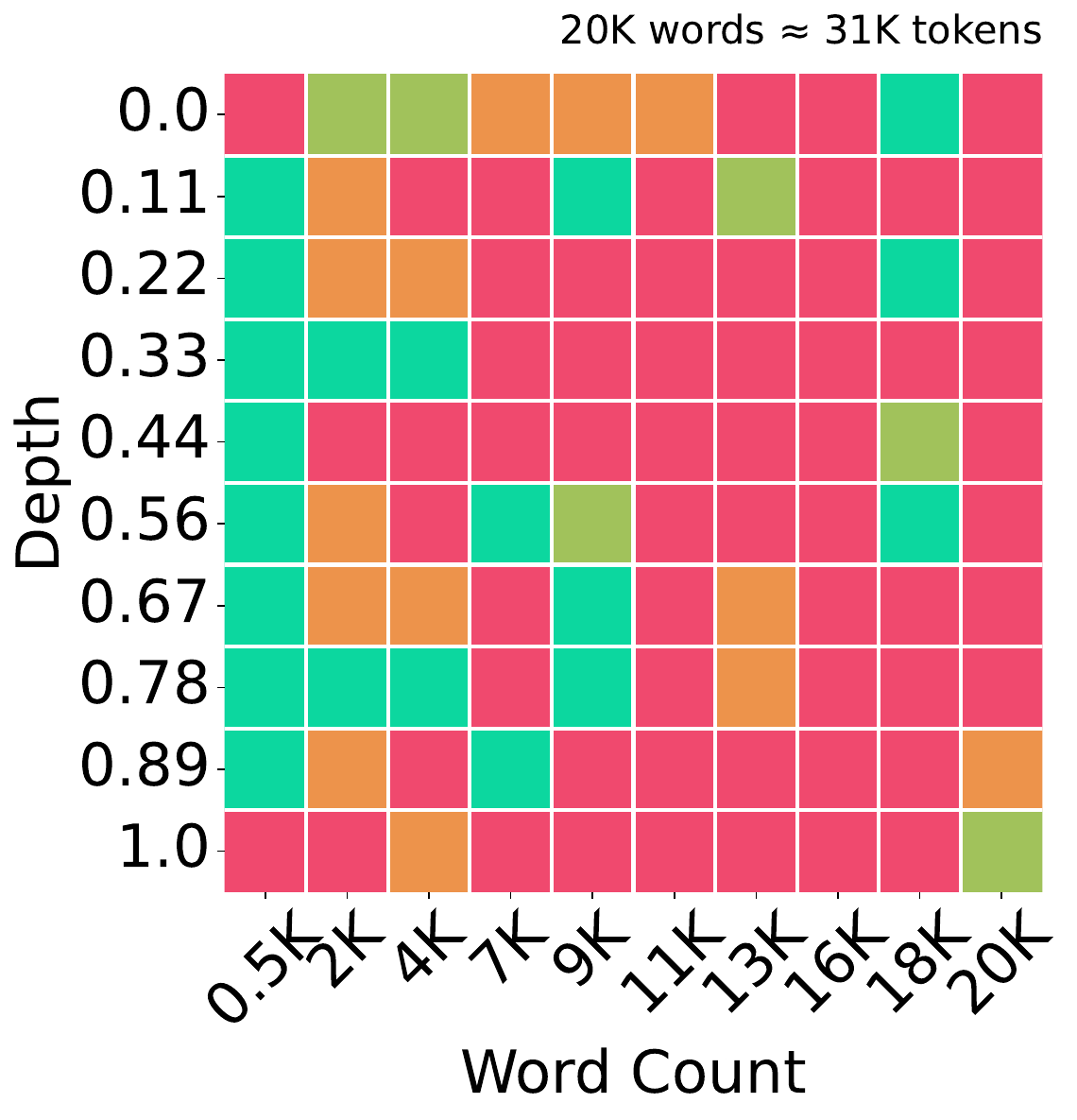}
	\end{minipage}%
}
\subfigure[4x Compression]{
\centering
	\begin{minipage}[t]{0.23\linewidth}
		\includegraphics[width=\linewidth]{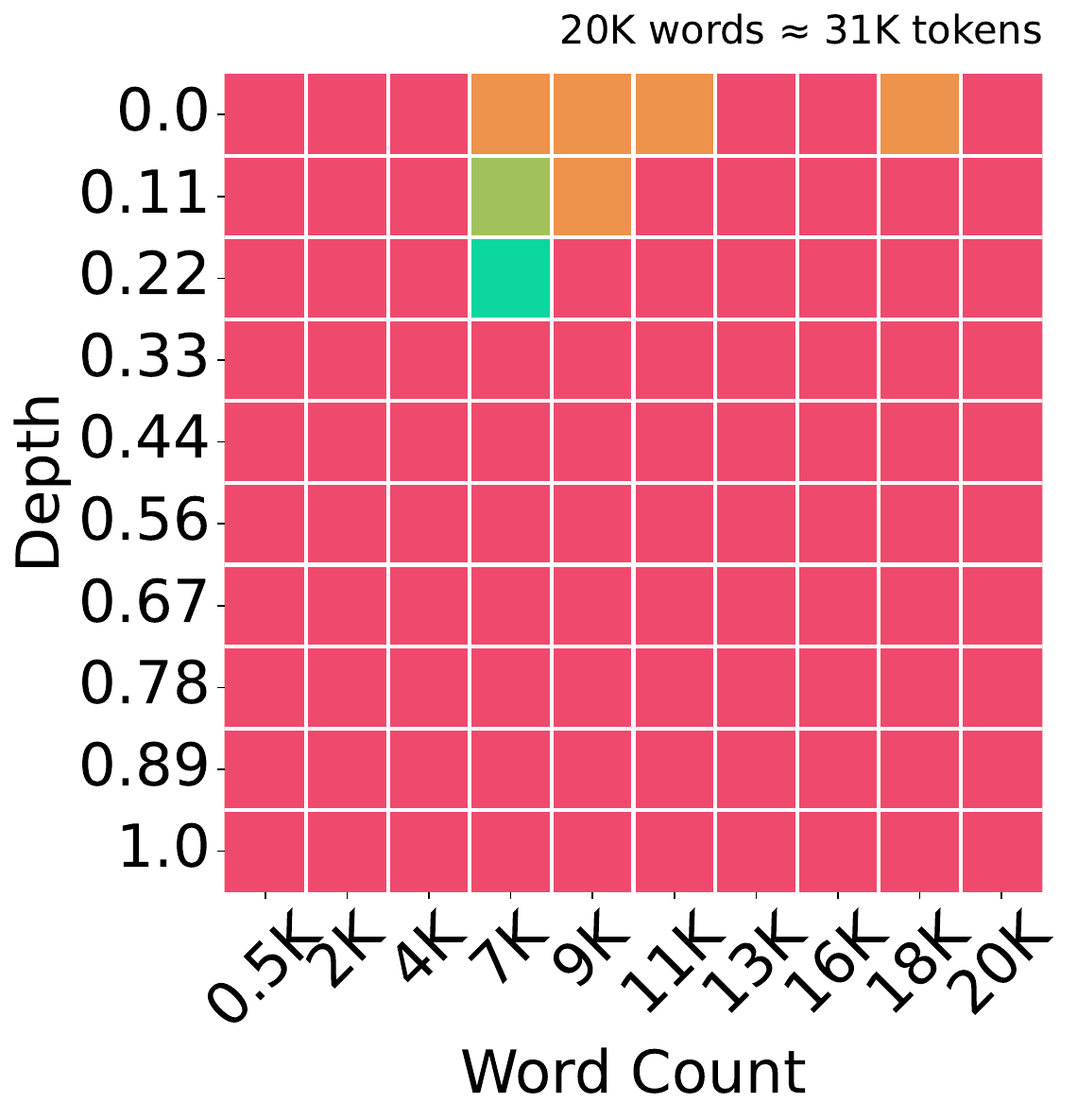}
	\end{minipage}%
}
\subfigure[6x Compression]{
\centering
	\begin{minipage}[t]{0.23\linewidth}
		\includegraphics[width=\linewidth]{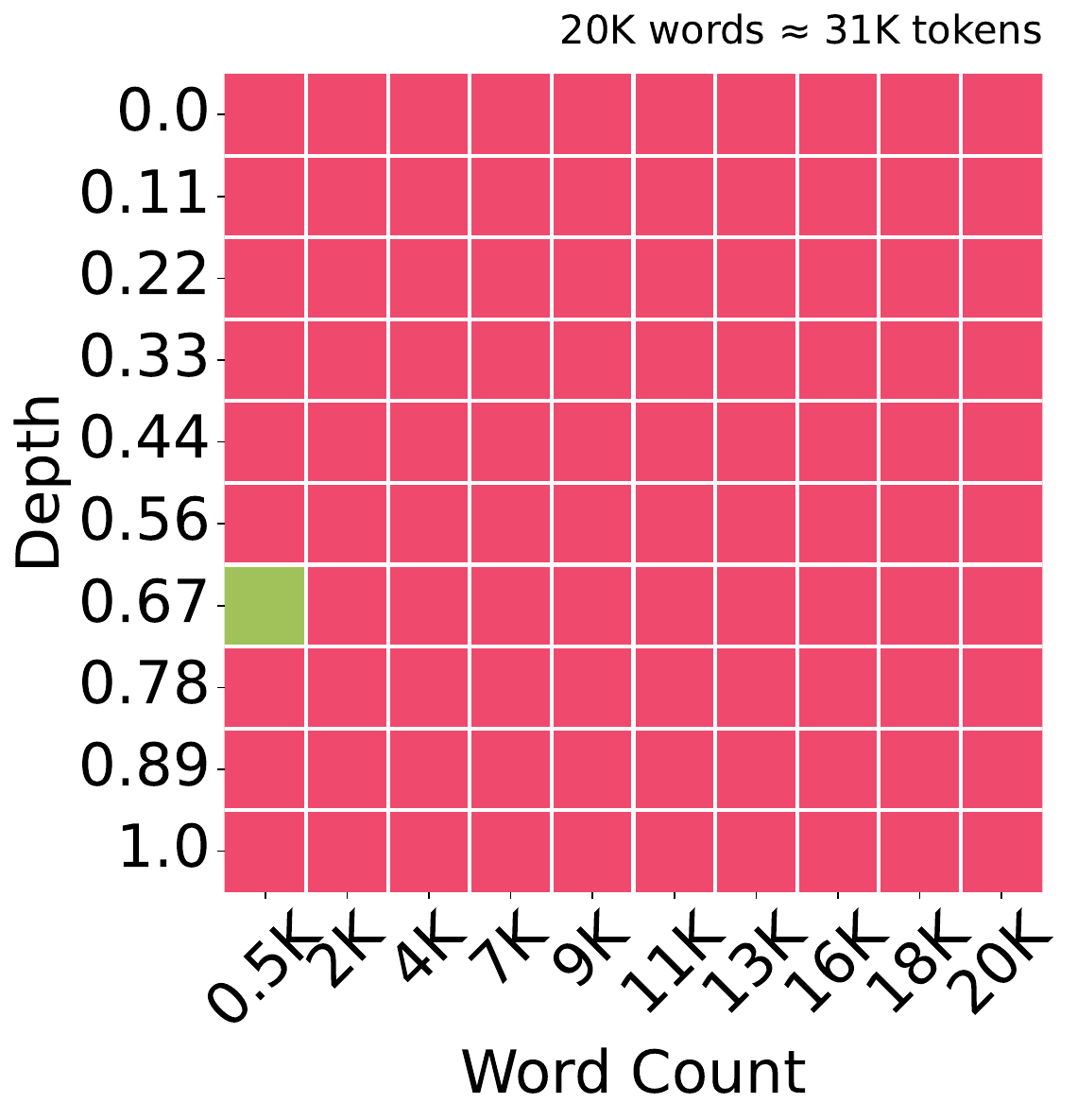}
	\end{minipage}%
}
\subfigure[8x Compression]{
\centering
	\begin{minipage}[t]{0.23\linewidth}
		\includegraphics[width=\linewidth]{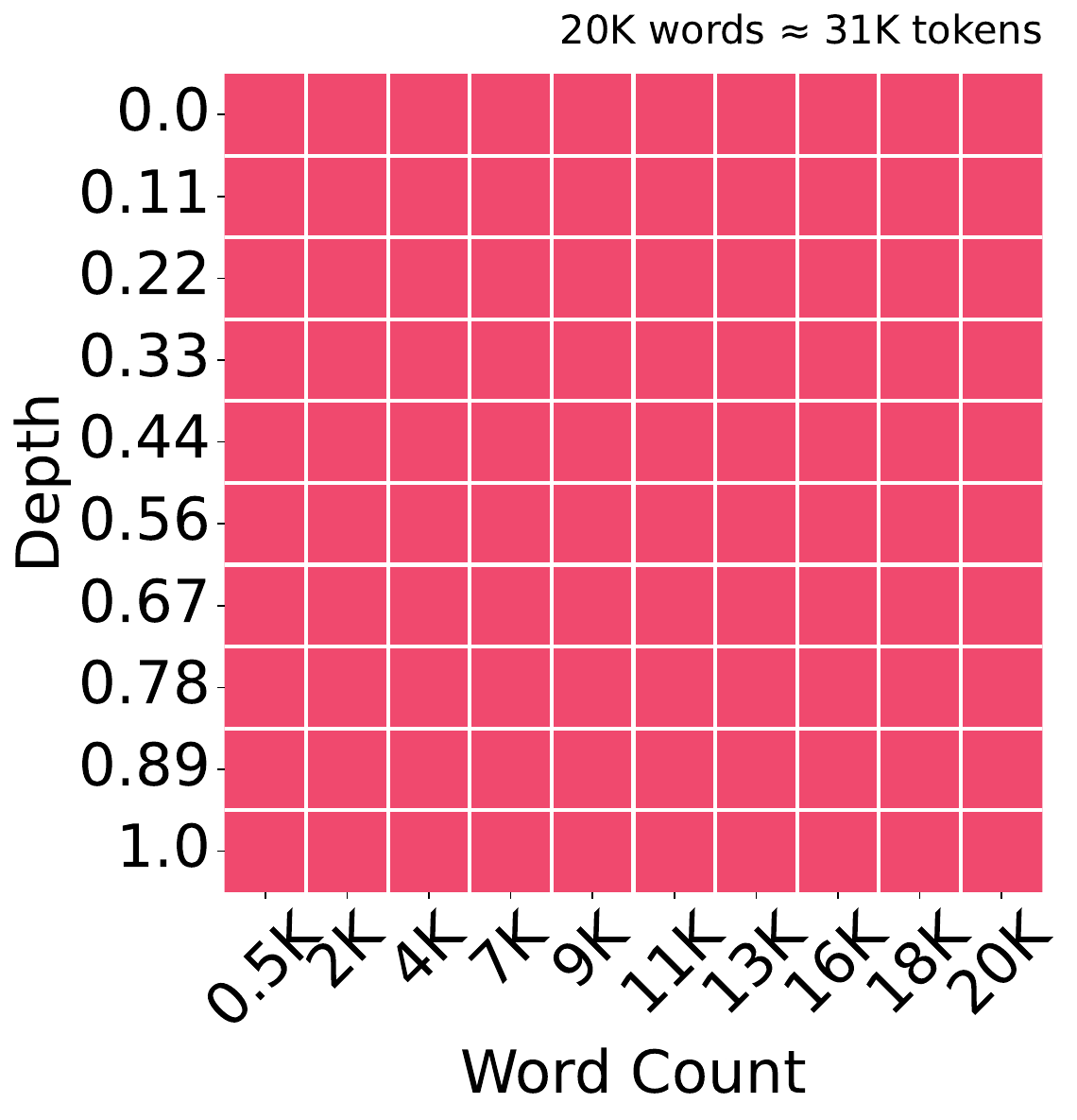}
	\end{minipage}%
}
\caption{LLMLingua on LongChat-7B-v1.5-32k with 4 different compression rates under needle test}
\label{fig:needle_llmlingua_longchat}
\end{figure*}

\begin{figure*}[h]
\setlength{\abovecaptionskip}{0mm}
\setlength{\belowcaptionskip}{0mm}
\centering
\subfigcapskip=-2mm
\subfigure[2x Compression]{
\centering
	\begin{minipage}[t]{0.23\linewidth}
		\includegraphics[width=\linewidth]{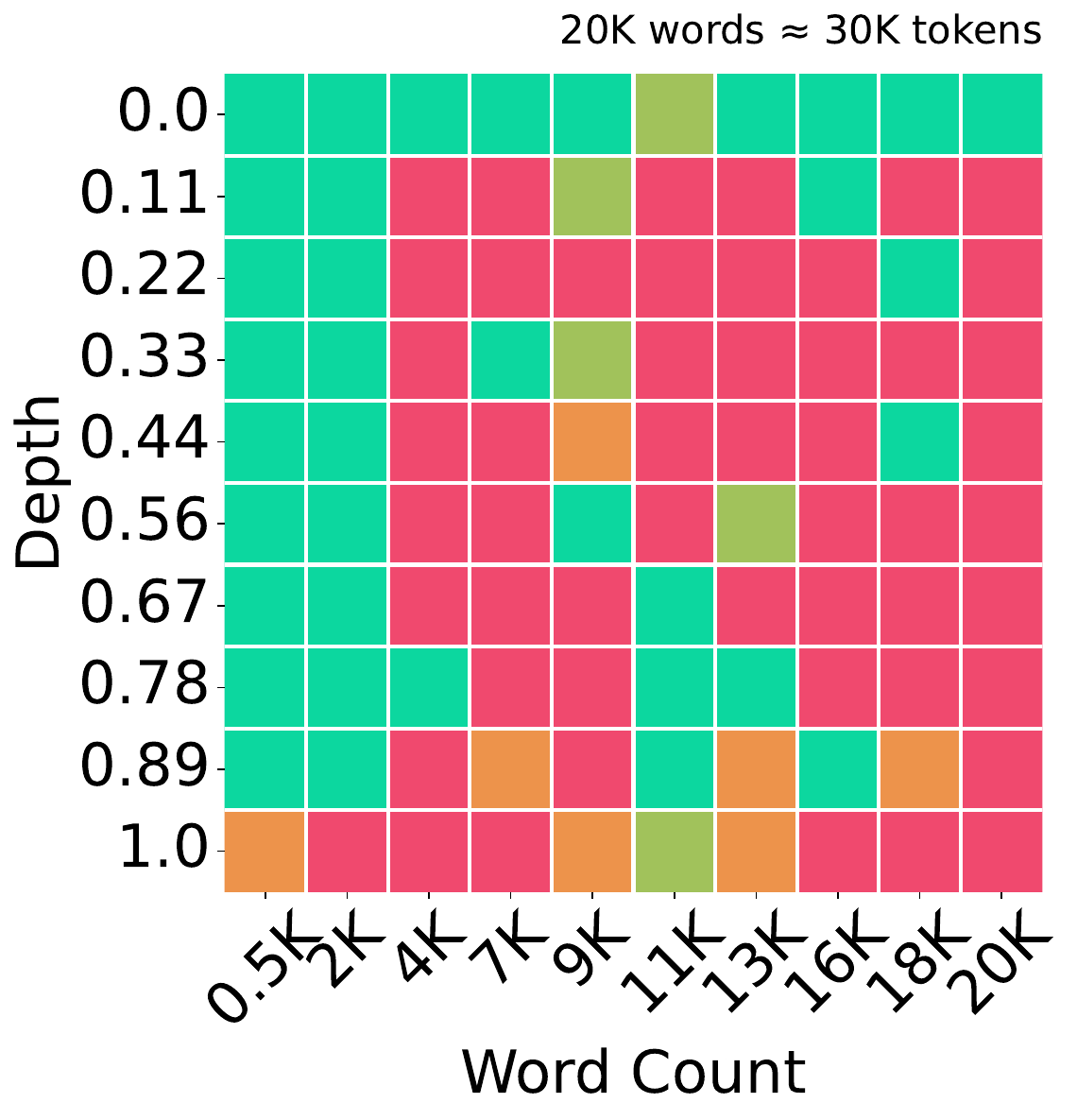}
	\end{minipage}%
}
\subfigure[4x Compression]{
\centering
	\begin{minipage}[t]{0.23\linewidth}
		\includegraphics[width=\linewidth]{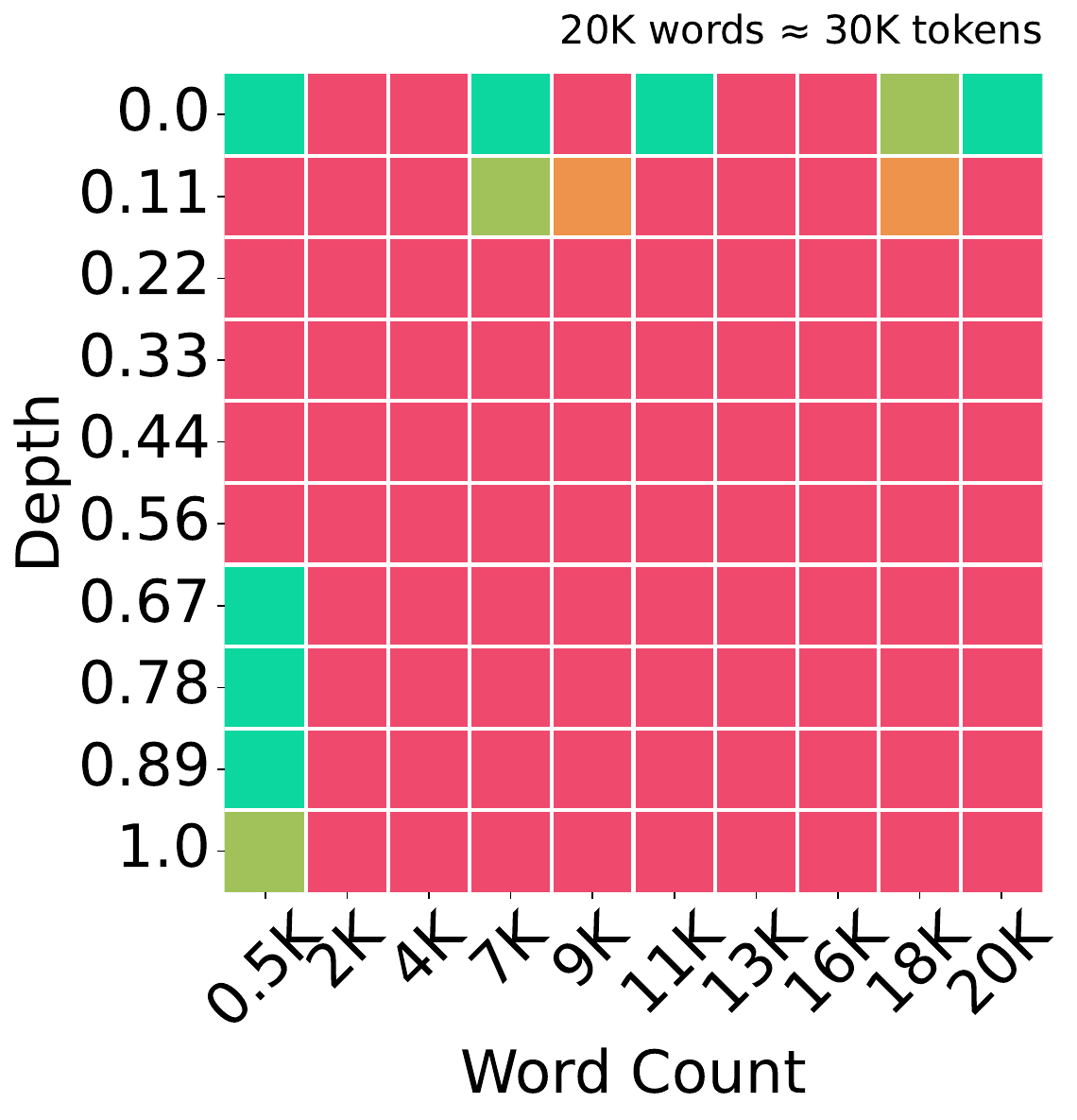}
	\end{minipage}%
}
\subfigure[6x Compression]{
\centering
	\begin{minipage}[t]{0.23\linewidth}
		\includegraphics[width=\linewidth]{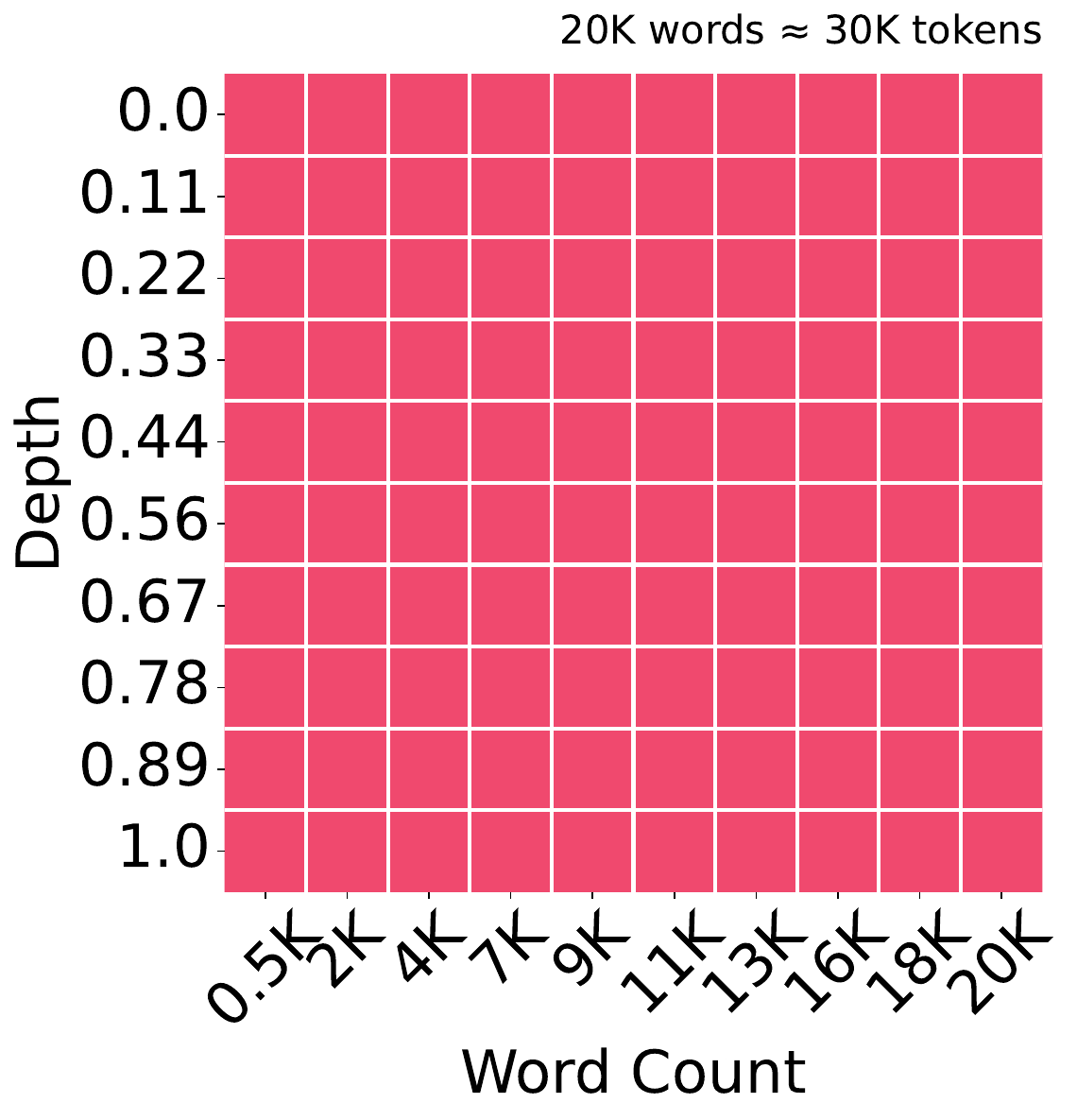}
	\end{minipage}%
}
\subfigure[8x Compression]{
\centering
	\begin{minipage}[t]{0.23\linewidth}
		\includegraphics[width=\linewidth]{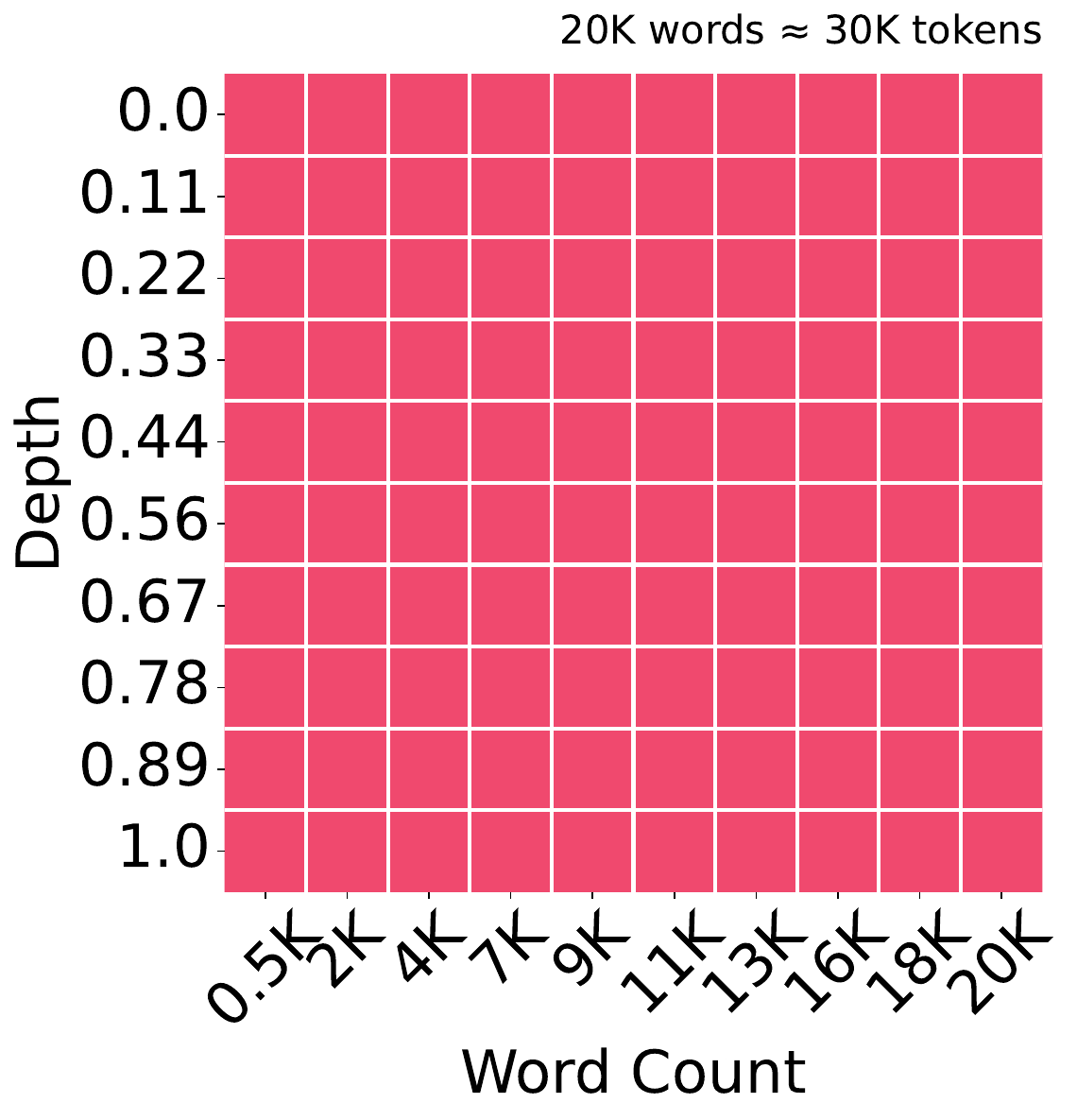}
	\end{minipage}%
}
\caption{LLMLingua on Mistral-7B-v0.2-Instruct with 4 different compression rates under needle test}
\label{fig:needle_llmlingua_mistral}
\end{figure*}

\begin{figure*}[h]
\setlength{\abovecaptionskip}{0mm}
\setlength{\belowcaptionskip}{0mm}
\centering
\subfigcapskip=-2mm
\subfigure[2x Compression]{
\centering
	\begin{minipage}[t]{0.23\linewidth}
		\includegraphics[width=\linewidth]{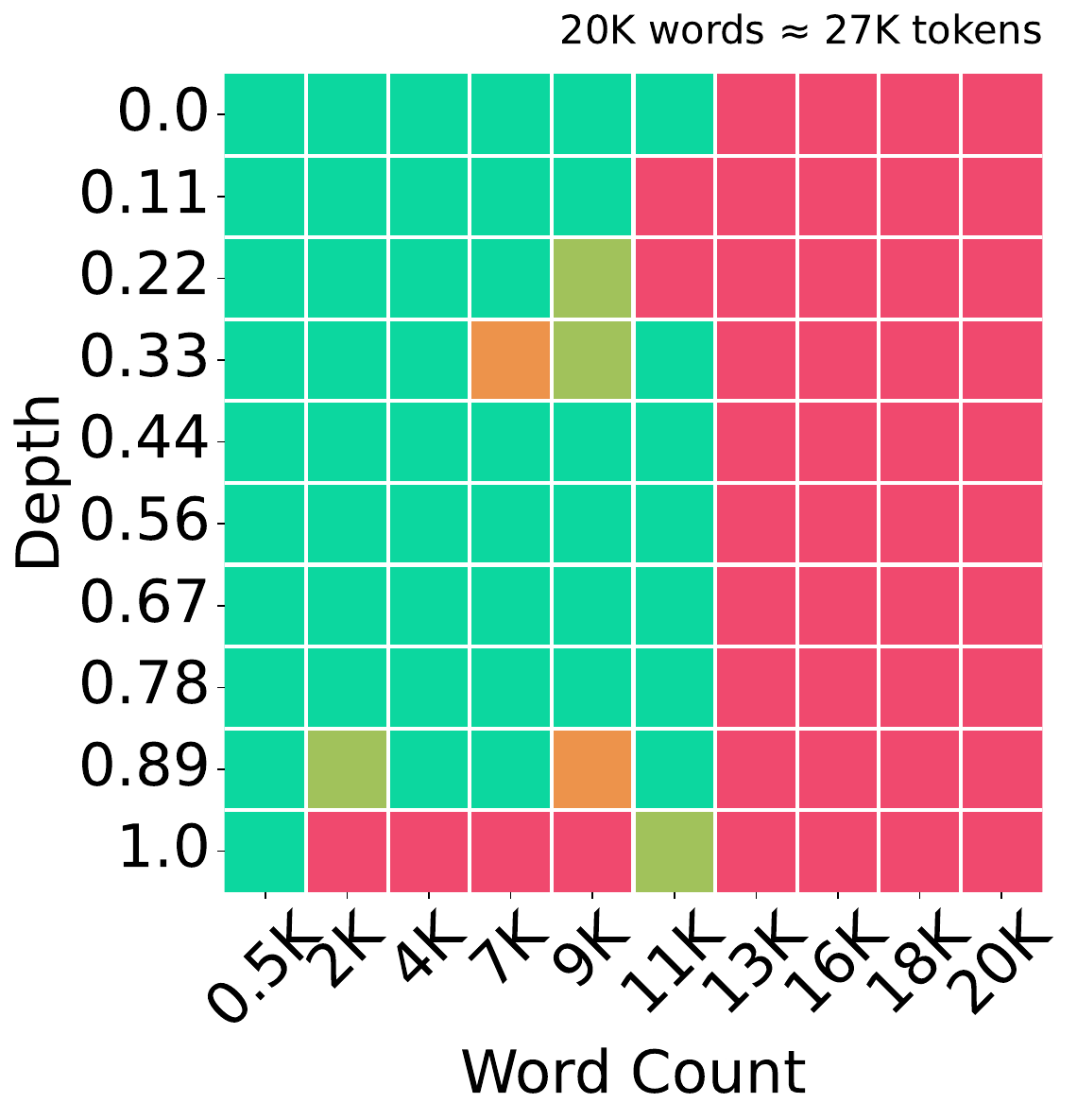}
	\end{minipage}%
}
\subfigure[4x Compression]{
\centering
	\begin{minipage}[t]{0.23\linewidth}
		\includegraphics[width=\linewidth]{figures/needle/llmlingua/llama/4x.pdf}
	\end{minipage}%
}
\subfigure[6x Compression]{
\centering
	\begin{minipage}[t]{0.23\linewidth}
		\includegraphics[width=\linewidth]{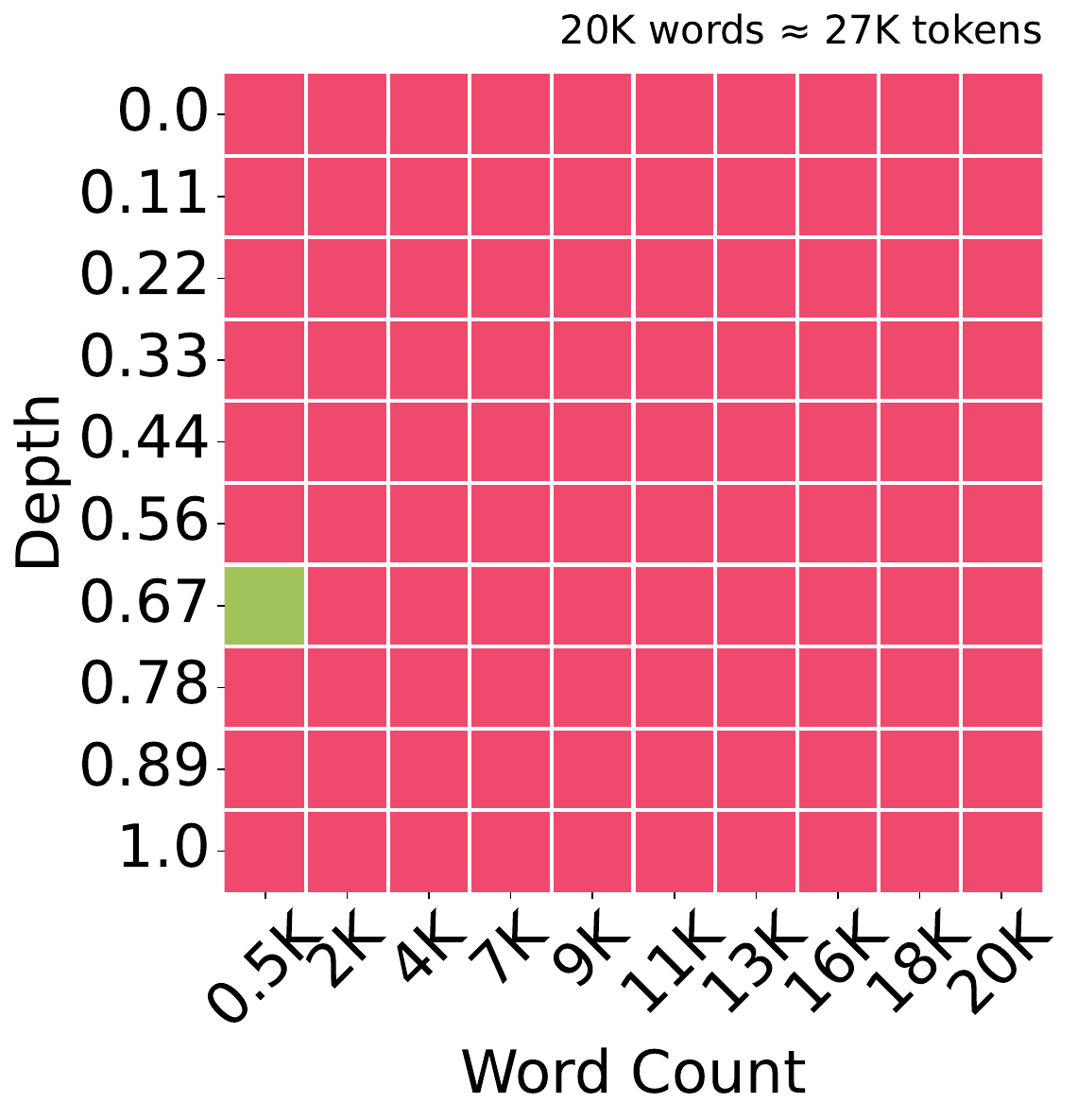}
	\end{minipage}%
}
\subfigure[8x Compression]{
\centering
	\begin{minipage}[t]{0.23\linewidth}
		\includegraphics[width=\linewidth]{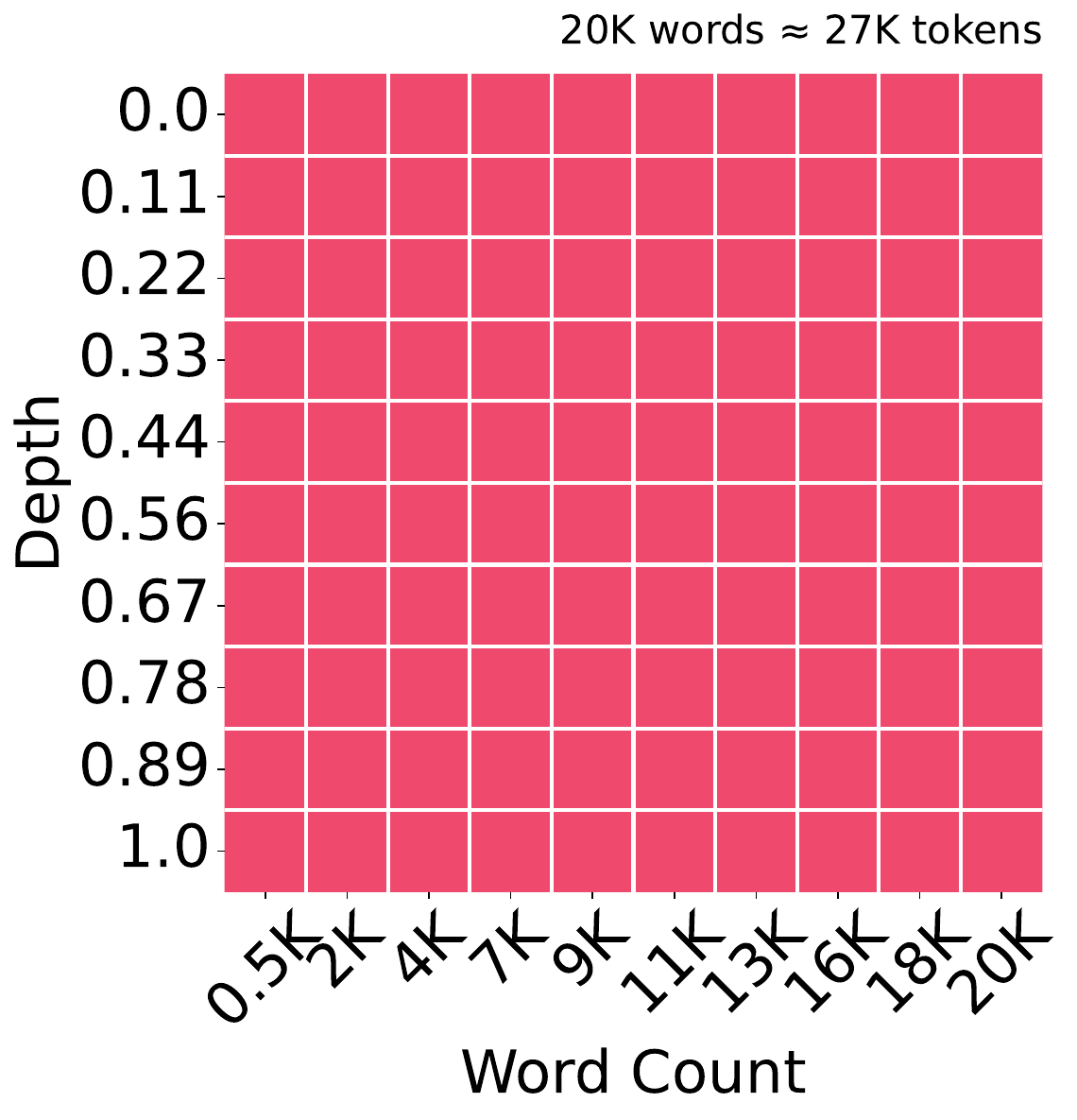}
	\end{minipage}%
}
\caption{LLMLingua on Llama-3-8B-Instruct with 4 different compression rates under needle test}
\label{fig:needle_llmlingua_llama}
\end{figure*}

\begin{figure*}[h]
\setlength{\abovecaptionskip}{0mm}
\setlength{\belowcaptionskip}{0mm}
\centering
\subfigcapskip=-2mm

\subfigure[Mamba-2.8B]{
\centering
    \begin{minipage}[t]{0.30\linewidth}
        \includegraphics[width=\linewidth]{figures/needle/mamba.pdf}
    \end{minipage}%
}
\subfigure[Mamba-Chat-2.8B]{
\centering
    \begin{minipage}[t]{0.30\linewidth}
        \includegraphics[width=\linewidth]{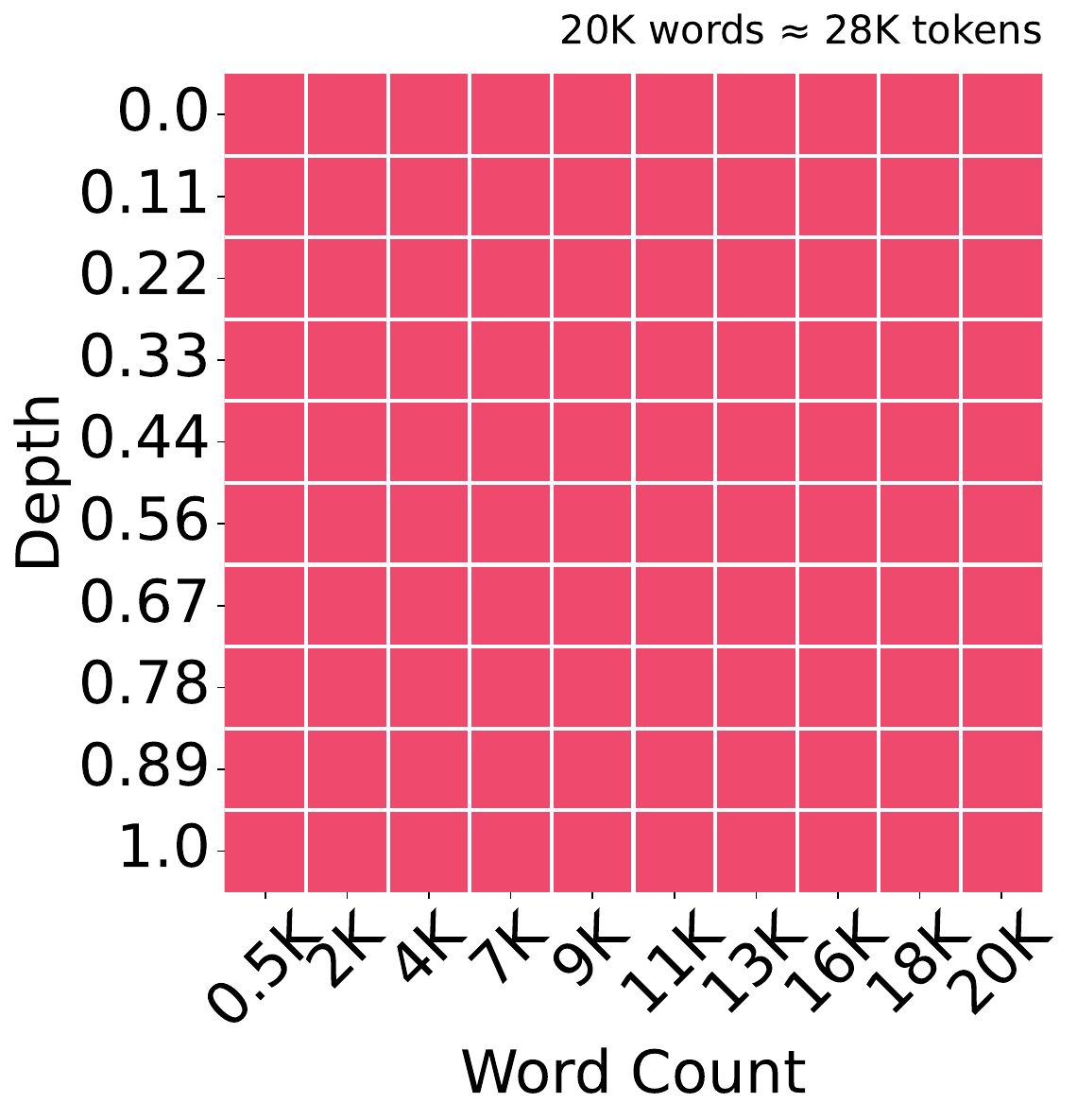}
    \end{minipage}%
}
\subfigure[RWKV-5-World-7B]{
\centering
    \begin{minipage}[t]{0.30\linewidth}
        \includegraphics[width=\linewidth]{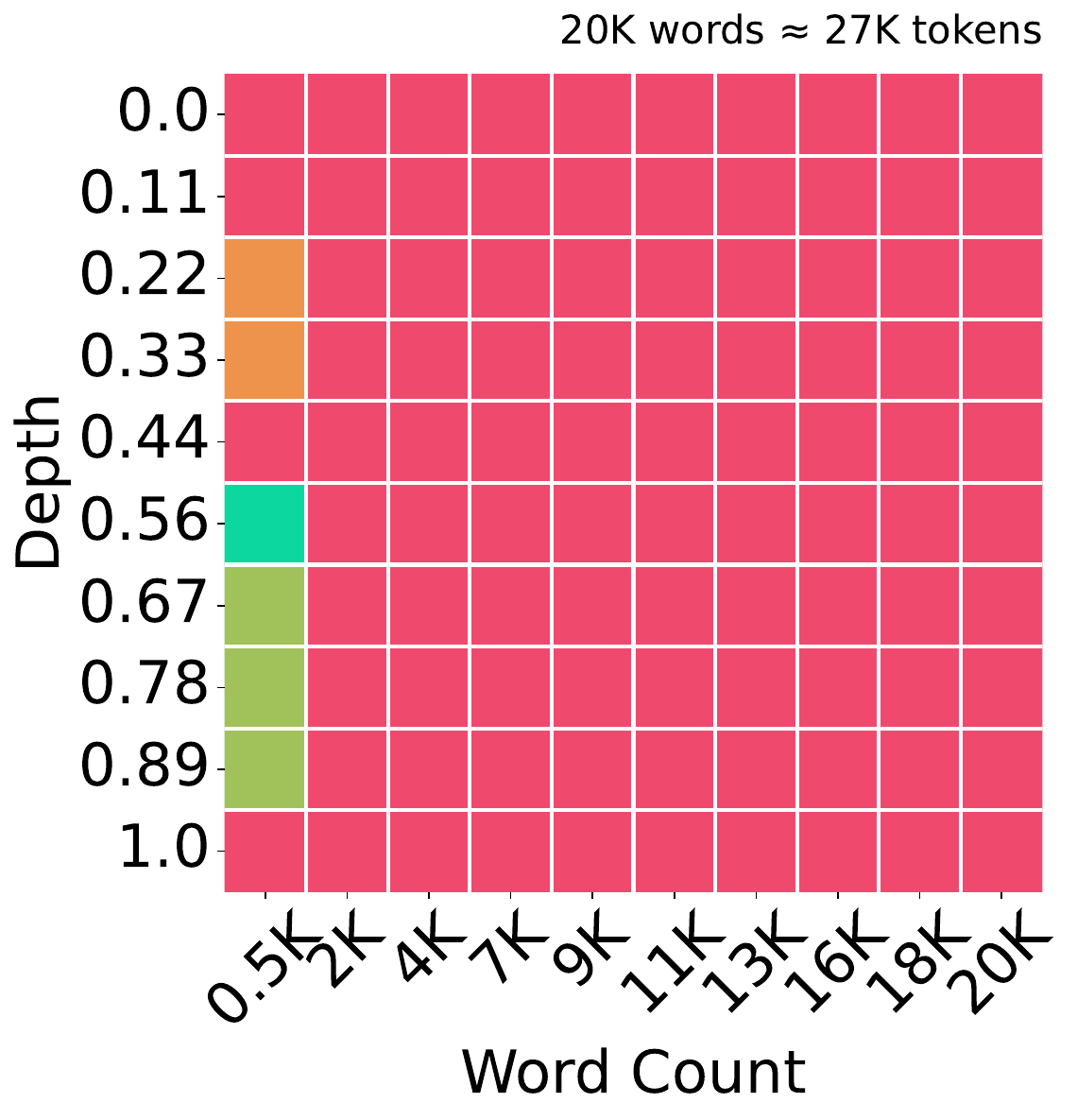}
    \end{minipage}%
}

\subfigure[RecurrentGemma-9B-it]{
\centering
    \begin{minipage}[t]{0.30\linewidth}
        \includegraphics[width=\linewidth]{figures/needle/recurrentgemma_9b.pdf}
    \end{minipage}%
}
\subfigure[RecurrentGemma-2B-it]{
\centering
    \begin{minipage}[t]{0.30\linewidth}
        \includegraphics[width=\linewidth]{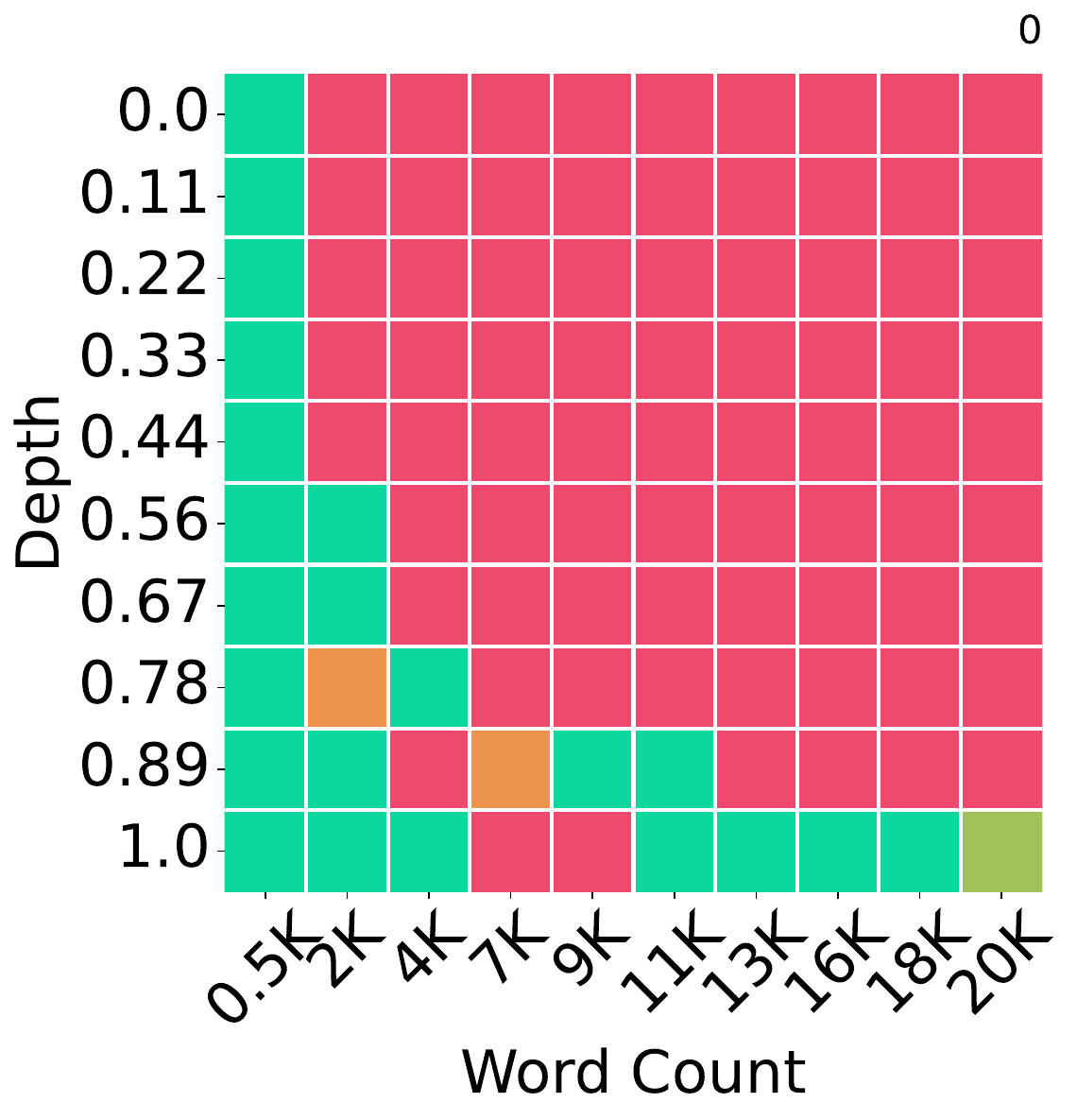}
    \end{minipage}%
}

\caption{Linear-time sequence models under needle test}
\label{fig:needle_lineartime}
\end{figure*}

\clearpage

\newdimen\abovecrulesep
\newdimen\belowcrulesep
\abovecrulesep=0pt
\belowcrulesep=0pt
\makeatletter
\patchcmd{\@@@cmidrule}{\aboverulesep}{\abovecrulesep}{}{}
\patchcmd{\@xcmidrule}{\belowrulesep}{\belowcrulesep}{}{}
\makeatother

\begin{table*}[h]
    \centering
    \caption{Full report of different compression methods on meta-llama/Meta-Llama-3-8B-Instruct across 15 datasets in LongBench}
    \resizebox{1.0\textwidth}{!}{
    \begin{tabular}{c|l|ccc ccc ccc ccc c cc c}
        \toprule
        \multirow{4}{*}{LLM} & \multirow{2}{*}{\diagbox[height=5.5em]{\raisebox{8pt}{Method}}{\raisebox{-8pt}{Dataset}}} & \multicolumn{3}{c}{Single-Document QA} & \multicolumn{3}{c}{Multi-Document QA} & \multicolumn{3}{c}{Summarization} & \multicolumn{3}{c}{Few-shot Learning} & Synthetic & \multicolumn{2}{c}{Code} & \multirow{4}{*}{Avg.} \\
        \cmidrule(lr){3-5} \cmidrule(lr){6-8} \cmidrule(lr){9-11} \cmidrule(lr){12-14} \cmidrule(lr){15-15} \cmidrule(lr){16-17}
        & & \rotatebox[origin=c]{30}{NarrativeQA} & \rotatebox[origin=c]{30}{Qasper} & \rotatebox[origin=c]{30}{MultiFieldQA} & \rotatebox[origin=c]{30}{HotpotQA} & \rotatebox[origin=c]{30}{2WikiMQA} & \rotatebox[origin=c]{30}{Musique} & \rotatebox[origin=c]{30}{GovReport} & \rotatebox[origin=c]{30}{QMSum} & \rotatebox[origin=c]{30}{MultiNews} & \rotatebox[origin=c]{30}{TREC} & \rotatebox[origin=c]{30}{TriviaQA} & \rotatebox[origin=c]{30}{SAMSum} & \rotatebox[origin=c]{30}{PassageRetrieval} & \rotatebox[origin=c]{30}{LCC} & \rotatebox[origin=c]{30}{RepoBench-P} \\
        \midrule
        \multirow{20}{*}{\rotatebox[origin=c]{90}{Meta-Llama-3-8B-Instruct}} 
        & Baseline & 21.7 & 44.3 & 44.5 & 46.6 & 36.4 & 21.5 & 29.9 & 22.6 & 27.7 & 74.0 & 90.6 & 42.7 & 67.0 & 57.2 & 51.2 & 45.2 \\
        & \cellcolor{LightCyan}KIVI-2bit & \cellcolor{LightCyan}21.4 & \cellcolor{LightCyan}43.1 & \cellcolor{LightCyan}44.2 & \cellcolor{LightCyan}46.8 & \cellcolor{LightCyan}37.0 & \cellcolor{LightCyan}20.6 & \cellcolor{LightCyan}29.8 & \cellcolor{LightCyan}22.1 & \cellcolor{LightCyan}27.5 & \cellcolor{LightCyan}74.5 & \cellcolor{LightCyan}90.5 & \cellcolor{LightCyan}42.5 & \cellcolor{LightCyan}67.5 & \cellcolor{LightCyan}50.8 & \cellcolor{LightCyan}46.7 & \cellcolor{LightCyan}44.3 \\
        & KIVI-4bit & 21.0 & 44.8 & 44.6 & 47.0 & 36.5 & 21.4 & 30.1 & 22.5 & 28.0 & 74.5 & 90.3 & 43.1 & 66.5 & 57.3 & 52.0 & 45.3 \\
        & \cellcolor{LightCyan}FlexGen-4bit &  \cellcolor{LightCyan}20.9 & \cellcolor{LightCyan}44.0 & \cellcolor{LightCyan}44.5 & \cellcolor{LightCyan}43.3 & \cellcolor{LightCyan}33.5 & \cellcolor{LightCyan}20.5 & \cellcolor{LightCyan}29.7 & \cellcolor{LightCyan}22.0 & \cellcolor{LightCyan}27.7 & \cellcolor{LightCyan}73.0 & \cellcolor{LightCyan}90.5 & \cellcolor{LightCyan}42.2 & \cellcolor{LightCyan}65.5 & \cellcolor{LightCyan}59.7 & \cellcolor{LightCyan}50.7 & \cellcolor{LightCyan}44.5 \\
        & InfLLM-2x & 19.7 & 38.5 & 37.3 & 40.9 & 30.8 & 20.9 & 29.9 & 20.7 & 26.5 & 69.0 & 91.2 & 42.5 & 57.5 & 57.6 & 54.0 & 42.5 \\
        & \cellcolor{LightCyan}InfLLM-4x & \cellcolor{LightCyan}18.1 & \cellcolor{LightCyan}28.0 & \cellcolor{LightCyan}35.1 & \cellcolor{LightCyan}36.6 & \cellcolor{LightCyan}23.0 & \cellcolor{LightCyan}14.4 & \cellcolor{LightCyan}29.6 & \cellcolor{LightCyan}19.8 & \cellcolor{LightCyan}25.5 & \cellcolor{LightCyan}61.0 & \cellcolor{LightCyan}88.8 & \cellcolor{LightCyan}42.0 & \cellcolor{LightCyan}37.5 & \cellcolor{LightCyan}56.9 & \cellcolor{LightCyan}58.3 & \cellcolor{LightCyan}38.3 \\
        & InfLLM-6x & 19.3 & 24.1 & 29.8 & 35.9 & 19.3 & 15.0 & 28.8 & 19.4 & 24.6 & 56.0 & 85.4 & 41.8 & 29.5 & 60.2 & 58.3 & 36.5 \\
        & \cellcolor{LightCyan}InfLLM-8x & \cellcolor{LightCyan}14.2 & \cellcolor{LightCyan}22.0 & \cellcolor{LightCyan}27.0 & \cellcolor{LightCyan}33.1 & \cellcolor{LightCyan}20.5 & \cellcolor{LightCyan}9.3 & \cellcolor{LightCyan}27.5 & \cellcolor{LightCyan}19.1 & \cellcolor{LightCyan}24.4 & \cellcolor{LightCyan}58.0 & \cellcolor{LightCyan}82.0 & \cellcolor{LightCyan}40.9 & \cellcolor{LightCyan}18.0 & \cellcolor{LightCyan}61.4 & \cellcolor{LightCyan}58.4 & \cellcolor{LightCyan}34.4 \\
        & StreamLLM-2x & 17.3 & 33.5 & 27.6 & 37.0 & 30.3 & 19.0 & 28.1 & 20.1 & 25.5 & 68.0 & 90.4 & 41.1 & 34.0 & 55.0 & 56.2 & 38.9 \\
        & \cellcolor{LightCyan}StreamLLM-4x & \cellcolor{LightCyan}17.4 & \cellcolor{LightCyan}23.0 & \cellcolor{LightCyan}21.1 & \cellcolor{LightCyan}29.8 & \cellcolor{LightCyan}24.7 & \cellcolor{LightCyan}12.0 & \cellcolor{LightCyan}25.9 & \cellcolor{LightCyan}19.5 & \cellcolor{LightCyan}22.6 & \cellcolor{LightCyan}60.5 & \cellcolor{LightCyan}85.7 & \cellcolor{LightCyan}40.5 & \cellcolor{LightCyan}21.0 & \cellcolor{LightCyan}55.0 & \cellcolor{LightCyan}57.2 & \cellcolor{LightCyan}34.4 \\
        & StreamLLM-6x & 15.7 & 18.7 & 17.8 & 26.1 & 19.3 & 10.7 & 24.7 & 18.6 & 20.7 & 58.0 & 82.2 & 40.2 & 14.5 & 59.4 & 58.5 & 32.3 \\
        & \cellcolor{LightCyan}StreamLLM-8x & \cellcolor{LightCyan}13.1 & \cellcolor{LightCyan}16.6 & \cellcolor{LightCyan}17.4 & \cellcolor{LightCyan}25.7 & \cellcolor{LightCyan}18.5 & \cellcolor{LightCyan}9.8 & \cellcolor{LightCyan}23.4 & \cellcolor{LightCyan}18.2 & \cellcolor{LightCyan}19.9 & \cellcolor{LightCyan}55.5 & \cellcolor{LightCyan}72.3 & \cellcolor{LightCyan}39.9 & \cellcolor{LightCyan}8.0 & \cellcolor{LightCyan}60.4 & \cellcolor{LightCyan}55.8 & \cellcolor{LightCyan}30.3 \\

        & $\mathrm{H_2O}$-2x & 21.5 & 42.6 & 43.2 & 46.4 & 36.5 & 21.5 & 28.1 & 22.1 & 26.0 & 74.0 & 90.6 & 42.8 & 66.0 & 57.2 & 51.6 & 44.7 \\
        & \cellcolor{LightCyan}$\mathrm{H_2O}$-4x & \cellcolor{LightCyan}21.8 & \cellcolor{LightCyan}41.2 & \cellcolor{LightCyan}41.8 & \cellcolor{LightCyan}46.8 & \cellcolor{LightCyan}36.9 & \cellcolor{LightCyan}21.5 & \cellcolor{LightCyan}25.7 & \cellcolor{LightCyan}21.4 & \cellcolor{LightCyan}23.7 & \cellcolor{LightCyan}74.0 & \cellcolor{LightCyan}90.6 & \cellcolor{LightCyan}42.4 & \cellcolor{LightCyan}66.0 & \cellcolor{LightCyan}55.1 & \cellcolor{LightCyan}51.2 & \cellcolor{LightCyan}44.0 \\
        & $\mathrm{H_2O}$-6x & 21.5 & 38.3 & 41.8 & 46.8 & 36.8 & 21.7 & 24.5 & 21.1 & 22.5 & 74.0 & 90.5 & 42.6 & 66.0 & 55.1 & 51.1 & 43.6 \\
        & \cellcolor{LightCyan}$\mathrm{H_2O}$-8x & \cellcolor{LightCyan}21.3 & \cellcolor{LightCyan}37.8 & \cellcolor{LightCyan}42.1 & \cellcolor{LightCyan}46.6 & \cellcolor{LightCyan}36.9 & \cellcolor{LightCyan}21.5 & \cellcolor{LightCyan}23.7 & \cellcolor{LightCyan}21.1 & \cellcolor{LightCyan}21.9 & \cellcolor{LightCyan}74.0 & \cellcolor{LightCyan}90.5 & \cellcolor{LightCyan}42.9 & \cellcolor{LightCyan}65.5 & \cellcolor{LightCyan}54.6 & \cellcolor{LightCyan}50.8 & \cellcolor{LightCyan}43.4 \\
        
        & LLMLingua2-2x & 8.0 & 39.1 & 41.0 & 42.5 & 33.9 & 18.1 & 25.8 & 19.8 & 26.6 & 15.5 & 63.9 & 36.5 & 68.0 & 25.8 & 37.9 & 33.5 \\
        & \cellcolor{LightCyan}LLMLingua2-4x & \cellcolor{LightCyan}13.0 & \cellcolor{LightCyan}33.2 & \cellcolor{LightCyan}33.3 & \cellcolor{LightCyan}43.8 & \cellcolor{LightCyan}24.2 & \cellcolor{LightCyan}24.4 & \cellcolor{LightCyan}25.3 & \cellcolor{LightCyan}22.4 & \cellcolor{LightCyan}24.7 & \cellcolor{LightCyan}4.4 & \cellcolor{LightCyan}79.1 & \cellcolor{LightCyan}34.4 & \cellcolor{LightCyan}22.5 & \cellcolor{LightCyan}19.9 & \cellcolor{LightCyan}44.6 & \cellcolor{LightCyan}29.9 \\
        & LLMLingua2-6x & 17.1 & 34.0 & 26.2 & 40.2 & 20.4 & 18.5 & 25.0 & 21.7 & 23.6 & 2.8 & 76.8 & 34.1 & 18.0 & 17.5 & 45.2 & 28.1 \\
        & \cellcolor{LightCyan}LLMLingua2-8x & \cellcolor{LightCyan}19.8 & \cellcolor{LightCyan}28.9 & \cellcolor{LightCyan}23.4 & \cellcolor{LightCyan}35.2 & \cellcolor{LightCyan}23.4 & \cellcolor{LightCyan}17.5 & \cellcolor{LightCyan}24.1 & \cellcolor{LightCyan}21.6 & \cellcolor{LightCyan}22.9 & \cellcolor{LightCyan}0.0 & \cellcolor{LightCyan}76.1 & \cellcolor{LightCyan}34.6 & \cellcolor{LightCyan}13.0 & \cellcolor{LightCyan}16.1 & \cellcolor{LightCyan}47.7 & \cellcolor{LightCyan}26.9 \\
        \bottomrule
    \end{tabular}
    }
    \label{tab:appendix_llama}
\end{table*}

\begin{table*}[h]
    \centering
    \caption{Full report of different compression methods on mistralai/Mistral-7B-Instruct-v0.2 across 15 datasets in LongBench}
    \resizebox{1.0\textwidth}{!}{
    \begin{tabular}{c|l|ccc ccc ccc ccc c cc c}
        \toprule
        \multirow{4}{*}{LLM} & \multirow{2}{*}{\diagbox[height=5.5em]{\raisebox{8pt}{Method}}{\raisebox{-8pt}{Dataset}}} & \multicolumn{3}{c}{Single-Document QA} & \multicolumn{3}{c}{Multi-Document QA} & \multicolumn{3}{c}{Summarization} & \multicolumn{3}{c}{Few-shot Learning} & Synthetic & \multicolumn{2}{c}{Code} & \multirow{4}{*}{Avg.} \\
        \cmidrule(lr){3-5} \cmidrule(lr){6-8} \cmidrule(lr){9-11} \cmidrule(lr){12-14} \cmidrule(lr){15-15} \cmidrule(lr){16-17}
        & & \rotatebox[origin=c]{30}{NarrativeQA} & \rotatebox[origin=c]{30}{Qasper} & \rotatebox[origin=c]{30}{MultiFieldQA} & \rotatebox[origin=c]{30}{HotpotQA} & \rotatebox[origin=c]{30}{2WikiMQA} & \rotatebox[origin=c]{30}{Musique} & \rotatebox[origin=c]{30}{GovReport} & \rotatebox[origin=c]{30}{QMSum} & \rotatebox[origin=c]{30}{MultiNews} & \rotatebox[origin=c]{30}{TREC} & \rotatebox[origin=c]{30}{TriviaQA} & \rotatebox[origin=c]{30}{SAMSum} & \rotatebox[origin=c]{30}{PassageRetrieval} & \rotatebox[origin=c]{30}{LCC} & \rotatebox[origin=c]{30}{RepoBench-P} \\
        \midrule
        \multirow{20}{*}{\rotatebox[origin=c]{90}{Mistral-7B-Instruct-v0.2}} 
        & Baseline & 21.0 & 29.4 & 47.1 & 36.4 & 21.9 & 19.1 & 32.5 & 24.2 & 27.1 & 71.0 & 86.2 & 43.0 & 89.3 & 55.1 & 53.0 & 43.8 \\
        & \cellcolor{LightCyan}KIVI-2bit & \cellcolor{LightCyan}20.6 & \cellcolor{LightCyan}28.4 & \cellcolor{LightCyan}44.9 & \cellcolor{LightCyan}35.5 & \cellcolor{LightCyan}20.7 & \cellcolor{LightCyan}17.9 & \cellcolor{LightCyan}32.5 & \cellcolor{LightCyan}23.5 & \cellcolor{LightCyan}26.7 & \cellcolor{LightCyan}71.0 & \cellcolor{LightCyan}86.0 & \cellcolor{LightCyan}43.5 & \cellcolor{LightCyan}80.8 & \cellcolor{LightCyan}54.7 & \cellcolor{LightCyan}52.8 & \cellcolor{LightCyan}42.6 \\
        & KIVI-4bit & 21.0 & 29.5 & 46.6 & 36.2 & 21.7 & 19.6 & 32.9 & 24.0 & 26.9 & 71.0 & 86.2 & 43.4 & 89.4 & 54.9 & 53.0 & 43.8 \\
        & \cellcolor{LightCyan}FlexGen-4bit & \cellcolor{LightCyan}22.2 & \cellcolor{LightCyan}29.9 & \cellcolor{LightCyan}47.0 & \cellcolor{LightCyan}34.8 & \cellcolor{LightCyan}21.6 & \cellcolor{LightCyan}16.9 & \cellcolor{LightCyan}32.4 & \cellcolor{LightCyan}24.0 & \cellcolor{LightCyan}26.9 & \cellcolor{LightCyan}69.5 & \cellcolor{LightCyan}86.4 & \cellcolor{LightCyan}42.6 & \cellcolor{LightCyan}83.0 & \cellcolor{LightCyan}54.4 & \cellcolor{LightCyan}53.0 & \cellcolor{LightCyan}43.0 \\
        & InfLLM-2x & 21.6 & 24.2 & 46.1 & 35.0 & 20.9 & 18.3 & 31.0 & 23.4 & 25.9 & 67.5 & 86.7 & 41.2 & 65.8 & 54.8 & 53.6 & 41.1 \\
        & \cellcolor{LightCyan}InfLLM-4x & \cellcolor{LightCyan}20.9 & \cellcolor{LightCyan}16.8 & \cellcolor{LightCyan}38.4 & \cellcolor{LightCyan}33.9 & \cellcolor{LightCyan}19.2 & \cellcolor{LightCyan}18.2 & \cellcolor{LightCyan}29.6 & \cellcolor{LightCyan}22.2 & \cellcolor{LightCyan}24.7 & \cellcolor{LightCyan}60.5 & \cellcolor{LightCyan}88.3 & \cellcolor{LightCyan}41.3 & \cellcolor{LightCyan}41.4 & \cellcolor{LightCyan}52.8 & \cellcolor{LightCyan}55.3 & \cellcolor{LightCyan}37.5 \\
        & InfLLM-6x & 19.9 & 14.6 & 36.9 & 31.8 & 16.4 & 14.7 & 29.1 & 22.1 & 23.8 & 57.0 & 87.4 & 40.4 & 32.6 & 53.2 & 53.6 & 35.6 \\
        & \cellcolor{LightCyan}InfLLM-8x & \cellcolor{LightCyan}20.9 & \cellcolor{LightCyan}12.8 & \cellcolor{LightCyan}33.0 & \cellcolor{LightCyan}29.1 & \cellcolor{LightCyan}16.2 & \cellcolor{LightCyan}13.3 & \cellcolor{LightCyan}27.9 & \cellcolor{LightCyan}21.2 & \cellcolor{LightCyan}23.8 & \cellcolor{LightCyan}60.0 & \cellcolor{LightCyan}86.0 & \cellcolor{LightCyan}40.1 & \cellcolor{LightCyan}26.2 & \cellcolor{LightCyan}54.1 & \cellcolor{LightCyan}53.5 & \cellcolor{LightCyan}34.5 \\
        & StreamLLM-2x & 20.7 & 20.6 & 32.6 & 32.3 & 19.0 & 14.7 & 29.9 & 21.6 & 24.4 & 66.5 & 87.0 & 40.0 & 47.1 & 52.3 & 53.7 & 37.5 \\
        & \cellcolor{LightCyan}StreamLLM-4x & \cellcolor{LightCyan}19.7 & \cellcolor{LightCyan}15.1 & \cellcolor{LightCyan}25.4 & \cellcolor{LightCyan}27.7 & \cellcolor{LightCyan}17.4 & \cellcolor{LightCyan}14.6 & \cellcolor{LightCyan}27.4 & \cellcolor{LightCyan}20.2 & \cellcolor{LightCyan}22.1 & \cellcolor{LightCyan}61.0 & \cellcolor{LightCyan}83.7 & \cellcolor{LightCyan}39.2 & \cellcolor{LightCyan}31.6 & \cellcolor{LightCyan}51.8 & \cellcolor{LightCyan}55.9 & \cellcolor{LightCyan}34.2 \\
        & StreamLLM-6x & 17.8 & 12.8 & 24.1 & 24.7 & 13.2 & 10.0 & 25.4 & 20.3 & 20.5 & 59.0 & 81.8 & 38.0 & 25.3 & 52.9 & 56.8 & 32.2 \\
        & \cellcolor{LightCyan}StreamLLM-8x & \cellcolor{LightCyan}16.8 & \cellcolor{LightCyan}11.3 & \cellcolor{LightCyan}22.9 & \cellcolor{LightCyan}22.8 & \cellcolor{LightCyan}12.0 & \cellcolor{LightCyan}10.7 & \cellcolor{LightCyan}24.6 & \cellcolor{LightCyan}19.8 & \cellcolor{LightCyan}19.7 & \cellcolor{LightCyan}56.5 & \cellcolor{LightCyan}79.6 & \cellcolor{LightCyan}38.8 & \cellcolor{LightCyan}16.9 & \cellcolor{LightCyan}53.8 & \cellcolor{LightCyan}56.0 & \cellcolor{LightCyan}30.8 \\

        & $\mathrm{H_2O}$-2x & 21.4 & 27.4 & 47.0 & 36.1 & 20.8 & 19.3 & 31.2 & 23.5 & 25.8 & 71.0 & 86.2 & 43.2 & 87.7 & 54.7 & 52.9 & 43.2 \\
        & \cellcolor{LightCyan}$\mathrm{H_2O}$-4x & \cellcolor{LightCyan}21.6 & \cellcolor{LightCyan}24.9 & \cellcolor{LightCyan}44.8 & \cellcolor{LightCyan}35.0 & \cellcolor{LightCyan}19.0 & \cellcolor{LightCyan}17.5 & \cellcolor{LightCyan}28.6 & \cellcolor{LightCyan}22.8 & \cellcolor{LightCyan}24.2 & \cellcolor{LightCyan}71.0 & \cellcolor{LightCyan}86.7 & \cellcolor{LightCyan}43.8 & \cellcolor{LightCyan}82.9 & \cellcolor{LightCyan}53.9 & \cellcolor{LightCyan}52.3 & \cellcolor{LightCyan}41.9 \\
        & $\mathrm{H_2O}$-6x & 21.5 & 22.7 & 42.9 & 34.5 & 17.2 & 16.6 & 27.1 & 22.5 & 23.2 & 71.0 & 86.4 & 43.5 & 82.0 & 53.1 & 51.9 & 41.1 \\
        & \cellcolor{LightCyan}$\mathrm{H_2O}$-8x & \cellcolor{LightCyan}20.9 & \cellcolor{LightCyan}21.4 & \cellcolor{LightCyan}41.1 & \cellcolor{LightCyan}32.8 & \cellcolor{LightCyan}16.8 & \cellcolor{LightCyan}15.9 & \cellcolor{LightCyan}26.2 & \cellcolor{LightCyan}22.6 & \cellcolor{LightCyan}23.0 & \cellcolor{LightCyan}71.0 & \cellcolor{LightCyan}86.3 & \cellcolor{LightCyan}43.8 & \cellcolor{LightCyan}79.5 & \cellcolor{LightCyan}52.8 & \cellcolor{LightCyan}51.8 & \cellcolor{LightCyan}40.4 \\
        
        & LLMLingua2-2x & 20.2 & 26.8 & 38.9 & 34.7 & 16.8 & 17.7 & 29.9 & 23.6 & 25.8 & 19.0 & 81.2 & 36.5 & 54.9 & 21.9 & 41.5 & 32.6 \\
        & \cellcolor{LightCyan}LLMLingua2-4x & \cellcolor{LightCyan}18.0 & \cellcolor{LightCyan}22.1 & \cellcolor{LightCyan}35.0 & \cellcolor{LightCyan}31.6 & \cellcolor{LightCyan}16.4 & \cellcolor{LightCyan}15.9 & \cellcolor{LightCyan}26.9 & \cellcolor{LightCyan}22.7 & \cellcolor{LightCyan}24.1 & \cellcolor{LightCyan}3.5 & \cellcolor{LightCyan}80.0 & \cellcolor{LightCyan}34.0 & \cellcolor{LightCyan}14.0 & \cellcolor{LightCyan}18.9 & \cellcolor{LightCyan}47.3 & \cellcolor{LightCyan}27.4 \\
        & LLMLingua2-6x & 15.7 & 18.4 & 29.5 & 25.7 & 15.5 & 11.0 & 26.0 & 21.2 & 22.7 & 2.0 & 80.7 & 34.0 & 8.9 & 19.1 & 50.3 & 25.4 \\
        & \cellcolor{LightCyan}LLMLingua2-8x & \cellcolor{LightCyan}15.3 & \cellcolor{LightCyan}16.7 & \cellcolor{LightCyan}26.9 & \cellcolor{LightCyan}23.9 & \cellcolor{LightCyan}15.1 & \cellcolor{LightCyan}8.9 & \cellcolor{LightCyan}25.2 & \cellcolor{LightCyan}21.4 & \cellcolor{LightCyan}22.1 & \cellcolor{LightCyan}0.5 & \cellcolor{LightCyan}81.5 & \cellcolor{LightCyan}33.5 & \cellcolor{LightCyan}8.0 & \cellcolor{LightCyan}19.3 & \cellcolor{LightCyan}51.7 & \cellcolor{LightCyan}24.7 \\
        \bottomrule
    \end{tabular}
    }
    \label{tab:appendix_mistral}
\end{table*}

\begin{table*}[h]
    \centering
    \caption{Full report of different compression methods on lmsys/longchat-7b-v1.5-32k across 15 datasets in LongBench}
    \resizebox{1.0\textwidth}{!}{
    \begin{tabular}{c|l|ccc ccc ccc ccc c cc c}
        \toprule
        \multirow{4}{*}{LLM} & \multirow{2}{*}{\diagbox[height=5.5em]{\raisebox{8pt}{Method}}{\raisebox{-8pt}{Dataset}}} & \multicolumn{3}{c}{Single-Document QA} & \multicolumn{3}{c}{Multi-Document QA} & \multicolumn{3}{c}{Summarization} & \multicolumn{3}{c}{Few-shot Learning} & Synthetic & \multicolumn{2}{c}{Code} & \multirow{4}{*}{Avg.} \\
        \cmidrule(lr){3-5} \cmidrule(lr){6-8} \cmidrule(lr){9-11} \cmidrule(lr){12-14} \cmidrule(lr){15-15} \cmidrule(lr){16-17}
        & & \rotatebox[origin=c]{30}{NarrativeQA} & \rotatebox[origin=c]{30}{Qasper} & \rotatebox[origin=c]{30}{MultiFieldQA} & \rotatebox[origin=c]{30}{HotpotQA} & \rotatebox[origin=c]{30}{2WikiMQA} & \rotatebox[origin=c]{30}{Musique} & \rotatebox[origin=c]{30}{GovReport} & \rotatebox[origin=c]{30}{QMSum} & \rotatebox[origin=c]{30}{MultiNews} & \rotatebox[origin=c]{30}{TREC} & \rotatebox[origin=c]{30}{TriviaQA} & \rotatebox[origin=c]{30}{SAMSum} & \rotatebox[origin=c]{30}{PassageRetrieval} & \rotatebox[origin=c]{30}{LCC} & \rotatebox[origin=c]{30}{RepoBench-P} \\
        \midrule
        \multirow{18}{*}{\rotatebox[origin=c]{90}{LongChat-7b-v1.5-32K}} 
        & Baseline & 20.9 & 29.4 & 43.1 & 33.0 & 24.1 & 14.7 & 30.8 & 22.8 & 26.6 & 66.5 & 84.0 & 40.9 & 30.5 & 52.9 & 56.8 & 38.5 \\
        & \cellcolor{LightCyan}KIVI-2bit & \cellcolor{LightCyan}20.9 & \cellcolor{LightCyan}29.0 & \cellcolor{LightCyan}41.0 & \cellcolor{LightCyan}32.8 & \cellcolor{LightCyan}22.8 & \cellcolor{LightCyan}13.7 & \cellcolor{LightCyan}30.7 & \cellcolor{LightCyan}22.4 & \cellcolor{LightCyan}26.4 & \cellcolor{LightCyan}66.5 & \cellcolor{LightCyan}83.2 & \cellcolor{LightCyan}41.2 & \cellcolor{LightCyan}32.2 & \cellcolor{LightCyan}52.4 & \cellcolor{LightCyan}55.4 & \cellcolor{LightCyan}38.0 \\
        & KIVI-4bit & 21.0 & 28.9 & 43.3 & 33.1 & 24.9 & 14.7 & 31.1 & 22.7 & 26.5 & 67.0 & 83.9 & 40.8 & 31.5 & 52.2 & 56.3 & 38.5 \\
        & \cellcolor{LightCyan}FlexGen-4bit & \cellcolor{LightCyan}20.5 & \cellcolor{LightCyan}30.4 & \cellcolor{LightCyan}43.2 & \cellcolor{LightCyan}33.7 & \cellcolor{LightCyan}23.8 & \cellcolor{LightCyan}13.9 & \cellcolor{LightCyan}31.7 & \cellcolor{LightCyan}22.9 & \cellcolor{LightCyan}26.5 & \cellcolor{LightCyan}66.0 & \cellcolor{LightCyan}81.5 & \cellcolor{LightCyan}40.9 & \cellcolor{LightCyan}31.5 & \cellcolor{LightCyan}50.4 & \cellcolor{LightCyan}56.3 & \cellcolor{LightCyan}38.2 \\
        & InfLLM-2x & 19.1 & 28.4 & 40.0 & 30.1 & 26.6 & 13.2 & 31.1 & 21.7 & 24.6 & 60.0 & 84.1 & 11.3 & 19.5 & 48.9 & 54.9 & 34.2 \\
        & \cellcolor{LightCyan}InfLLM-4x & \cellcolor{LightCyan}17.6 & \cellcolor{LightCyan}21.2 & \cellcolor{LightCyan}33.9 & \cellcolor{LightCyan}32.4 & \cellcolor{LightCyan}25.4 & \cellcolor{LightCyan}14.4 & \cellcolor{LightCyan}29.4 & \cellcolor{LightCyan}21.5 & \cellcolor{LightCyan}22.3 & \cellcolor{LightCyan}55.0 & \cellcolor{LightCyan}84.3 & \cellcolor{LightCyan}9.9 & \cellcolor{LightCyan}9.5 & \cellcolor{LightCyan}41.4 & \cellcolor{LightCyan}52.1 & \cellcolor{LightCyan}31.4 \\
        & InfLLM-6x & 16.5 & 17.0 & 29.8 & 30.6 & 24.6 & 15.2 & 28.4 & 21.0 & 21.1 & 55.5 & 82.4 & 8.7 & 10.0 & 41.4 & 51.6 & 30.2 \\
        & \cellcolor{LightCyan}InfLLM-8x & \cellcolor{LightCyan}14.9 & \cellcolor{LightCyan}16.5 & \cellcolor{LightCyan}29.1 & \cellcolor{LightCyan}26.4 & \cellcolor{LightCyan}23.1 & \cellcolor{LightCyan}15.4 & \cellcolor{LightCyan}26.6 & \cellcolor{LightCyan}20.7 & \cellcolor{LightCyan}20.5 & \cellcolor{LightCyan}48.5 & \cellcolor{LightCyan}78.8 & \cellcolor{LightCyan}8.7 & \cellcolor{LightCyan}6.0 & \cellcolor{LightCyan}41.6 & \cellcolor{LightCyan}50.8 & \cellcolor{LightCyan}28.5 \\
        & StreamLLM-2x & 18.8 & 26.3 & 26.6 & 29.1 & 24.8 & 11.5 & 28.5 & 20.8 & 23.3 & 61.0 & 82.9 & 9.6 & 25.0 & 43.9 & 54.6 & 32.5 \\
        & \cellcolor{LightCyan}StreamLLM-4x & \cellcolor{LightCyan}18.1 & \cellcolor{LightCyan}19.1 & \cellcolor{LightCyan}22.4 & \cellcolor{LightCyan}29.5 & \cellcolor{LightCyan}25.1 & \cellcolor{LightCyan}11.6 & \cellcolor{LightCyan}25.5 & \cellcolor{LightCyan}20.9 & \cellcolor{LightCyan}20.4 & \cellcolor{LightCyan}54.5 & \cellcolor{LightCyan}81.1 & \cellcolor{LightCyan}11.4 & \cellcolor{LightCyan}13.5 & \cellcolor{LightCyan}49.7 & \cellcolor{LightCyan}54.2 & \cellcolor{LightCyan}30.5 \\
        & StreamLLM-6x & 17.7 & 17.7 & 23.3 & 26.0 & 26.8 & 11.5 & 23.4 & 20.4 & 18.2 & 54.5 & 78.1 & 9.9 & 10.5 & 49.5 & 53.4 & 29.4 \\
        & \cellcolor{LightCyan}StreamLLM-8x & \cellcolor{LightCyan}13.9 & \cellcolor{LightCyan}16.7 & \cellcolor{LightCyan}21.9 & \cellcolor{LightCyan}24.3 & \cellcolor{LightCyan}26.4 & \cellcolor{LightCyan}13.6 & \cellcolor{LightCyan}21.6 & \cellcolor{LightCyan}19.8 & \cellcolor{LightCyan}17.1 & \cellcolor{LightCyan}49.0 & \cellcolor{LightCyan}73.9 & \cellcolor{LightCyan}10.4 & \cellcolor{LightCyan}10.0 & \cellcolor{LightCyan}52.4 & \cellcolor{LightCyan}51.4 & \cellcolor{LightCyan}28.2 \\

        & $\mathrm{H_2O}$-2x & 20.9 & 27.1 & 35.0 & 30.8 & 22.6 & 12.8 & 28.0 & 21.9 & 24.0 & 66.0 & 82.1 & 39.8 & 30.5 & 59.5 & 56.0 & 37.1 \\
        & \cellcolor{LightCyan}$\mathrm{H_2O}$-4x & \cellcolor{LightCyan}21.4 & \cellcolor{LightCyan}25.3 & \cellcolor{LightCyan}32.1 & \cellcolor{LightCyan}30.6 & \cellcolor{LightCyan}22.9 & \cellcolor{LightCyan}12.2 & \cellcolor{LightCyan}23.0 & \cellcolor{LightCyan}21.7 & \cellcolor{LightCyan}21.1 & \cellcolor{LightCyan}65.5 & \cellcolor{LightCyan}80.6 & \cellcolor{LightCyan}39.7 & \cellcolor{LightCyan}28.5 & \cellcolor{LightCyan}56.2 & \cellcolor{LightCyan}54.3 & \cellcolor{LightCyan}35.7 \\
        & $\mathrm{H_2O}$-6x & 21.1 & 23.7 & 32.1 & 29.7 & 21.6 & 12.7 & 21.5 & 21.4 & 19.9 & 65.5 & 81.0 & 39.7 & 28.0 & 53.2 & 53.4 & 35.0 \\
        & \cellcolor{LightCyan}$\mathrm{H_2O}$-8x & \cellcolor{LightCyan}20.4 & \cellcolor{LightCyan}22.4 & \cellcolor{LightCyan}32.7 & \cellcolor{LightCyan}29.3 & \cellcolor{LightCyan}21.2 & \cellcolor{LightCyan}12.3 & \cellcolor{LightCyan}20.1 & \cellcolor{LightCyan}20.9 & \cellcolor{LightCyan}18.5 & \cellcolor{LightCyan}65.5 & \cellcolor{LightCyan}80.5 & \cellcolor{LightCyan}38.8 & \cellcolor{LightCyan}28.5 & \cellcolor{LightCyan}50.2 & \cellcolor{LightCyan}52.7 & \cellcolor{LightCyan}34.3 \\
        
        & LLMLingua2-2x & 13.3 & 27.5 & 36.3 & 28.5 & 25.4 & 13.0 & 28.5 & 22.3 & 25.6 & 6.0 & 65.3 & 34.8 & 19.5 & 16.1 & 49.0 & 27.4 \\
        & \cellcolor{LightCyan}LLMLingua2-4x & \cellcolor{LightCyan}14.4 & \cellcolor{LightCyan}26.3 & \cellcolor{LightCyan}30.7 & \cellcolor{LightCyan}27.2 & \cellcolor{LightCyan}24.0 & \cellcolor{LightCyan}10.7 & \cellcolor{LightCyan}25.1 & \cellcolor{LightCyan}22.0 & \cellcolor{LightCyan}23.4 & \cellcolor{LightCyan}1.0 & \cellcolor{LightCyan}61.9 & \cellcolor{LightCyan}32.0 & \cellcolor{LightCyan}5.5 & \cellcolor{LightCyan}16.0 & \cellcolor{LightCyan}47.7 & \cellcolor{LightCyan}24.5 \\
        & LLMLingua2-6x & 14.7 & 25.6 & 27.6 & 24.3 & 24.7 & 11.7 & 23.8 & 21.6 & 22.4 & 0.0 & 64.4 & 32.6 & 5.0 & 15.2 & 48.6 & 24.1 \\
        & \cellcolor{LightCyan}LLMLingua2-8x & \cellcolor{LightCyan}14.6 & \cellcolor{LightCyan}24.4 & \cellcolor{LightCyan}24.9 & \cellcolor{LightCyan}23.8 & \cellcolor{LightCyan}23.5 & \cellcolor{LightCyan}11.2 & \cellcolor{LightCyan}23.1 & \cellcolor{LightCyan}21.4 & \cellcolor{LightCyan}21.4 & \cellcolor{LightCyan}0.5 & \cellcolor{LightCyan}66.6 & \cellcolor{LightCyan}31.5 & \cellcolor{LightCyan}6.5 & \cellcolor{LightCyan}16.7 & \cellcolor{LightCyan}48.3 & \cellcolor{LightCyan}23.9 \\
        \bottomrule
    \end{tabular}
    }
    \label{tab:appendix_longchat}
\end{table*}

\begin{table*}[h]
    \centering
    \caption{Full report of linear-time sequence models and mixed architecture across 15 datasets in LongBench}
    \resizebox{1.0\textwidth}{!}{
    \begin{tabular}{c|l|ccc ccc ccc ccc c cc c}
        \toprule
        \multirow{4}{*}{LLM} & \multirow{2}{*}{\diagbox[height=5.5em]{\raisebox{8pt}{Method}}{\raisebox{-8pt}{Dataset}}} & \multicolumn{3}{c}{Single-Document QA} & \multicolumn{3}{c}{Multi-Document QA} & \multicolumn{3}{c}{Summarization} & \multicolumn{3}{c}{Few-shot Learning} & Synthetic & \multicolumn{2}{c}{Code} & \multirow{4}{*}{Avg.} \\
        \cmidrule(lr){3-5} \cmidrule(lr){6-8} \cmidrule(lr){9-11} \cmidrule(lr){12-14} \cmidrule(lr){15-15} \cmidrule(lr){16-17}
        & & \rotatebox[origin=c]{30}{NarrativeQA} & \rotatebox[origin=c]{30}{Qasper} & \rotatebox[origin=c]{30}{MultiFieldQA} & \rotatebox[origin=c]{30}{HotpotQA} & \rotatebox[origin=c]{30}{2WikiMQA} & \rotatebox[origin=c]{30}{Musique} & \rotatebox[origin=c]{30}{GovReport} & \rotatebox[origin=c]{30}{QMSum} & \rotatebox[origin=c]{30}{MultiNews} & \rotatebox[origin=c]{30}{TREC} & \rotatebox[origin=c]{30}{TriviaQA} & \rotatebox[origin=c]{30}{SAMSum} & \rotatebox[origin=c]{30}{PassageRetrieval} & \rotatebox[origin=c]{30}{LCC} & \rotatebox[origin=c]{30}{RepoBench-P} \\
        \midrule
        \multirow{3}{*}{Mamba} 
        & Mamba-2.8B & 2.7 & 5.8 & 13.3 & 6.2 & 9.2 & 3.6 & 17.9 & 16.6 & 22.9 & 50.0 & 54.4 & 12.5 & 1.2 & 50.6 & 44.5 & 20.8 \\
        & \cellcolor{LightCyan}Mamba-Chat-2.8B & \cellcolor{LightCyan}3.2 & \cellcolor{LightCyan}6.3 & \cellcolor{LightCyan}17.9 & \cellcolor{LightCyan}7.3 & \cellcolor{LightCyan}9.5 & \cellcolor{LightCyan}4.0 & \cellcolor{LightCyan}21.5 & \cellcolor{LightCyan}18.5 & \cellcolor{LightCyan}23.5 & \cellcolor{LightCyan}45.5 & \cellcolor{LightCyan}43.2 & \cellcolor{LightCyan}23.9 & \cellcolor{LightCyan}3.7 & \cellcolor{LightCyan}50.5 & \cellcolor{LightCyan}45.0 & \cellcolor{LightCyan}21.6 \\
        & Mamba2-2.7B & 2.6 & 5.6 & 14.4 & 7.9 & 8.7 & 3.5 & 20.3 & 17.8 & 24.8 & 45.0 & 58.2 & 18.4 & 4.1 & 54.4 & 45.5 & 22.1 \\
        \midrule
        \multirow{1}{*}{RWKV} 
        & \cellcolor{LightCyan}RWKV-5-World-7B & \cellcolor{LightCyan}3.3 & \cellcolor{LightCyan}9.8 & \cellcolor{LightCyan}16.2 & \cellcolor{LightCyan}6.5 & \cellcolor{LightCyan}7.6 & \cellcolor{LightCyan}2.2 & \cellcolor{LightCyan}21.5 & \cellcolor{LightCyan}16.2 & \cellcolor{LightCyan}17.9 & \cellcolor{LightCyan}61.0 & \cellcolor{LightCyan}77.2 & \cellcolor{LightCyan}18.9 & \cellcolor{LightCyan}4.5 & \cellcolor{LightCyan}36.2 & \cellcolor{LightCyan}31.8 & \cellcolor{LightCyan}22.1 \\
        \midrule
        \multirow{2}{*}{R-Gemma} 
        & R-Gemma-2B-it & 12.0 & 16.2 & 26.0 & 9.8 & 10.8 & 4.3 & 20.7 & 20.0 & 22.1 & 52.0 & 63.3 & 23.6 & 4.0 & 57.0 & 50.3 & 26.1 \\
        & \cellcolor{LightCyan}R-Gemma-9B-it & \cellcolor{LightCyan}15.4 & \cellcolor{LightCyan}25.8 & \cellcolor{LightCyan}32.3 & \cellcolor{LightCyan}25.4 & \cellcolor{LightCyan}27.3 & \cellcolor{LightCyan}13.0 & \cellcolor{LightCyan}24.6 & \cellcolor{LightCyan}18.1 & \cellcolor{LightCyan}23.0 & \cellcolor{LightCyan}60.5 & \cellcolor{LightCyan}70.5 & \cellcolor{LightCyan}32.3 & \cellcolor{LightCyan}9.0 & \cellcolor{LightCyan}64.1 & \cellcolor{LightCyan}57.5 & \cellcolor{LightCyan}33.2 \\
        \bottomrule
    \end{tabular}
}
    \label{tab:appendix_rnn}
\end{table*}






\end{document}